\pdfoutput=1

\documentclass[11pt]{article}

\usepackage{coling}

\usepackage{times}
\usepackage{latexsym}

\usepackage[T1]{fontenc}

\usepackage[utf8]{inputenc}

\usepackage{microtype}

\usepackage{inconsolata}

\usepackage{graphicx}

\usepackage{danudefs}
\usepackage{algorithmic}
\usepackage{algorithm}
\usepackage{color}
\usepackage{booktabs}
\usepackage{xspace}
\usepackage{url}
\usepackage{braket}
\usepackage[T1]{fontenc}
\usepackage[utf8]{inputenc}
\usepackage{graphicx}
\usepackage{enumitem}
\usepackage{subcaption}
\usepackage{multirow}
\usepackage{multicol}
\usepackage{comment}

\usepackage[english]{babel}
\usepackage{hyperref}
\addto\captionsenglish{%
}
\addto\extrasenglish{%
}

\usepackage{acronym}

\usepackage{etoolbox}
\makeatletter
\newif\if@in@acrolist
\AtBeginEnvironment{acronym}{\@in@acrolisttrue}
\newrobustcmd{\LU}[2]{\if@in@acrolist#1\else#2\fi}

\newcommand{\ACF}[1]{{\@in@acrolisttrue\acf{#1}}}
\makeatother

\acrodef{MLM}[MLM]{Masked Language Model}
\acrodefplural{MLM}[MLMs]{\LU{M}{m}asked \LU{L}{l}anguage \LU{M}{m}odels}
\acrodef{LLM}[LLM]{Large Language Model}
\acrodefplural{LLM}[LLMs]{Large Language Models}
\acrodef{SoTA}[SoTA]{state-of-the-art}
\acrodef{ICL}{\LU{I}{i}n-cotext \LU{L}{l}earning}
\acrodef{SCD}{\LU{S}{s}emantic \LU{C}{c}hange \LU{D}{d}etection}
\acrodef{WiC}{Word-in-Context}
\acrodef{ITML}[ITML]{Information-Theoretic Metric Learning}
\acrodef{SDML}[SDML]{Semantic Distance Metric Learning}
\acrodef{CWE}{\LU{C}{c}ontextualised \LU{W}{w}ord \LU{E}{e}mbedding}
\acrodefplural{CWE}[CWEs]{\LU{C}{c}ontextualised \LU{W}{w}ord \LU{E}{e}mbeddings}
\acrodef{SCWE}[SCWE]{Sense-aware Contextualised Word Embedding}
\acrodefplural{SCWE}[SCWEs]{\LU{S}{s}ense-aware \LU{C}{c}ontextualised \LU{W}{w}ord \LU{E}{e}mbeddings}
\acrodef{PCA}[PCA]{Principal Component Analysis}
\acrodef{ICA}[ICA]{Independent Component Analysis}
\acrodef{ROC}[ROC]{Receiver Operating Characteristic}
\acrodef{AUC}[AUC]{Area Under the Curve}

\title{Investigating the Contextualised Word Embedding Dimensions Specified for Contextual and Temporal Semantic Changes}

\author{Taichi Aida \\
  Tokyo Metropolitan University \\
  {\tt aida-taichi@ed.tmu.ac.jp}
  \And Danushka Bollegala \\
  University of Liverpool \\
  {\tt danushka@liverpool.ac.uk}}

\begin{document}
\maketitle
\begin{abstract}
The \Acp{SCWE} encode semantic changes of words within the \ac{CWE} spaces.
Despite the superior performance of \Acp{SCWE} in contextual/temporal \ac{SCD} benchmarks, it remains unclear as to how the meaning changes are encoded in the embedding space.
To study this, we compare pre-trained \acp{CWE} and their fine-tuned versions on contextual and temporal semantic change benchmarks under \ac{PCA} and \ac{ICA} transformations.
Our experimental results reveal
(a) although there exist a smaller number of axes that are specific to semantic changes of words in the pre-trained \ac{CWE} space, this information gets distributed across all dimensions when fine-tuned, and
(b) in contrast to prior work studying the geometry of \acp{CWE}, we find that \ac{PCA} to better represent semantic changes than \ac{ICA} within the top 10\% of axes.
These findings encourage the development of more efficient \ac{SCD} methods with a small number of \ac{SCD}-aware dimensions.\footnote{Source code is available at \url{https://github.com/LivNLP/svp-dims} .}

\end{abstract}

\section{Introduction}
\label{sec:intro}


Meaning of a word is a dynamic phenomenon that is both \emph{contextual} (i.e. depends on the context in which the word is used)~\cite{pilehvar-camacho-collados-2019-wic} as well as \emph{temporal} (i.e. the meaning of a word can change over time)~\cite{tahmasebia-etal-2021-survey}.
A large body of methods have been proposed to represent the meaning of a word in a given context~\cite{BERT,conneau-etal-2020-xlmroberta,yi-zhou-2021-learning,rachinskiy-arefyev-2021-glossreader,periti-etal-2024-automatically}, or within a given time period~\cite{hamilton-etal-2016-diachronic,rosenfeld-erk-2018-deep,aida-etal-2021-comprehensive,rosin-etal-2022-time,aida-bollegala-2023-unsupervised,Tang2023-yu, fedorova-etal-2024-definition}. 
In particular, \acp{SCWE} such as XL-LEXEME~\cite{cassotti-etal-2023-xl} obtained by fine-tuning \acp{MLM} such as XLM-RoBERTa~\cite{conneau-etal-2020-xlmroberta} on \ac{WiC}~\cite{pilehvar-camacho-collados-2019-wic} have reported superior performance in \ac{SCD} benchmarks~\cite{cassotti-etal-2023-xl,aida-bollegala-2023-swap,periti-tahmasebi-2024-systematic,aida-bollegala-2024-semantic}, implying that semantic changes can be accurately inferred from \acp{SCWE}.

\begin{table*}[t!]
    \centering
    \small
    \resizebox{\textwidth}{!}{
    \begin{tabular}{lp{40mm}p{40mm}l} \toprule
        \multicolumn{1}{c}{Type of Semantic Change} & \multicolumn{2}{c}{Instances} & \multicolumn{1}{c}{Label} \\ \midrule
        \multirow{4}{*}{Contextual} & \dots two points on a \textit{plane} lies \dots & \dots the \textit{plane} graph as the X-Y \dots & \multirow{2}{*}{True (Same meanings)} \\
        & He lived on a worldly \textit{plane}. & \dots the \textit{plane} graph as the X-Y \dots & \multirow{2}{*}{False (Different meanings)} \\ \hline
        Temporal & 
        \begin{minipage}{40mm}
            \begin{itemize}
                \setlength{\leftskip}{-4mm}
                \item \dots this is a horizontal \textit{plane}, and \dots
                \item \dots because it is parallel with the ground \textit{plane} \dots
                \item \dots this is a horizontal \textit{plane}, \dots
            \end{itemize}
        \end{minipage} &
        \begin{minipage}{40mm}
            \begin{itemize}
                \setlength{\leftskip}{-4mm}
                \item \dots as the \textit{plane} settled down at \dots
                \item \dots 558 combat \textit{planes} and 4,000 tanks.
                \item The President's \textit{plane} landed at \dots
            \end{itemize}
        \end{minipage} & 
        True (Semantically Changed) \\\bottomrule
    \end{tabular}
    }
    \caption{Examples of contextual/temporal semantic change tasks. In contextual semantic change tasks, models predict the meanings of a target word (e.g. plane) in \textbf{each pair of sentences in the same time period}. On the other hand,  in temporal semantic change tasks, models predict the meaning of a target word (e.g. plane) from \textbf{sets of sentences across different time periods}.}
    \label{tab:scd_example}
\end{table*}

Despite the empirical success, to the best of our knowledge, no prior work has investigated \textbf{whether there are dedicated dimensions in the XL-LEXEME embedding space specified for the semantic changes of the words} it represents.
In this paper, we study this problem from two complementary directions.
First, in \autoref{sec:contextual}, we investigate the embedding dimensions specific to the contextual semantic changes of words using \ac{WiC} benchmarks~\cite{pilehvar-camacho-collados-2019-wic,raganato-etal-2020-xlwic,martelli-etal-2021-mclwic,liu-etal-2021-am2ico} as the evaluation task.
Second, in \autoref{sec:temporal}, we investigate the embedding dimensions specific to the temporal semantic changes of words on  SemEval-2020 Task 1~\cite{schlechtweg-etal-2020-semeval} benchmark.
In each setting, we compare pre-trained \acp{CWE} and the \acp{SCWE} obtained by fine-tuning on \ac{WiC} using \ac{PCA} and \ac{ICA}, which have been used in prior work investigating dimensions in \acp{CWE}~\cite{yamagiwa-etal-2023-discovering}.
Our investigations reveal several interesting novel insights that will be useful when developing accurate and efficient low-dimensional \ac{SCD} methods as follows.
\begin{itemize}[leftmargin=*]
    \item \ac{PCA} discovers contextual/temporal semantic change-aware axes within the top 10\% of the transformed axes better than \ac{ICA}.
    \item In pre-trained embeddings, we identify a small number of axes that are specified for contextual/temporal semantic changes, while such axes are uniformly distributed in the fine-tuned embeddings.
    \item Semantic change aware dimensions report comparable or superior performance over using all dimensions in \ac{SCD} benchmarks.
\end{itemize}

\section{Task Description}
\label{sec:example_scd}
In this section, we explain the two types of semantic changes of words considered in the paper:
(a) contextual semantic changes and (b) temporal semantic changes.
\paragraph{Contextual Semantic Change Detection Task} involves predicting whether the meaning of a word in a given pair of sentences are the same~\cite{pilehvar-camacho-collados-2019-wic}. For example, an ambiguous word can express different meanings in different contexts, which is considered under contextual semantic changes. Models are required to make a prediction for each pair of sentences.
\paragraph{Temporal Semantic Change Detection Task} involves predicting the meanings of a word in given sets of sentences across different time periods~\cite{schlechtweg-etal-2020-semeval}. A word that was used in a different meaning in the past can be associated with novel meanings later on, which is considered as a temporal semantic change of that word. Models predict whether the meaning of the word has changed over time by comparing the given sets of sentences.
\paragraph{Models} For the \textbf{Contextual Semantic Change Detection Task}, contextual word embeddings~\cite{BERT,conneau-etal-2020-xlmroberta} are the primary choice, as they effectively capture word meanings based on sentence context. For the \textbf{Temporal Semantic Change Detection Task}, both static~\cite{kim-etal-2014-temporal,kulkarni-etal-2015-statistically,hamilton-etal-2016-diachronic,yao-etal-2018-dynamic,aida-etal-2021-comprehensive} and contextual~\cite{rosenfeld-erk-2018-deep,kutuzov-giulianelli-2020-uio,laicher-etal-2021-explaining,aida-bollegala-2023-unsupervised} embeddings can be applied. Notably, sense-aware contextual embeddings trained specifically for contextual semantic change tasks have achieved superior performance, demonstrating their broader applicability~\cite{cassotti-etal-2023-xl,aida-bollegala-2024-semantic}.

Both types of semantic changes are common and even the same word can undergo both types of semantic changes as shown in \autoref{tab:scd_example} for the word \textit{plane}.
The contextual semantic change task requires models to be sensitive to the context within just two given sentences, whereas
the temporal semantic change task requires models to account for the semantic changes of words across two different time periods.

\begin{figure*}[t]
    \centering
    \begin{minipage}[b]{0.65\columnwidth}
        \centering
        \includegraphics[width=0.85\columnwidth]{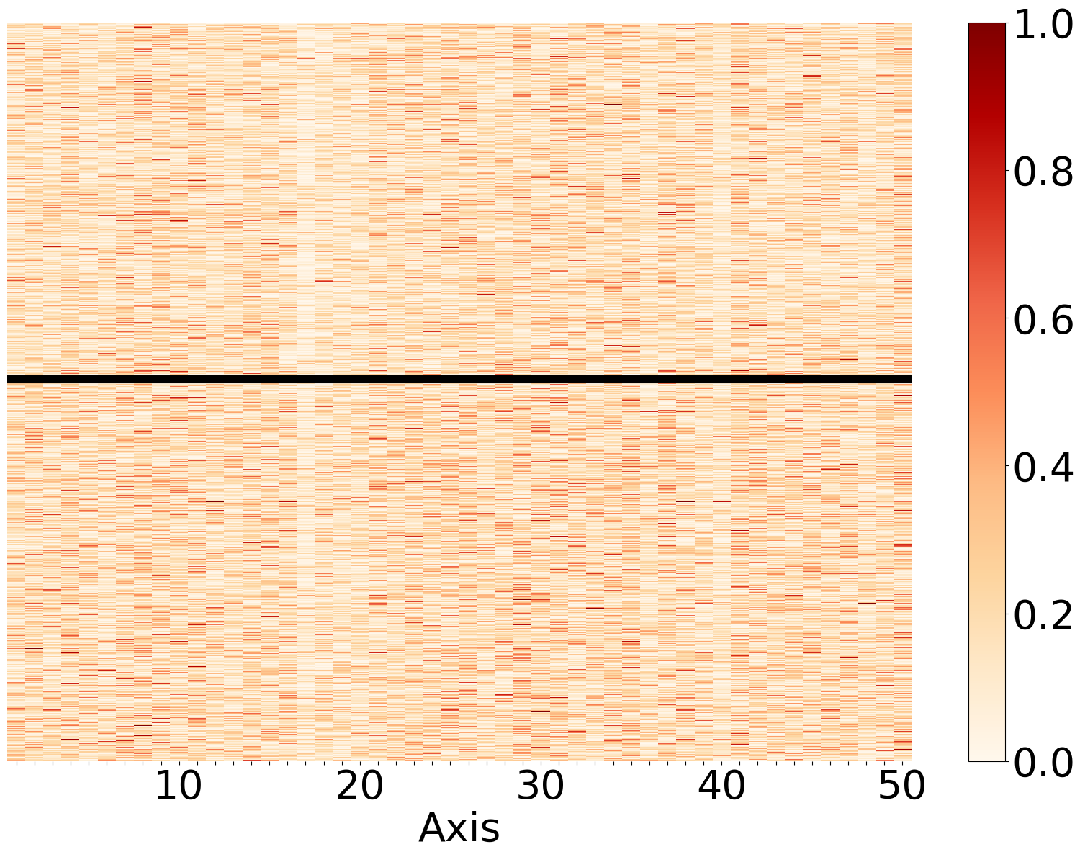}
        \subcaption{Pre-trained \ac{CWE}, Raw}
        \label{fig:wic_instance_pretrained_raw}
    \end{minipage}
    \begin{minipage}[b]{0.65\columnwidth}
        \centering
        \includegraphics[width=0.85\columnwidth]{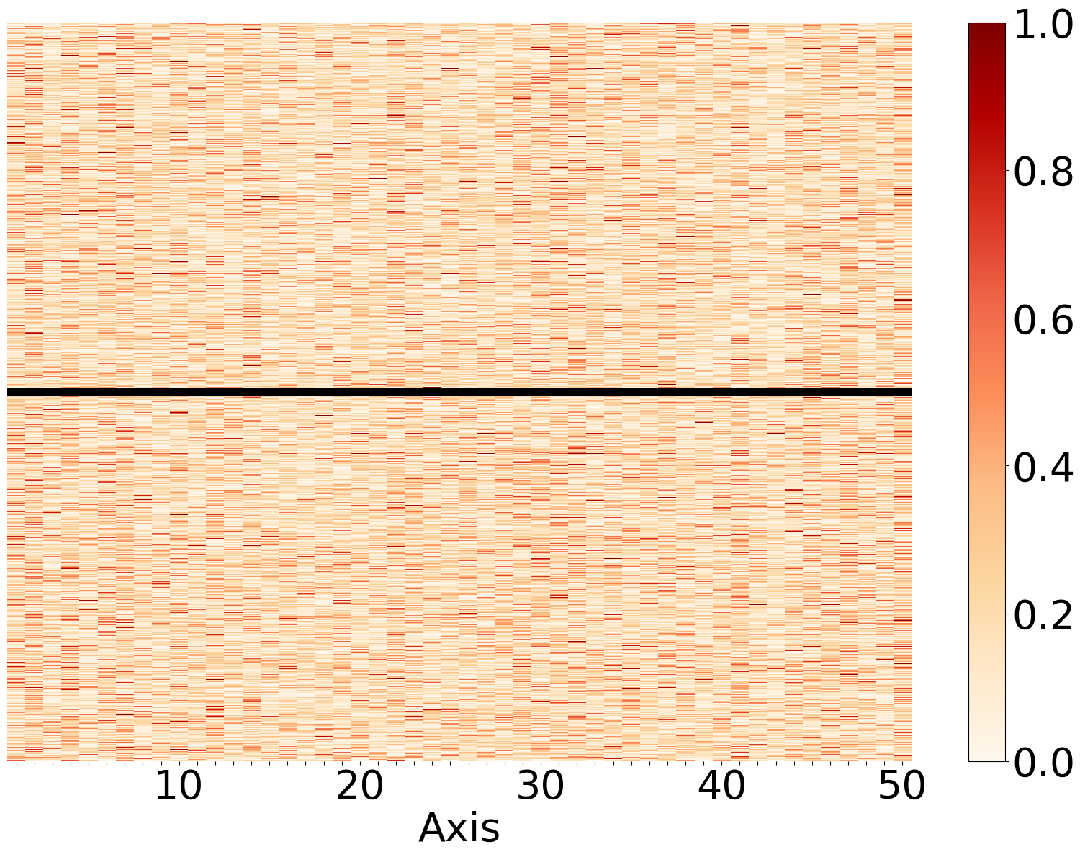}
        \subcaption{Pre-trained \ac{CWE}, PCA}
        \label{fig:wic_instance_pretrained_pca}
    \end{minipage}
    \begin{minipage}[b]{0.65\columnwidth}
        \centering
        \includegraphics[width=0.85\columnwidth]{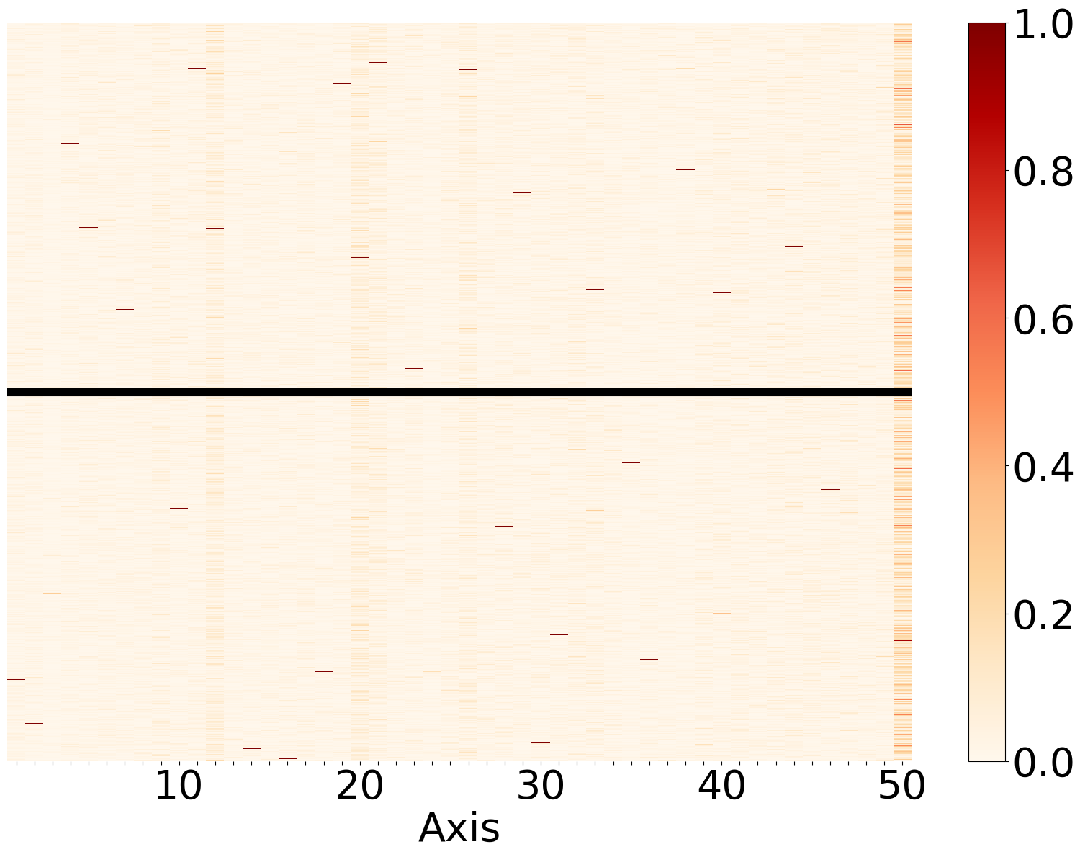}
        \subcaption{Pre-trained \ac{CWE}, ICA}
        \label{fig:wic_instance_pretrained_ica}
    \end{minipage} \\
    \begin{minipage}[b]{0.65\columnwidth}
        \centering
        \includegraphics[width=0.85\columnwidth]{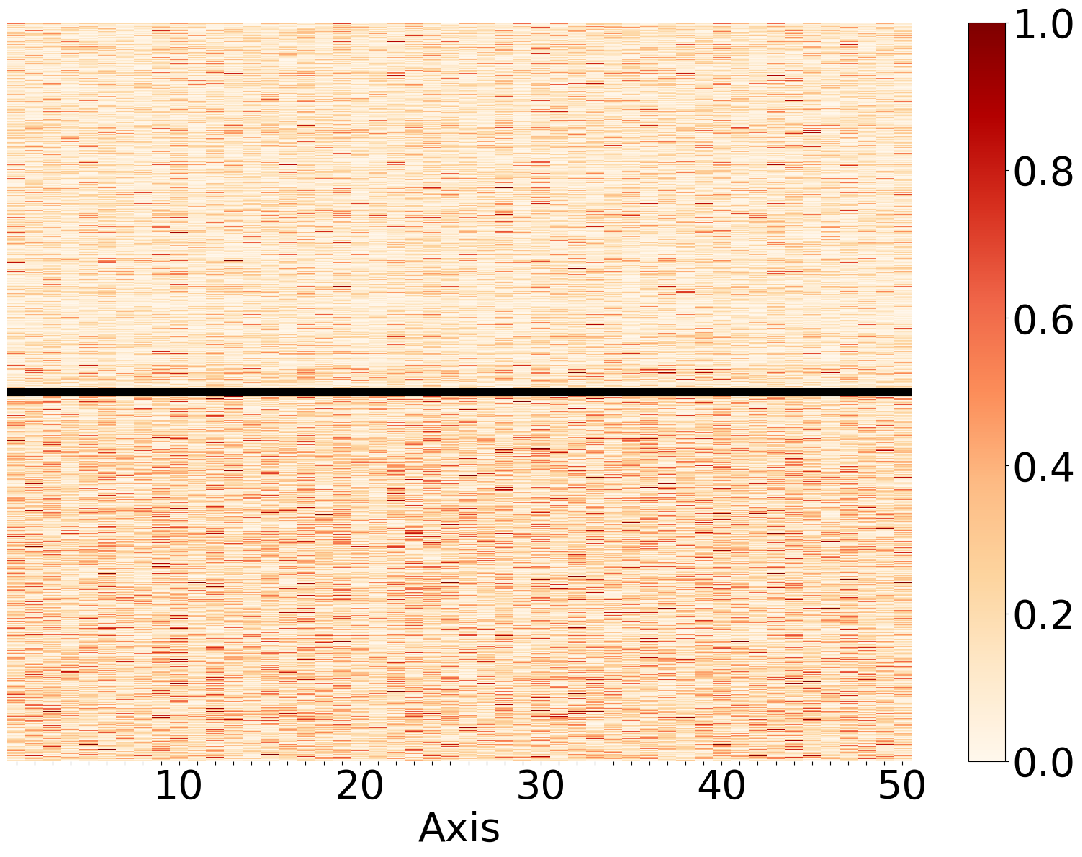}
        \subcaption{Fine-tuned \ac{SCWE}, Raw}
        \label{fig:wic_instance_finetuned_raw}
    \end{minipage}
    \begin{minipage}[b]{0.65\columnwidth}
        \centering
        \includegraphics[width=0.85\columnwidth]{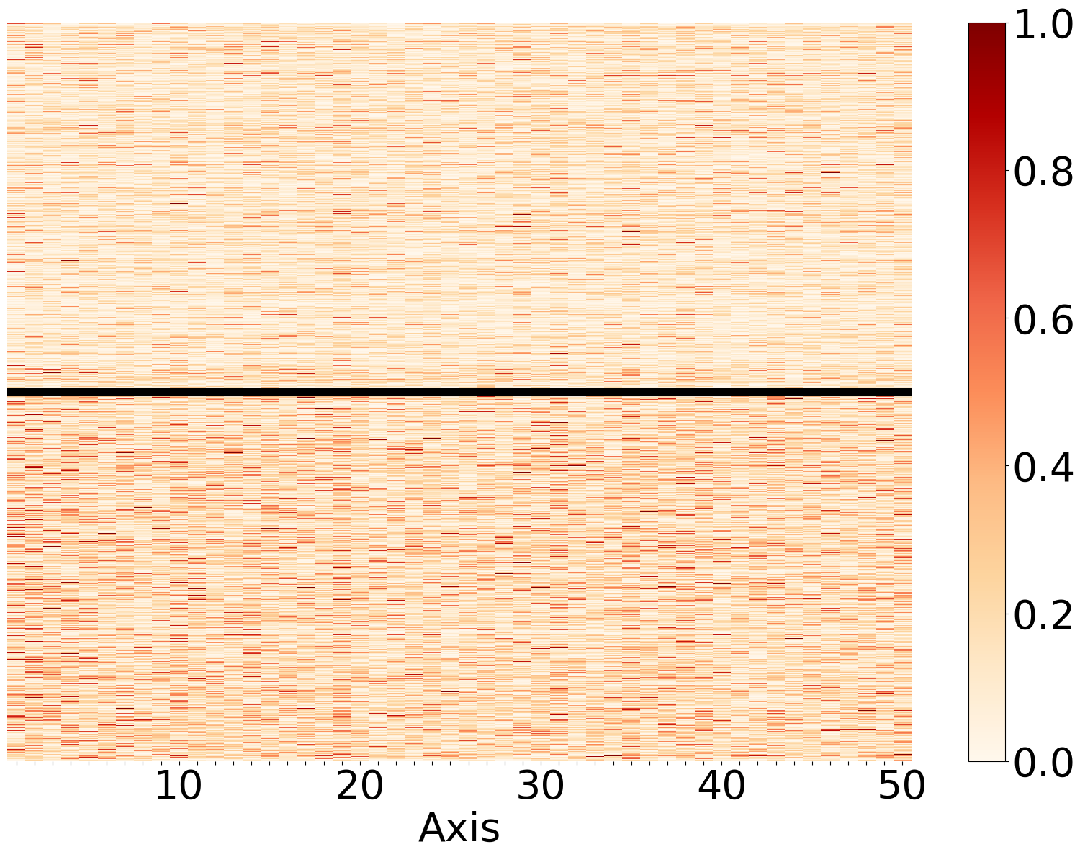}
        \subcaption{Fine-tuned \ac{SCWE}, PCA}
        \label{fig:wic_instance_finetuned_pca}
    \end{minipage}
    \begin{minipage}[b]{0.65\columnwidth}
        \centering
        \includegraphics[width=0.85\columnwidth]{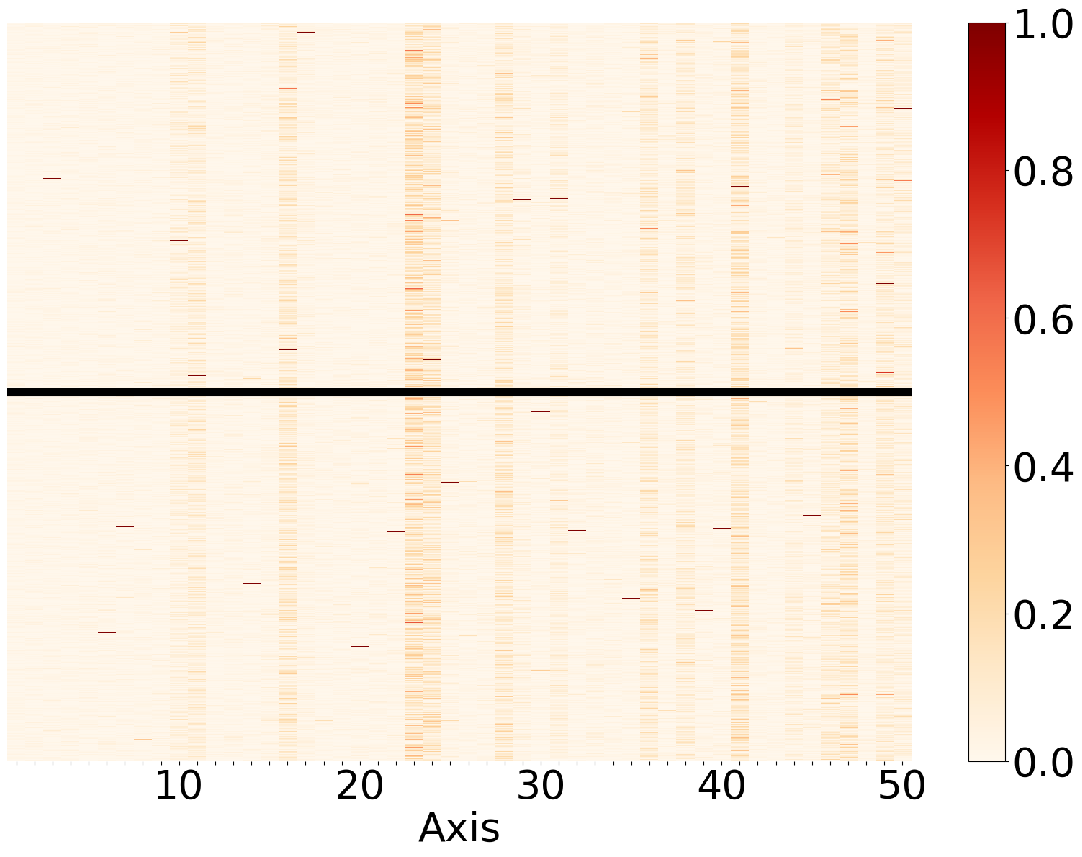}
        \subcaption{Fine-tuned \ac{SCWE}, ICA}
        \label{fig:wic_instance_finetuned_ica}
    \end{minipage}
    \caption{Visualisation of the top-50 dimensions of pre-trained \acp{CWE} (XLM-RoBERTa) and \acp{SCWE} (XL-LEXEME) for each instance in the English \ac{WiC} dataset, where the difference of vectors is calculated for (a/d) \textbf{Raw} vectors, (b/e) \ac{PCA}-transformed axes, and (c/f) \ac{ICA}-transformed axes. In each figure, the upper/lower half uses instances for the True/False labels. While the \textbf{Raw} dimensions display the information from the 0th to the 49th dimensions in the original order, the same observations are found in all dimensions.}
    \label{fig:wic_instance}
    \vspace{-5mm}
\end{figure*}

\section{Contextual Semantic Changes}
\label{sec:contextual}
We first investigate the existence of axes specific to contextual semantic changes. 
Recall that XL-LEXEME is fine-tuned from XLM-RoBERTa on \ac{WiC} datasets.
Therefore, the emergence of any semantic change-aware axes due to fine-tuning can be investigated using contextual semantic change benchmarks. 
We use the test split of the English \ac{WiC}~\cite{pilehvar-camacho-collados-2019-wic}, XL-WiC~\citep{raganato-etal-2020-xlwic}, MCL-WiC~\citep{martelli-etal-2021-mclwic}, and AM$^2$iCo~\citep{liu-etal-2021-am2ico} datasets for evaluations.\footnote{Due to the page limitations, results for other datasets than the English WiC are shown in \autoref{app_sec:full_results}.}
Data statistics are in \autoref{app_sec:data}.

\paragraph{RQ1: When do the contextual \ac{SCD}-aware axes emerge?}
To investigate whether contextual semantic change-aware axes were already present in the pre-trained \acp{CWE}, or do they emerge during the fine-tuning step, for each sentence-pair in \ac{WiC} datasets, we compute the difference between the two target word embeddings obtained from the pre-trained XLM-RoBERTa (\acp{CWE}) and the fine-tuned XL-LEXEME (\acp{SCWE}).
To obtain the sets of target word embeddings, we follow \citet{cassotti-etal-2023-xl} by using a Sentence-BERT~\cite{reimers-gurevych-2019-sentence} architecture.
We conduct this analysis for the non-transformed original axes (indicated as \textbf{Raw} here onwards), as well as for the \ac{PCA}/\ac{ICA}-transformed axes in order to investigate whether such transformations can discover the axes specified for contextual semantic changes as proposed by \citet{yamagiwa-etal-2023-discovering}.\footnote{As in \citet{yamagiwa-etal-2023-discovering}, we used PCA and FastICA provided in scikit-learn \url{https://scikit-learn.org/} .}
In this paper, \ac{PCA}/\ac{ICA}-transformed axes are sorted by the experimental variance ratio/skewness, and this process is consistently applied where \ac{PCA} or \ac{ICA} is used.
If a particular axis is sensitive to contextual semantic changes, it will take similar values in the two target word embeddings, thus having a near-zero value in their subtraction.

To address RQ1, we visualised the difference vectors for sentence pairs where the target word takes the \emph{same} meaning in the two sentences (True) vs. \emph{different} meanings (False).
This visualisation was performed by following steps: (a) we prepared Raw or PCA/ICA-transformed axes; (b) for each WiC instance, which contains two sentences and a label, we calculated the difference between pair of sentences; (c) we normalised each axis (min=0 and max=1) for visualisation purposes. 

As shown in \autoref{fig:wic_instance}, we see that \textbf{the axes encoding contextual semantic changes are not obvious in the original \acp{CWE} after pre-training (\autoref{fig:wic_instance_pretrained_raw}), but materialise during the fine-tuning process (\autoref{fig:wic_instance_finetuned_raw}).}
Similar trends are observed with \ac{PCA}-transformations 
(\Autoref{fig:wic_instance_finetuned_pca, fig:wic_instance_pretrained_pca}), whereas \ac{ICA} shows contrasting results (\Autoref{fig:wic_instance_finetuned_ica,fig:wic_instance_pretrained_ica}). 
In contrast to prior recommendations for using \ac{ICA} for analysing \ac{CWE} spaces~\cite{yamagiwa-etal-2023-discovering}, we find \ac{ICA} to be less sensitive to contextual semantic changes of words.
Interestingly, similar results have been shown in other languages/datasets(\autoref{app_sec:full_results}). \footnote{Our findings do not aim to claim the superiority of PCA over ICA but to explore the existence of task-specific axes. Experimental results show that for semantic change tasks, PCA provides more task-related axes because (a) PCA orders axes by importance (eigenvalue), making task-related axes more accessible, and (b) ICA-transformed axes require external sorting method based on skewness rather than importance. Prior research indicates that ICA can capture topic-related axes~\cite{yamagiwa-etal-2023-discovering}, suggesting that ICA may still hold potential for obtaining task-related axes. Further refinement of the approach remains as future research.}

\begin{figure}[t]
    \centering
    \begin{minipage}[b]{\columnwidth}
        \centering
        \includegraphics[width=0.9\columnwidth]{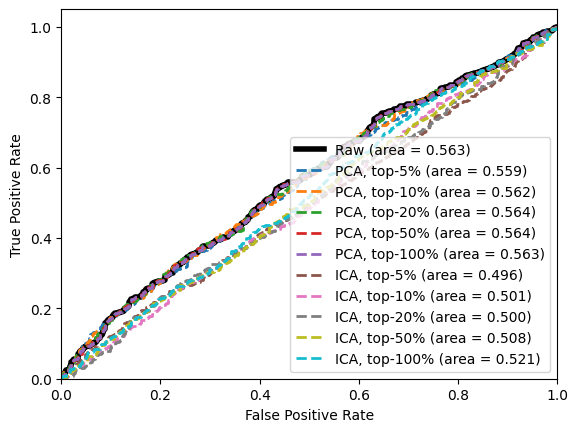}
        \subcaption{Pre-trained \ac{CWE} (XLM-RoBERTa)}
        \label{fig:wic_roc_pretrained}
    \end{minipage}
    \begin{minipage}[b]{\columnwidth}
        \centering
        \includegraphics[width=0.9\columnwidth]{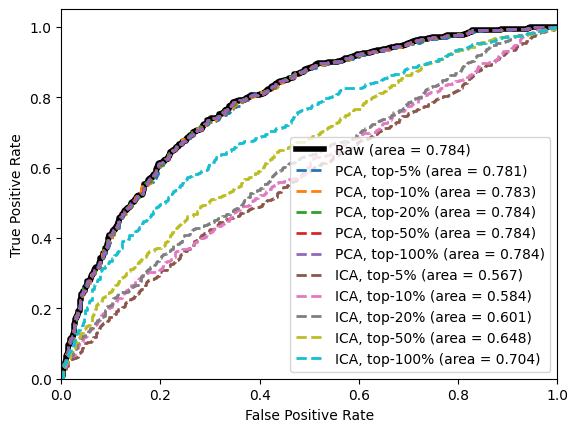}
        \subcaption{Fine-tuned \ac{SCWE} (XL-LEXEME)}
        \label{fig:wic_roc_finetuned}
    \end{minipage}
    \caption{The ROC curve on contextual semantic change task, the English WiC dataset.
    \textbf{Raw} indicates the performance of using full dimensions.
    \ac{PCA}/\ac{ICA} uses top-5/10/20/50/100\% of axes.}
    \label{fig:wic_roc}
    \vspace{-5mm}
\end{figure}

\paragraph{RQ2: Can top-$k$ \ac{PCA}/\ac{ICA}-transformed axes capture contextual semantic changes?}
\citet{yamagiwa-etal-2023-discovering} discovered that \ac{ICA}-transformed axes represent specific concepts and their linear combinations could represent more complex concepts (e.g. \textit{cars} $+$ \textit{italian} $=$ \textit{ferrari}).
Based on this finding, we investigate whether a combination of top-$k$ axes can collectively represent contextual semantic changes of words.
Specifically, we select the top-$k\%$ of the axes to represent a target word embedding.
We then compute the Euclidean distance between \acp{CWE} of the target word in each sentence for every test sentence-pair in the \ac{WiC} datasets.
We predict the target word to have the same meaning in the two sentences, if the Euclidean distance is below a threshold value.
We vary this threshold and report \ac{AUC} of \ac{ROC} curves, where higher \ac{AUC} values are desirable.
In \autoref{fig:wic_roc}, we show results for top $k \in \{5, 10, 20, 50, 100\}$ of the \ac{PCA}/\ac{ICA}-transformed axes and compare against the baseline that uses \emph{all} of the \textbf{Raw} dimensions.


For the pre-trained \acp{CWE} (\autoref{fig:wic_roc_pretrained}), we see that \textbf{Raw} reports slightly better \ac{AUC} than \ac{PCA}, but when fine-tuned (\autoref{fig:wic_roc_finetuned}) \ac{PCA} matches \textbf{Raw} even by using less than 10\% of the axes.
On the other hand, \ac{ICA} reports lower \ac{AUC} values than both \textbf{Raw} and \ac{PCA} in both models.
These results indicate that \ac{PCA} is better suited for discovering axes specified for contextual semantic changes than \ac{ICA}.
We suspect that although \ac{ICA} is able to retrieve concepts such as topics~\cite{yamagiwa-etal-2023-discovering}, it is less fluent when discovering task-specific axes that require the consideration of different types of information.
In conclusion, (1) contextual semantic change-aware axes emerge during fine-tuning, and (2) they are discovered by \ac{PCA} even within 10\% of the principal components.
Notably, in other languages/datasets, similar trends have been observed (\autoref{app_sec:full_results}).
These results suggest that \textbf{contextual semantic change-aware dimensions can be observed within 10\% of the \ac{PCA}-transformed axes} across different languages.

\section{Temporal Semantic Changes}
\label{sec:temporal}

In contrast to contextual \ac{SCD}, temporal \ac{SCD} considers the problem of predicting whether a target word $w$ represents different meanings in two text corpora $C_1$ and $C_2$, sampled at different points in time.
For evaluations, we use the SemEval-2020 Task 1 dataset\footnote{Data statistics are in \autoref{app_sec:data}.}~\cite{schlechtweg-etal-2020-semeval}, which contains a manually rated set of target words for their temporal semantic changes in English, German, Swedish, and Latin.\footnote{Due space limitations, results for languages other than English are shown in \autoref{app_sec:full_results}.}

\begin{figure}[t]
    \centering
    \begin{minipage}[b]{\columnwidth}
        \centering
        \includegraphics[width=0.9\columnwidth]{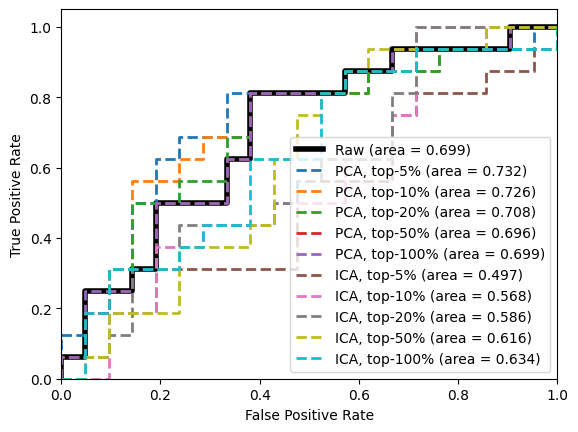}
        \subcaption{Pre-trained \ac{CWE} (XLM-RoBERTa)}
        \label{fig:scd_roc_pretrained}
    \end{minipage}
    \begin{minipage}[b]{\columnwidth}
        \centering
        \includegraphics[width=0.9\columnwidth]{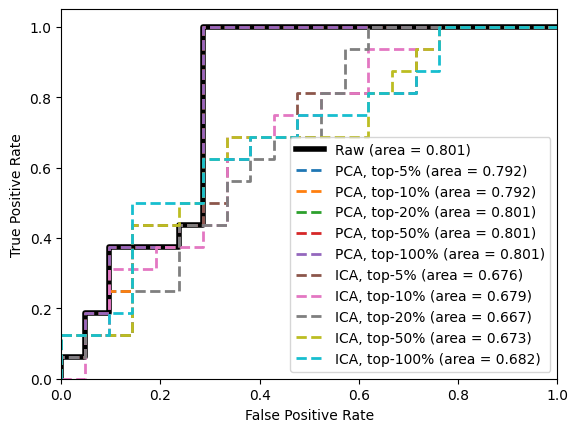}
        \subcaption{Fine-tuned \ac{SCWE} (XL-LEXEME)}
        \label{fig:scd_roc_finetuned}
    \end{minipage}
    \caption{The ROC curve on temporal semantic change task, SemEval-2020 Task 1 (English).
    \textbf{Raw} indicates the performance of using full dimensions.
    \ac{PCA}/\ac{ICA} uses top-5/10/20/50/100\% of axes.}
    \label{fig:scd_roc}
    \vspace{-5mm}
\end{figure}

\paragraph{RQ3: Can top-$k$ \ac{PCA}/\ac{ICA}-transformed axes capture temporal semantic changes?}
Similar to \autoref{fig:wic_roc}, we investigate whether \ac{PCA}/\ac{ICA} can discover axes specified for temporal semantic changes by considering the top-$k$\% of axes for $k \in \{5, 10, 20, 50, 100\}$.
We calculate the semantic change score of $w$ as the average pairwise Euclidean distance over the two sets of sentences containing the target word $w$ in $C_1$ and $C_2$ as conducted in previous work~\cite{kutuzov-giulianelli-2020-uio, laicher-etal-2021-explaining, cassotti-etal-2023-xl}.
Finally, $w$ is predicted to have its meaning changed between $C_1$ and $C_2$, if its semantic change score exceeds a pre-defined threshold. 
We vary this threshold and plot \ac{ROC} in \autoref{fig:scd_roc}.

In pre-trained \acp{CWE}, we can see that the use of the top 5\% to 20\% axes transformed by \ac{PCA} is more effective in temporal semantic change detection than when all of the \textbf{Raw} dimensions are used (\autoref{fig:scd_roc_pretrained}).
On the other hand, in fine-tuned \acp{SCWE}, \autoref{fig:scd_roc_finetuned} indicates that \ac{PCA}-transformed axes achieve the same \ac{AUC} scores as \textbf{Raw}, similar to the contextual semantic change (\autoref{fig:wic_roc_finetuned}).
Similar to the observation in contextual semantic change, \ac{ICA} returns the lowest performance.

\begin{figure}[t]
    \centering
    \begin{minipage}[b]{\columnwidth}
        \centering
        \includegraphics[width=0.9\columnwidth]{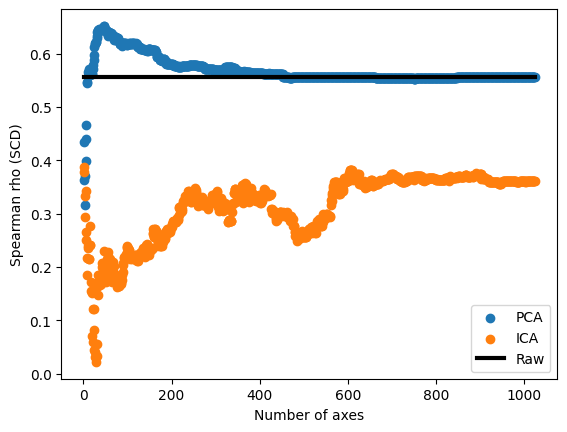}
        \subcaption{Pre-trained \ac{CWE} (XLM-RoBERTa)}
        \label{fig:scd_spearman_pretrained}
    \end{minipage} \\
    \begin{minipage}[b]{\columnwidth}
        \centering
        \includegraphics[width=0.9\columnwidth]{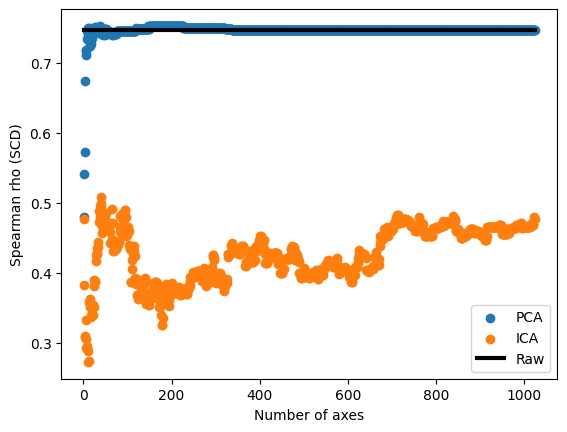}
        \subcaption{Fine-tuned \ac{SCWE} (XL-LEXEME)}
        \label{fig:scd_spearman_finetuned}
    \end{minipage}
    \caption{Spearman's rank correlation on temporal semantic change task, SemEval-2020 Task 1 (English).
    \textbf{Raw} indicates the performance of using full dimensions. 
    \ac{PCA}/\ac{ICA} cumulatively uses sorted axes.}
    \label{fig:scd_spearman}
    \vspace{-5mm}
\end{figure}


To further investigate whether the top \ac{PCA}/\ac{ICA} axes can explain the \emph{degree} of temporal semantic change, we measure the Spearman correlation between the semantic change scores and human ratings available in the SemEval-2020 Task 1 following the standard evaluation protocol for this task~\cite{rosin-etal-2022-time, rosin-radinsky-2022-temporal, aida-bollegala-2023-unsupervised, cassotti-etal-2023-xl, periti-tahmasebi-2024-systematic, aida-bollegala-2024-semantic}.
As shown in \autoref{fig:scd_spearman} for the pre-trained \acp{CWE} (\autoref{fig:scd_spearman_pretrained}), using only 10\% of the axes, \ac{PCA} outperforms \textbf{Raw} that uses all axes.
Moreover, for the fine-tuned \acp{SCWE} (\autoref{fig:scd_spearman_finetuned}), using only 10\% of the axes \ac{PCA} achieves the same performance as \textbf{Raw}.
However, \ac{ICA} consistently underperforms in both pre-trained and fine-tuned settings.
Importantly, we see similar trends in other languages (\autoref{app_sec:full_results}).
These results suggest that \textbf{temporal semantic change-aware dimensions can also be observed within 10\% of \ac{PCA}-transformed axes} across different languages.

\section{Conclusion}
%


We found that there exists a smaller number of axes that encode contextual and temporal semantic changes of words in \acp{MLM}, which are accurately discovered by \ac{PCA}.
These findings have several important practical implications.
First, it shows that \acp{MLM} can be compressed into efficient and accurate lower-dimensional embeddings when used for \ac{SCD} tasks.
Second, it suggests the possibility of efficiently updating a pre-trained \ac{MLM} to capture novel semantic associations of words since the \ac{MLM} was first trained, by updating only a smaller number of dimensions. 


\section*{Limitations}
In this paper, we limited experiments to XLM-RoBERTa based \ac{MLM} models.
These models are all fine-tuned on \ac{WiC} datasets and have reported \ac{SoTA} performance in \ac{SCD} benchmarks.
We consider it would be important to further validate the findings reported in this paper 
using other embedding models and across multiple downstream applications.

\section*{Ethical Considerations}
In this paper, we focus on investigating the existence of dedicated dimensions capturing contextual/temporal semantic changes of words.
For the best of our knowledge, no ethical issues have been reported for the \ac{WiC} and \ac{SCD} datasets we used in our experiments.
On the other hand, we also used publicly available pre-trained/fine-tuned \acp{MLM}, some of which are known to encode and potentially amplify unfair social biases~\citep{basta-etal-2019-evaluating}.
Whether such social biases are influenced by the dimension selection methods we consider in the paper must be carefully evaluated before deploying any \acp{MLM} in downstream applications.

\section*{Acknowledgements}
Taichi Aida would like to acknowledge the support by JST, the establishment of university fellowships towards the creation of science technology innovation, Grant Number JPMJFS2139.

\bibliography{anthology,custom}

\begin{thebibliography}{31}
\providecommand{\natexlab}[1]{#1}

\bibitem[{Aida and Bollegala(2023{\natexlab{a}})}]{aida-bollegala-2023-swap}
Taichi Aida and Danushka Bollegala. 2023{\natexlab{a}}.
\newblock \href {https://doi.org/10.18653/v1/2023.findings-emnlp.520} {Swap and
  predict {--} predicting the semantic changes in words across corpora by
  context swapping}.
\newblock In \emph{Findings of the Association for Computational Linguistics:
  EMNLP 2023}, pages 7753--7772, Singapore. Association for Computational
  Linguistics.

\bibitem[{Aida and
  Bollegala(2023{\natexlab{b}})}]{aida-bollegala-2023-unsupervised}
Taichi Aida and Danushka Bollegala. 2023{\natexlab{b}}.
\newblock \href {https://doi.org/10.18653/v1/2023.findings-acl.429}
  {Unsupervised semantic variation prediction using the distribution of sibling
  embeddings}.
\newblock In \emph{Findings of the Association for Computational Linguistics:
  ACL 2023}, pages 6868--6882, Toronto, Canada. Association for Computational
  Linguistics.

\bibitem[{Aida and Bollegala(2024)}]{aida-bollegala-2024-semantic}
Taichi Aida and Danushka Bollegala. 2024.
\newblock \href {https://doi.org/10.18653/v1/2024.findings-acl.451} {A semantic
  distance metric learning approach for lexical semantic change detection}.
\newblock In \emph{Findings of the Association for Computational Linguistics:
  ACL 2024}, pages 7570--7584, Bangkok, Thailand. Association for Computational
  Linguistics.

\bibitem[{Aida et~al.(2021)Aida, Komachi, Ogiso, Takamura, and
  Mochihashi}]{aida-etal-2021-comprehensive}
Taichi Aida, Mamoru Komachi, Toshinobu Ogiso, Hiroya Takamura, and Daichi
  Mochihashi. 2021.
\newblock \href {https://aclanthology.org/2021.paclic-1.3} {A comprehensive
  analysis of {PMI}-based models for measuring semantic differences}.
\newblock In \emph{Proceedings of the 35th Pacific Asia Conference on Language,
  Information and Computation}, pages 21--31, Shanghai, China. Association for
  Computational Lingustics.

\bibitem[{Basta et~al.(2019)Basta, Costa-juss{\`a}, and
  Casas}]{basta-etal-2019-evaluating}
Christine Basta, Marta~R. Costa-juss{\`a}, and Noe Casas. 2019.
\newblock \href {https://doi.org/10.18653/v1/W19-3805} {Evaluating the
  underlying gender bias in contextualized word embeddings}.
\newblock In \emph{Proceedings of the First Workshop on Gender Bias in Natural
  Language Processing}, pages 33--39, Florence, Italy. Association for
  Computational Linguistics.

\bibitem[{Cassotti et~al.(2023)Cassotti, Siciliani, DeGemmis, Semeraro, and
  Basile}]{cassotti-etal-2023-xl}
Pierluigi Cassotti, Lucia Siciliani, Marco DeGemmis, Giovanni Semeraro, and
  Pierpaolo Basile. 2023.
\newblock \href {https://doi.org/10.18653/v1/2023.acl-short.135}
  {{XL}-{LEXEME}: {W}i{C} pretrained model for cross-lingual {LEX}ical
  s{EM}antic chang{E}}.
\newblock In \emph{Proceedings of the 61st Annual Meeting of the Association
  for Computational Linguistics (Volume 2: Short Papers)}, pages 1577--1585,
  Toronto, Canada. Association for Computational Linguistics.

\bibitem[{Conneau et~al.(2020)Conneau, Khandelwal, Goyal, Chaudhary, Wenzek,
  Guzm{\'a}n, Grave, Ott, Zettlemoyer, and
  Stoyanov}]{conneau-etal-2020-xlmroberta}
Alexis Conneau, Kartikay Khandelwal, Naman Goyal, Vishrav Chaudhary, Guillaume
  Wenzek, Francisco Guzm{\'a}n, Edouard Grave, Myle Ott, Luke Zettlemoyer, and
  Veselin Stoyanov. 2020.
\newblock \href {https://doi.org/10.18653/v1/2020.acl-main.747} {Unsupervised
  cross-lingual representation learning at scale}.
\newblock In \emph{Proceedings of the 58th Annual Meeting of the Association
  for Computational Linguistics}, pages 8440--8451, Online. Association for
  Computational Linguistics.

\bibitem[{Devlin et~al.(2019)Devlin, Chang, Lee, and Toutanova}]{BERT}
Jacob Devlin, Ming-Wei Chang, Kenton Lee, and Kristina Toutanova. 2019.
\newblock \href {https://doi.org/10.18653/v1/N19-1423} {{BERT}: Pre-training of
  deep bidirectional transformers for language understanding}.
\newblock In \emph{Proceedings of the 2019 Conference of the North {A}merican
  Chapter of the Association for Computational Linguistics: Human Language
  Technologies, Volume 1 (Long and Short Papers)}, pages 4171--4186,
  Minneapolis, Minnesota. Association for Computational Linguistics.

\bibitem[{Fedorova et~al.(2024)Fedorova, Kutuzov, and
  Scherrer}]{fedorova-etal-2024-definition}
Mariia Fedorova, Andrey Kutuzov, and Yves Scherrer. 2024.
\newblock \href {https://doi.org/10.18653/v1/2024.findings-acl.339} {Definition
  generation for lexical semantic change detection}.
\newblock In \emph{Findings of the Association for Computational Linguistics:
  ACL 2024}, pages 5712--5724, Bangkok, Thailand. Association for Computational
  Linguistics.

\bibitem[{Hamilton et~al.(2016)Hamilton, Leskovec, and
  Jurafsky}]{hamilton-etal-2016-diachronic}
William~L. Hamilton, Jure Leskovec, and Dan Jurafsky. 2016.
\newblock \href {https://doi.org/10.18653/v1/P16-1141} {Diachronic word
  embeddings reveal statistical laws of semantic change}.
\newblock In \emph{Proceedings of the 54th Annual Meeting of the Association
  for Computational Linguistics (Volume 1: Long Papers)}, pages 1489--1501,
  Berlin, Germany. Association for Computational Linguistics.

\bibitem[{Kim et~al.(2014)Kim, Chiu, Hanaki, Hegde, and
  Petrov}]{kim-etal-2014-temporal}
Yoon Kim, Yi-I Chiu, Kentaro Hanaki, Darshan Hegde, and Slav Petrov. 2014.
\newblock \href {https://doi.org/10.3115/v1/W14-2517} {Temporal analysis of
  language through neural language models}.
\newblock In \emph{Proceedings of the {ACL} 2014 Workshop on Language
  Technologies and Computational Social Science}, pages 61--65, Baltimore, MD,
  USA. Association for Computational Linguistics.

\bibitem[{Kulkarni et~al.(2015)Kulkarni, Al-Rfou, Perozzi, and
  Skiena}]{kulkarni-etal-2015-statistically}
Vivek Kulkarni, Rami Al-Rfou, Bryan Perozzi, and Steven Skiena. 2015.
\newblock Statistically significant detection of linguistic change.
\newblock In \emph{WWW 2015}, pages 625--635.

\bibitem[{Kutuzov and Giulianelli(2020)}]{kutuzov-giulianelli-2020-uio}
Andrey Kutuzov and Mario Giulianelli. 2020.
\newblock \href {https://doi.org/10.18653/v1/2020.semeval-1.14}
  {{U}i{O}-{U}v{A} at {S}em{E}val-2020 task 1: Contextualised embeddings for
  lexical semantic change detection}.
\newblock In \emph{Proceedings of the Fourteenth Workshop on Semantic
  Evaluation}, pages 126--134, Barcelona (online). International Committee for
  Computational Linguistics.

\bibitem[{Laicher et~al.(2021)Laicher, Kurtyigit, Schlechtweg, Kuhn, and
  Schulte~im Walde}]{laicher-etal-2021-explaining}
Severin Laicher, Sinan Kurtyigit, Dominik Schlechtweg, Jonas Kuhn, and Sabine
  Schulte~im Walde. 2021.
\newblock \href {https://doi.org/10.18653/v1/2021.eacl-srw.25} {Explaining and
  improving {BERT} performance on lexical semantic change detection}.
\newblock In \emph{Proceedings of the 16th Conference of the European Chapter
  of the Association for Computational Linguistics: Student Research Workshop},
  pages 192--202, Online. Association for Computational Linguistics.

\bibitem[{Liu et~al.(2021)Liu, Ponti, McCarthy, Vuli{\'c}, and
  Korhonen}]{liu-etal-2021-am2ico}
Qianchu Liu, Edoardo~Maria Ponti, Diana McCarthy, Ivan Vuli{\'c}, and Anna
  Korhonen. 2021.
\newblock \href {https://doi.org/10.18653/v1/2021.emnlp-main.571} {{AM}2i{C}o:
  Evaluating word meaning in context across low-resource languages with
  adversarial examples}.
\newblock In \emph{Proceedings of the 2021 Conference on Empirical Methods in
  Natural Language Processing}, pages 7151--7162, Online and Punta Cana,
  Dominican Republic. Association for Computational Linguistics.

\bibitem[{Martelli et~al.(2021)Martelli, Kalach, Tola, and
  Navigli}]{martelli-etal-2021-mclwic}
Federico Martelli, Najla Kalach, Gabriele Tola, and Roberto Navigli. 2021.
\newblock \href {https://doi.org/10.18653/v1/2021.semeval-1.3}
  {{S}em{E}val-2021 task 2: Multilingual and cross-lingual word-in-context
  disambiguation ({MCL}-{W}i{C})}.
\newblock In \emph{Proceedings of the 15th International Workshop on Semantic
  Evaluation (SemEval-2021)}, pages 24--36, Online. Association for
  Computational Linguistics.

\bibitem[{Periti et~al.(2024)Periti, Alfter, and
  Tahmasebi}]{periti-etal-2024-automatically}
Francesco Periti, David Alfter, and Nina Tahmasebi. 2024.
\newblock \href {https://doi.org/10.18653/v1/2024.emnlp-main.776}
  {Automatically generated definitions and their utility for modeling word
  meaning}.
\newblock In \emph{Proceedings of the 2024 Conference on Empirical Methods in
  Natural Language Processing}, pages 14008--14026, Miami, Florida, USA.
  Association for Computational Linguistics.

\bibitem[{Periti and Tahmasebi(2024)}]{periti-tahmasebi-2024-systematic}
Francesco Periti and Nina Tahmasebi. 2024.
\newblock \href {https://doi.org/10.18653/v1/2024.naacl-long.240} {A systematic
  comparison of contextualized word embeddings for lexical semantic change}.
\newblock In \emph{Proceedings of the 2024 Conference of the North American
  Chapter of the Association for Computational Linguistics: Human Language
  Technologies (Volume 1: Long Papers)}, pages 4262--4282, Mexico City, Mexico.
  Association for Computational Linguistics.

\bibitem[{Pilehvar and
  Camacho-Collados(2019)}]{pilehvar-camacho-collados-2019-wic}
Mohammad~Taher Pilehvar and Jose Camacho-Collados. 2019.
\newblock \href {https://doi.org/10.18653/v1/N19-1128} {{W}i{C}: the
  word-in-context dataset for evaluating context-sensitive meaning
  representations}.
\newblock In \emph{Proceedings of the 2019 Conference of the North {A}merican
  Chapter of the Association for Computational Linguistics: Human Language
  Technologies, Volume 1 (Long and Short Papers)}, pages 1267--1273,
  Minneapolis, Minnesota. Association for Computational Linguistics.

\bibitem[{Rachinskiy and Arefyev(2021)}]{rachinskiy-arefyev-2021-glossreader}
Maxim Rachinskiy and Nikolay Arefyev. 2021.
\newblock \href {https://doi.org/10.18653/v1/2021.semeval-1.100}
  {{G}loss{R}eader at {S}em{E}val-2021 task 2: Reading definitions improves
  contextualized word embeddings}.
\newblock In \emph{Proceedings of the 15th International Workshop on Semantic
  Evaluation (SemEval-2021)}, pages 756--762, Online. Association for
  Computational Linguistics.

\bibitem[{Raganato et~al.(2020)Raganato, Pasini, Camacho-Collados, and
  Pilehvar}]{raganato-etal-2020-xlwic}
Alessandro Raganato, Tommaso Pasini, Jose Camacho-Collados, and Mohammad~Taher
  Pilehvar. 2020.
\newblock \href {https://doi.org/10.18653/v1/2020.emnlp-main.584}
  {{XL}-{W}i{C}: A multilingual benchmark for evaluating semantic
  contextualization}.
\newblock In \emph{Proceedings of the 2020 Conference on Empirical Methods in
  Natural Language Processing (EMNLP)}, pages 7193--7206, Online. Association
  for Computational Linguistics.

\bibitem[{Reimers and Gurevych(2019)}]{reimers-gurevych-2019-sentence}
Nils Reimers and Iryna Gurevych. 2019.
\newblock \href {https://doi.org/10.18653/v1/D19-1410} {Sentence-{BERT}:
  Sentence embeddings using {S}iamese {BERT}-networks}.
\newblock In \emph{Proceedings of the 2019 Conference on Empirical Methods in
  Natural Language Processing and the 9th International Joint Conference on
  Natural Language Processing (EMNLP-IJCNLP)}, pages 3982--3992, Hong Kong,
  China. Association for Computational Linguistics.

\bibitem[{Rosenfeld and Erk(2018)}]{rosenfeld-erk-2018-deep}
Alex Rosenfeld and Katrin Erk. 2018.
\newblock \href {https://doi.org/10.18653/v1/N18-1044} {Deep neural models of
  semantic shift}.
\newblock In \emph{Proceedings of the 2018 Conference of the North {A}merican
  Chapter of the Association for Computational Linguistics: Human Language
  Technologies, Volume 1 (Long Papers)}, pages 474--484, New Orleans,
  Louisiana. Association for Computational Linguistics.

\bibitem[{Rosin et~al.(2022)Rosin, Guy, and Radinsky}]{rosin-etal-2022-time}
Guy~D. Rosin, Ido Guy, and Kira Radinsky. 2022.
\newblock \href {https://doi.org/10.1145/3488560.3498529} {Time masking for
  temporal language models}.
\newblock In \emph{Proceedings of the Fifteenth ACM International Conference on
  Web Search and Data Mining}, WSDM '22, pages 833--841, New York, NY, USA.
  Association for Computing Machinery.

\bibitem[{Rosin and Radinsky(2022)}]{rosin-radinsky-2022-temporal}
Guy~D. Rosin and Kira Radinsky. 2022.
\newblock \href {https://doi.org/10.18653/v1/2022.findings-naacl.112} {Temporal
  attention for language models}.
\newblock In \emph{Findings of the Association for Computational Linguistics:
  NAACL 2022}, pages 1498--1508, Seattle, United States. Association for
  Computational Linguistics.

\bibitem[{Schlechtweg et~al.(2020)Schlechtweg, McGillivray, Hengchen,
  Dubossarsky, and Tahmasebi}]{schlechtweg-etal-2020-semeval}
Dominik Schlechtweg, Barbara McGillivray, Simon Hengchen, Haim Dubossarsky, and
  Nina Tahmasebi. 2020.
\newblock \href {https://doi.org/10.18653/v1/2020.semeval-1.1}
  {{S}em{E}val-2020 task 1: Unsupervised lexical semantic change detection}.
\newblock In \emph{Proceedings of the Fourteenth Workshop on Semantic
  Evaluation}, pages 1--23, Barcelona (online). International Committee for
  Computational Linguistics.

\bibitem[{Tahmasebi et~al.(2021)Tahmasebi, Borina, and
  Jatowtb}]{tahmasebia-etal-2021-survey}
Nina Tahmasebi, Lars Borina, and Adam Jatowtb. 2021.
\newblock Survey of computational approaches to lexical semantic change
  detection.
\newblock \emph{Computational approaches to semantic change}, 6:1.

\bibitem[{Tang et~al.(2023)Tang, Zhou, and Bollegala}]{Tang2023-yu}
Xiaohang Tang, Yi~Zhou, and Danushka Bollegala. 2023.
\newblock \href {https://aclanthology.org/2023.acl-long.520.pdf} {Learning
  dynamic contextualised word embeddings via template-based temporal
  adaptation}.
\newblock In \emph{Proceedings of the 61st Annual Meeting of the Association
  for Computational Linguistics (Volume 1: Long Papers)}, pages 9352--9369,
  Stroudsburg, PA, USA. Association for Computational Linguistics.

\bibitem[{Yamagiwa et~al.(2023)Yamagiwa, Oyama, and
  Shimodaira}]{yamagiwa-etal-2023-discovering}
Hiroaki Yamagiwa, Momose Oyama, and Hidetoshi Shimodaira. 2023.
\newblock \href {https://doi.org/10.18653/v1/2023.emnlp-main.283} {Discovering
  universal geometry in embeddings with {ICA}}.
\newblock In \emph{Proceedings of the 2023 Conference on Empirical Methods in
  Natural Language Processing}, pages 4647--4675, Singapore. Association for
  Computational Linguistics.

\bibitem[{Yao et~al.(2018)Yao, Sun, Ding, Rao, and
  Xiong}]{yao-etal-2018-dynamic}
Zijun Yao, Yifan Sun, Weicong Ding, Nikhil Rao, and Hui Xiong. 2018.
\newblock \href {https://doi.org/10.1145/3159652.3159703} {Dynamic word
  embeddings for evolving semantic discovery}.
\newblock In \emph{WSDM 2018}, page 673–681.

\bibitem[{Zhou and Bollegala(2021)}]{yi-zhou-2021-learning}
Yi~Zhou and Danushka Bollegala. 2021.
\newblock \href {https://aclanthology.org/2021.paclic-1.52} {Learning
  sense-specific static embeddings using contextualised word embeddings as a
  proxy}.
\newblock In \emph{Proceedings of the 35th Pacific Asia Conference on Language,
  Information and Computation}, pages 493--502, Shanghai, China. Association
  for Computational Lingustics.

\end{thebibliography}

\appendix

\section{Data Statistics}
\label{app_sec:data}
Full statistics of contextual and temporal \ac{SCD} benchmarks are shown in \autoref{tab:data_wic} and \autoref{tab:data_scd}.\footnote{WiC, XL-WiC, and MCL-WiC are licensed under the Creative Commons Attribution-NonCommercial 4.0 License, while AM$^2$iCo and SemEval-2020 Task 1 are licensed under the Creative Commons Attribution 4.0 International License.}

\begin{table}[t]
    \centering
    \small{
    \begin{tabular}{llrrr} \toprule
        Dataset & Language  & \#Train & \#Dev & \#Test \\ \midrule
        \multicolumn{5}{l}{\textbf{Monolingual}} \\ \midrule
        WiC     & English   & 5.4k    & 6.4k  & 1.4k   \\ \midrule 
        \multirow{3}{*}{XL-WiC}  
                & German    & 48k     & 8.9k  & 1.1k   \\
                & French    & 39k     & 8.6k  & 22k    \\
                & Italian   & 1.1k    & 0.2k  & 0.6k   \\ \midrule
        \multirow{5}{*}{MCL-WiC} 
                & Arabic    & \multicolumn{1}{c}{-} 
                                      & 0.5k  & 0.5k   \\
                & English   & 4.0k    & 0.5k  & 0.5k   \\ 
                & French    & \multicolumn{1}{c}{-}
                                      & 0.5k  & 0.5k   \\
                & Russian   & \multicolumn{1}{c}{-}       
                                      & 0.5k  & 0.5k   \\
                & Chinese   & \multicolumn{1}{c}{-}      
                                      & 0.5k  & 0.5k   \\ \midrule
        \multicolumn{5}{l}{\textbf{Cross-lingual}} \\ \midrule
        \multirow{10}{*}{AM$^2$iCo} 
                & German    & 50k     & 0.5k  & 1.0k \\
                & Russian   & 28k     & 0.5k  & 1.0k   \\
                & Japanese  & 16k     & 0.5k  & 1.0k   \\
                & Chinese   & 13k     & 0.5k  & 1.0k   \\
                & Arabic    & 9.6k    & 0.5k  & 1.0k   \\
                & Korean    & 7.0k    & 0.5k  & 1.0k   \\
                & Finnish   & 6.3k    & 0.5k  & 1.0k   \\
                & Turkish   & 3.9k    & 0.5k  & 1.0k   \\
                & Indonesian& 1.6k    & 0.5k  & 1.0k   \\
                & Basque    & 1.0k    & 0.5k  & 1.0k   \\ \bottomrule
    \end{tabular}
    }
    \caption{Statistics of the contextual \ac{SCD} benchmarks used in the fine-tuning for XL-LEXEME. \#Train, \#Dev, and \#Test show the number of instances. AM$^2$iCo is a cross-lingual contextual \ac{SCD} benchmark, where the second language in each pair is English.}
    \label{tab:data_wic}
\end{table}

\begin{table}[t]
    \centering
    \small{
    \begin{tabular}{lccrrr} \toprule
        Language & Time Period & \#Targets           & \#Tokens \\ \midrule
        \multirow{2}{*}{English} 
                 & 1810--1860  & \multirow{2}{*}{37} & 6.5M \\
                 & 1960--2010  &                     & 6.7M \\
        \multirow{2}{*}{German}
                 & 1800--1899  & \multirow{2}{*}{48} & 70.2M \\
                 & 1946--1990  &                     & 72.3M \\
        \multirow{2}{*}{Swedish} 
                 & 1790--1830  & \multirow{2}{*}{31} & 71.0M \\
                 & 1895--1903  &                     & 110.0M \\
        \multirow{2}{*}{Latin} 
                 & B.C. 200--0 & \multirow{2}{*}{40} & 1.7M \\
                 & 0--2000     &                     & 9.4M \\ \midrule
    \end{tabular}
    }
    \caption{Statistics of the temporal \ac{SCD} benchmark, SemEval-2020 Task 1. \#Targets and \#Tokens show the number of target words and tokens, respectively.}
    \label{tab:data_scd}
\end{table}

\section{Full Results}
\label{app_sec:full_results}
\begin{figure*}[t]
    \centering
    \begin{minipage}[b]{0.65\columnwidth}
        \centering
        \includegraphics[width=0.85\columnwidth]{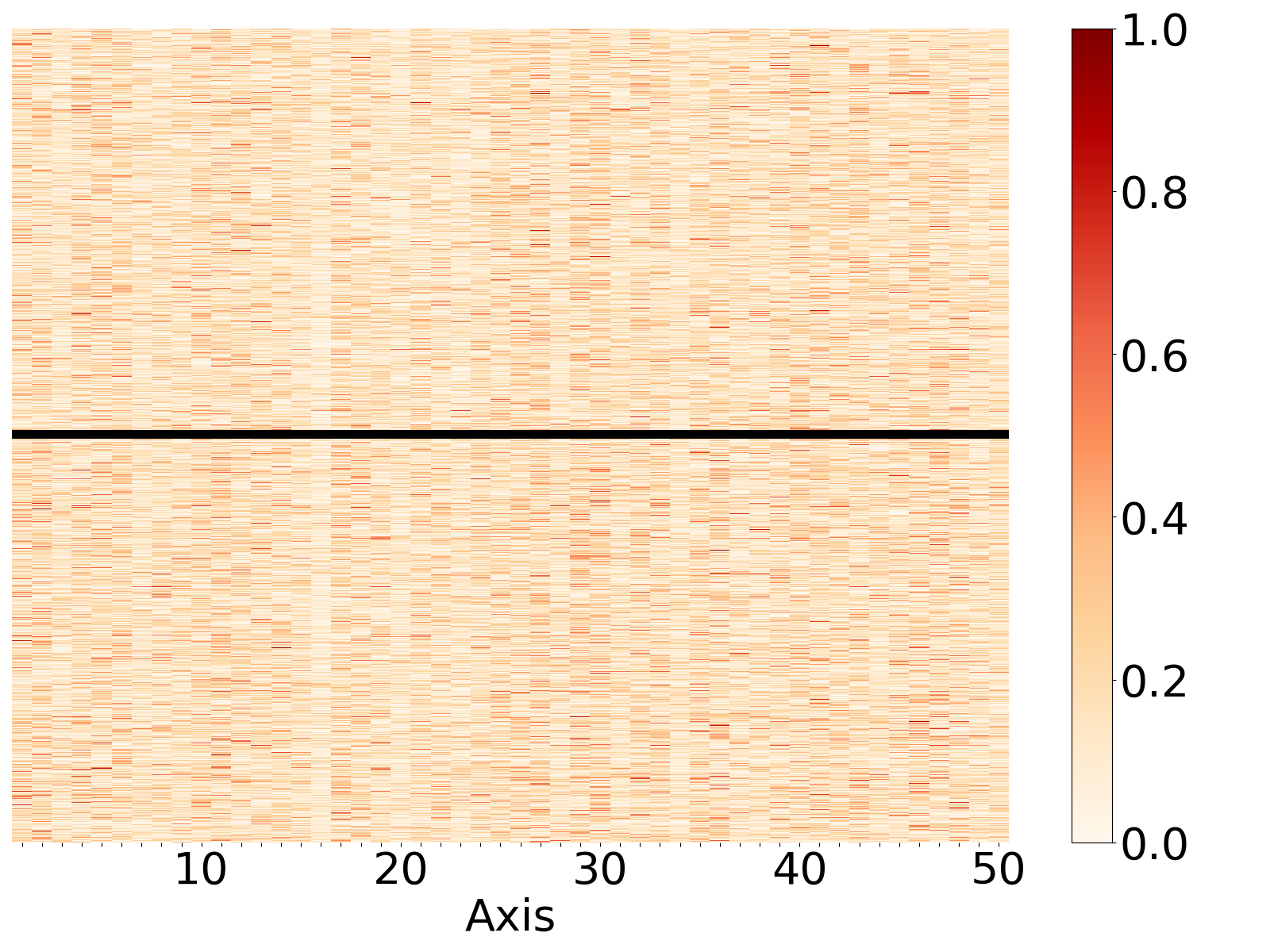}
        \subcaption{Pre-trained \ac{CWE}, Raw}
        \label{fig:wic_xlwic_de_instances_raw_pretrained}
    \end{minipage}
    \begin{minipage}[b]{0.65\columnwidth}
        \centering
        \includegraphics[width=0.85\columnwidth]{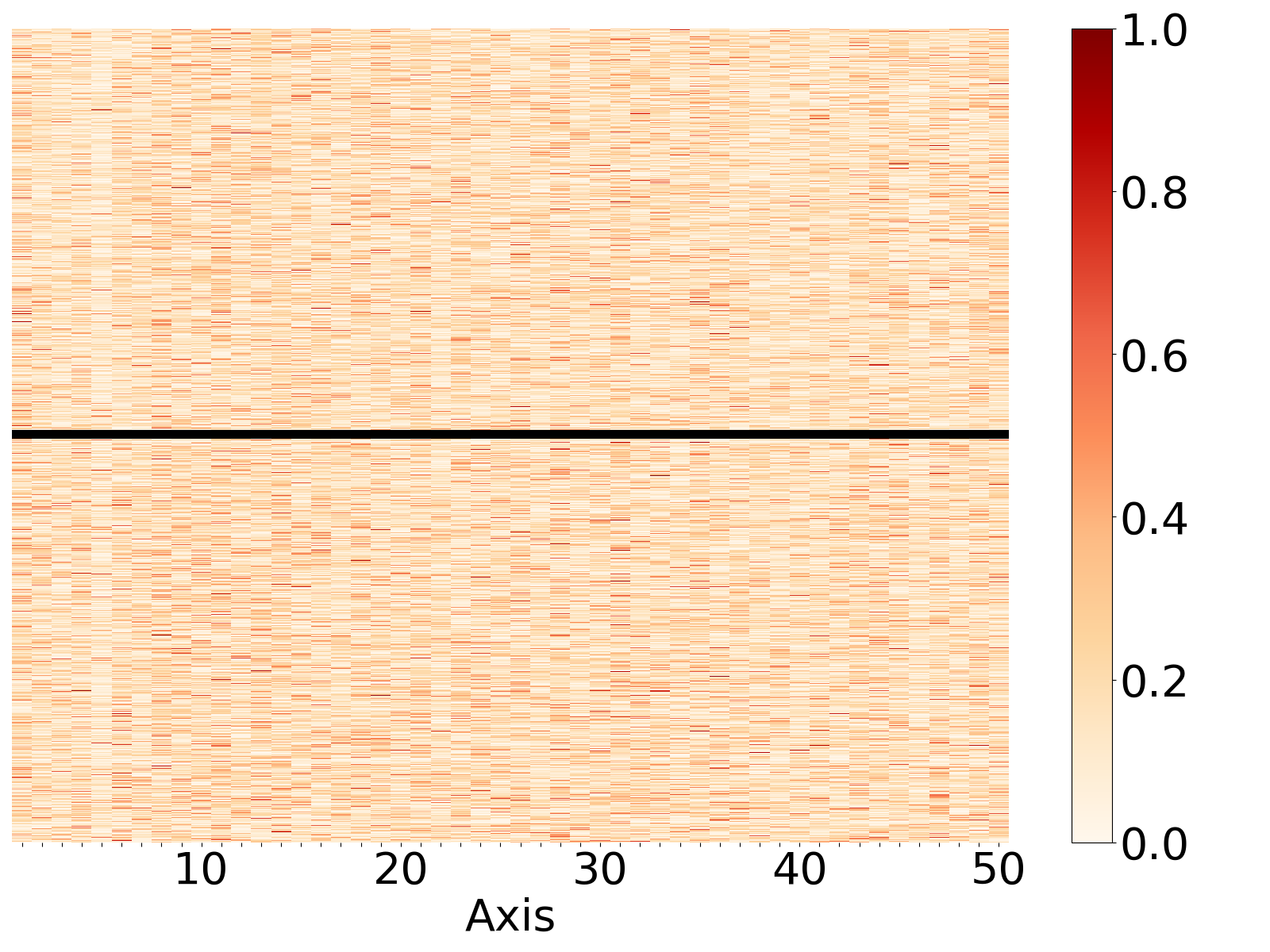}
        \subcaption{Pre-trained \ac{CWE}, PCA}
        \label{fig:wic_xlwic_de_instances_pca_pretrained}
    \end{minipage}
    \begin{minipage}[b]{0.65\columnwidth}
        \centering
        \includegraphics[width=0.85\columnwidth]{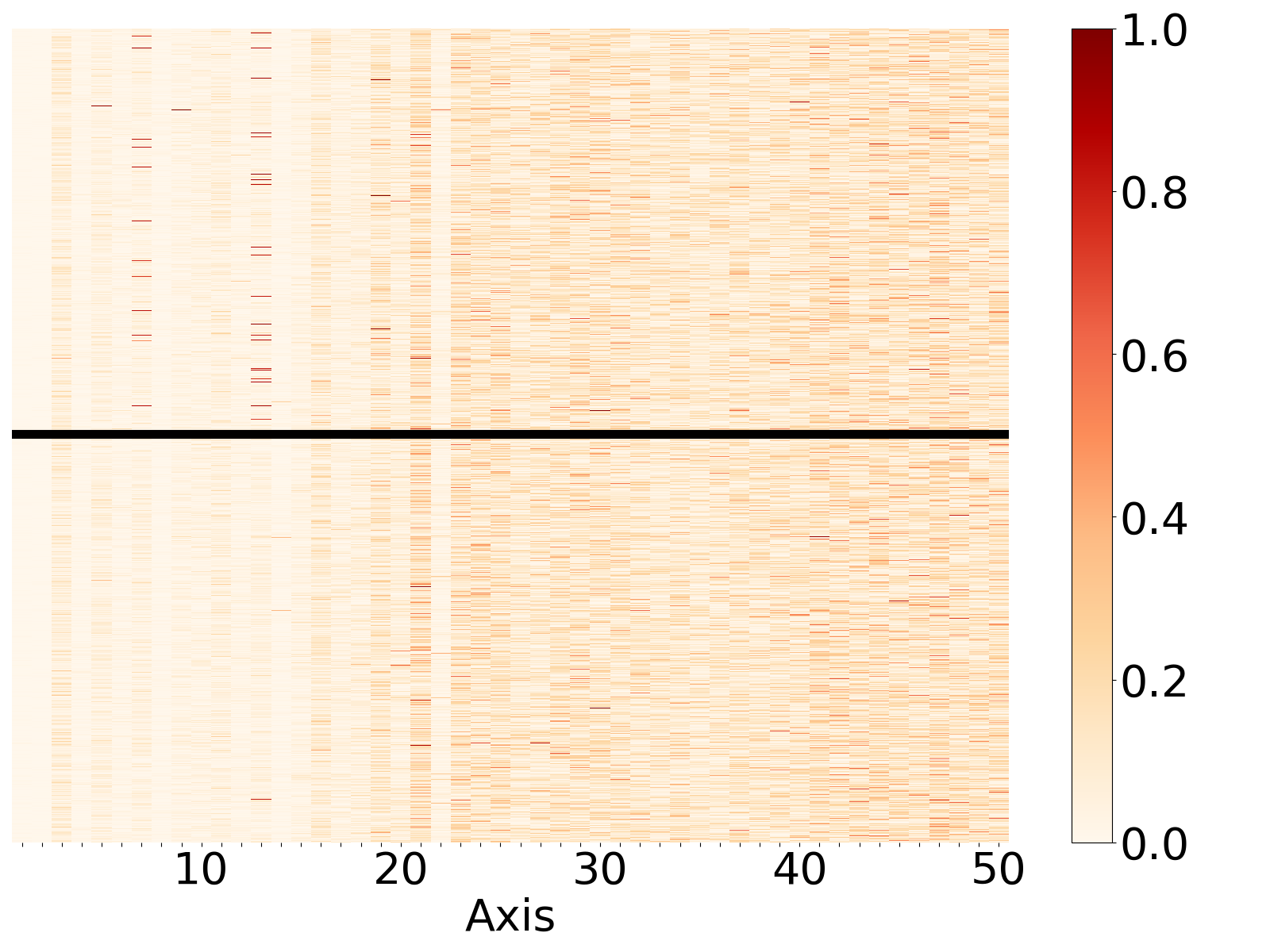}
        \subcaption{Pre-trained \ac{CWE}, ICA}
        \label{fig:wic_xlwic_de_instances_ica_pretrained}
    \end{minipage} \\
    \begin{minipage}[b]{0.65\columnwidth}
        \centering
        \includegraphics[width=0.85\columnwidth]{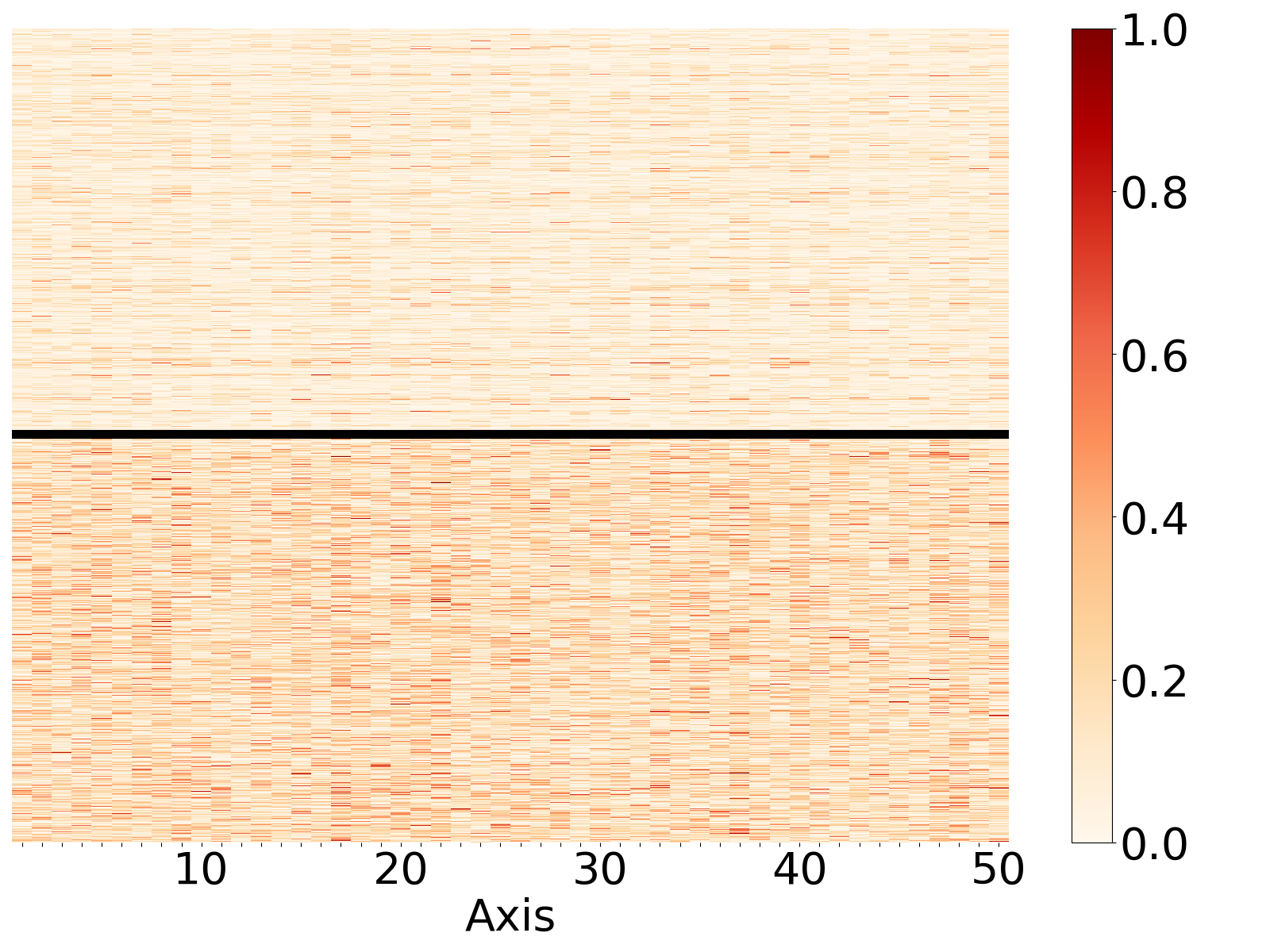}
        \subcaption{Fine-tuned \ac{SCWE}, Raw}
        \label{fig:wic_xlwic_de_instances_raw_finetuned}
    \end{minipage}
    \begin{minipage}[b]{0.65\columnwidth}
        \centering
        \includegraphics[width=0.85\columnwidth]{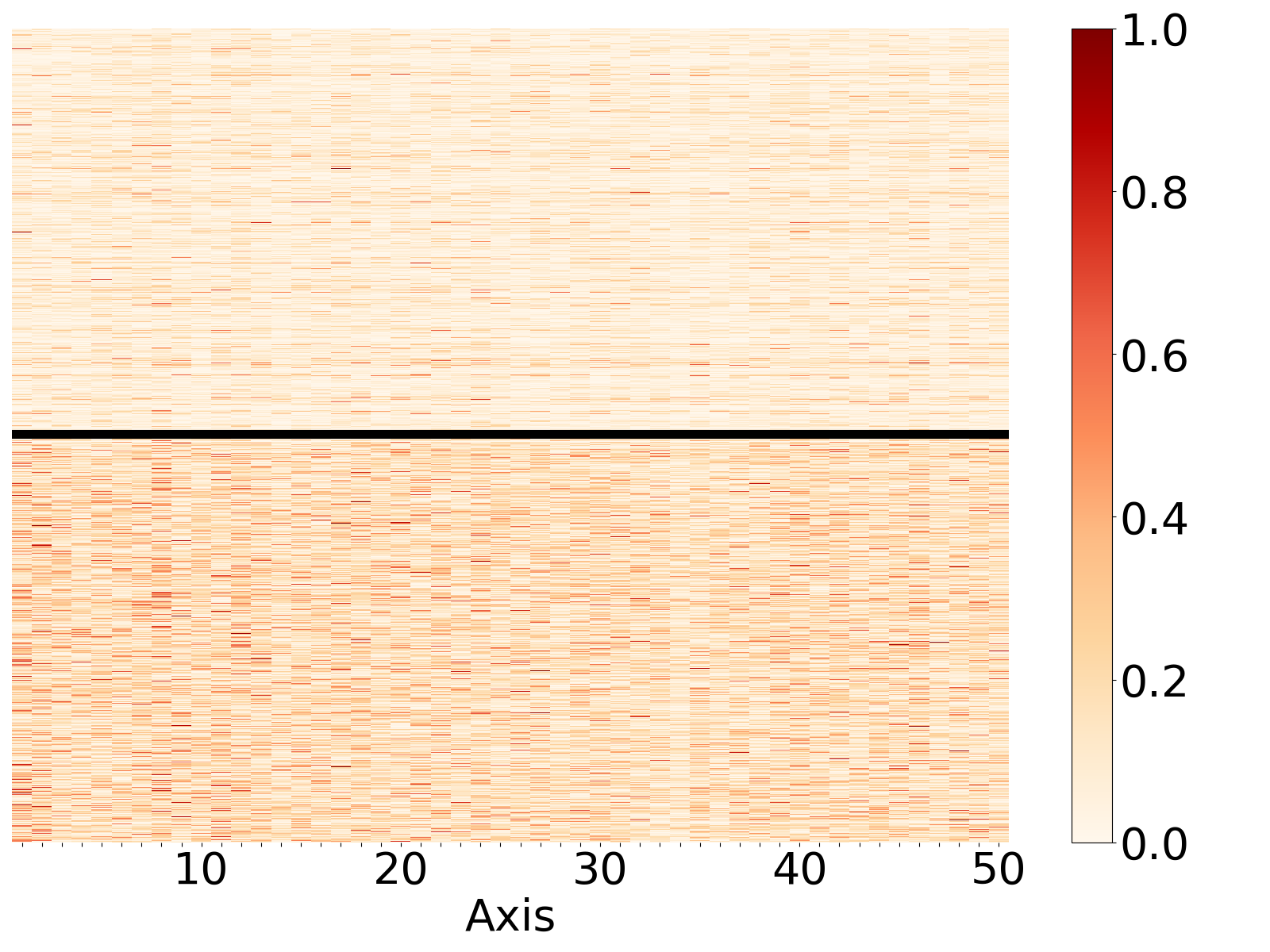}
        \subcaption{Fine-tuned \ac{SCWE}, PCA}
        \label{fig:wic_xlwic_de_instances_pca_finetuned}
    \end{minipage}
    \begin{minipage}[b]{0.65\columnwidth}
        \centering
        \includegraphics[width=0.85\columnwidth]{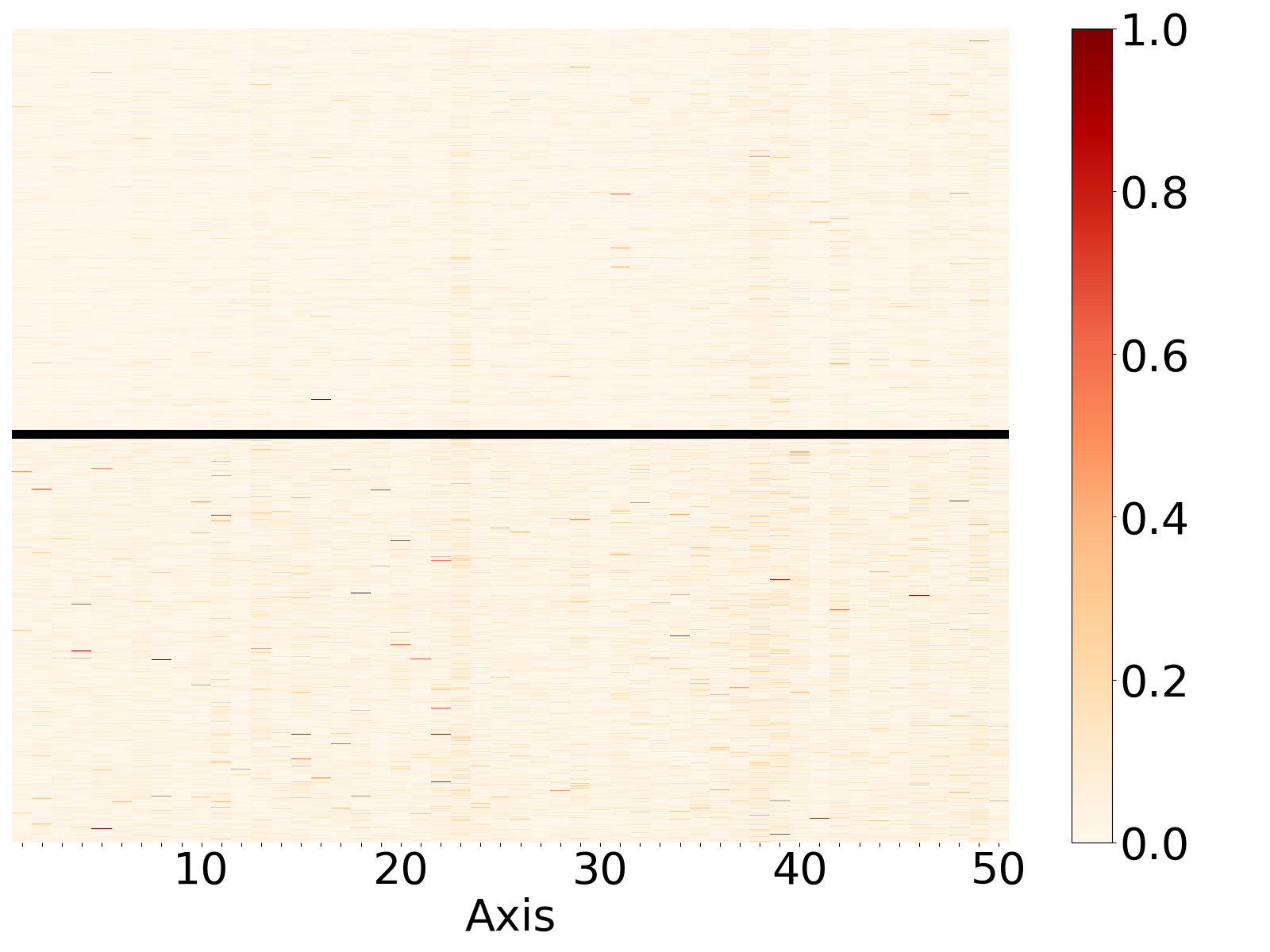}
        \subcaption{Fine-tuned \ac{SCWE}, ICA}
        \label{fig:wic_xlwic_de_instances_ica_finetuned}
    \end{minipage}
    \caption{Visualisation of the top-50 dimensions of pre-trained \acp{CWE} (XLM-RoBERTa) and \acp{SCWE} (XL-LEXEME) for each instance in XLWiC (German) dataset, where the difference of vectors is calculated for (a/d) \textbf{Raw} vectors, (b/e) \ac{PCA}-transformed axes, and (c/f) \ac{ICA}-transformed axes. In each figure, the upper/lower half uses instances for the True/False labels.}
    \label{fig:wic_instance_xlwic_de}
\end{figure*}

\begin{figure*}[t]
    \centering
    \begin{minipage}[b]{0.65\columnwidth}
        \centering
        \includegraphics[width=0.85\columnwidth]{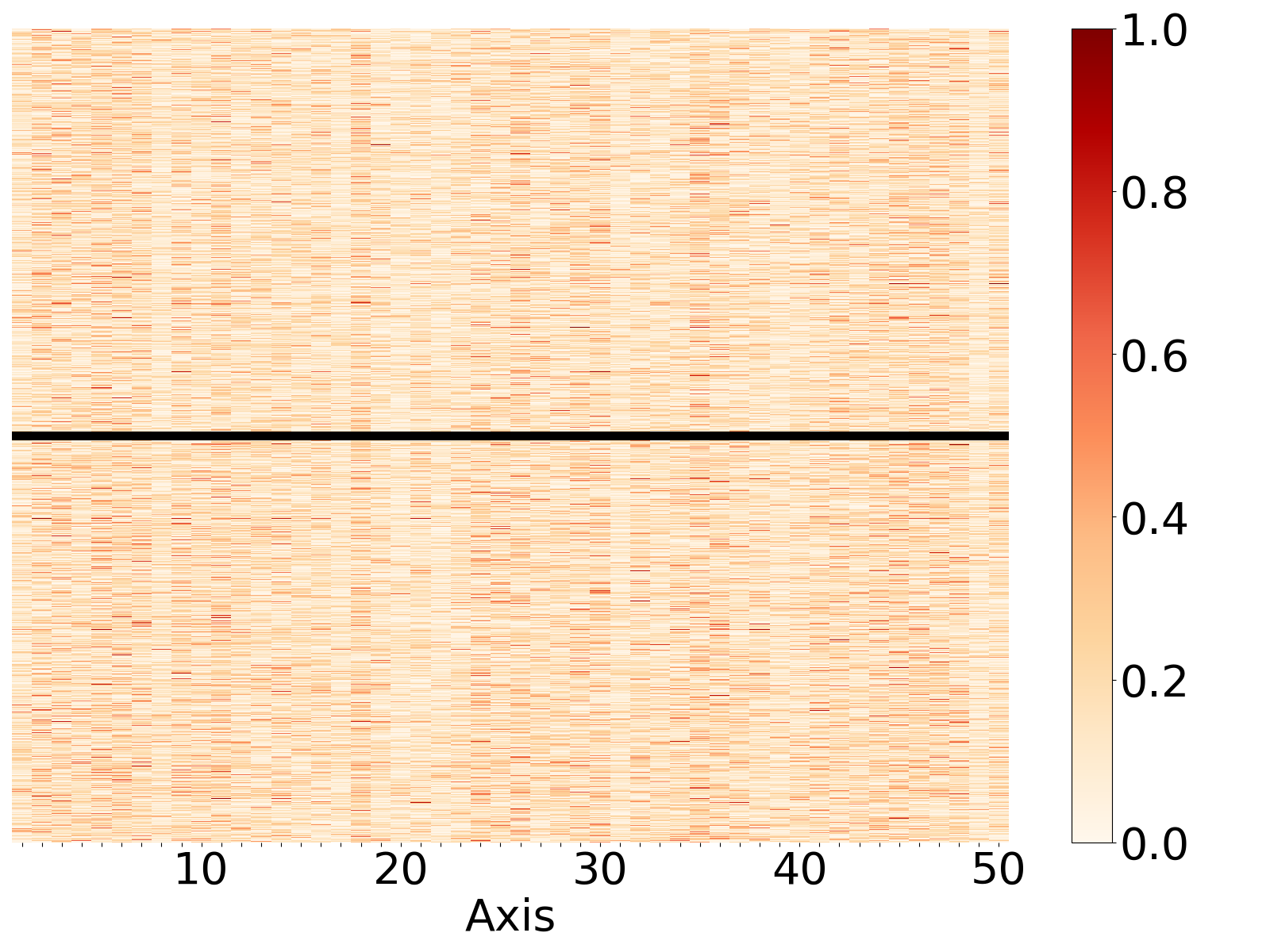}
        \subcaption{Pre-trained \ac{CWE}, Raw}
        \label{fig:wic_xlwic_fr_instances_raw_pretrained}
    \end{minipage}
    \begin{minipage}[b]{0.65\columnwidth}
        \centering
        \includegraphics[width=0.85\columnwidth]{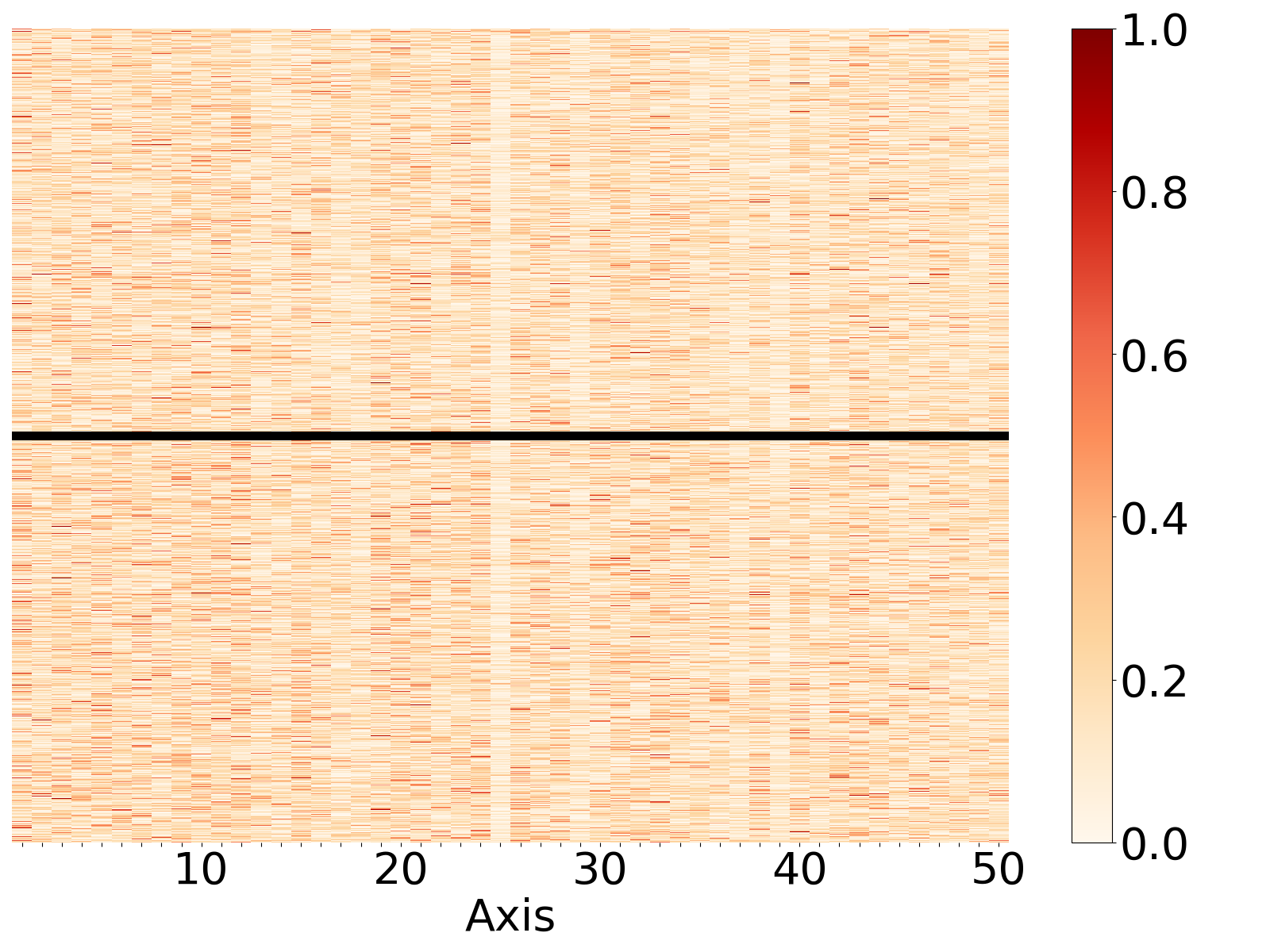}
        \subcaption{Pre-trained \ac{CWE}, PCA}
        \label{fig:wic_xlwic_fr_instances_pca_pretrained}
    \end{minipage}
    \begin{minipage}[b]{0.65\columnwidth}
        \centering
        \includegraphics[width=0.85\columnwidth]{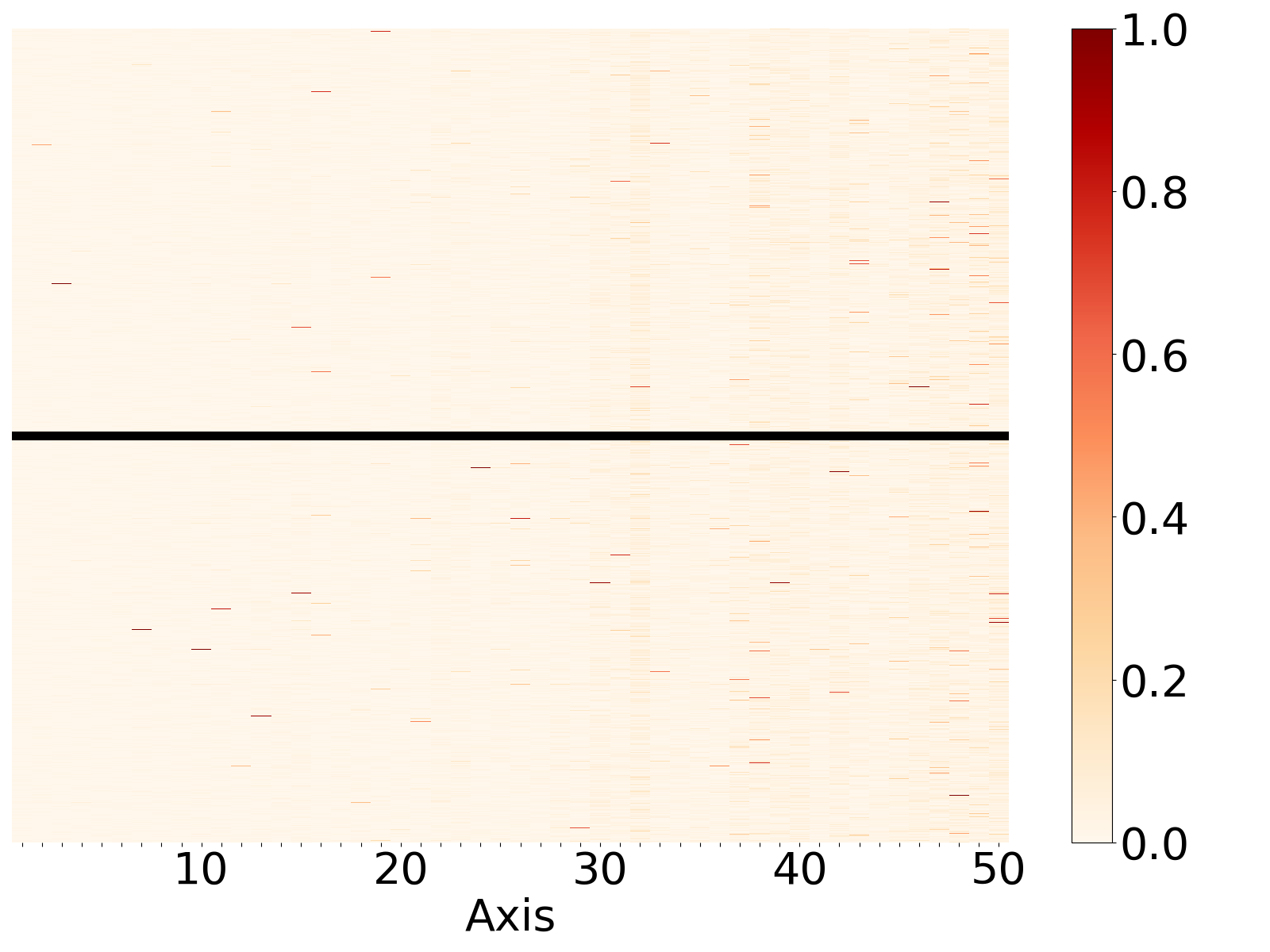}
        \subcaption{Pre-trained \ac{CWE}, ICA}
        \label{fig:wic_xlwic_fr_instances_ica_pretrained}
    \end{minipage} \\
    \begin{minipage}[b]{0.65\columnwidth}
        \centering
        \includegraphics[width=0.85\columnwidth]{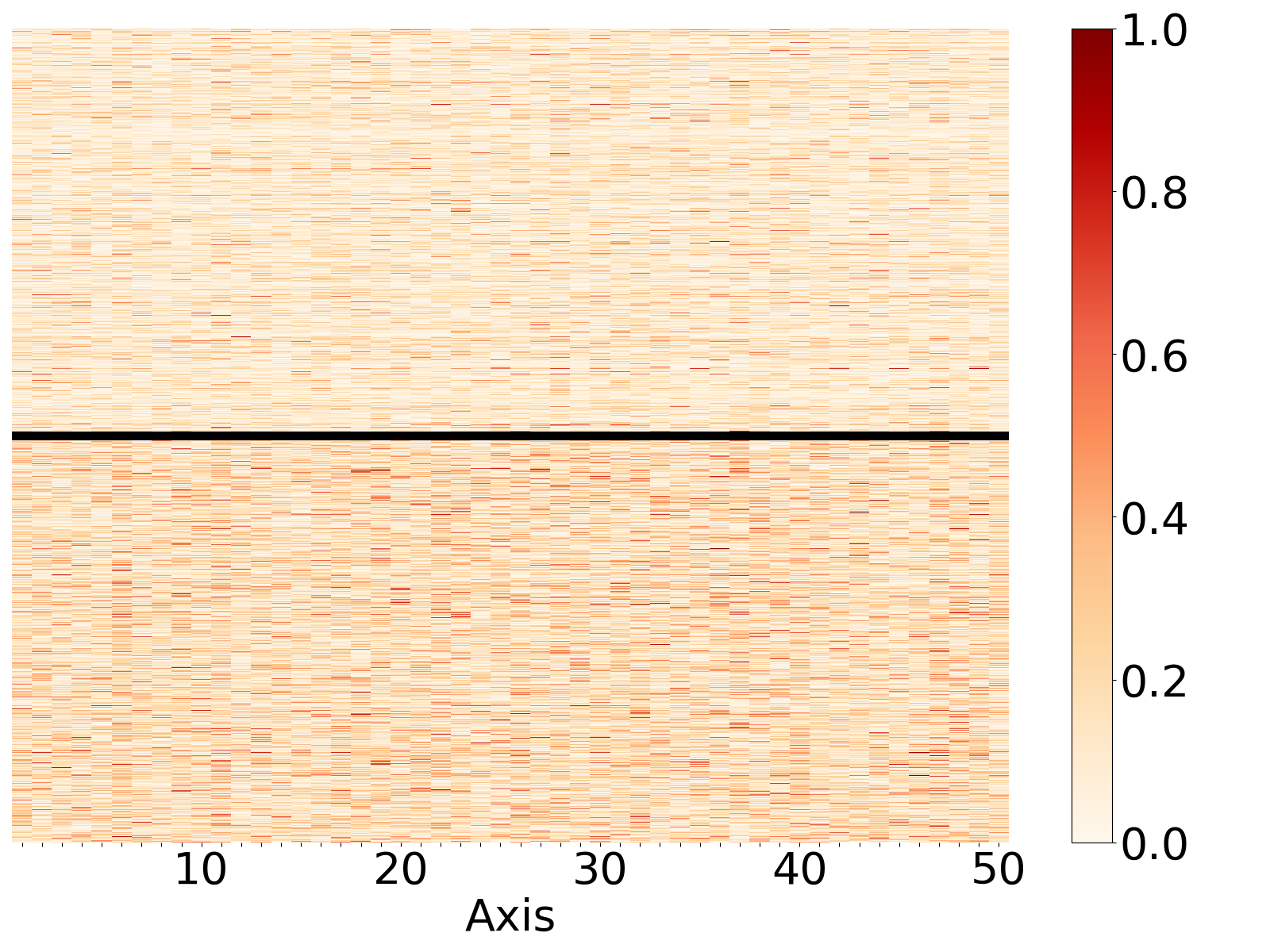}
        \subcaption{Fine-tuned \ac{SCWE}, Raw}
        \label{fig:wic_xlwic_fr_instances_raw_finetuned}
    \end{minipage}
    \begin{minipage}[b]{0.65\columnwidth}
        \centering
        \includegraphics[width=0.85\columnwidth]{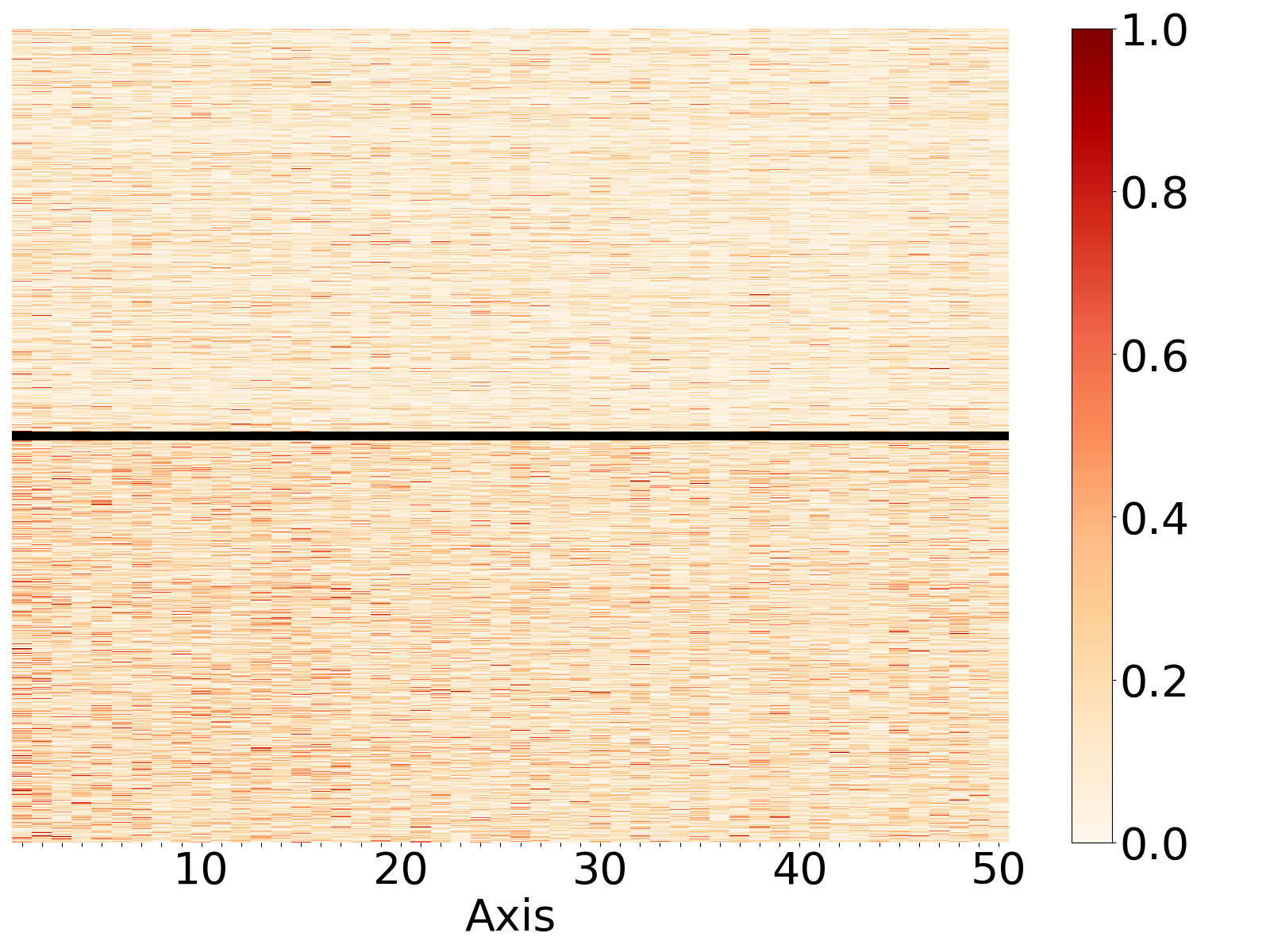}
        \subcaption{Fine-tuned \ac{SCWE}, PCA}
        \label{fig:wic_xlwic_fr_instances_pca_finetuned}
    \end{minipage}
    \begin{minipage}[b]{0.65\columnwidth}
        \centering
        \includegraphics[width=0.85\columnwidth]{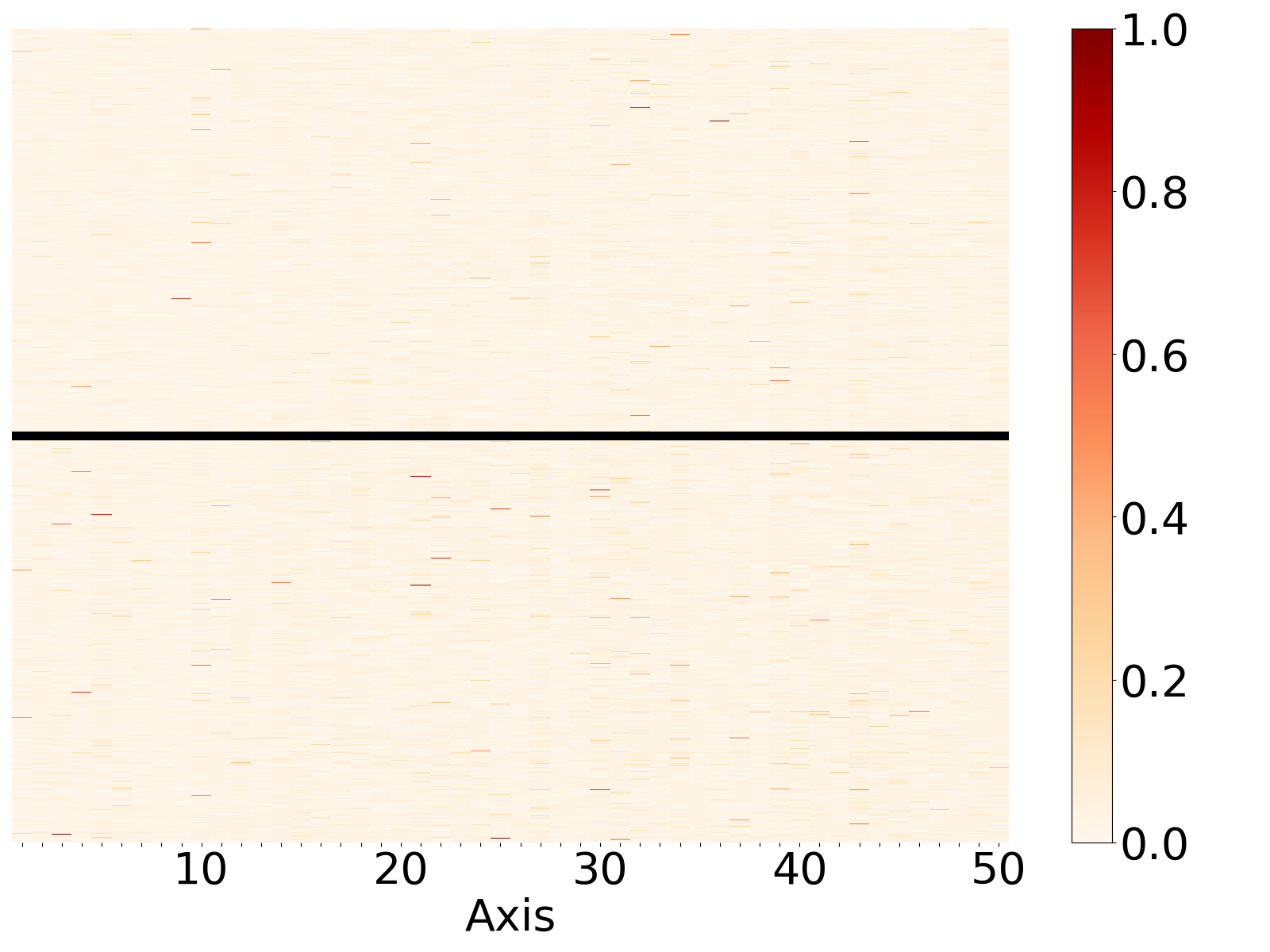}
        \subcaption{Fine-tuned \ac{SCWE}, ICA}
        \label{fig:wic_xlwic_fr_instances_ica_finetuned}
    \end{minipage}
    \caption{Visualisation of the top-50 dimensions of pre-trained \acp{CWE} (XLM-RoBERTa) and \acp{SCWE} (XL-LEXEME) for each instance in XLWiC (French) dataset, where the difference of vectors is calculated for (a/d) \textbf{Raw} vectors, (b/e) \ac{PCA}-transformed axes, and (c/f) \ac{ICA}-transformed axes. In each figure, the upper/lower half uses instances for the True/False labels.}
    \label{fig:wic_instance_xlwic_fr}
\end{figure*}

\begin{figure*}[t]
    \centering
    \begin{minipage}[b]{0.65\columnwidth}
        \centering
        \includegraphics[width=0.85\columnwidth]{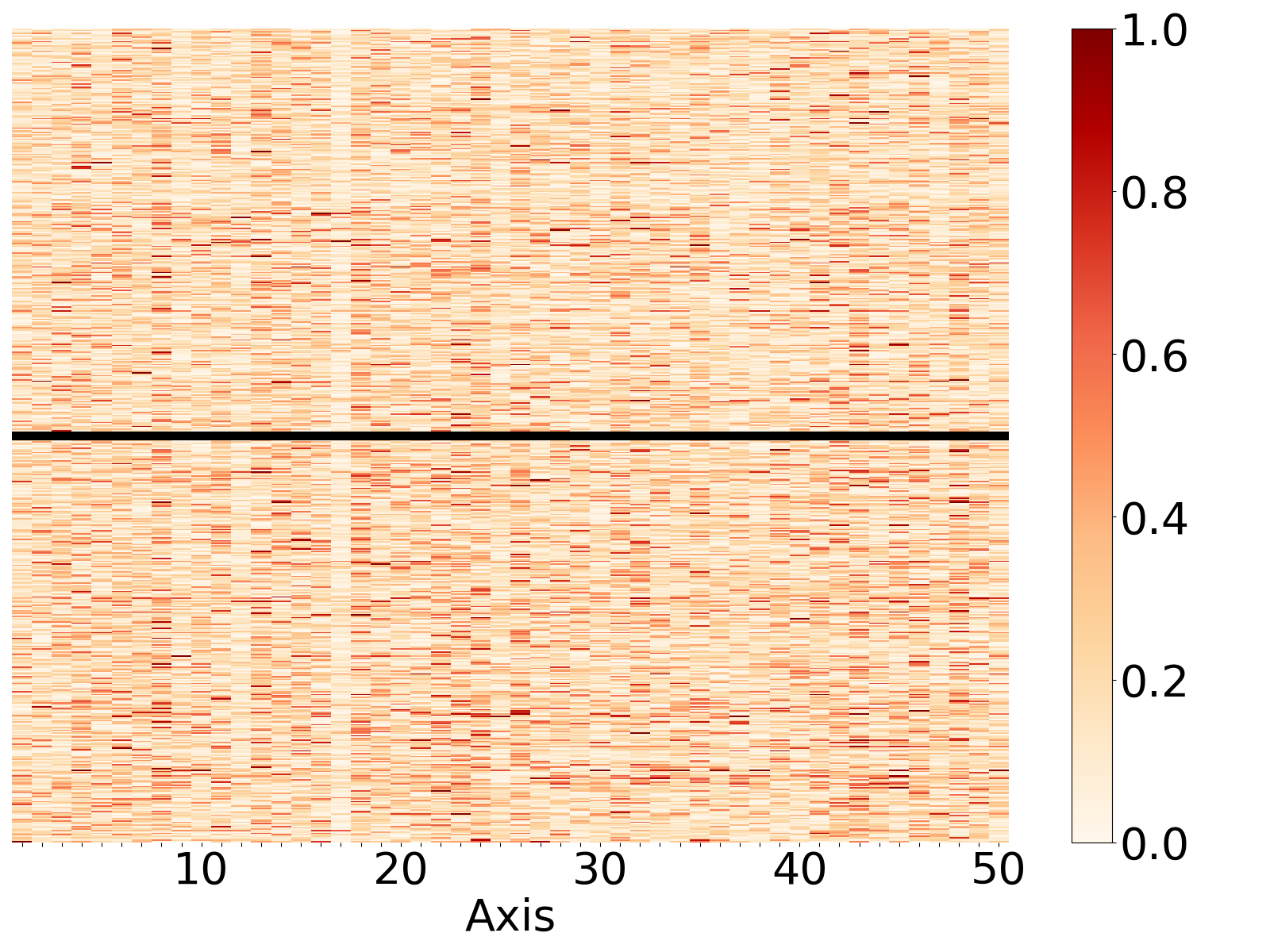}
        \subcaption{Pre-trained \ac{CWE}, Raw}
        \label{fig:wic_xlwic_it_instances_raw_pretrained}
    \end{minipage}
    \begin{minipage}[b]{0.65\columnwidth}
        \centering
        \includegraphics[width=0.85\columnwidth]{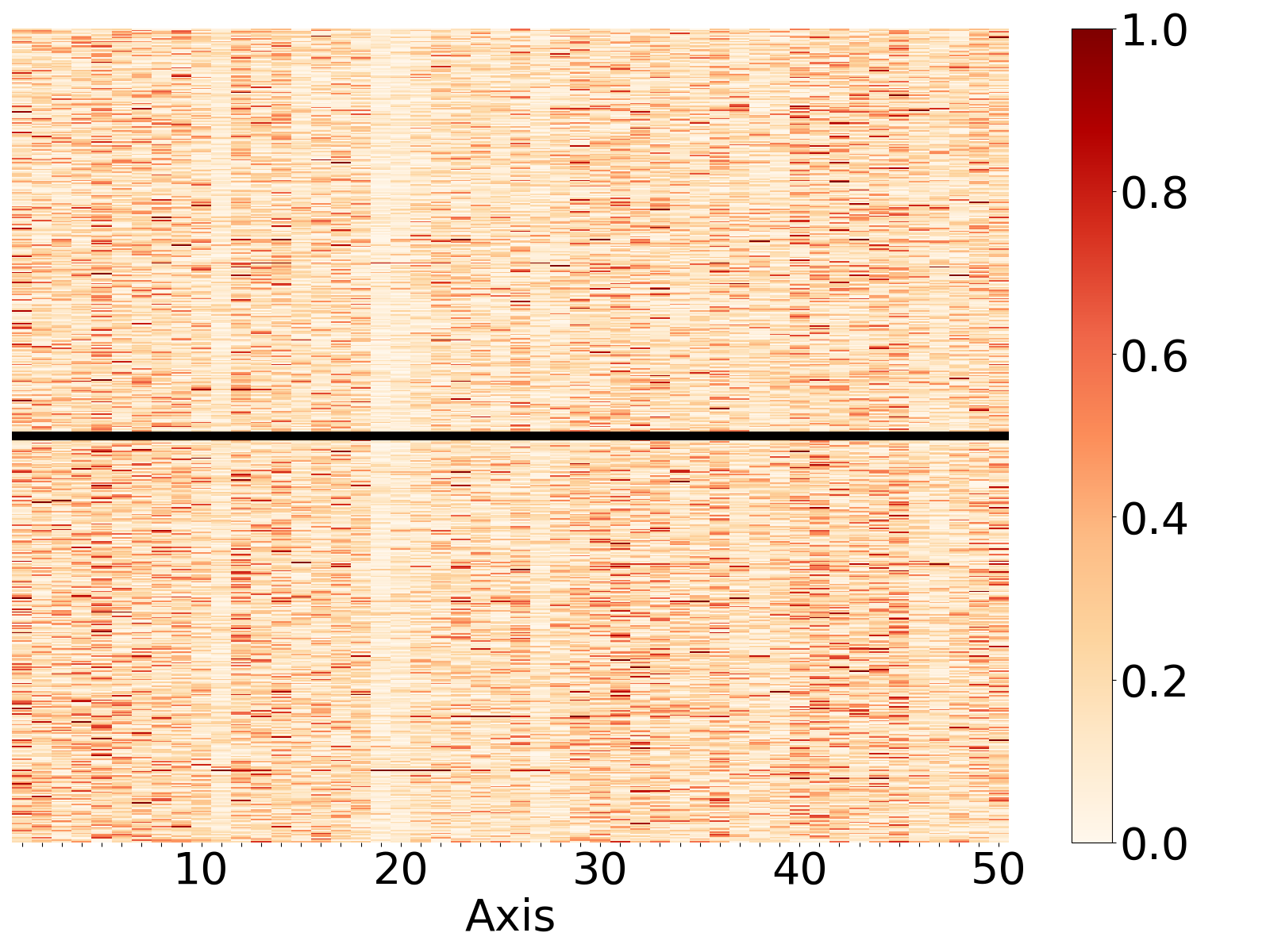}
        \subcaption{Pre-trained \ac{CWE}, PCA}
        \label{fig:wic_xlwic_it_instances_pca_pretrained}
    \end{minipage}
    \begin{minipage}[b]{0.65\columnwidth}
        \centering
        \includegraphics[width=0.85\columnwidth]{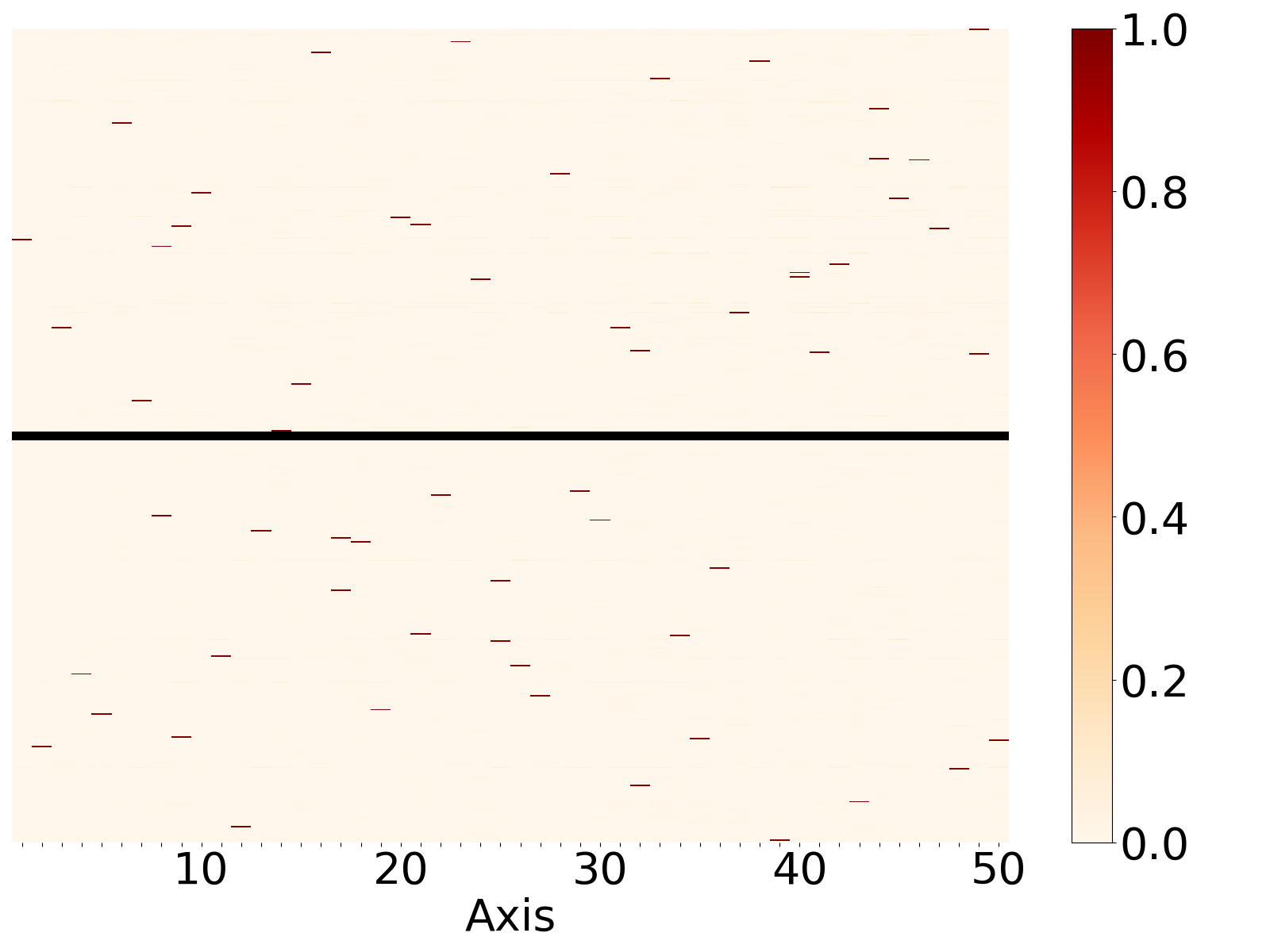}
        \subcaption{Pre-trained \ac{CWE}, ICA}
        \label{fig:wic_xlwic_it_instances_ica_pretrained}
    \end{minipage} \\
    \begin{minipage}[b]{0.65\columnwidth}
        \centering
        \includegraphics[width=0.85\columnwidth]{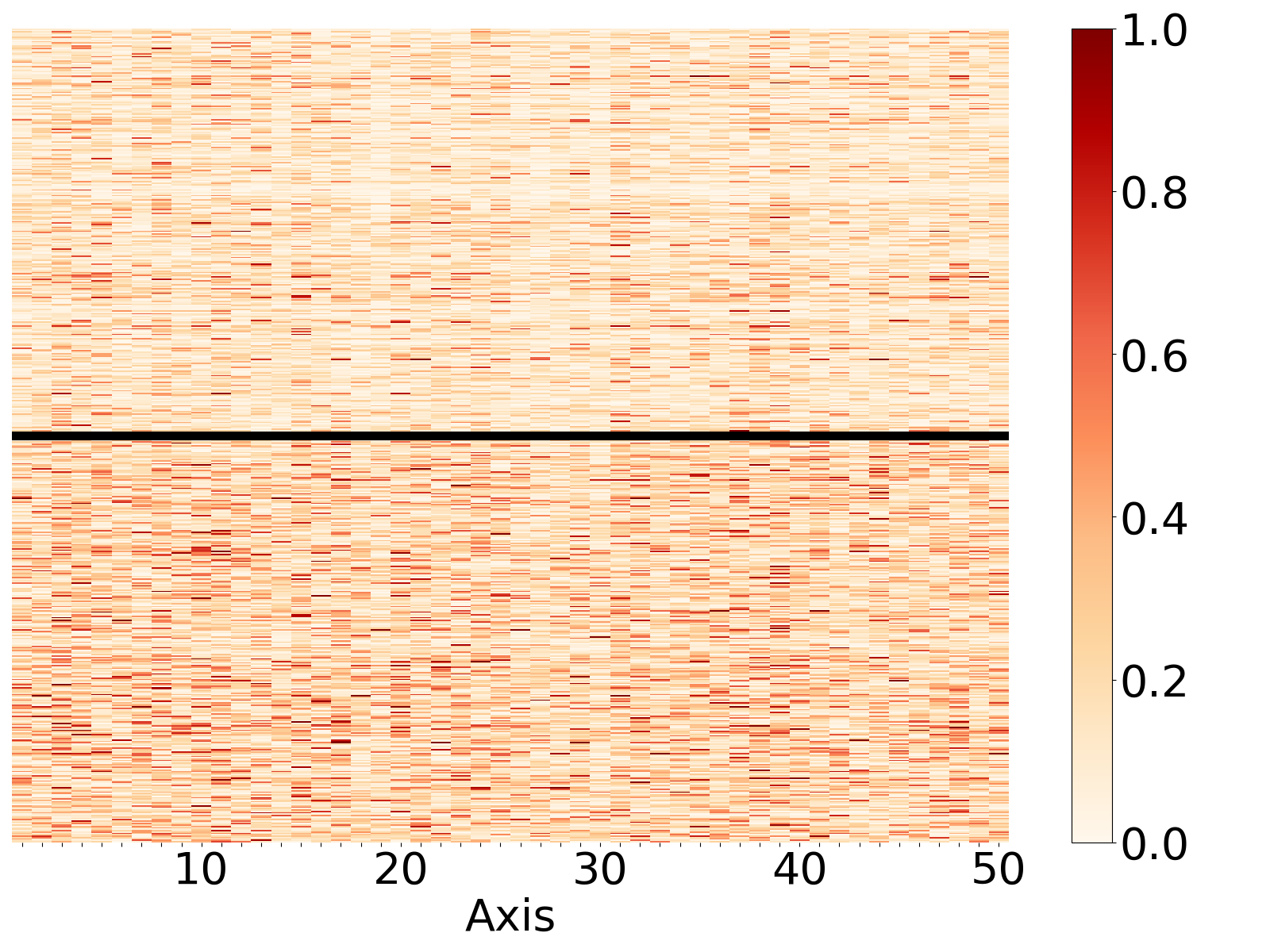}
        \subcaption{Fine-tuned \ac{SCWE}, Raw}
        \label{fig:wic_xlwic_it_instances_raw_finetuned}
    \end{minipage}
    \begin{minipage}[b]{0.65\columnwidth}
        \centering
        \includegraphics[width=0.85\columnwidth]{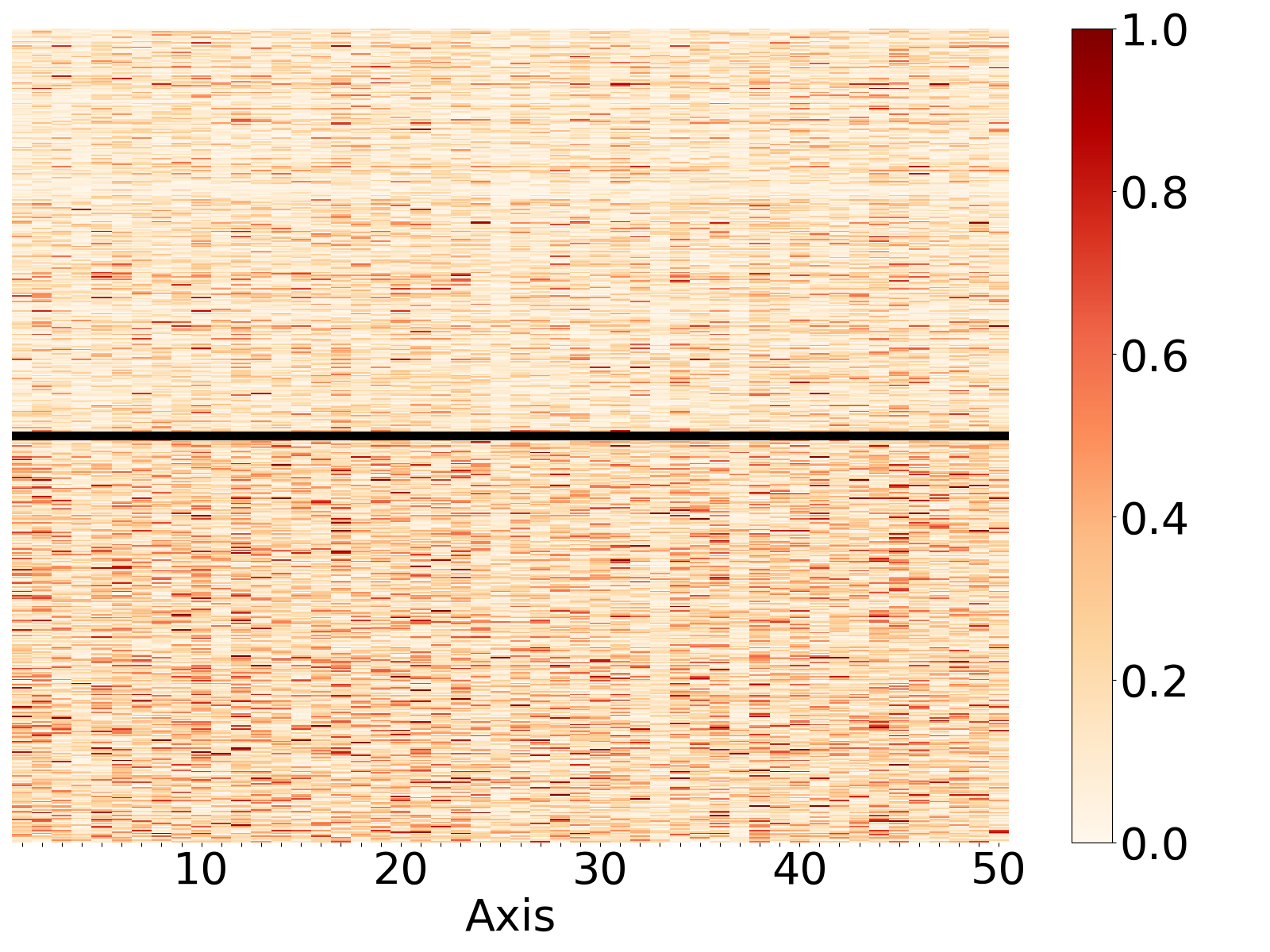}
        \subcaption{Fine-tuned \ac{SCWE}, PCA}
        \label{fig:wic_xlwic_it_instances_pca_finetuned}
    \end{minipage}
    \begin{minipage}[b]{0.65\columnwidth}
        \centering
        \includegraphics[width=0.85\columnwidth]{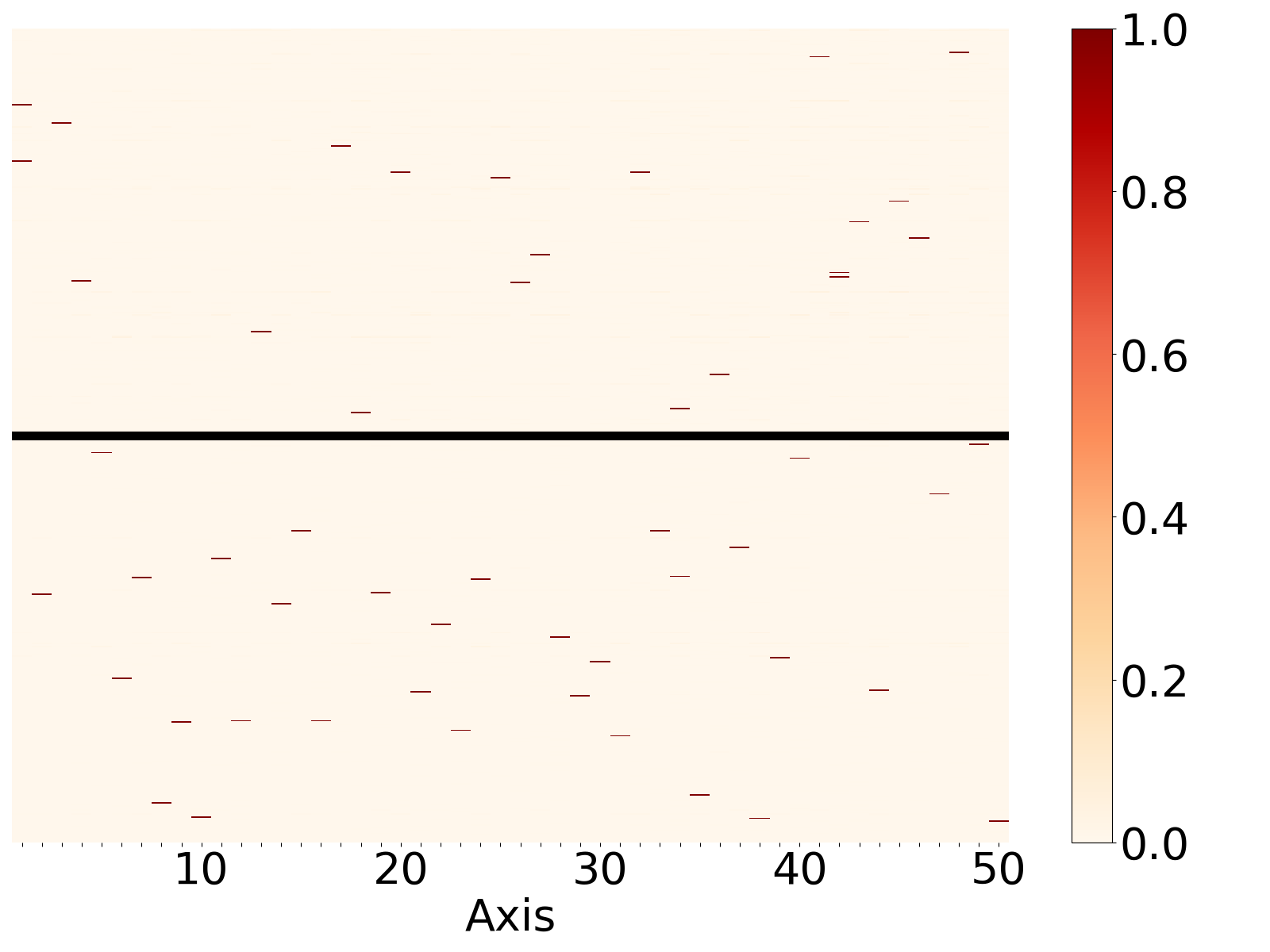}
        \subcaption{Fine-tuned \ac{SCWE}, ICA}
        \label{fig:wic_xlwic_it_instances_ica_finetuned}
    \end{minipage}
    \caption{Visualisation of the top-50 dimensions of pre-trained \acp{CWE} (XLM-RoBERTa) and \acp{SCWE} (XL-LEXEME) for each instance in XLWiC (Italian) dataset, where the difference of vectors is calculated for (a/d) \textbf{Raw} vectors, (b/e) \ac{PCA}-transformed axes, and (c/f) \ac{ICA}-transformed axes. In each figure, the upper/lower half uses instances for the True/False labels.}
    \label{fig:wic_instance_xlwic_it}
\end{figure*}

\begin{figure*}[t]
    \centering
    \begin{minipage}[b]{0.65\columnwidth}
        \centering
        \includegraphics[width=0.85\columnwidth]{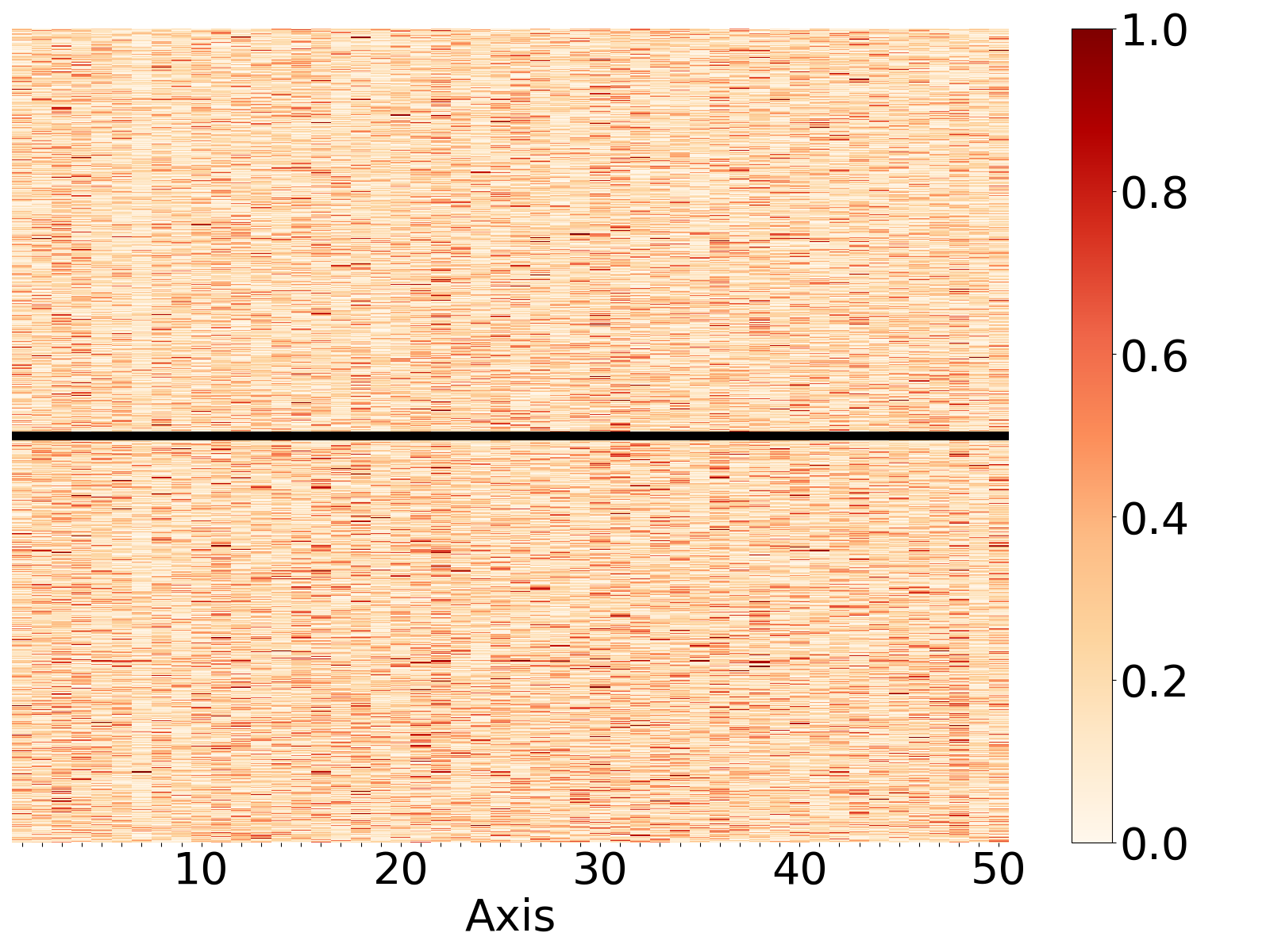}
        \subcaption{Pre-trained \ac{CWE}, Raw}
        \label{fig:wic_mclwic_ar_instances_raw_pretrained}
    \end{minipage}
    \begin{minipage}[b]{0.65\columnwidth}
        \centering
        \includegraphics[width=0.85\columnwidth]{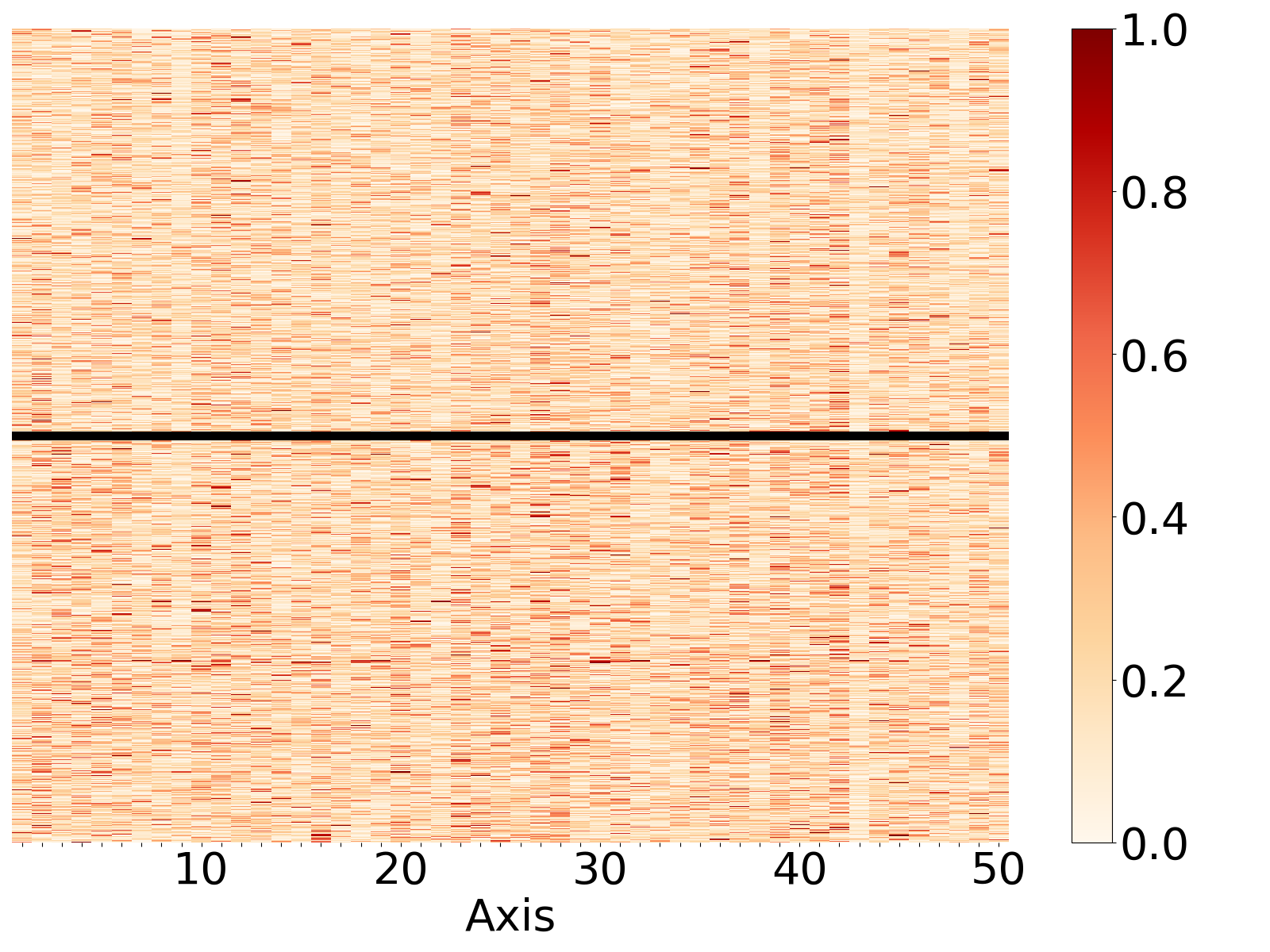}
        \subcaption{Pre-trained \ac{CWE}, PCA}
        \label{fig:wic_mclwic_ar_instances_pca_pretrained}
    \end{minipage}
    \begin{minipage}[b]{0.65\columnwidth}
        \centering
        \includegraphics[width=0.85\columnwidth]{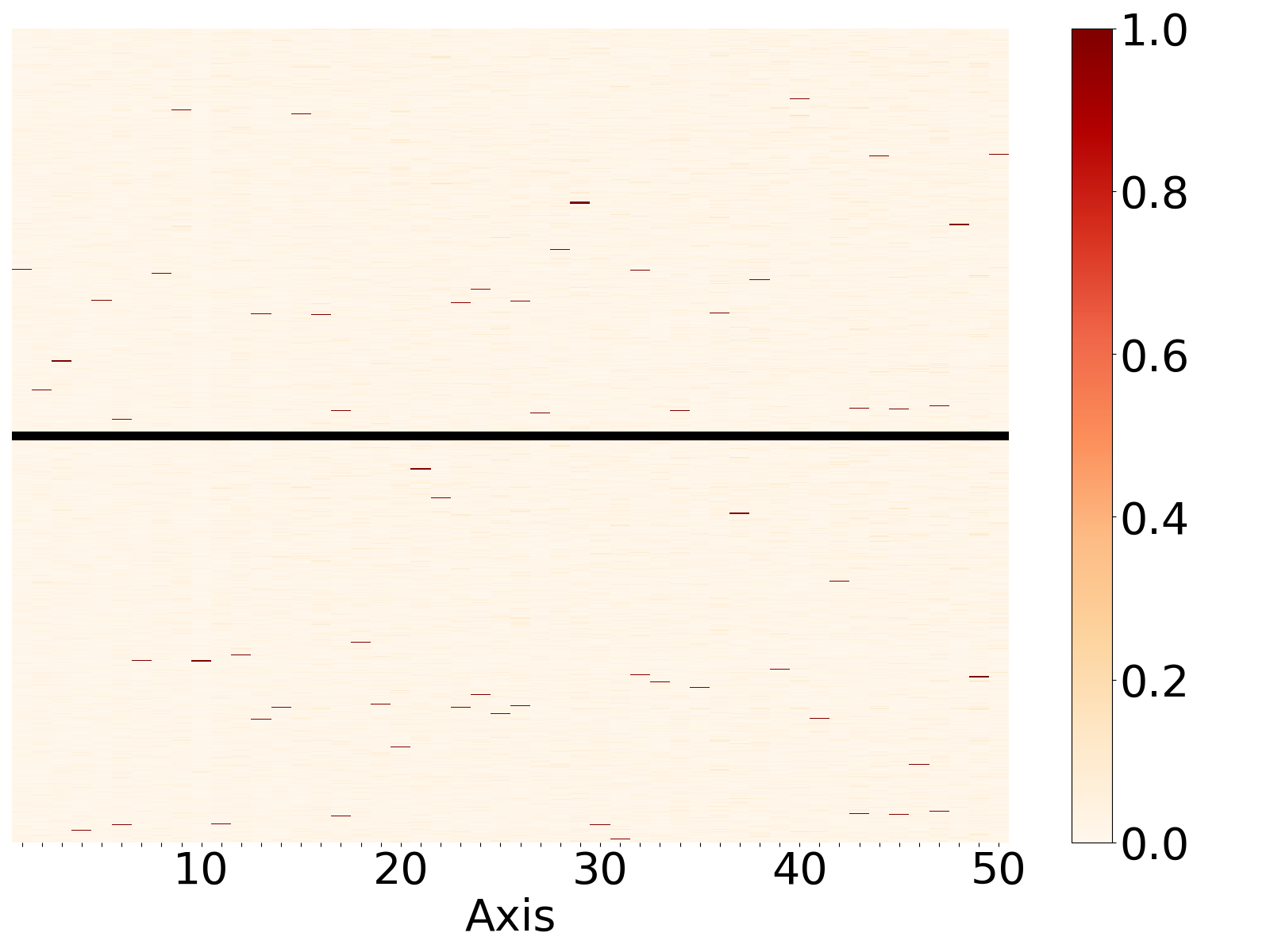}
        \subcaption{Pre-trained \ac{CWE}, ICA}
        \label{fig:wic_mclwic_ar_instances_ica_pretrained}
    \end{minipage} \\
    \begin{minipage}[b]{0.65\columnwidth}
        \centering
        \includegraphics[width=0.85\columnwidth]{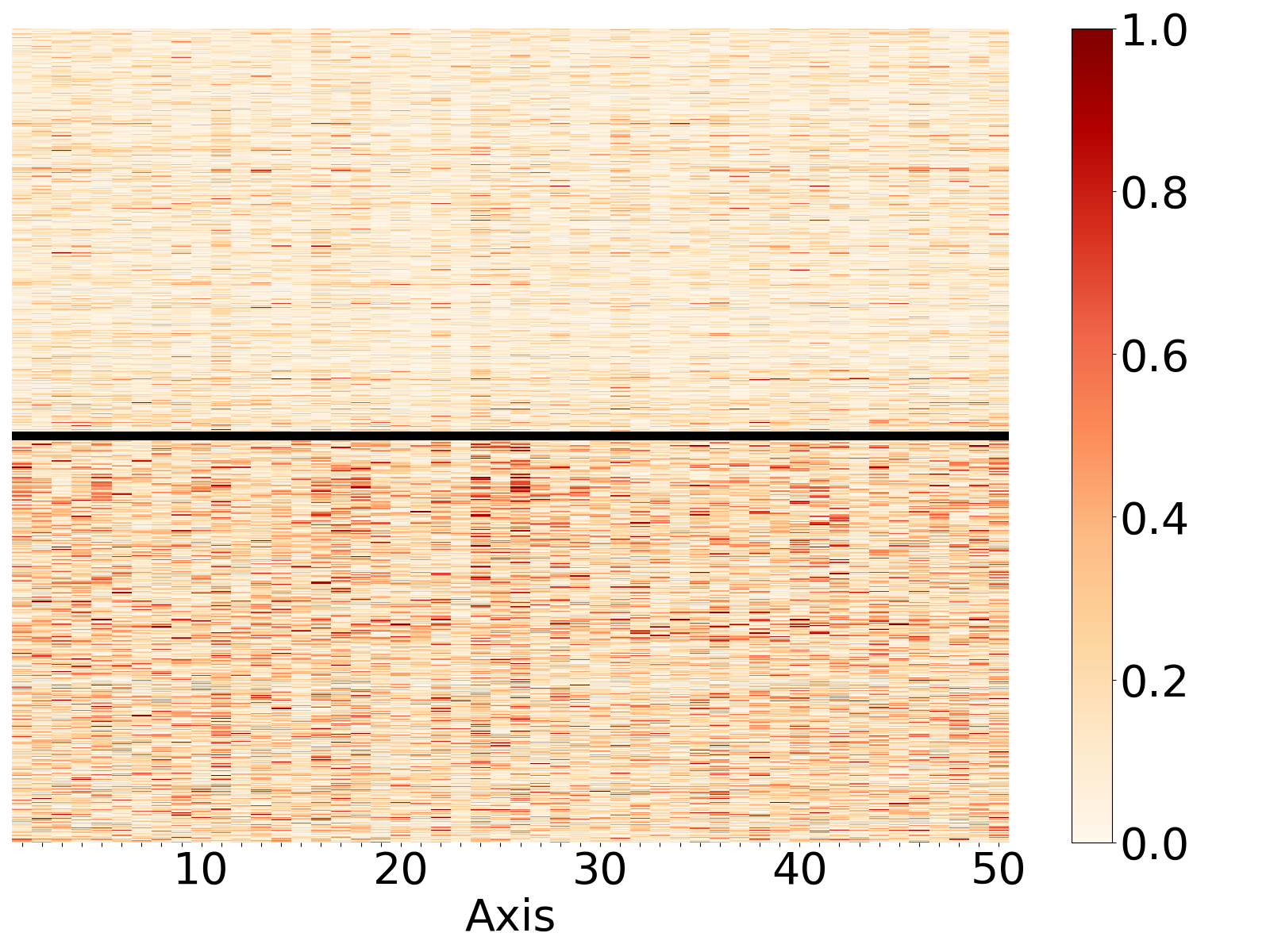}
        \subcaption{Fine-tuned \ac{SCWE}, Raw}
        \label{fig:wic_mclwic_ar_instances_raw_finetuned}
    \end{minipage}
    \begin{minipage}[b]{0.65\columnwidth}
        \centering
        \includegraphics[width=0.85\columnwidth]{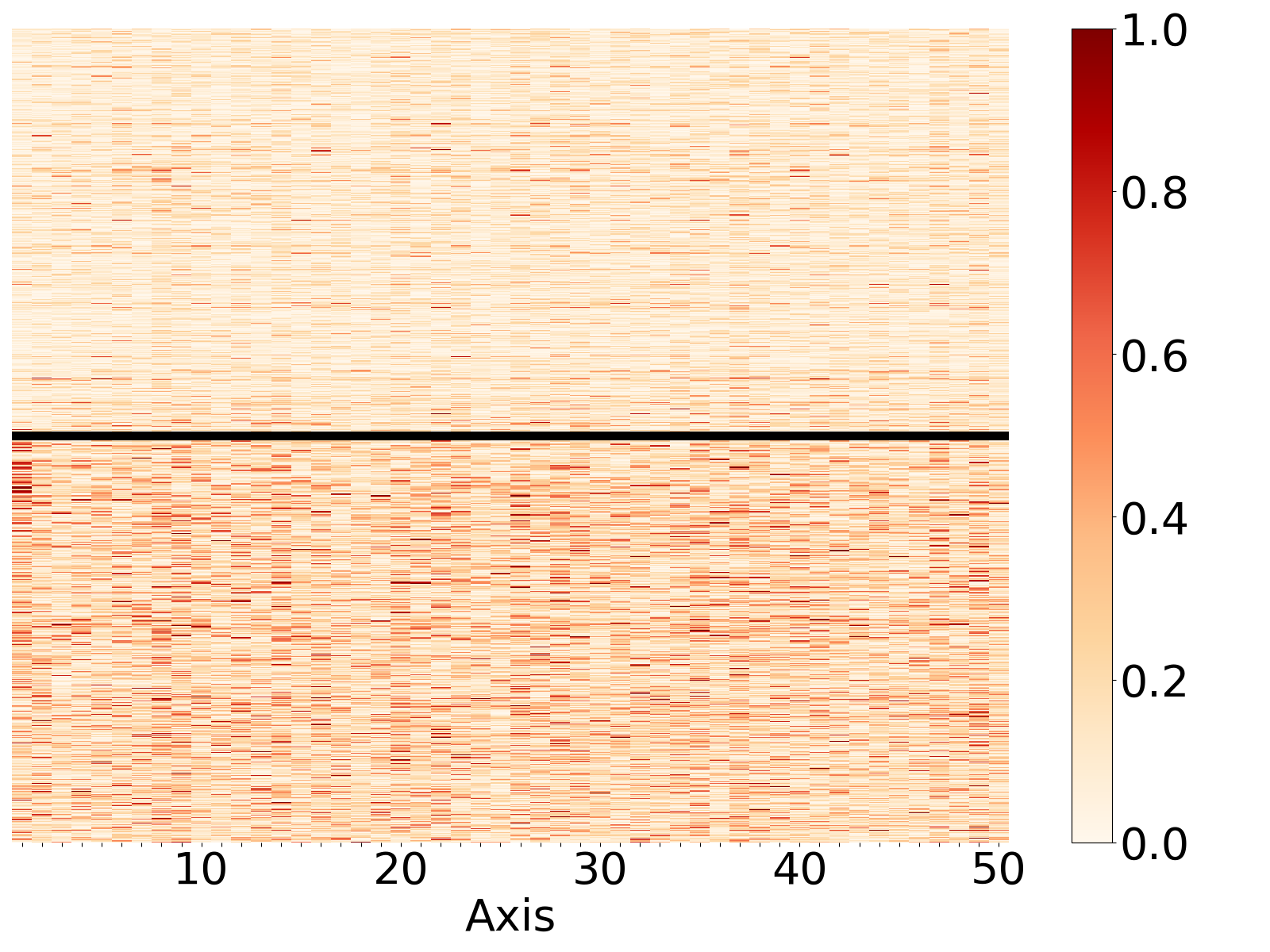}
        \subcaption{Fine-tuned \ac{SCWE}, PCA}
        \label{fig:wic_mclwic_ar_instances_pca_finetuned}
    \end{minipage}
    \begin{minipage}[b]{0.65\columnwidth}
        \centering
        \includegraphics[width=0.85\columnwidth]{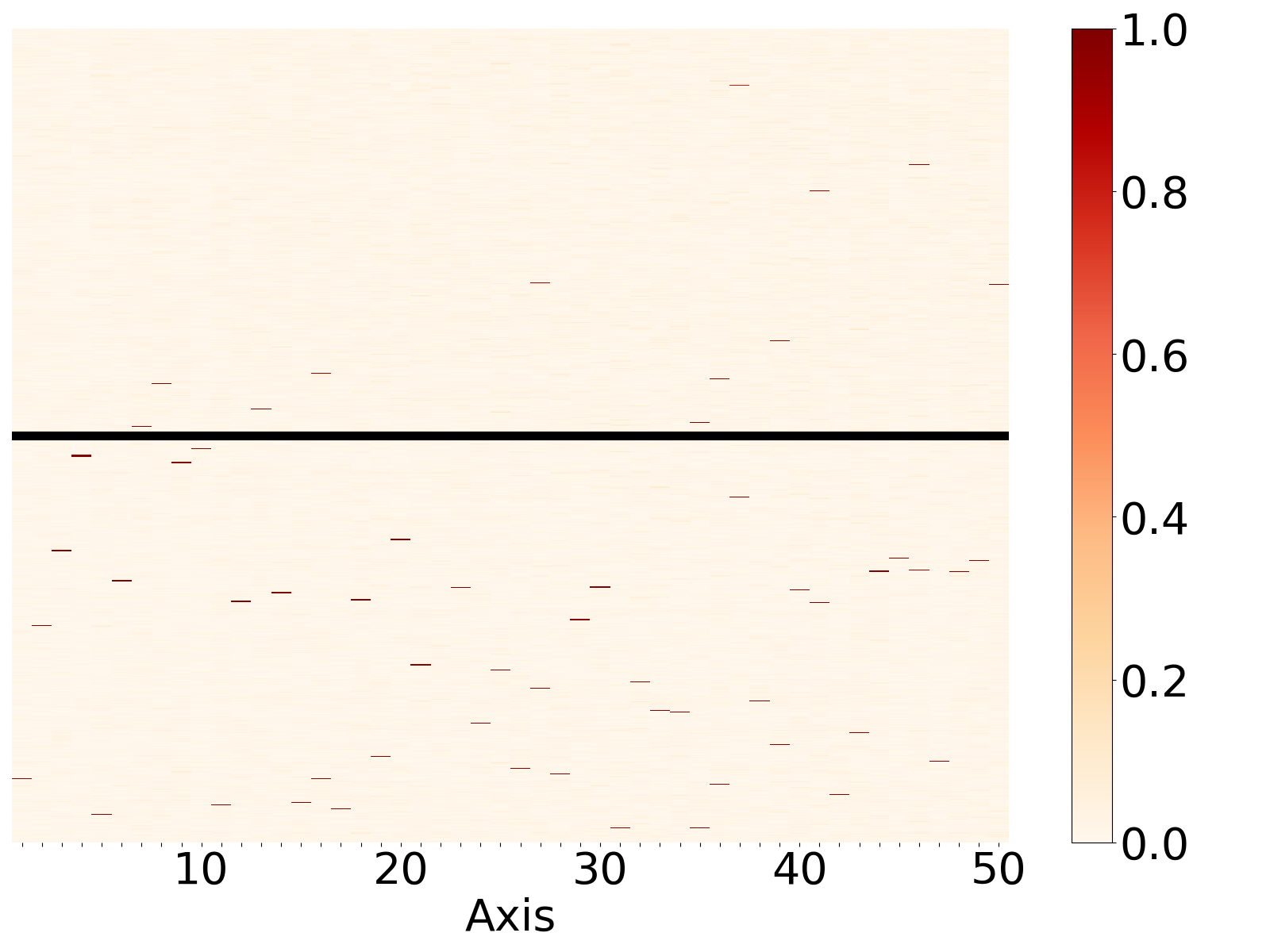}
        \subcaption{Fine-tuned \ac{SCWE}, ICA}
        \label{fig:wic_mclwic_ar_instances_ica_finetuned}
    \end{minipage}
    \caption{Visualisation of the top-50 dimensions of pre-trained \acp{CWE} (XLM-RoBERTa) and \acp{SCWE} (XL-LEXEME) for each instance in MCLWiC (Arabic) dataset, where the difference of vectors is calculated for (a/d) \textbf{Raw} vectors, (b/e) \ac{PCA}-transformed axes, and (c/f) \ac{ICA}-transformed axes. In each figure, the upper/lower half uses instances for the True/False labels.}
    \label{fig:wic_instance_mclwic_ar}
\end{figure*}

\begin{figure*}[t]
    \centering
    \begin{minipage}[b]{0.65\columnwidth}
        \centering
        \includegraphics[width=0.85\columnwidth]{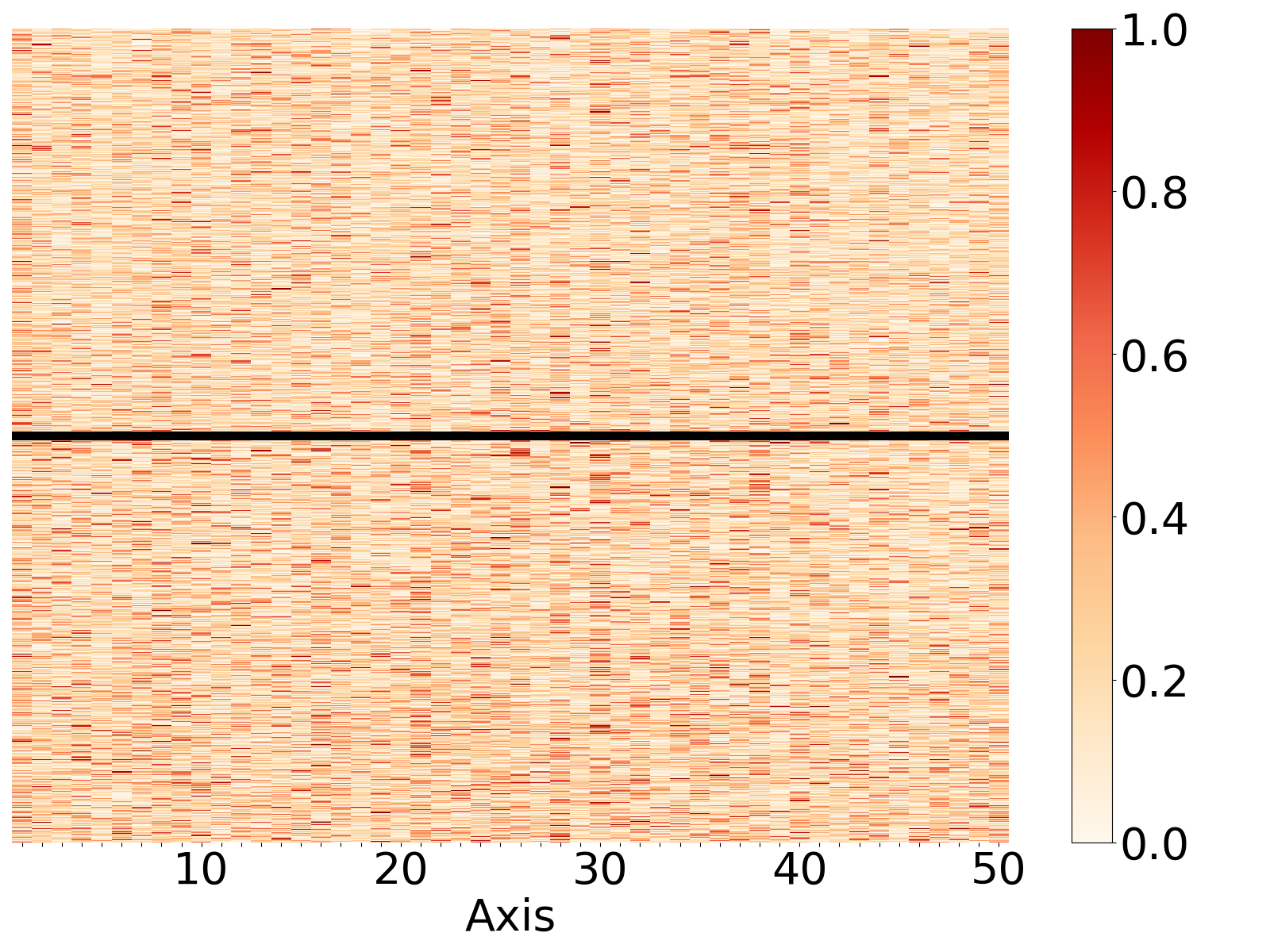}
        \subcaption{Pre-trained \ac{CWE}, Raw}
        \label{fig:wic_mclwic_en_instances_raw_pretrained}
    \end{minipage}
    \begin{minipage}[b]{0.65\columnwidth}
        \centering
        \includegraphics[width=0.85\columnwidth]{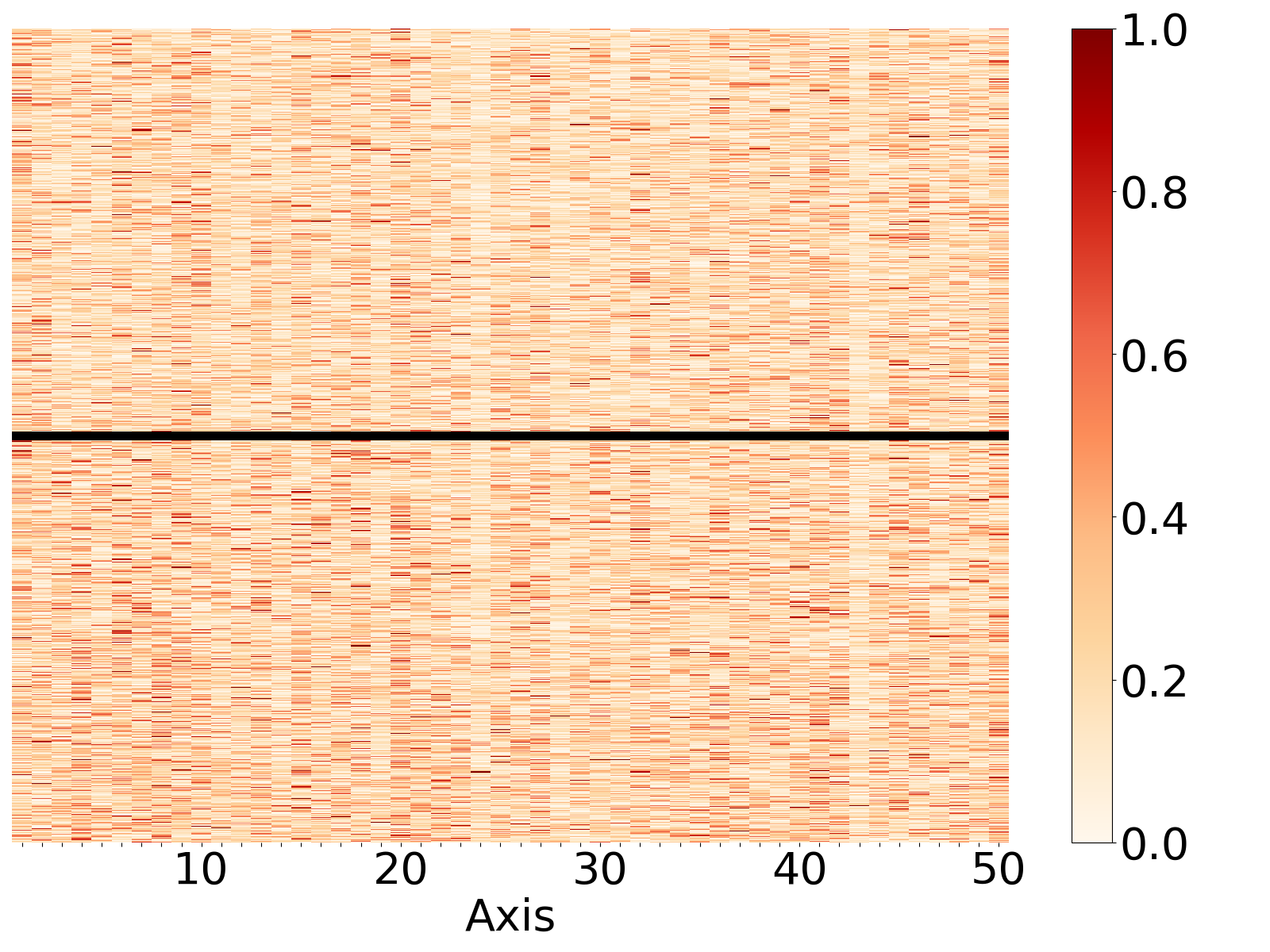}
        \subcaption{Pre-trained \ac{CWE}, PCA}
        \label{fig:wic_mclwic_en_instances_pca_pretrained}
    \end{minipage}
    \begin{minipage}[b]{0.65\columnwidth}
        \centering
        \includegraphics[width=0.85\columnwidth]{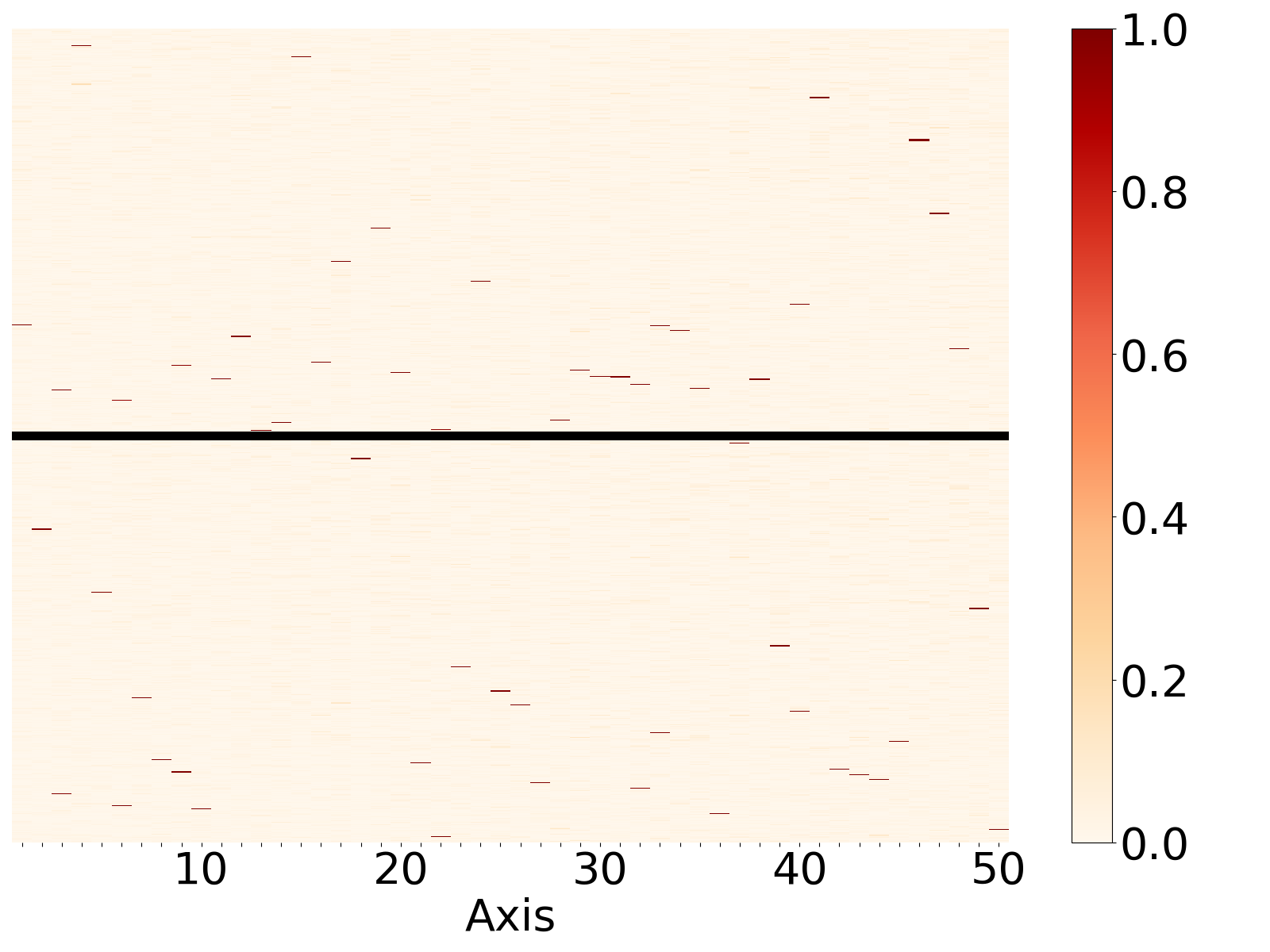}
        \subcaption{Pre-trained \ac{CWE}, ICA}
        \label{fig:wic_mclwic_en_instances_ica_pretrained}
    \end{minipage} \\
    \begin{minipage}[b]{0.65\columnwidth}
        \centering
        \includegraphics[width=0.85\columnwidth]{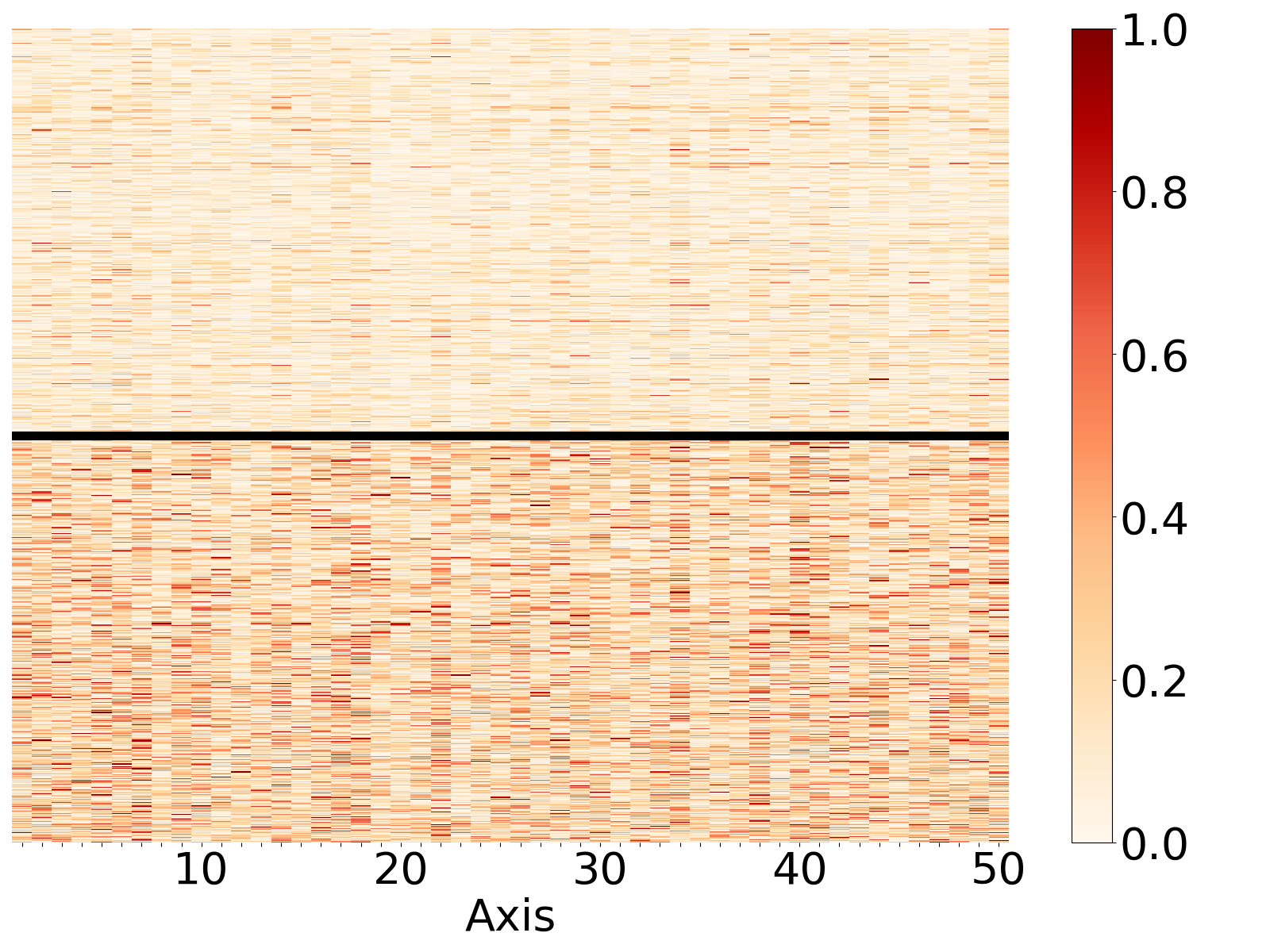}
        \subcaption{Fine-tuned \ac{SCWE}, Raw}
        \label{fig:wic_mclwic_en_instances_raw_finetuned}
    \end{minipage}
    \begin{minipage}[b]{0.65\columnwidth}
        \centering
        \includegraphics[width=0.85\columnwidth]{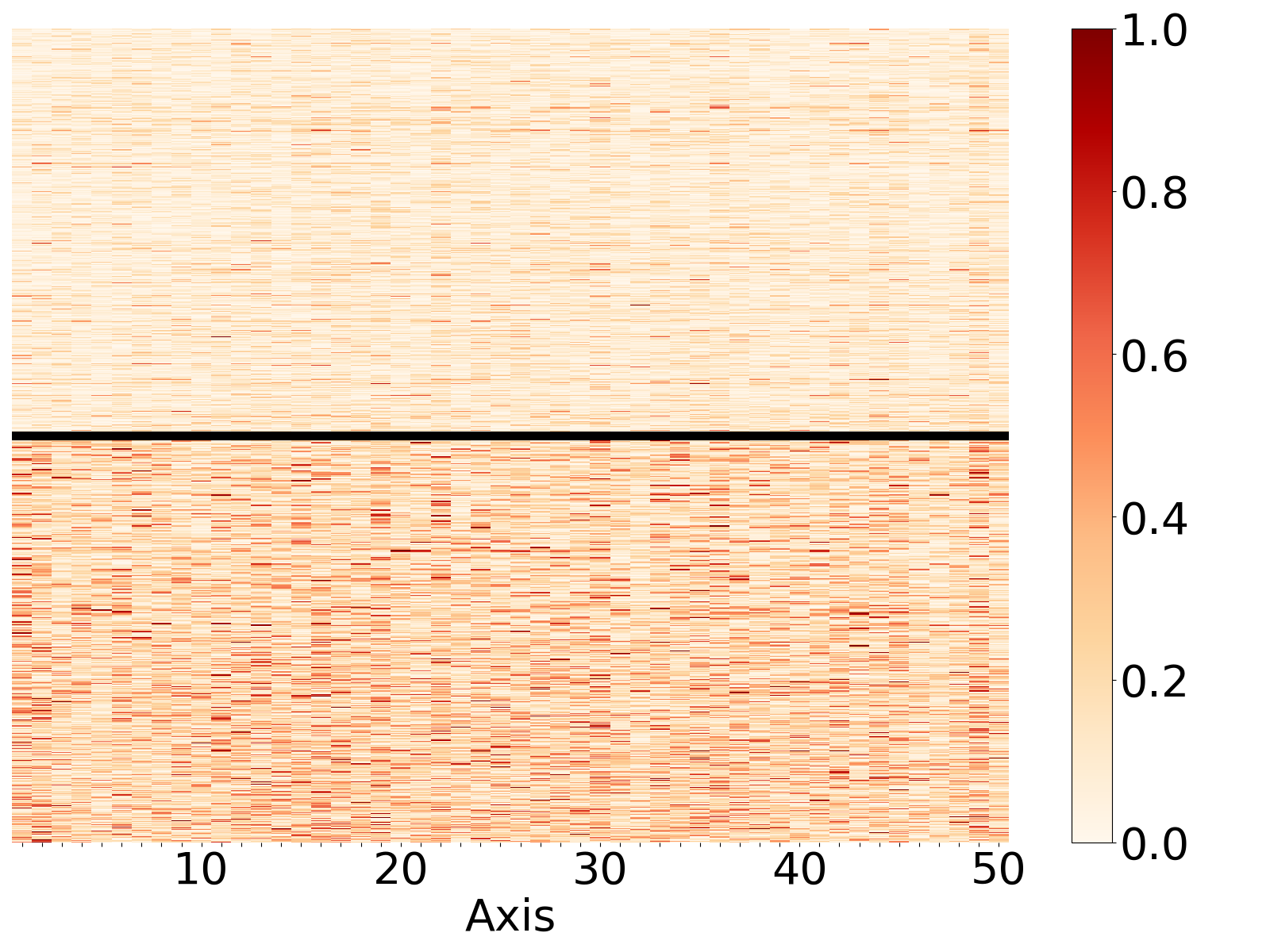}
        \subcaption{Fine-tuned \ac{SCWE}, PCA}
        \label{fig:wic_mclwic_en_instances_pca_finetuned}
    \end{minipage}
    \begin{minipage}[b]{0.65\columnwidth}
        \centering
        \includegraphics[width=0.85\columnwidth]{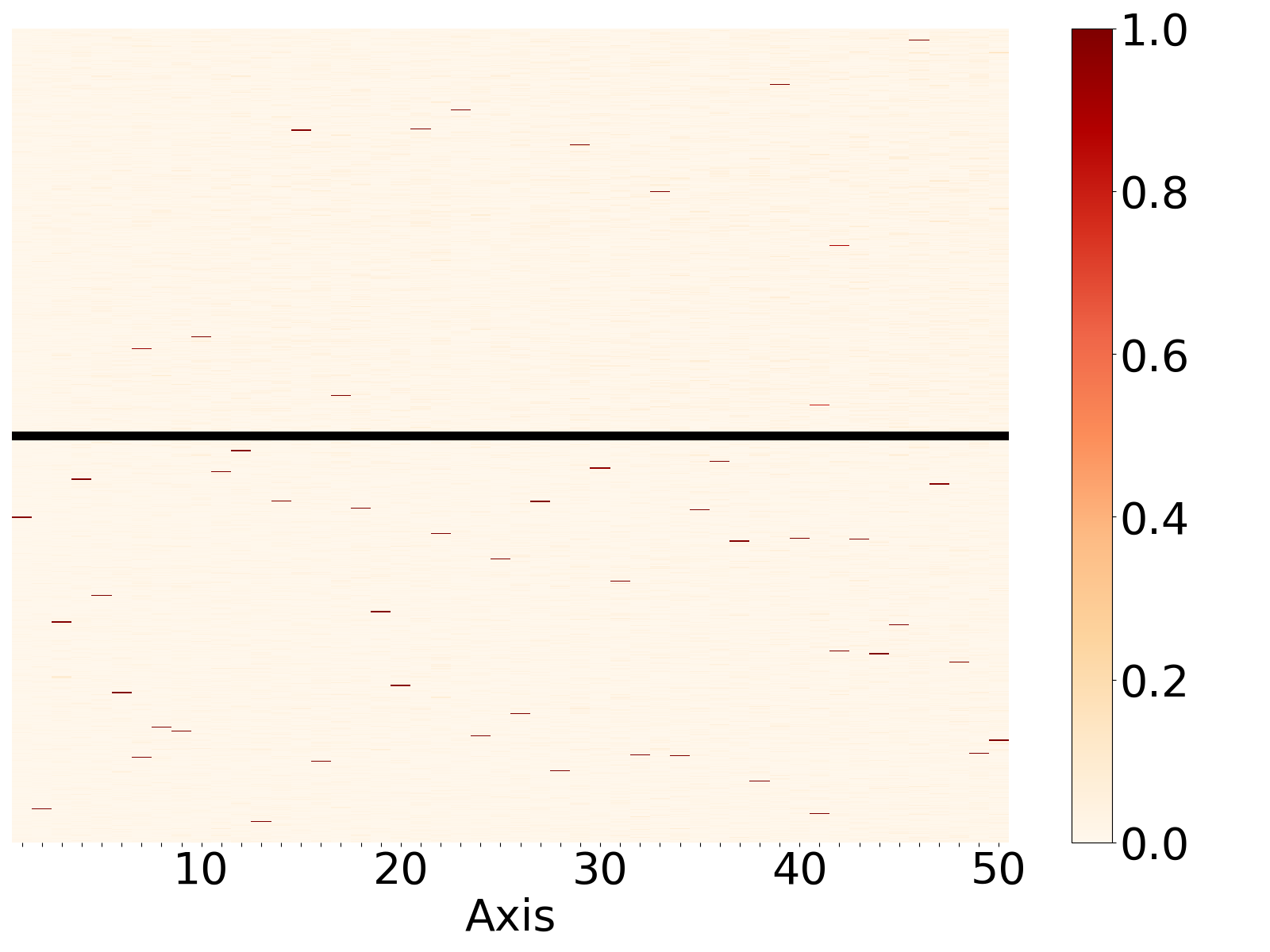}
        \subcaption{Fine-tuned \ac{SCWE}, ICA}
        \label{fig:wic_mclwic_en_instances_ica_finetuned}
    \end{minipage}
    \caption{Visualisation of the top-50 dimensions of pre-trained \acp{CWE} (XLM-RoBERTa) and \acp{SCWE} (XL-LEXEME) for each instance in MCLWiC (English) dataset, where the difference of vectors is calculated for (a/d) \textbf{Raw} vectors, (b/e) \ac{PCA}-transformed axes, and (c/f) \ac{ICA}-transformed axes. In each figure, the upper/lower half uses instances for the True/False labels.}
    \label{fig:wic_instance_mclwic_en}
\end{figure*}

\begin{figure*}[t]
    \centering
    \begin{minipage}[b]{0.65\columnwidth}
        \centering
        \includegraphics[width=0.85\columnwidth]{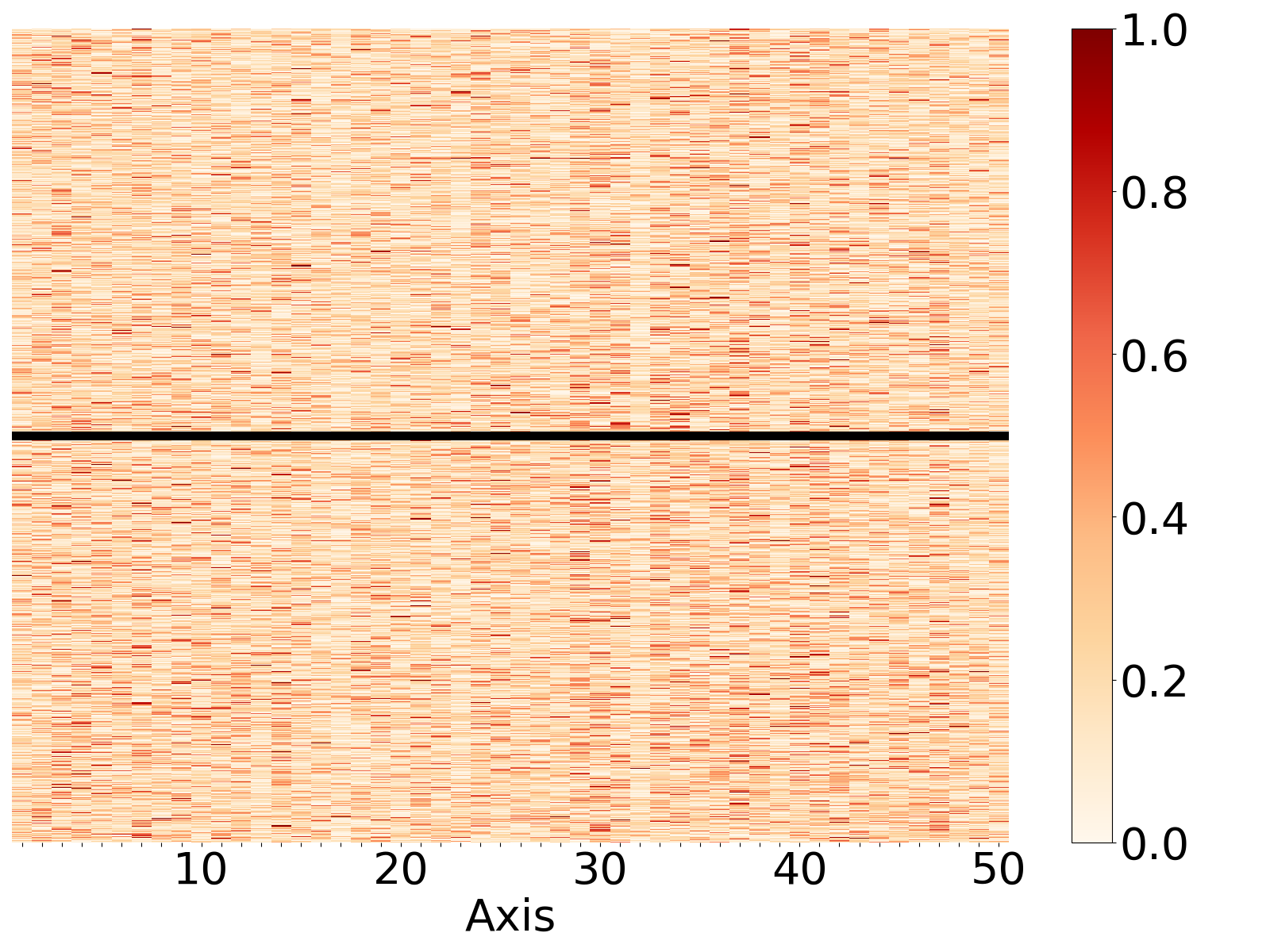}
        \subcaption{Pre-trained \ac{CWE}, Raw}
        \label{fig:wic_mclwic_fr_instances_raw_pretrained}
    \end{minipage}
    \begin{minipage}[b]{0.65\columnwidth}
        \centering
        \includegraphics[width=0.85\columnwidth]{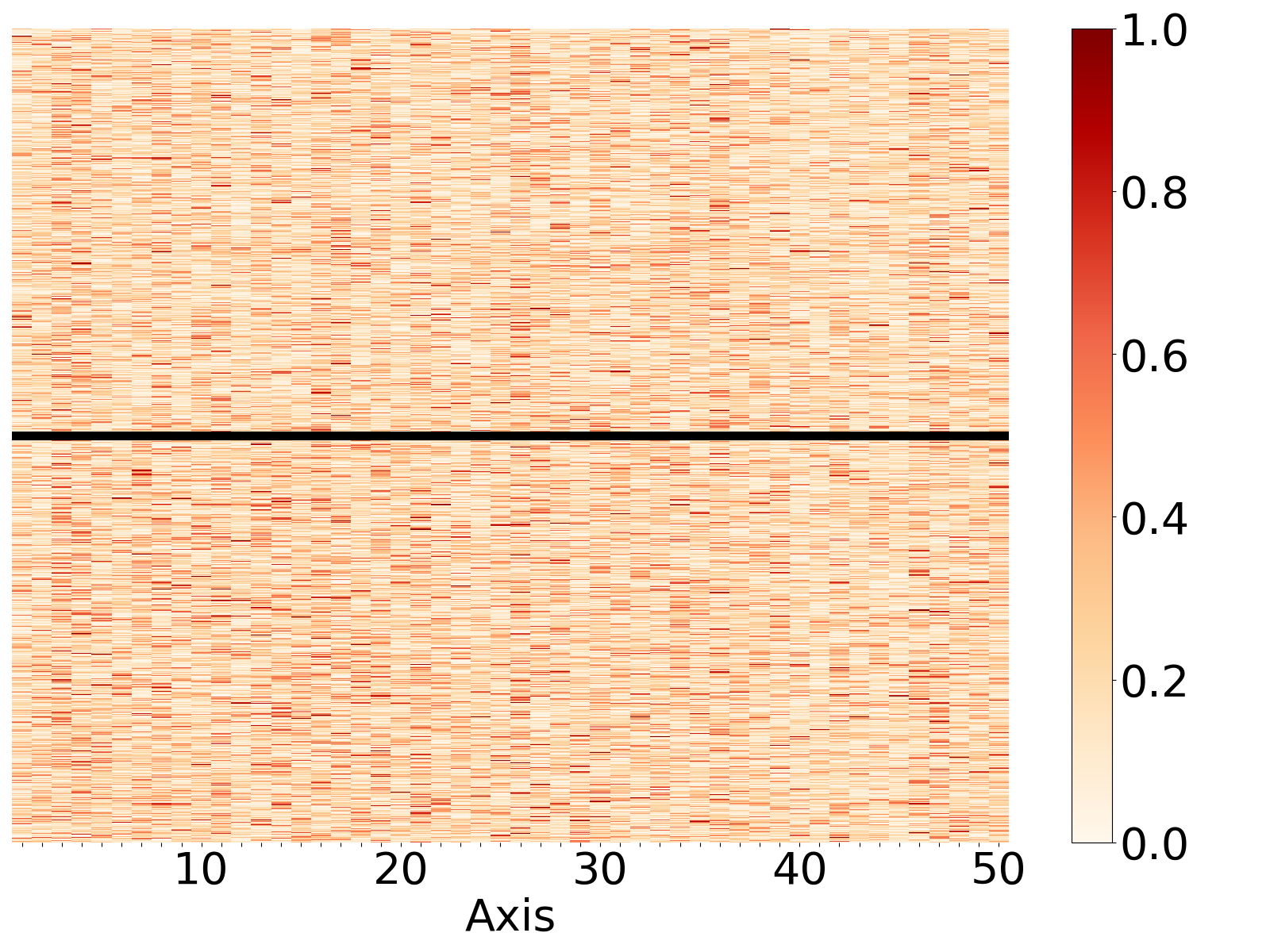}
        \subcaption{Pre-trained \ac{CWE}, PCA}
        \label{fig:wic_mclwic_fr_instances_pca_pretrained}
    \end{minipage}
    \begin{minipage}[b]{0.65\columnwidth}
        \centering
        \includegraphics[width=0.85\columnwidth]{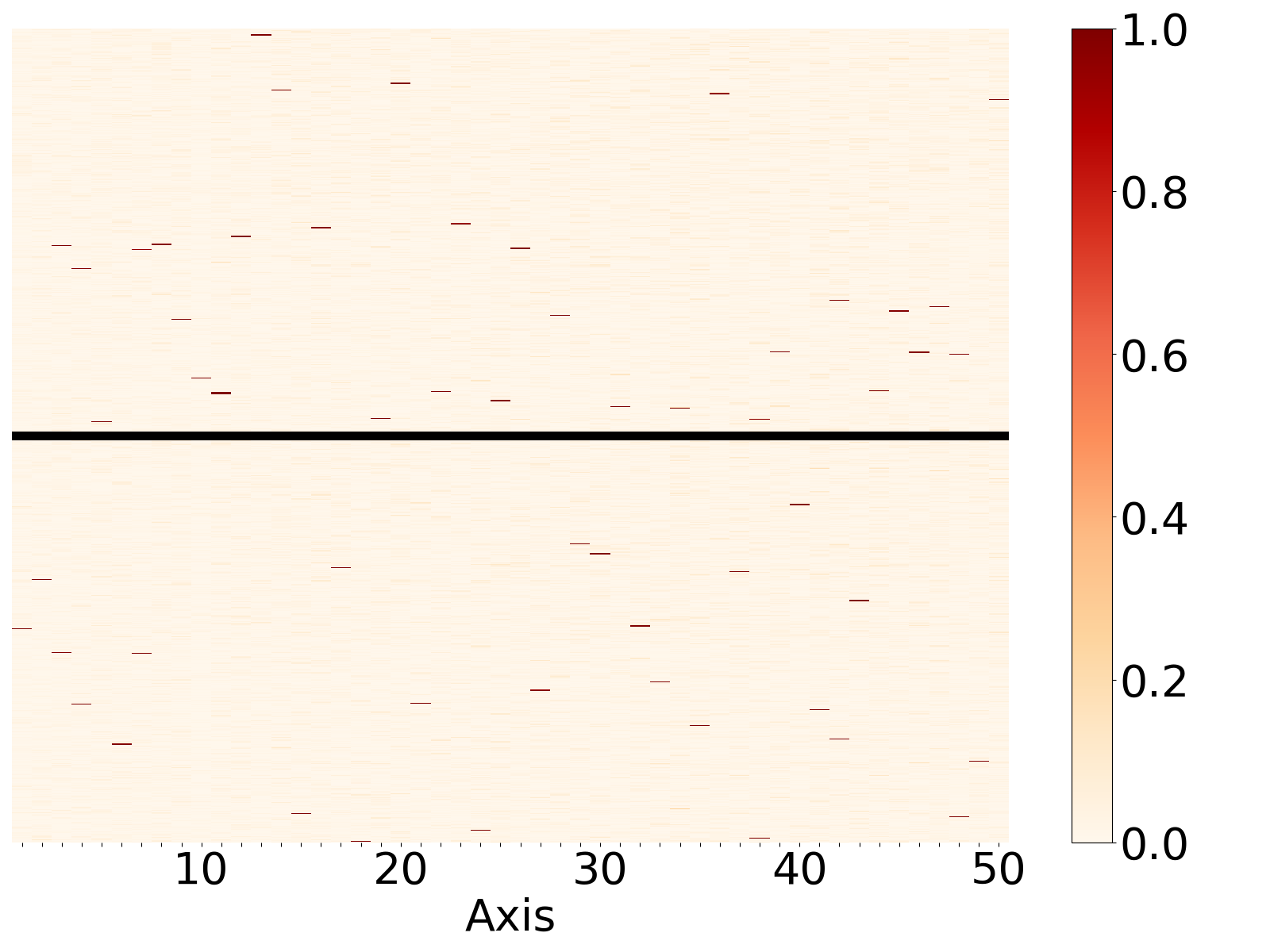}
        \subcaption{Pre-trained \ac{CWE}, ICA}
        \label{fig:wic_mclwic_fr_instances_ica_pretrained}
    \end{minipage} \\
    \begin{minipage}[b]{0.65\columnwidth}
        \centering
        \includegraphics[width=0.85\columnwidth]{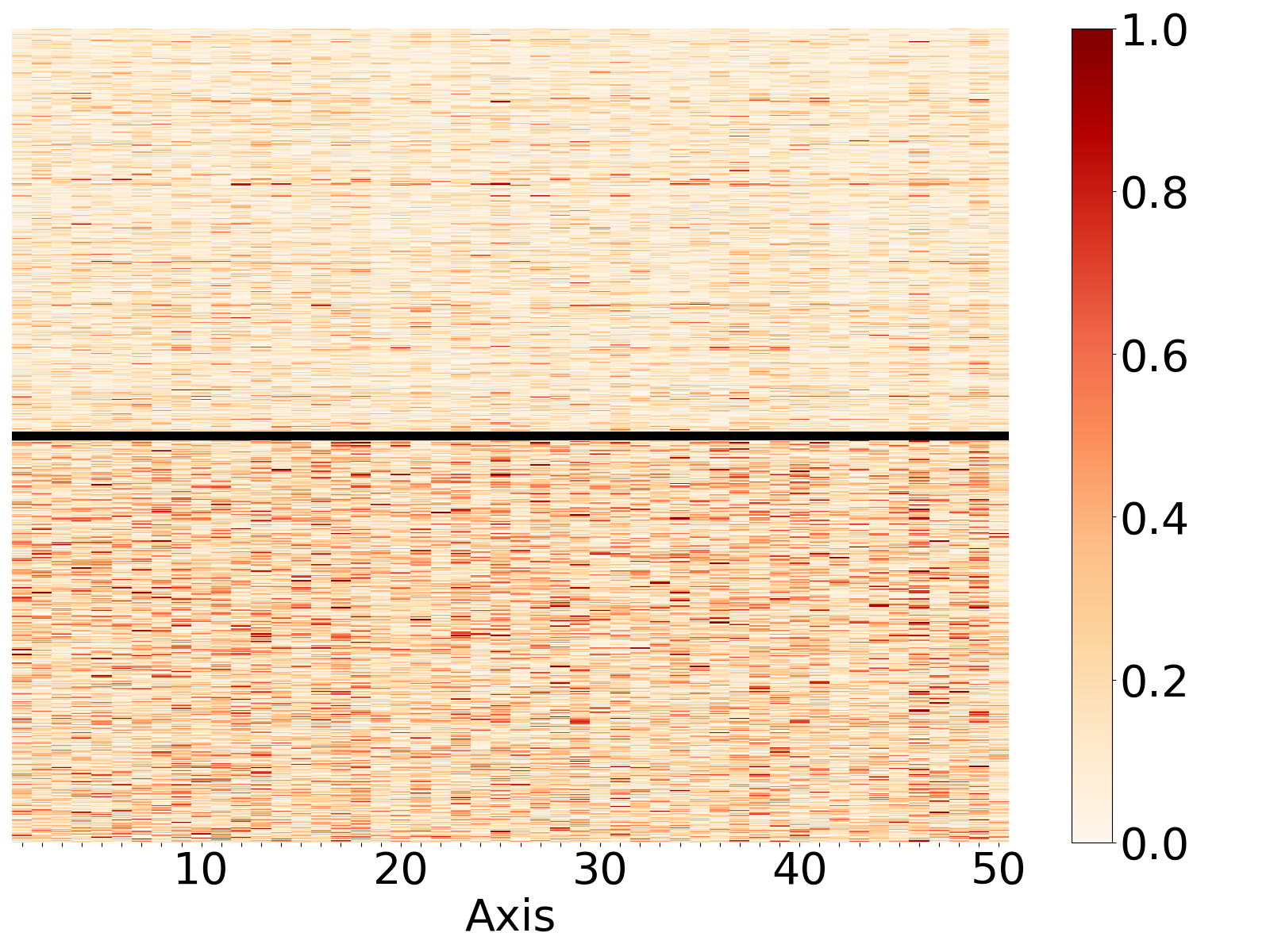}
        \subcaption{Fine-tuned \ac{SCWE}, Raw}
        \label{fig:wic_mclwic_fr_instances_raw_finetuned}
    \end{minipage}
    \begin{minipage}[b]{0.65\columnwidth}
        \centering
        \includegraphics[width=0.85\columnwidth]{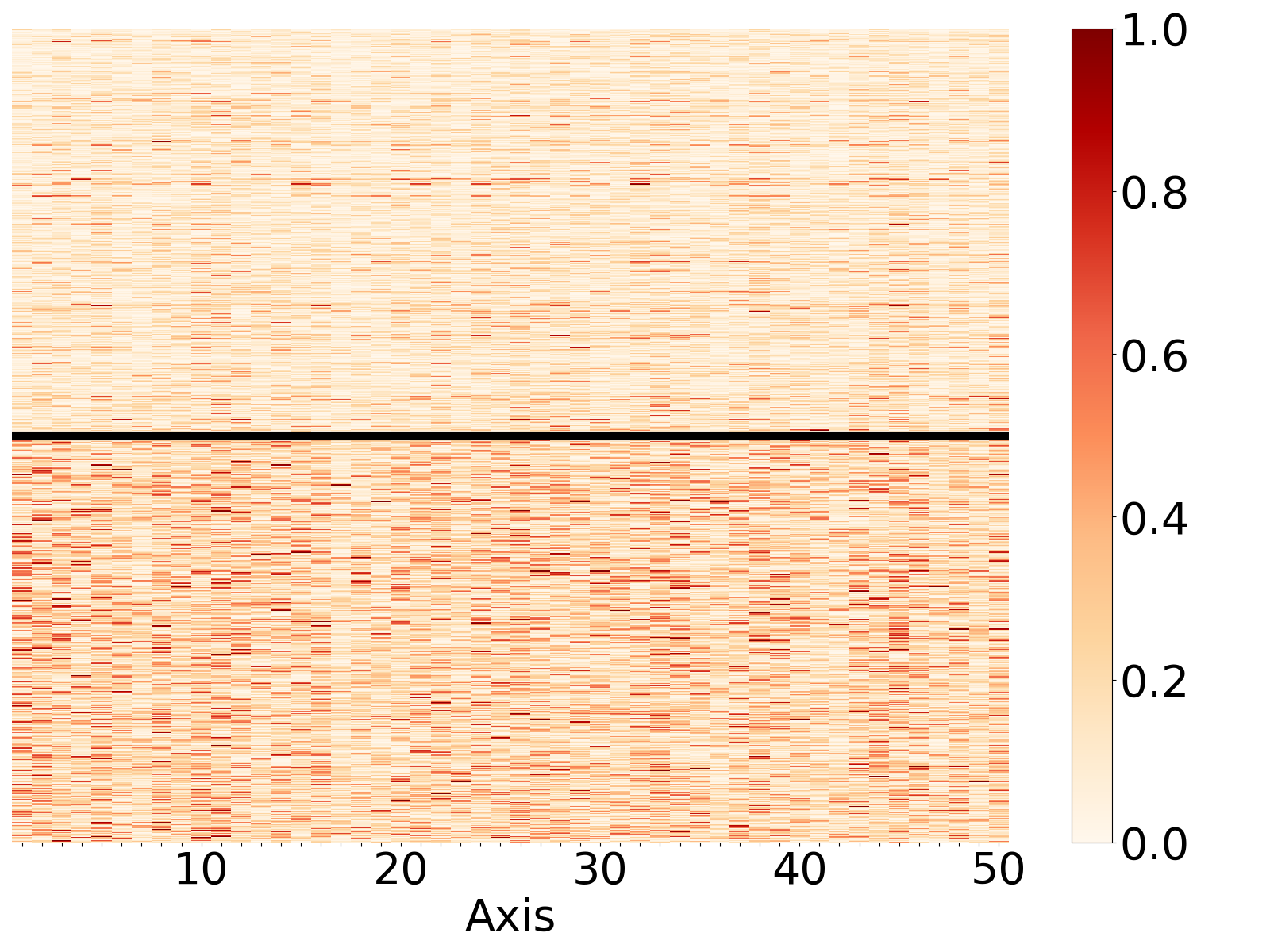}
        \subcaption{Fine-tuned \ac{SCWE}, PCA}
        \label{fig:wic_mclwic_fr_instances_pca_finetuned}
    \end{minipage}
    \begin{minipage}[b]{0.65\columnwidth}
        \centering
        \includegraphics[width=0.85\columnwidth]{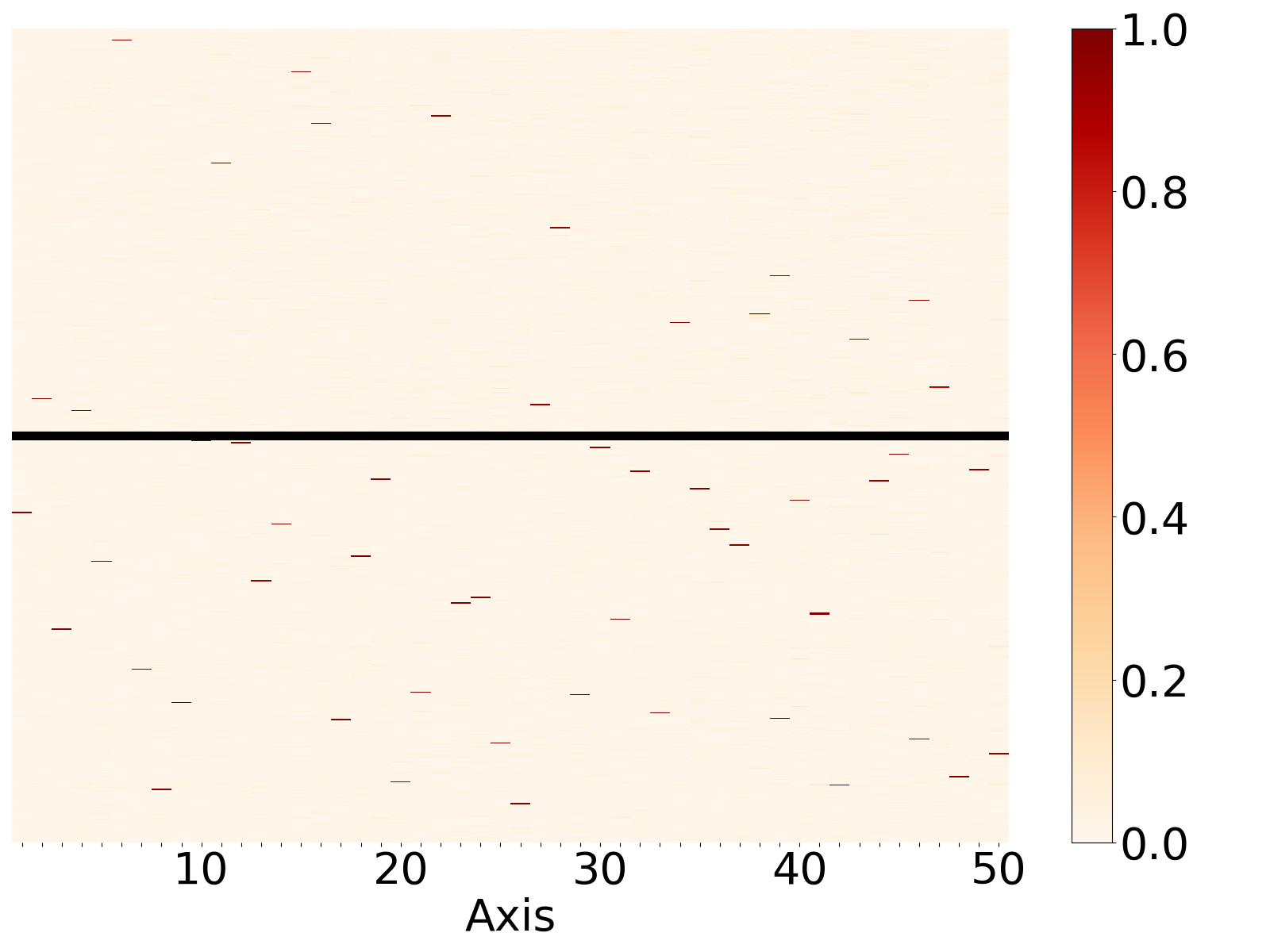}
        \subcaption{Fine-tuned \ac{SCWE}, ICA}
        \label{fig:wic_mclwic_fr_instances_ica_finetuned}
    \end{minipage}
    \caption{Visualisation of the top-50 dimensions of pre-trained \acp{CWE} (XLM-RoBERTa) and \acp{SCWE} (XL-LEXEME) for each instance in MCLWiC (French) dataset, where the difference of vectors is calculated for (a/d) \textbf{Raw} vectors, (b/e) \ac{PCA}-transformed axes, and (c/f) \ac{ICA}-transformed axes. In each figure, the upper/lower half uses instances for the True/False labels.}
    \label{fig:wic_instance_mclwic_fr}
\end{figure*}

\begin{figure*}[t]
    \centering
    \begin{minipage}[b]{0.65\columnwidth}
        \centering
        \includegraphics[width=0.85\columnwidth]{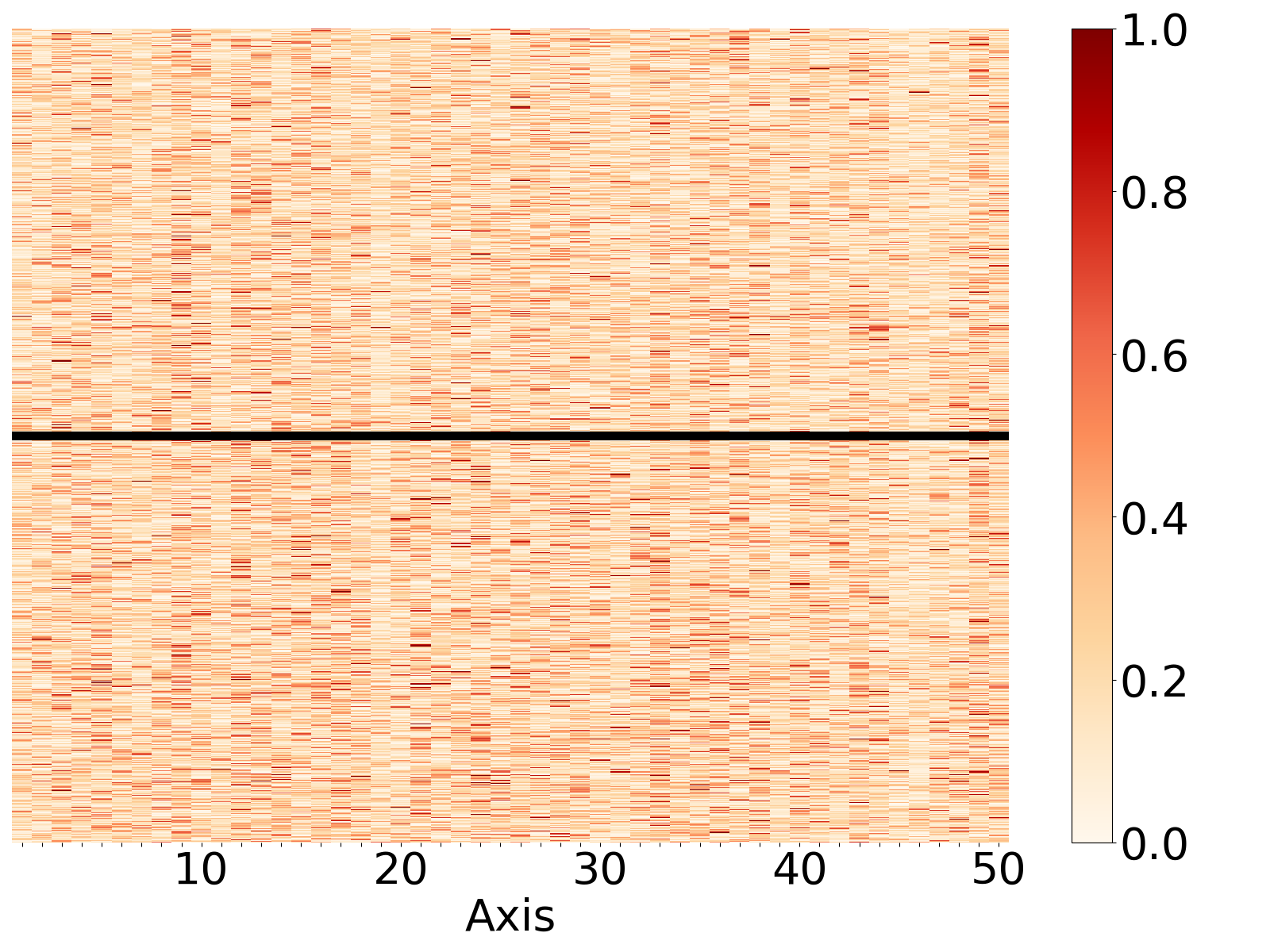}
        \subcaption{Pre-trained \ac{CWE}, Raw}
        \label{fig:wic_mclwic_ru_instances_raw_pretrained}
    \end{minipage}
    \begin{minipage}[b]{0.65\columnwidth}
        \centering
        \includegraphics[width=0.85\columnwidth]{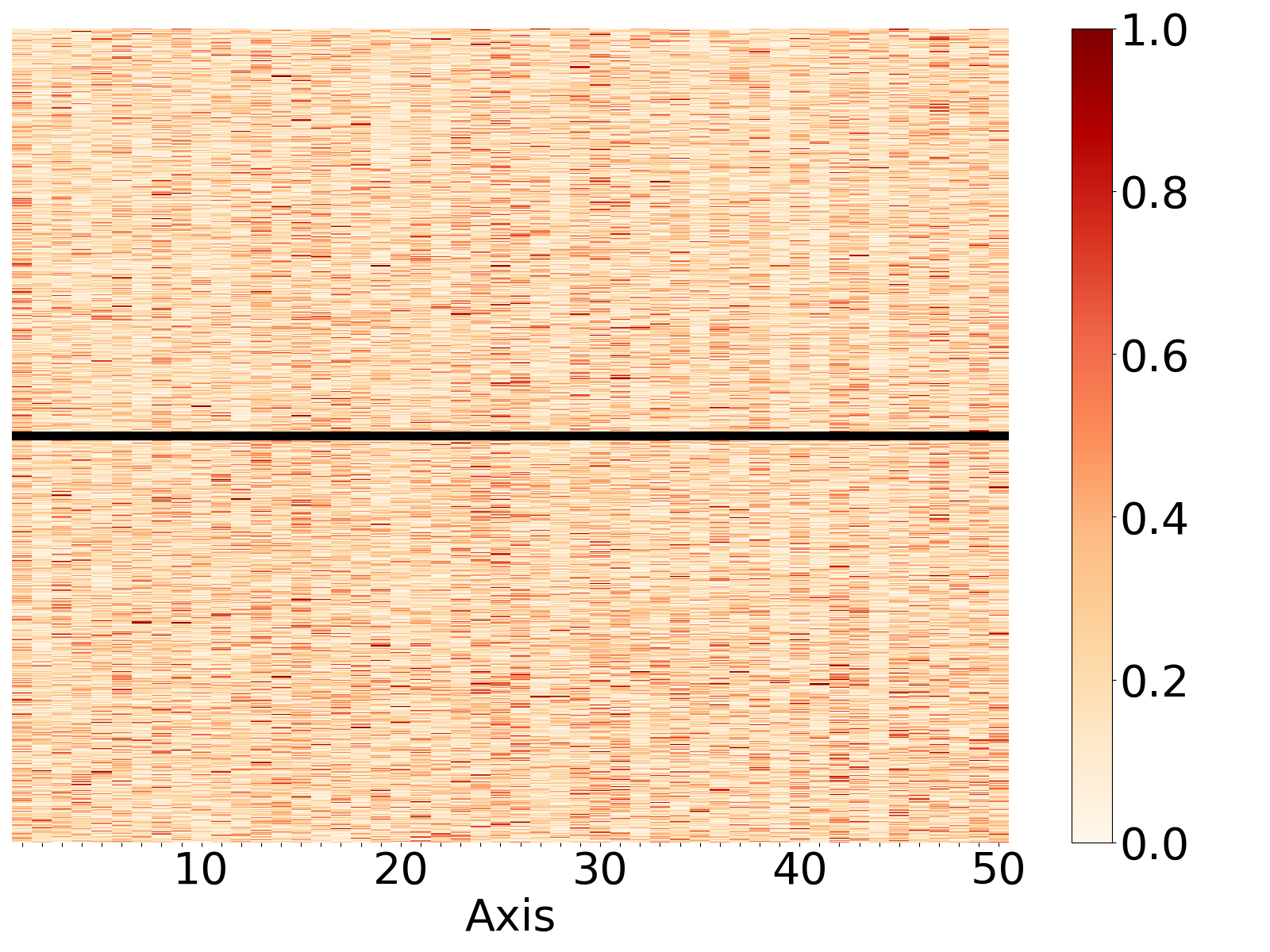}
        \subcaption{Pre-trained \ac{CWE}, PCA}
        \label{fig:wic_mclwic_ru_instances_pca_pretrained}
    \end{minipage}
    \begin{minipage}[b]{0.65\columnwidth}
        \centering
        \includegraphics[width=0.85\columnwidth]{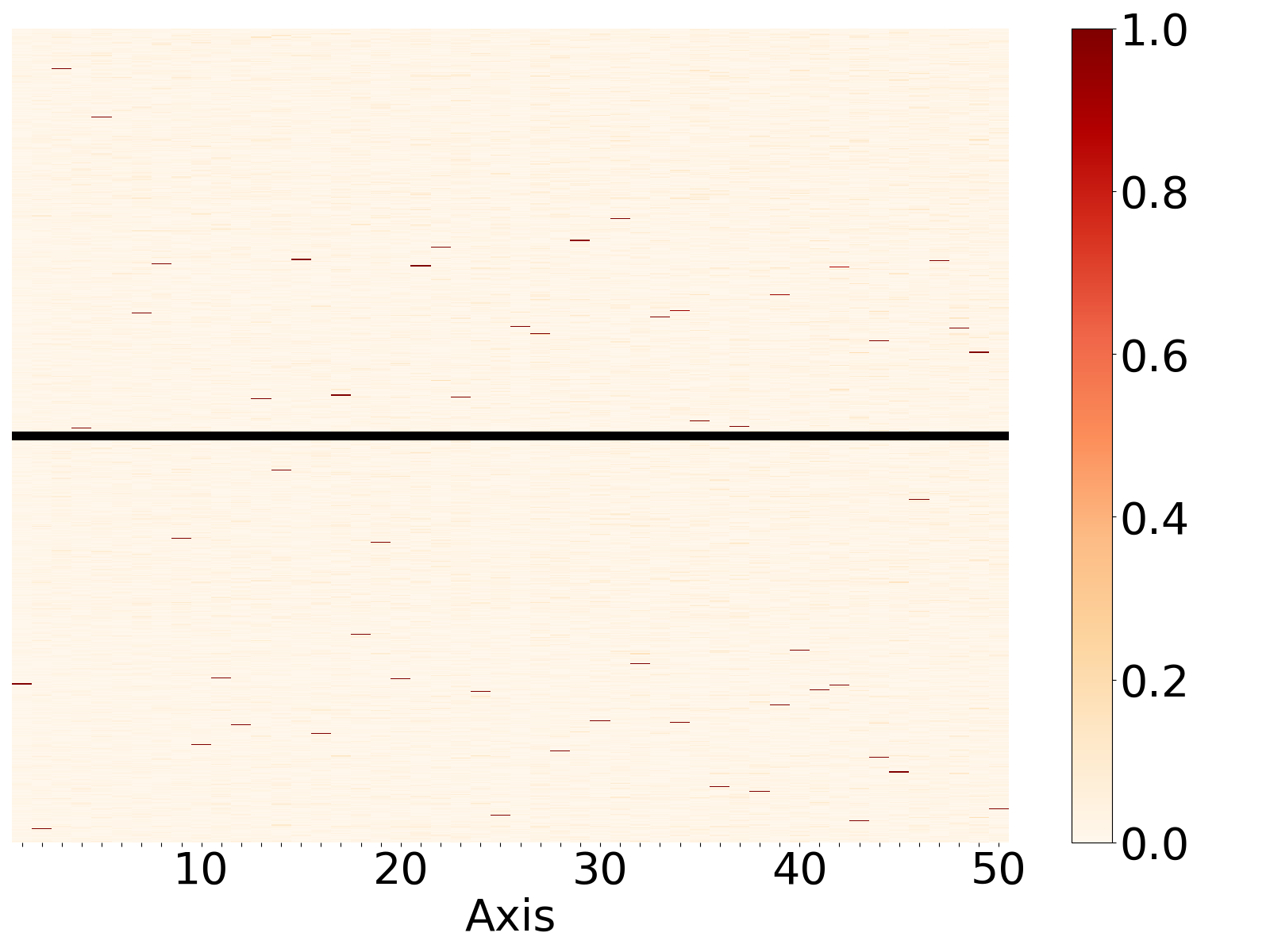}
        \subcaption{Pre-trained \ac{CWE}, ICA}
        \label{fig:wic_mclwic_ru_instances_ica_pretrained}
    \end{minipage} \\
    \begin{minipage}[b]{0.65\columnwidth}
        \centering
        \includegraphics[width=0.85\columnwidth]{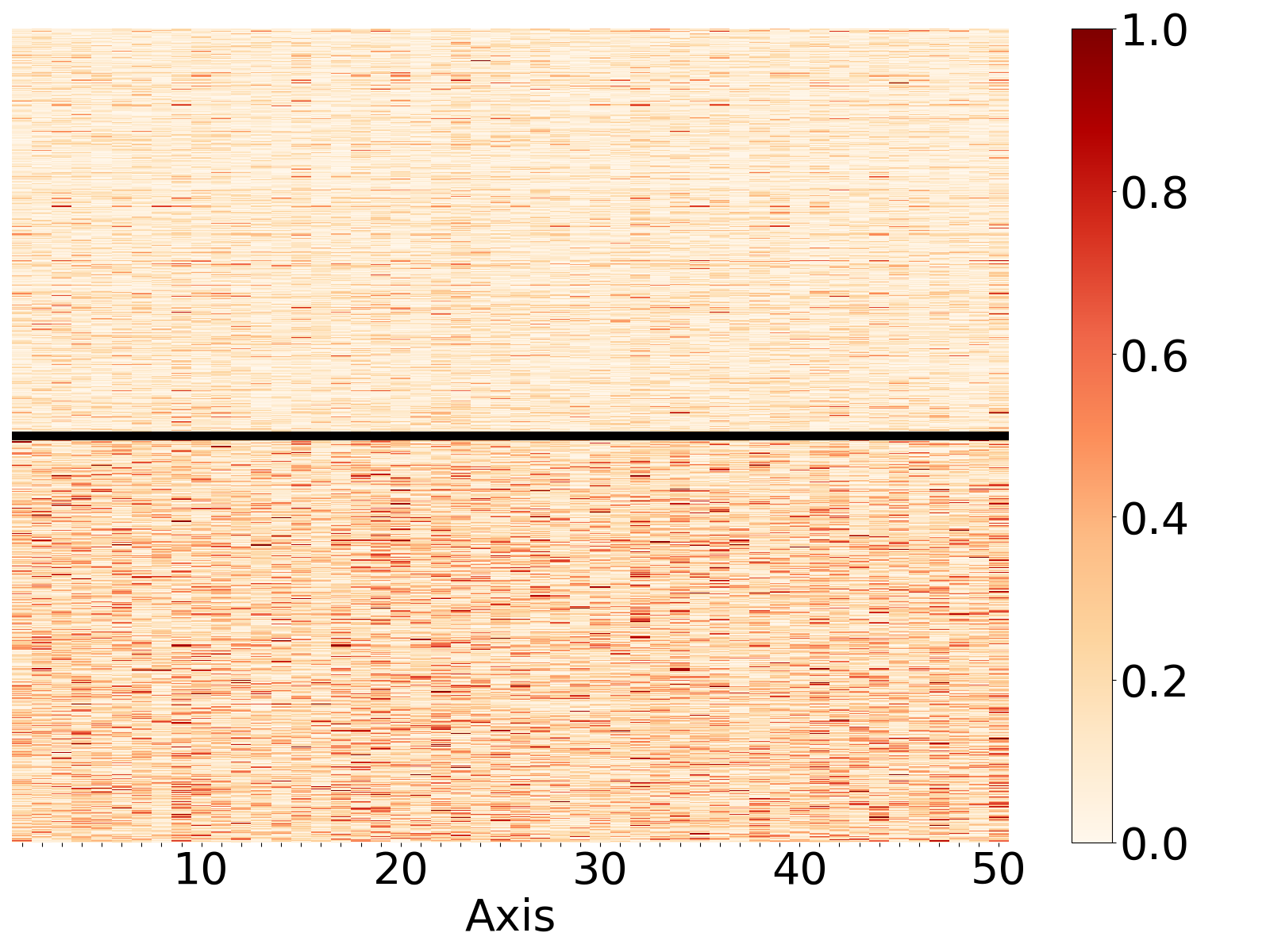}
        \subcaption{Fine-tuned \ac{SCWE}, Raw}
        \label{fig:wic_mclwic_ru_instances_raw_finetuned}
    \end{minipage}
    \begin{minipage}[b]{0.65\columnwidth}
        \centering
        \includegraphics[width=0.85\columnwidth]{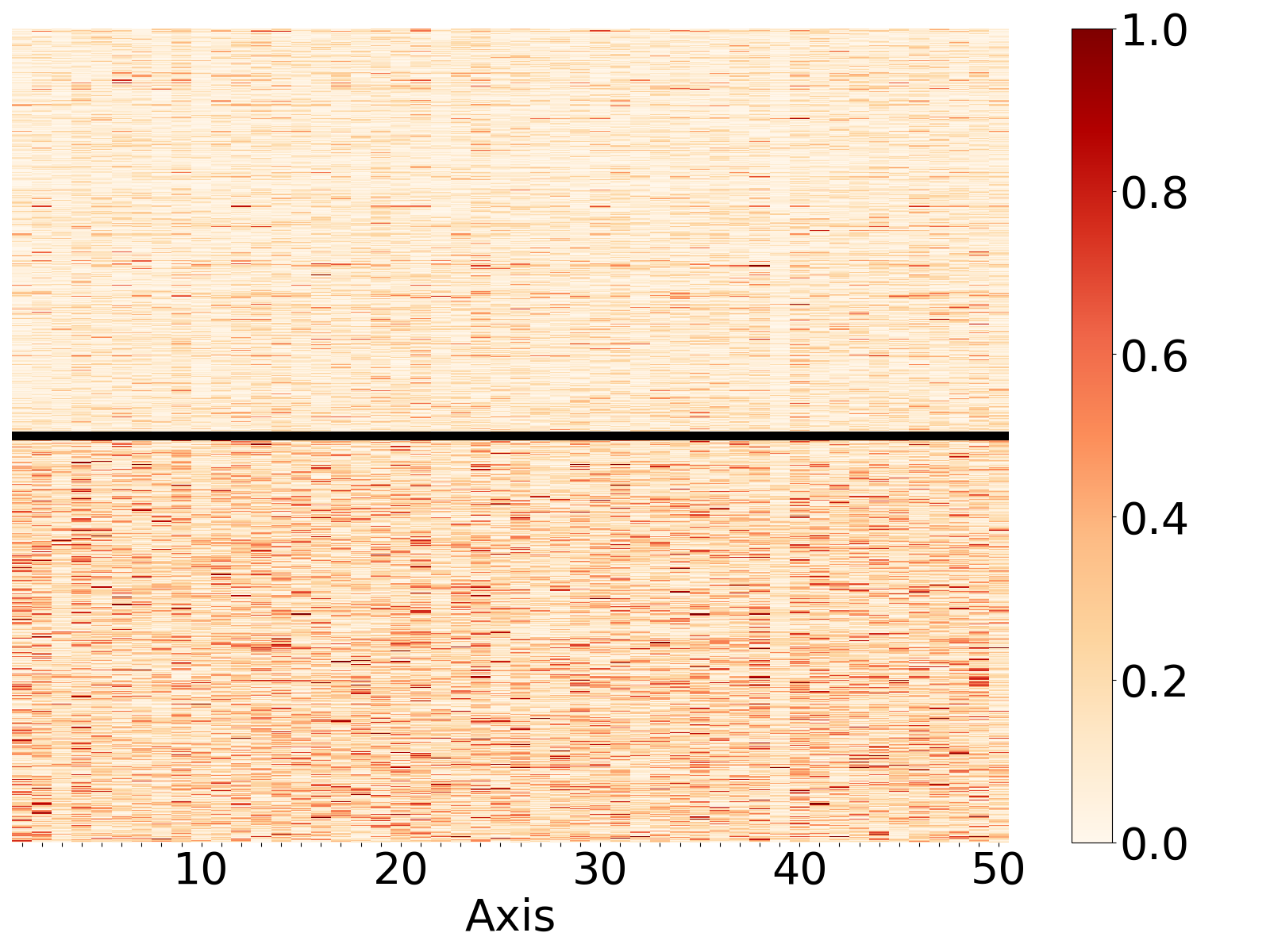}
        \subcaption{Fine-tuned \ac{SCWE}, PCA}
        \label{fig:wic_mclwic_ru_instances_pca_finetuned}
    \end{minipage}
    \begin{minipage}[b]{0.65\columnwidth}
        \centering
        \includegraphics[width=0.85\columnwidth]{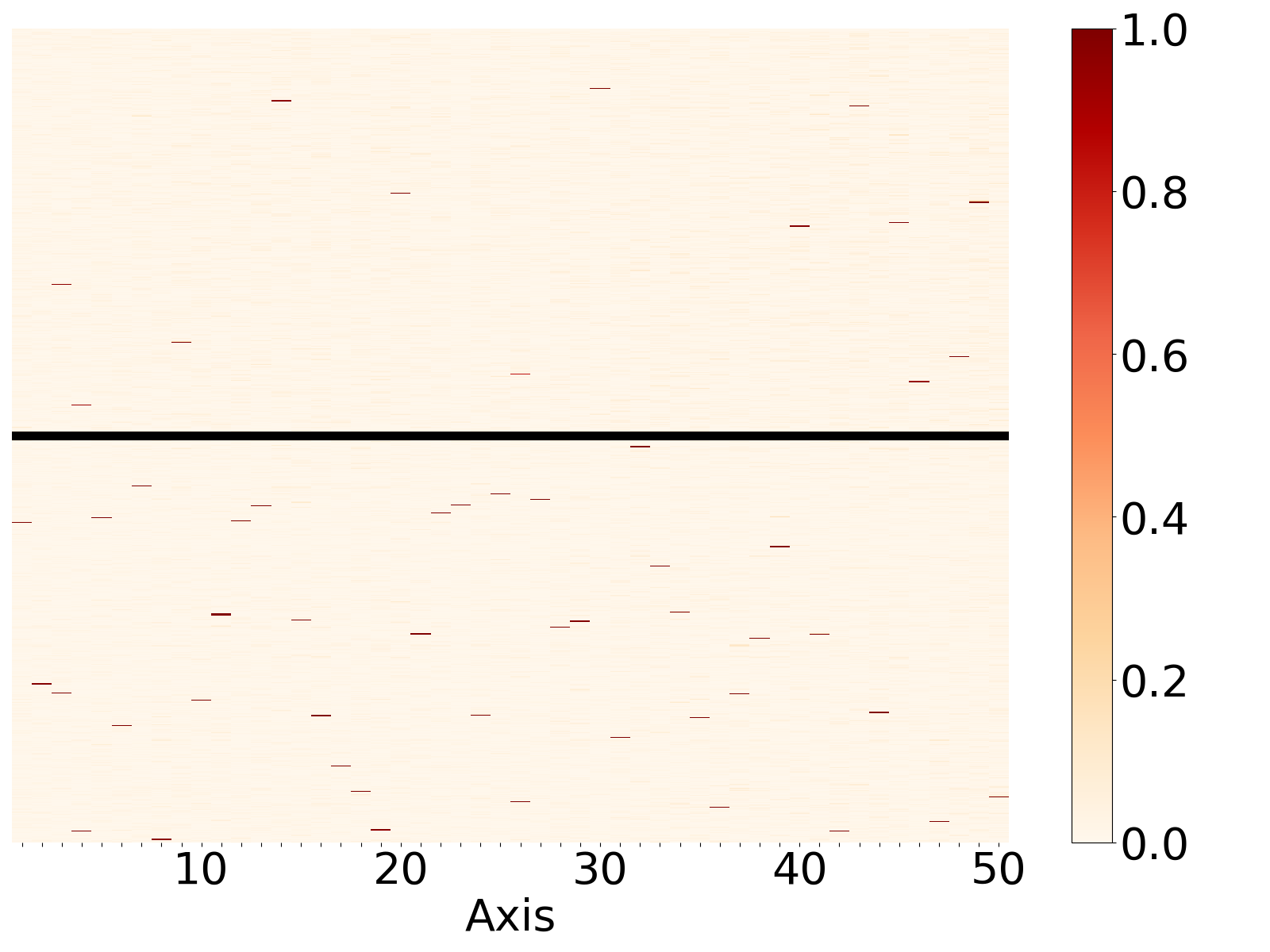}
        \subcaption{Fine-tuned \ac{SCWE}, ICA}
        \label{fig:wic_mclwic_ru_instances_ica_finetuned}
    \end{minipage}
    \caption{Visualisation of the top-50 dimensions of pre-trained \acp{CWE} (XLM-RoBERTa) and \acp{SCWE} (XL-LEXEME) for each instance in MCLWiC (Russian) dataset, where the difference of vectors is calculated for (a/d) \textbf{Raw} vectors, (b/e) \ac{PCA}-transformed axes, and (c/f) \ac{ICA}-transformed axes. In each figure, the upper/lower half uses instances for the True/False labels.}
    \label{fig:wic_instance_mclwic_ru}
\end{figure*}

\begin{figure*}[t]
    \centering
    \begin{minipage}[b]{0.65\columnwidth}
        \centering
        \includegraphics[width=0.85\columnwidth]{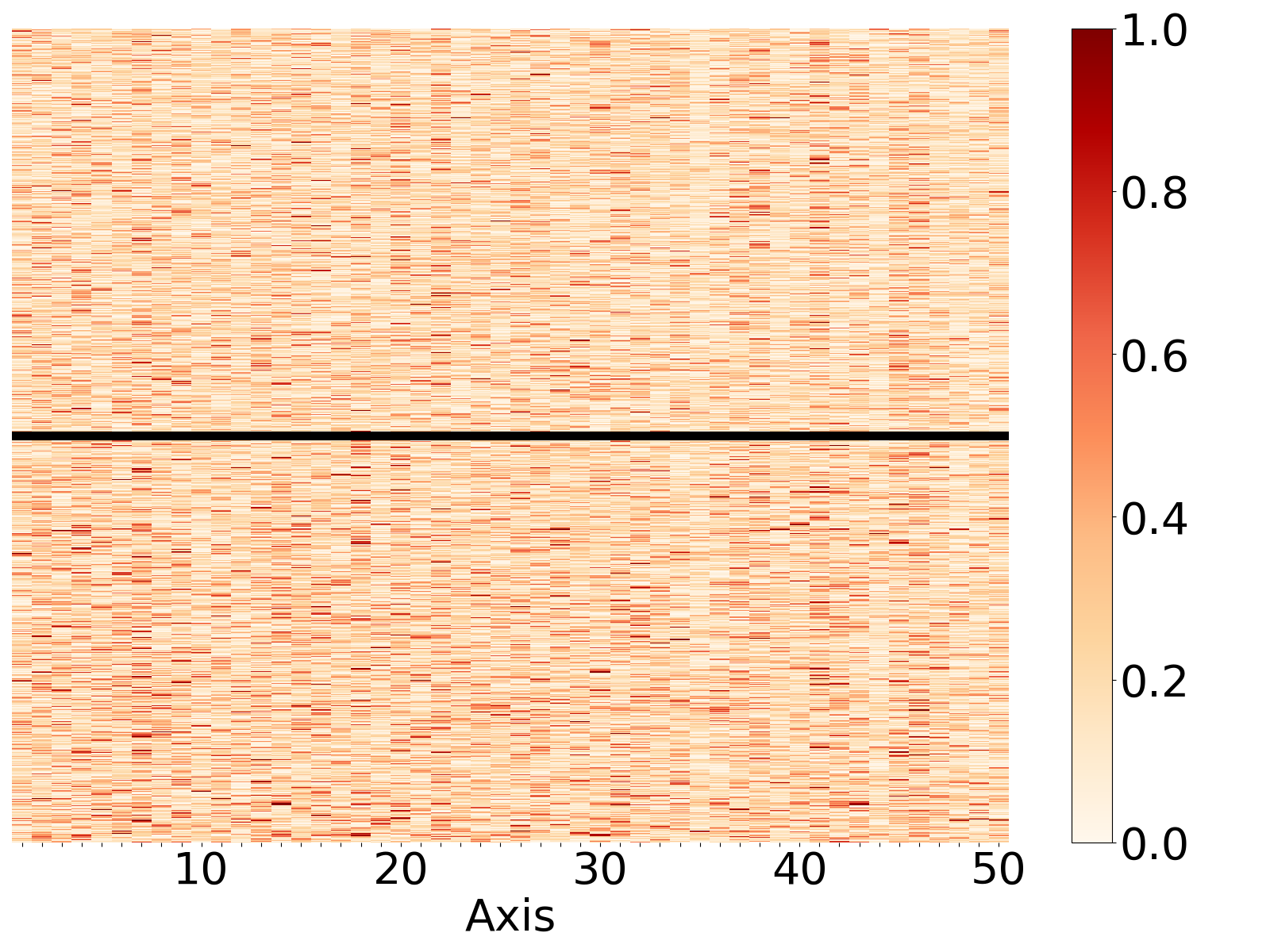}
        \subcaption{Pre-trained \ac{CWE}, Raw}
        \label{fig:wic_mclwic_zh_instances_raw_pretrained}
    \end{minipage}
    \begin{minipage}[b]{0.65\columnwidth}
        \centering
        \includegraphics[width=0.85\columnwidth]{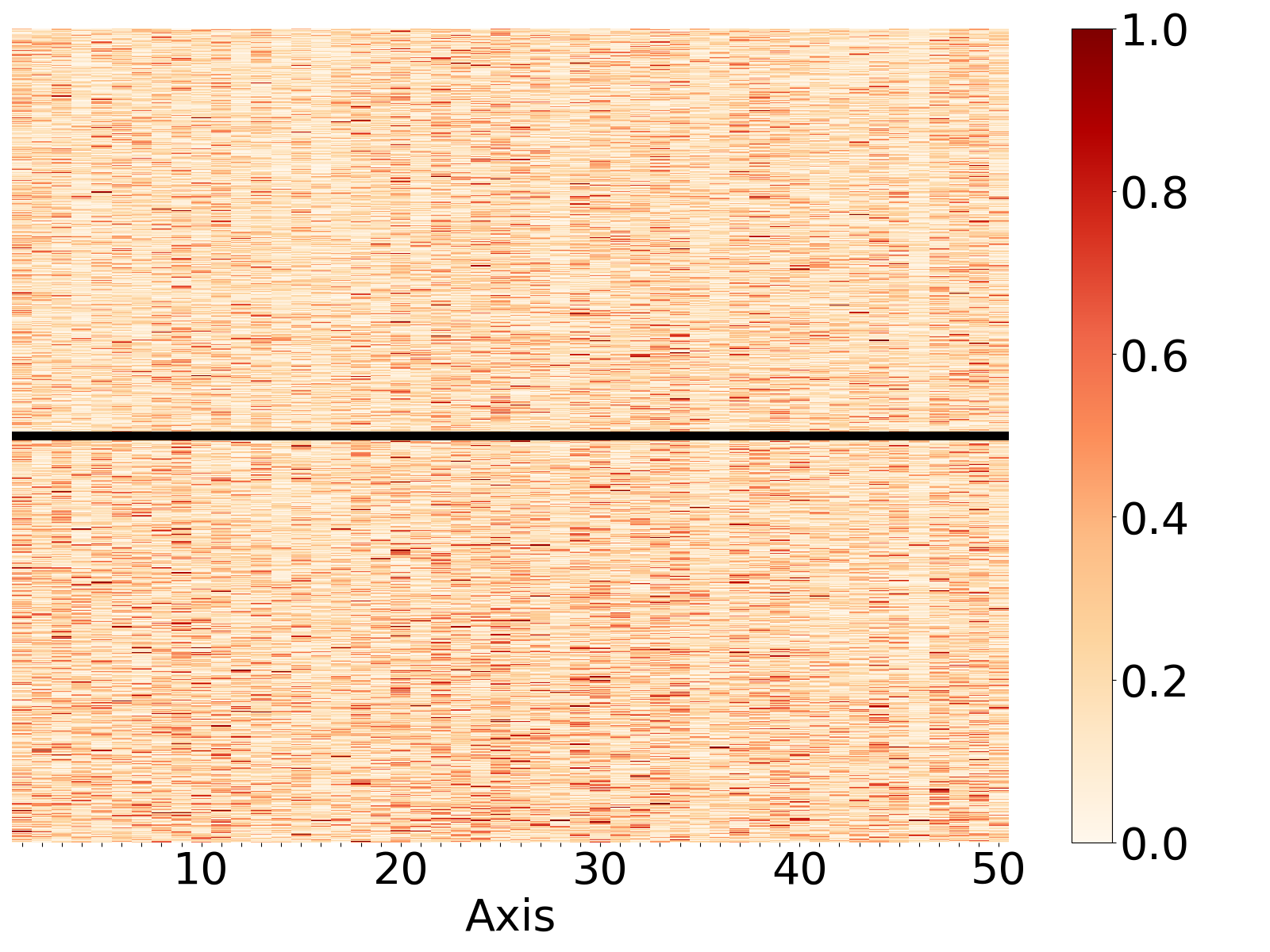}
        \subcaption{Pre-trained \ac{CWE}, PCA}
        \label{fig:wic_mclwic_zh_instances_pca_pretrained}
    \end{minipage}
    \begin{minipage}[b]{0.65\columnwidth}
        \centering
        \includegraphics[width=0.85\columnwidth]{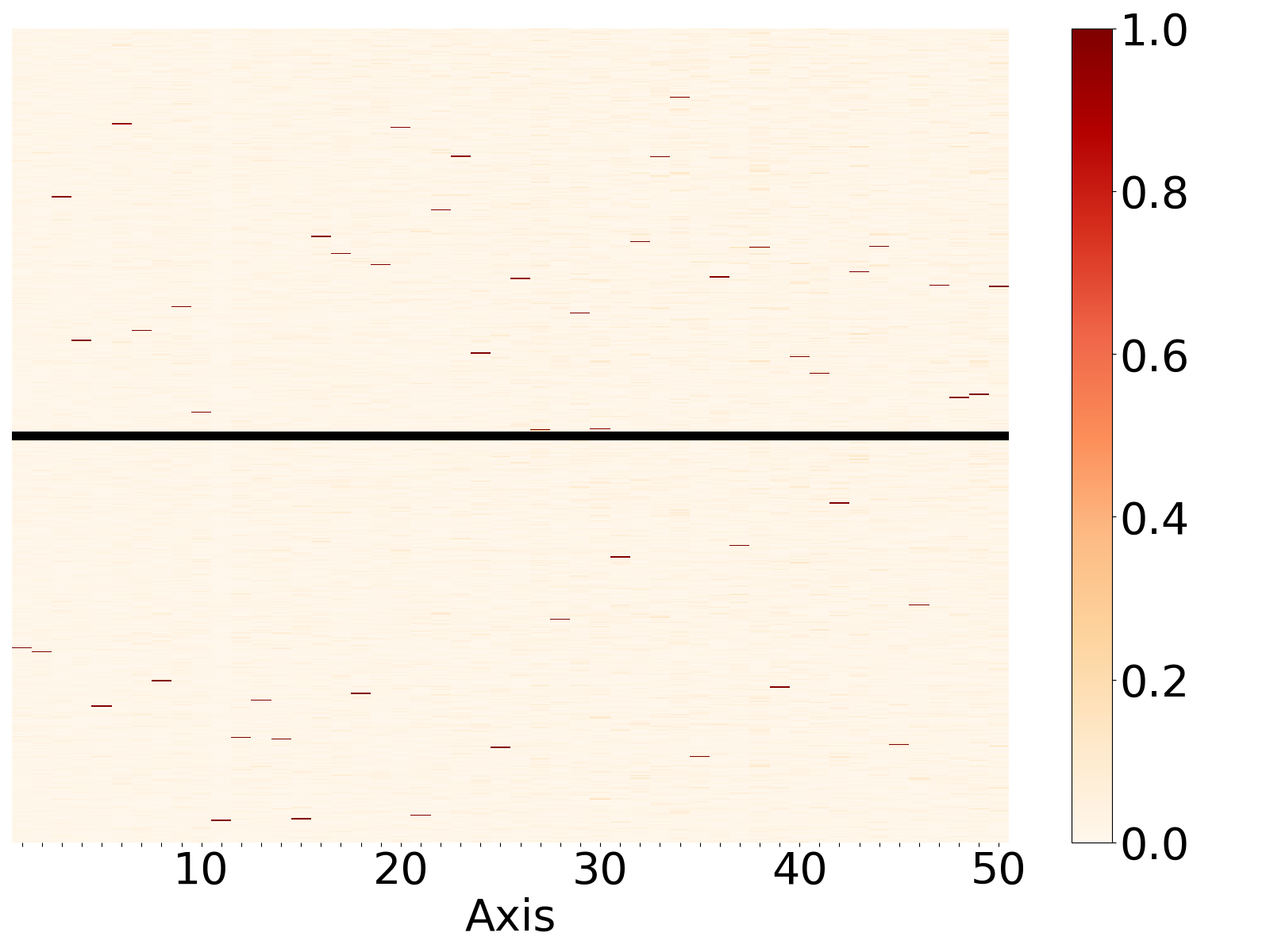}
        \subcaption{Pre-trained \ac{CWE}, ICA}
        \label{fig:wic_mclwic_zh_instances_ica_pretrained}
    \end{minipage} \\
    \begin{minipage}[b]{0.65\columnwidth}
        \centering
        \includegraphics[width=0.85\columnwidth]{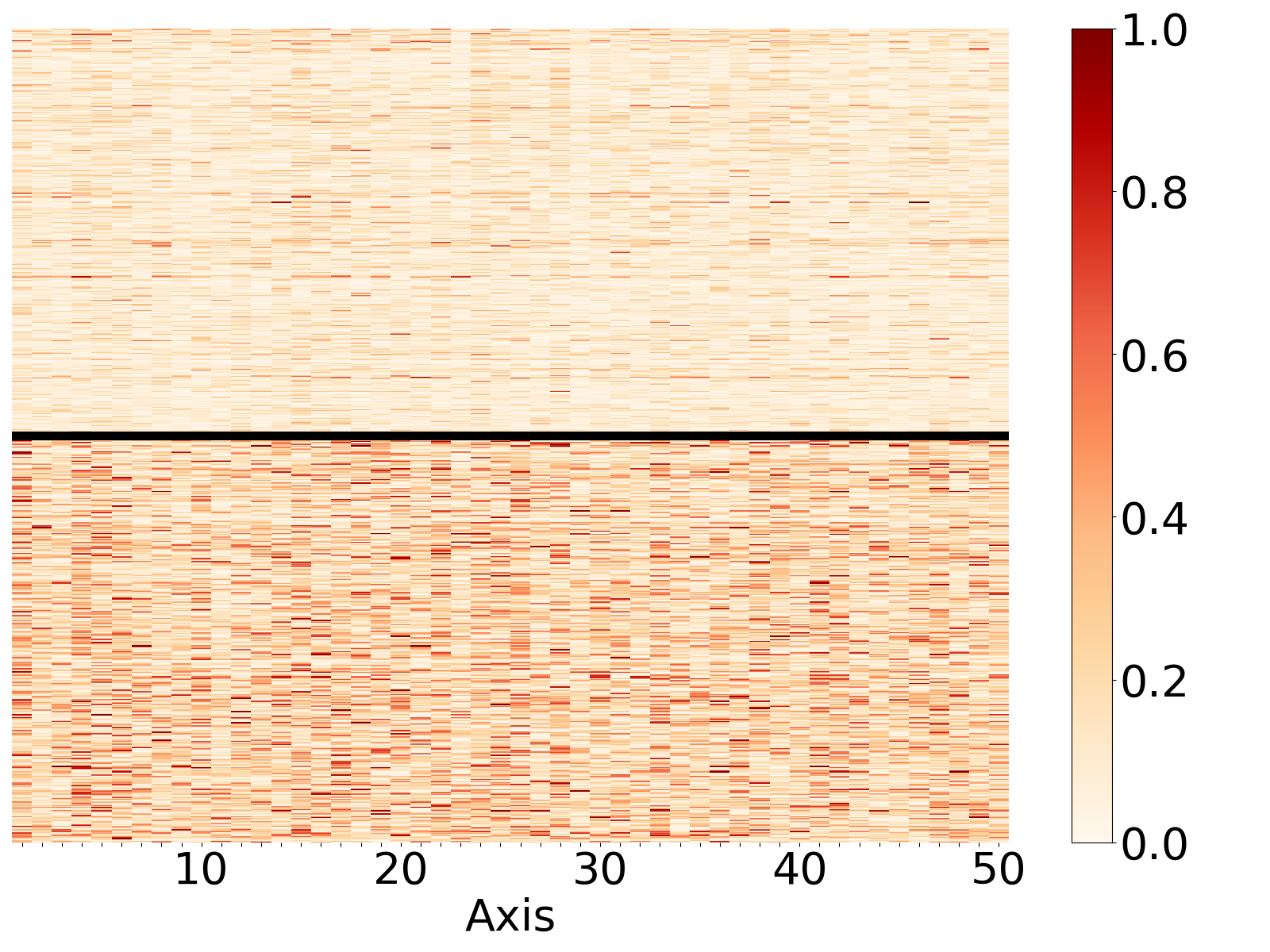}
        \subcaption{Fine-tuned \ac{SCWE}, Raw}
        \label{fig:wic_mclwic_zh_instances_raw_finetuned}
    \end{minipage}
    \begin{minipage}[b]{0.65\columnwidth}
        \centering
        \includegraphics[width=0.85\columnwidth]{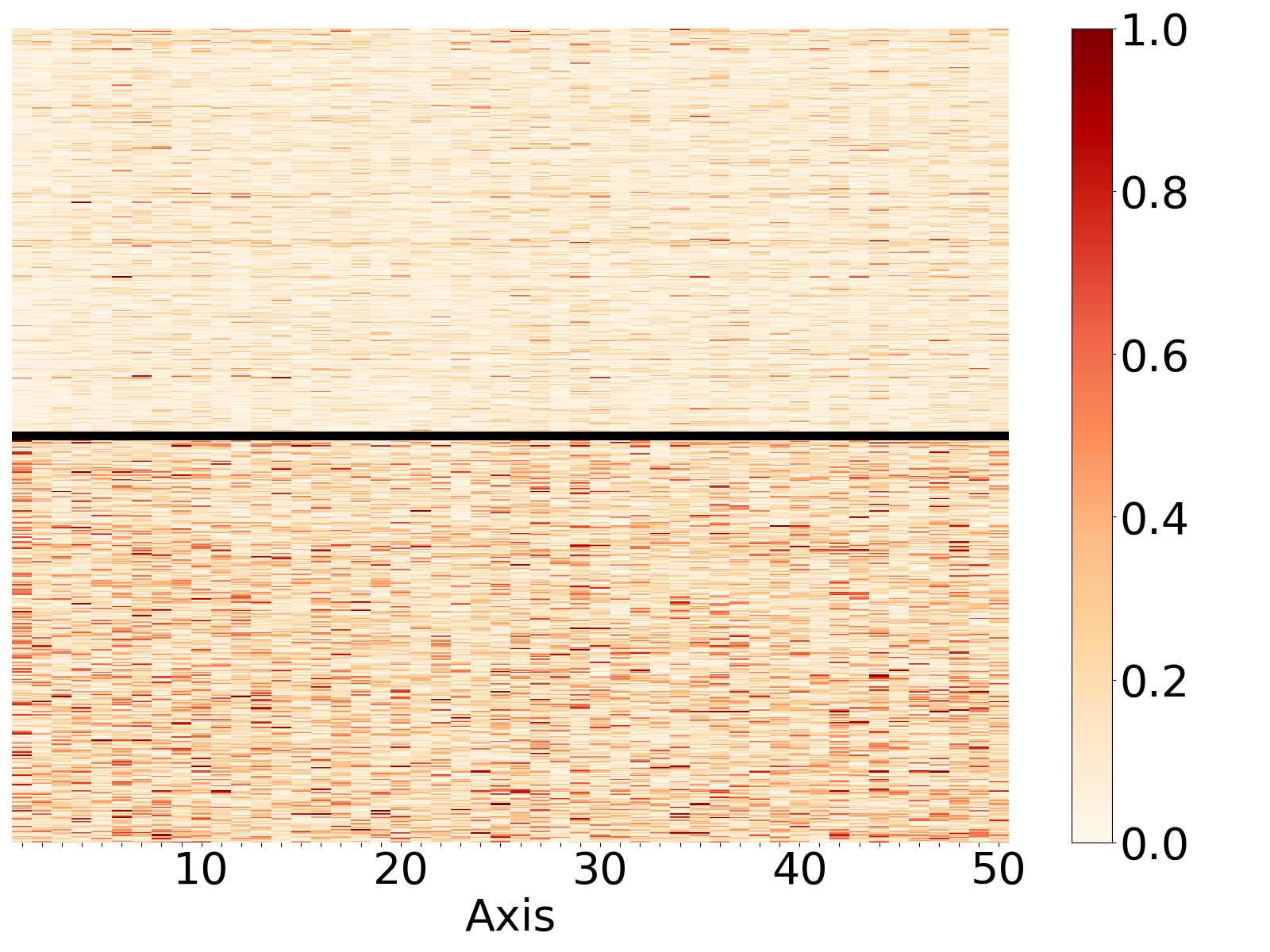}
        \subcaption{Fine-tuned \ac{SCWE}, PCA}
        \label{fig:wic_mclwic_zh_instances_pca_finetuned}
    \end{minipage}
    \begin{minipage}[b]{0.65\columnwidth}
        \centering
        \includegraphics[width=0.85\columnwidth]{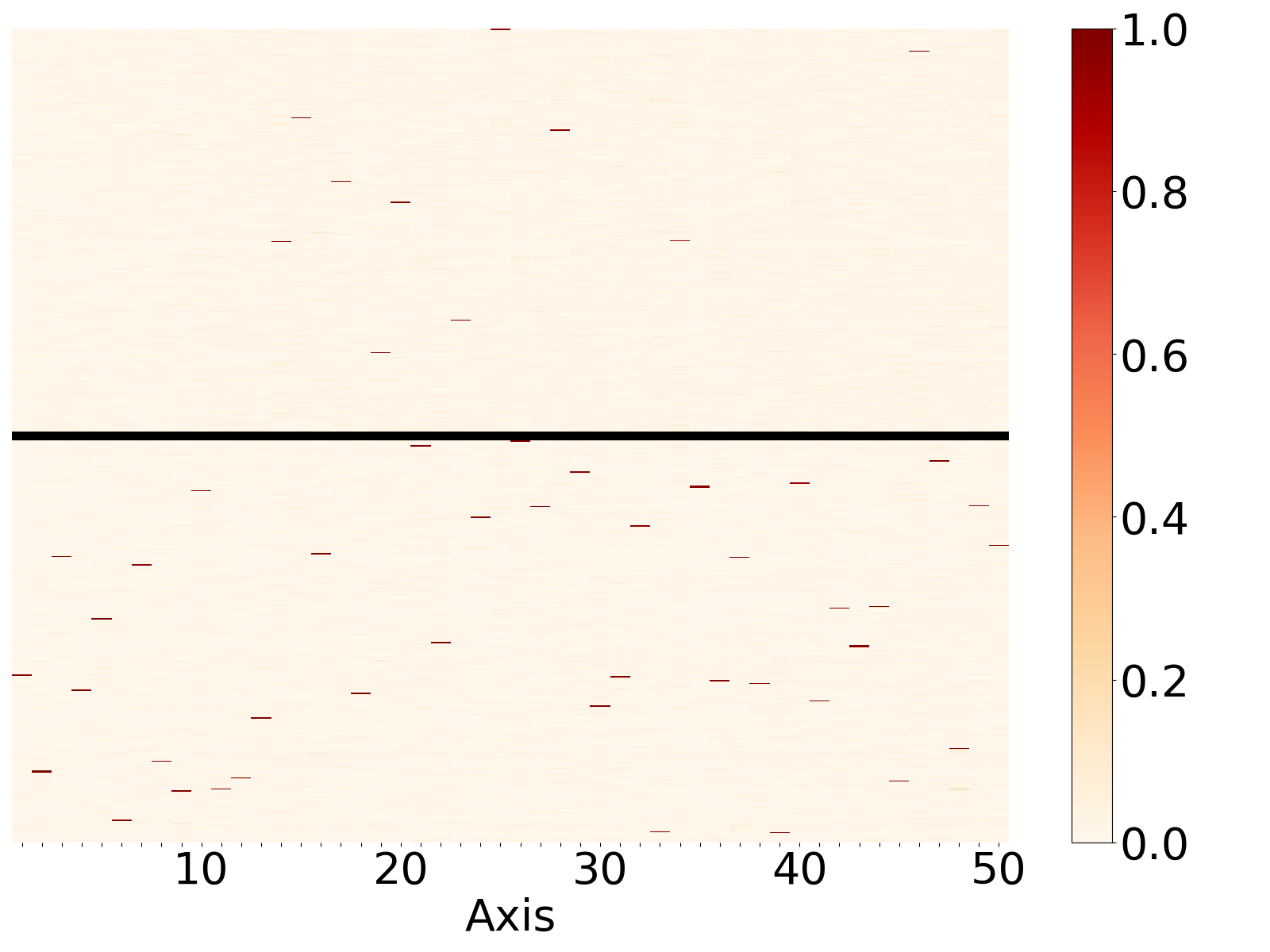}
        \subcaption{Fine-tuned \ac{SCWE}, ICA}
        \label{fig:wic_mclwic_zh_instances_ica_finetuned}
    \end{minipage}
    \caption{Visualisation of the top-50 dimensions of pre-trained \acp{CWE} (XLM-RoBERTa) and \acp{SCWE} (XL-LEXEME) for each instance in MCLWiC (Chinese) dataset, where the difference of vectors is calculated for (a/d) \textbf{Raw} vectors, (b/e) \ac{PCA}-transformed axes, and (c/f) \ac{ICA}-transformed axes. In each figure, the upper/lower half uses instances for the True/False labels.}
    \label{fig:wic_instance_mclwic_zh}
\end{figure*}

\begin{figure*}[t]
    \centering
    \begin{minipage}[b]{0.65\columnwidth}
        \centering
        \includegraphics[width=0.85\columnwidth]{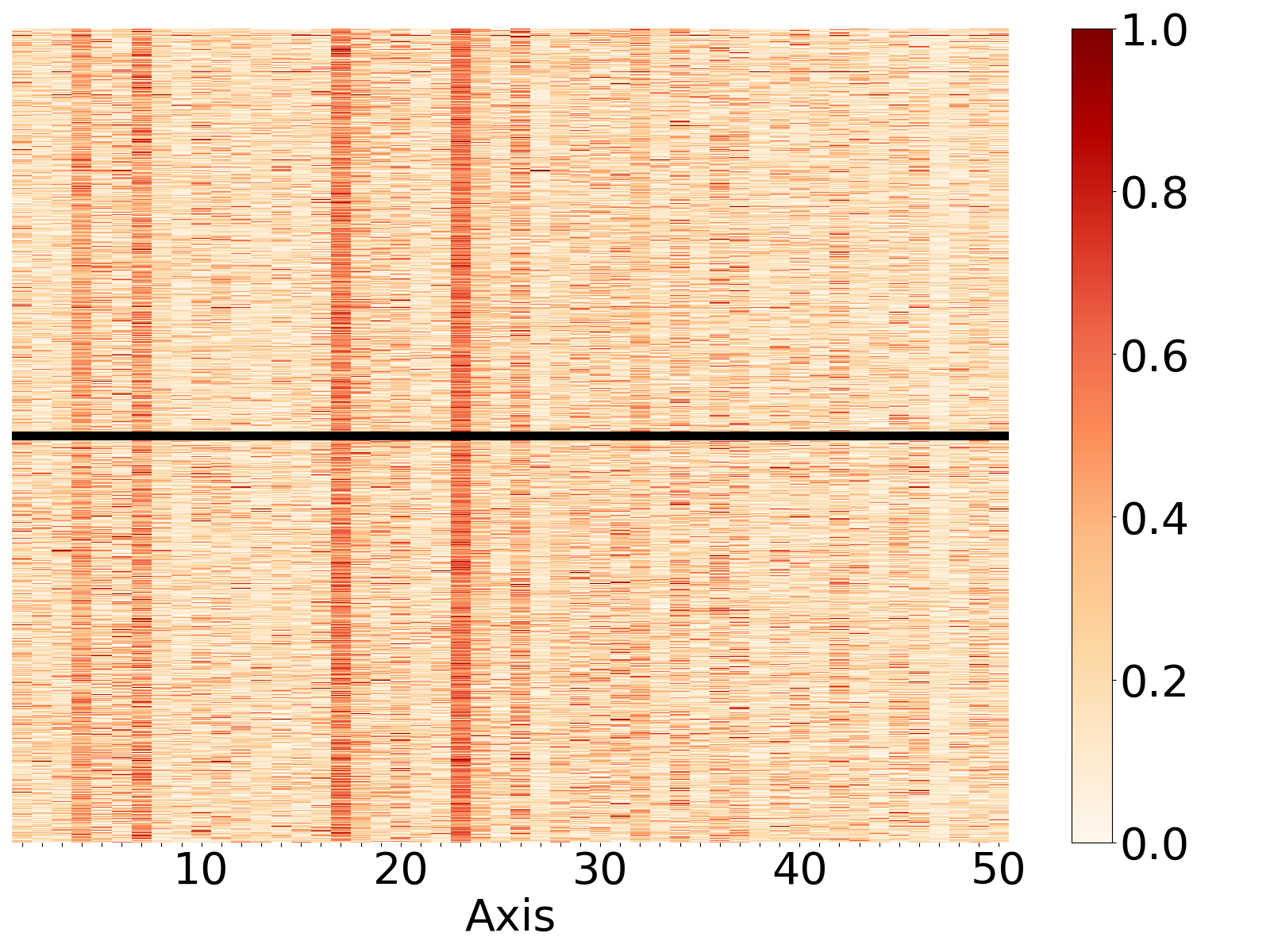}
        \subcaption{Pre-trained \ac{CWE}, Raw}
        \label{fig:wic_am2ico_de_instances_raw_pretrained}
    \end{minipage}
    \begin{minipage}[b]{0.65\columnwidth}
        \centering
        \includegraphics[width=0.85\columnwidth]{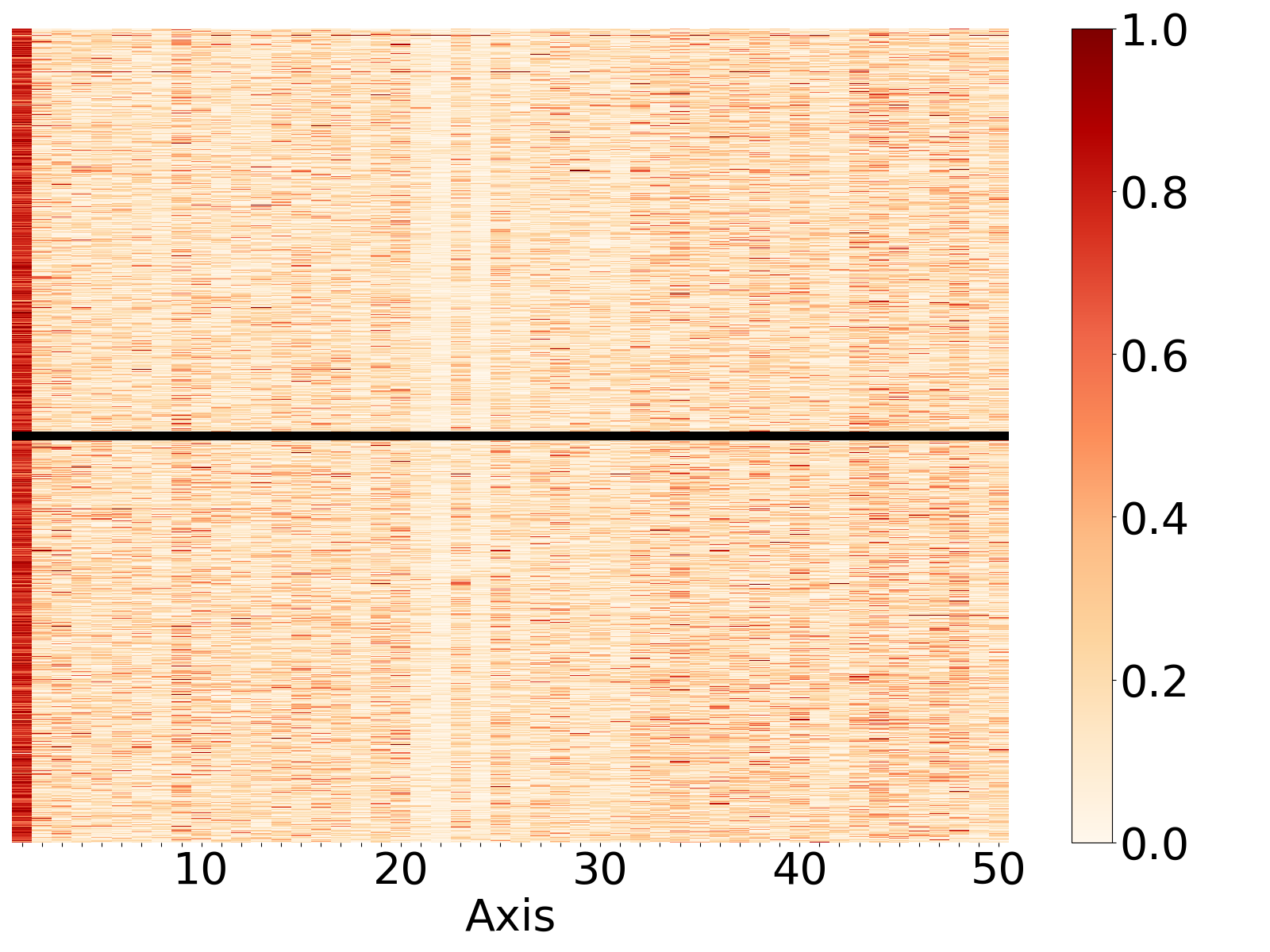}
        \subcaption{Pre-trained \ac{CWE}, PCA}
        \label{fig:wic_am2ico_de_instances_pca_pretrained}
    \end{minipage}
    \begin{minipage}[b]{0.65\columnwidth}
        \centering
        \includegraphics[width=0.85\columnwidth]{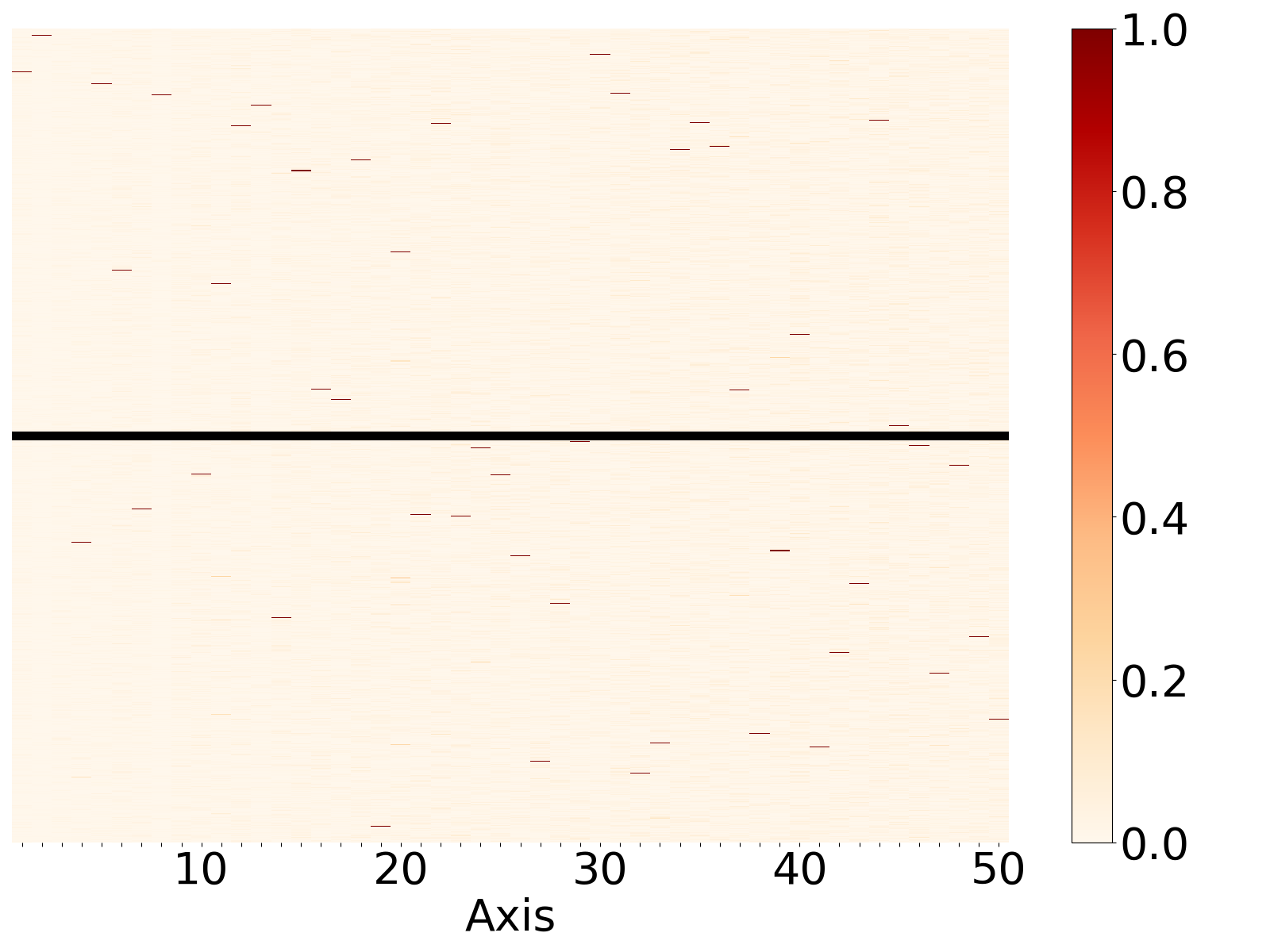}
        \subcaption{Pre-trained \ac{CWE}, ICA}
        \label{fig:wic_am2ico_de_instances_ica_pretrained}
    \end{minipage} \\
    \begin{minipage}[b]{0.65\columnwidth}
        \centering
        \includegraphics[width=0.85\columnwidth]{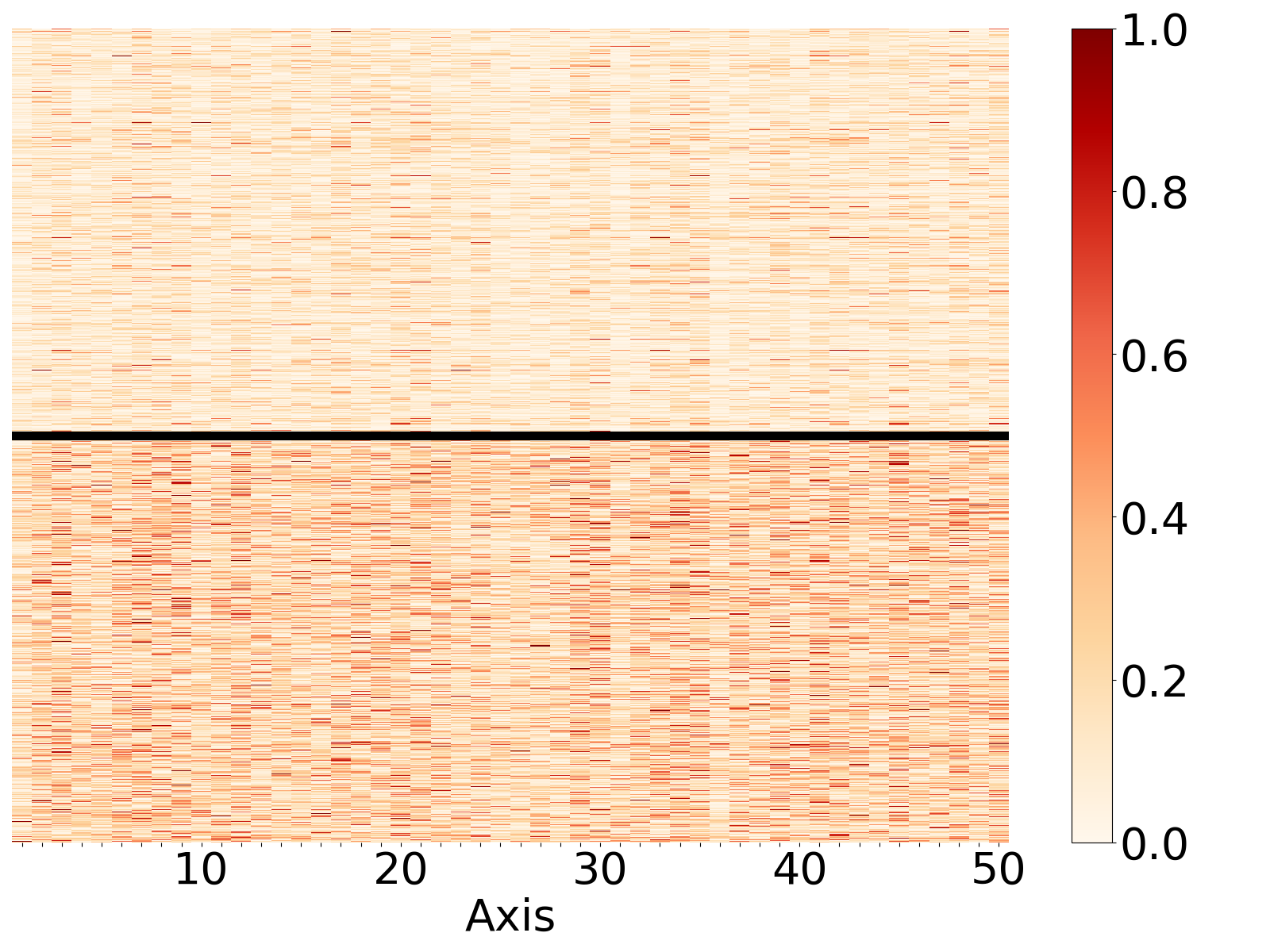}
        \subcaption{Fine-tuned \ac{SCWE}, Raw}
        \label{fig:wic_am2ico_de_instances_raw_finetuned}
    \end{minipage}
    \begin{minipage}[b]{0.65\columnwidth}
        \centering
        \includegraphics[width=0.85\columnwidth]{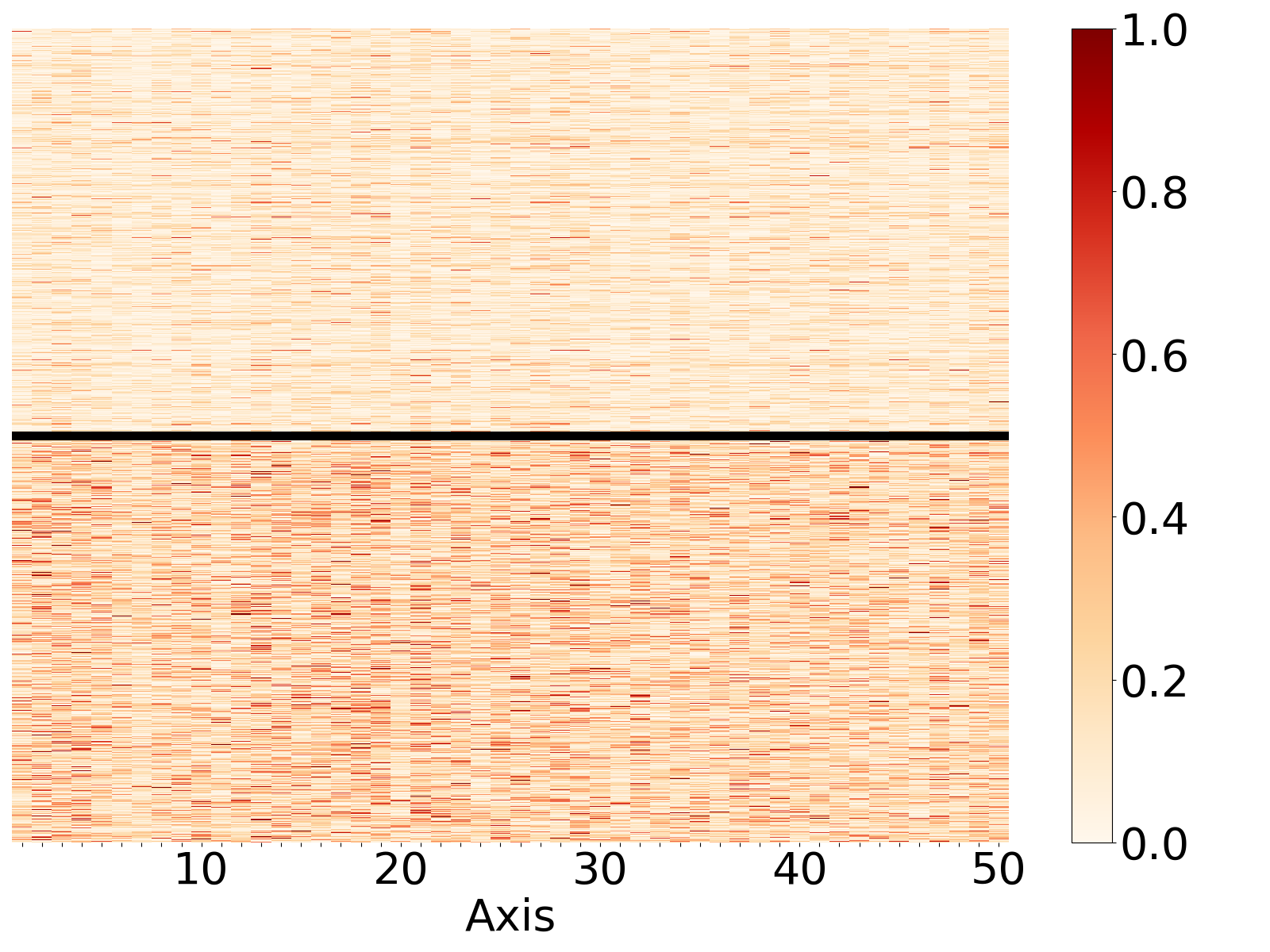}
        \subcaption{Fine-tuned \ac{SCWE}, PCA}
        \label{fig:wic_am2ico_de_instances_pca_finetuned}
    \end{minipage}
    \begin{minipage}[b]{0.65\columnwidth}
        \centering
        \includegraphics[width=0.85\columnwidth]{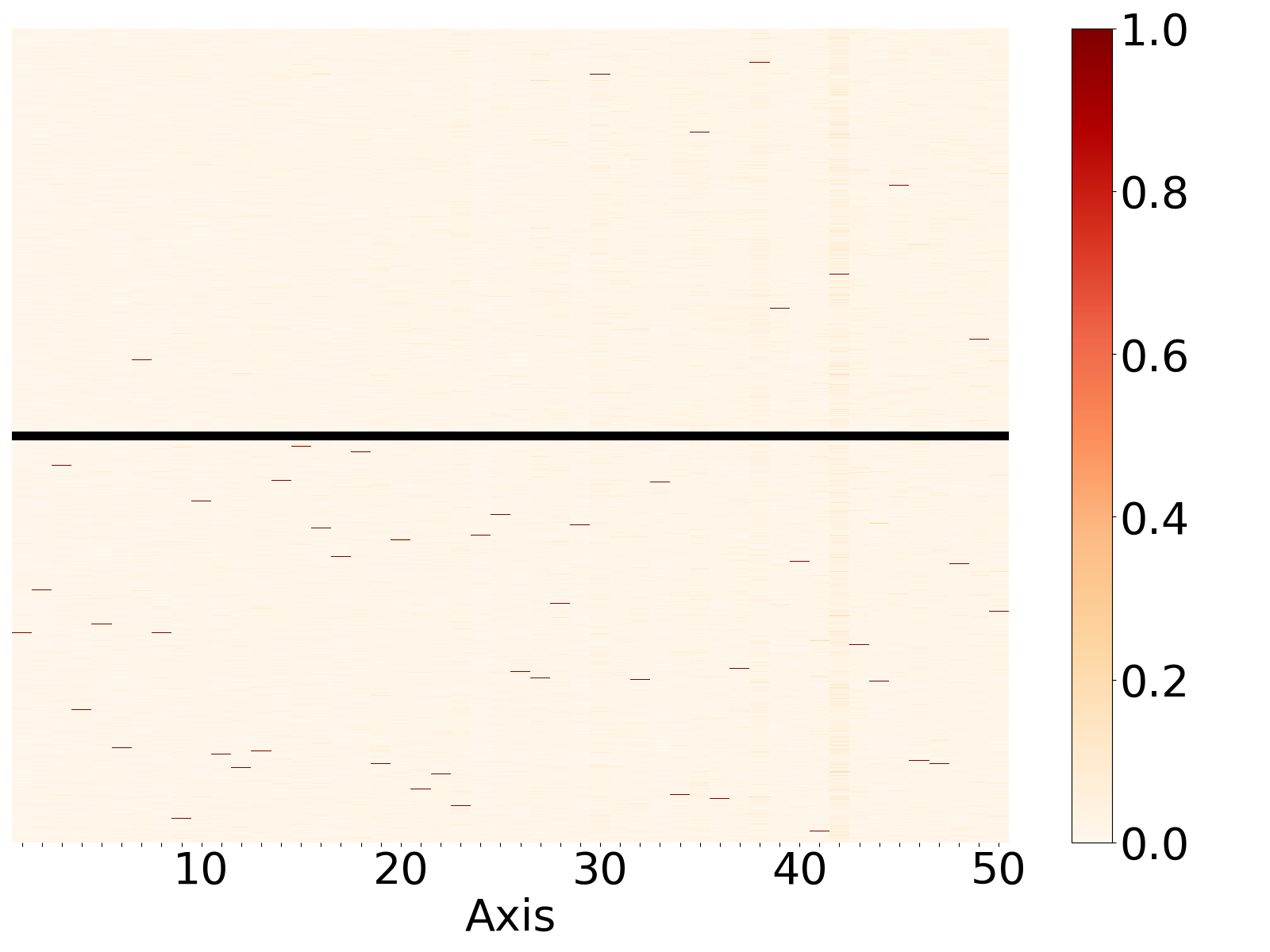}
        \subcaption{Fine-tuned \ac{SCWE}, ICA}
        \label{fig:wic_am2ico_de_instances_ica_finetuned}
    \end{minipage}
    \caption{Visualisation of the top-50 dimensions of pre-trained \acp{CWE} (XLM-RoBERTa) and \acp{SCWE} (XL-LEXEME) for each instance in AM$^2$iCo (German) dataset, where the difference of vectors is calculated for (a/d) \textbf{Raw} vectors, (b/e) \ac{PCA}-transformed axes, and (c/f) \ac{ICA}-transformed axes. In each figure, the upper/lower half uses instances for the True/False labels.}
    \label{fig:wic_instance_am2ico_de}
\end{figure*}

\begin{figure*}[t]
    \centering
    \begin{minipage}[b]{0.65\columnwidth}
        \centering
        \includegraphics[width=0.85\columnwidth]{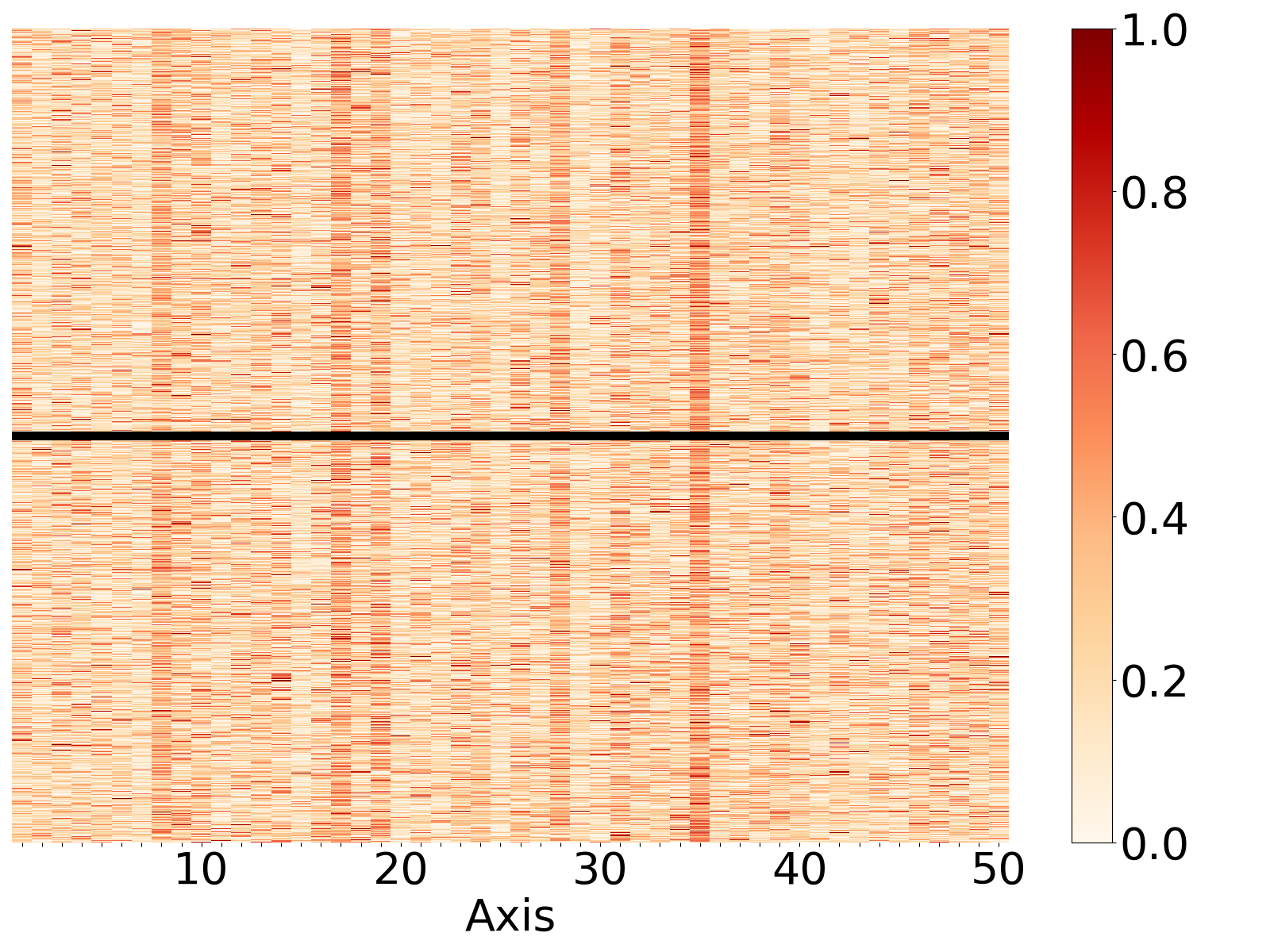}
        \subcaption{Pre-trained \ac{CWE}, Raw}
        \label{fig:wic_am2ico_ru_instances_raw_pretrained}
    \end{minipage}
    \begin{minipage}[b]{0.65\columnwidth}
        \centering
        \includegraphics[width=0.85\columnwidth]{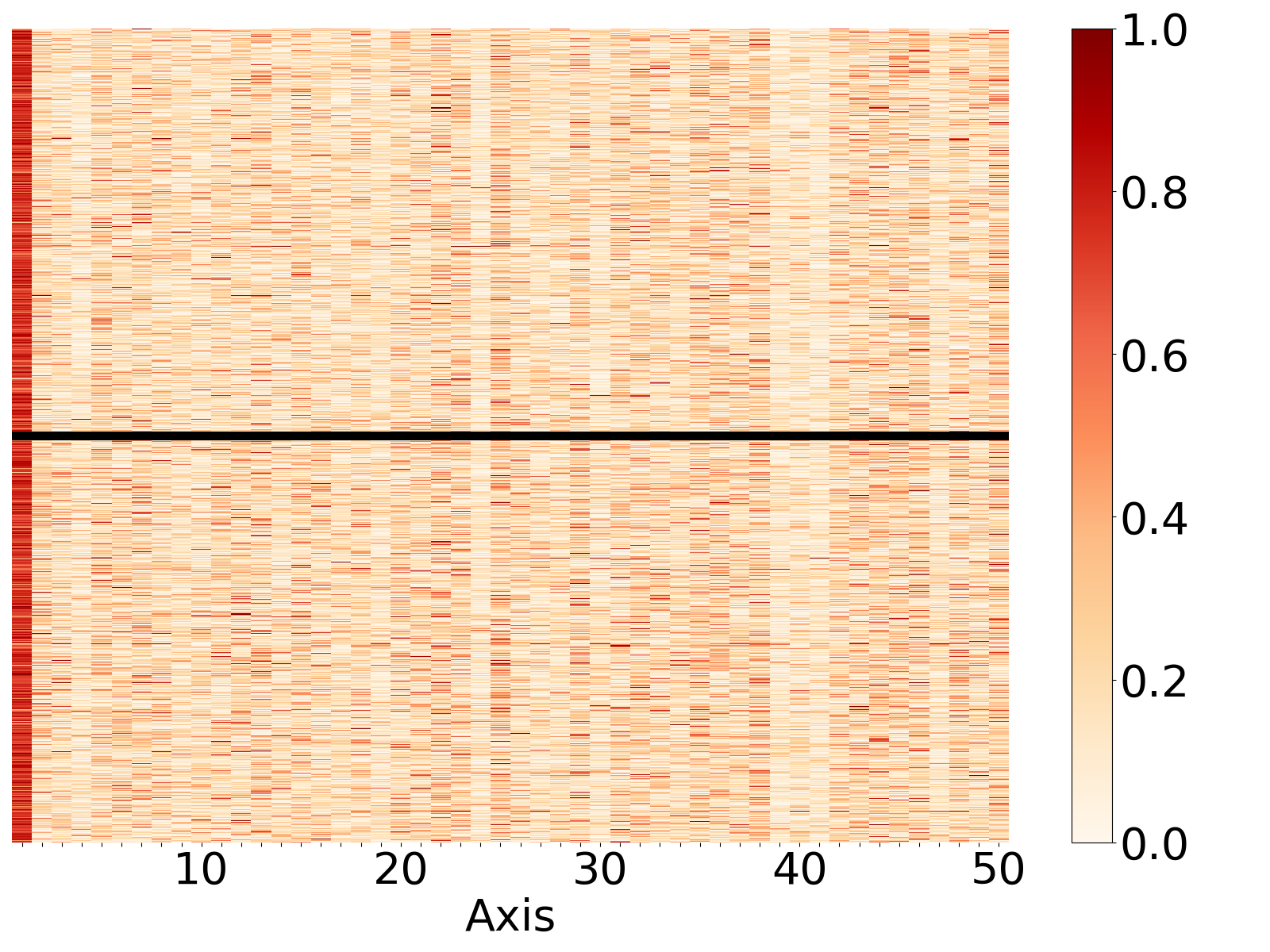}
        \subcaption{Pre-trained \ac{CWE}, PCA}
        \label{fig:wic_am2ico_ru_instances_pca_pretrained}
    \end{minipage}
    \begin{minipage}[b]{0.65\columnwidth}
        \centering
        \includegraphics[width=0.85\columnwidth]{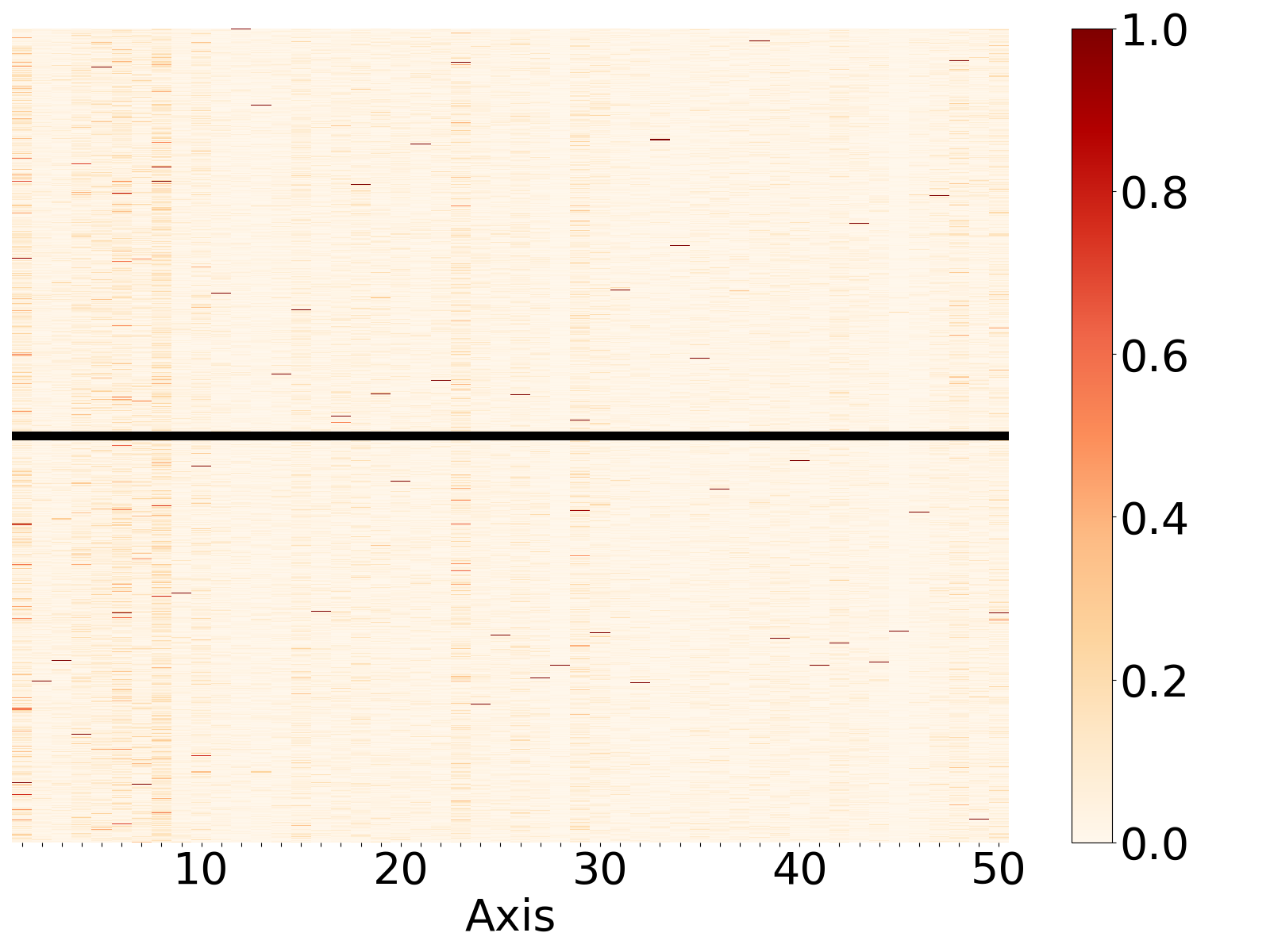}
        \subcaption{Pre-trained \ac{CWE}, ICA}
        \label{fig:wic_am2ico_ru_instances_ica_pretrained}
    \end{minipage} \\
    \begin{minipage}[b]{0.65\columnwidth}
        \centering
        \includegraphics[width=0.85\columnwidth]{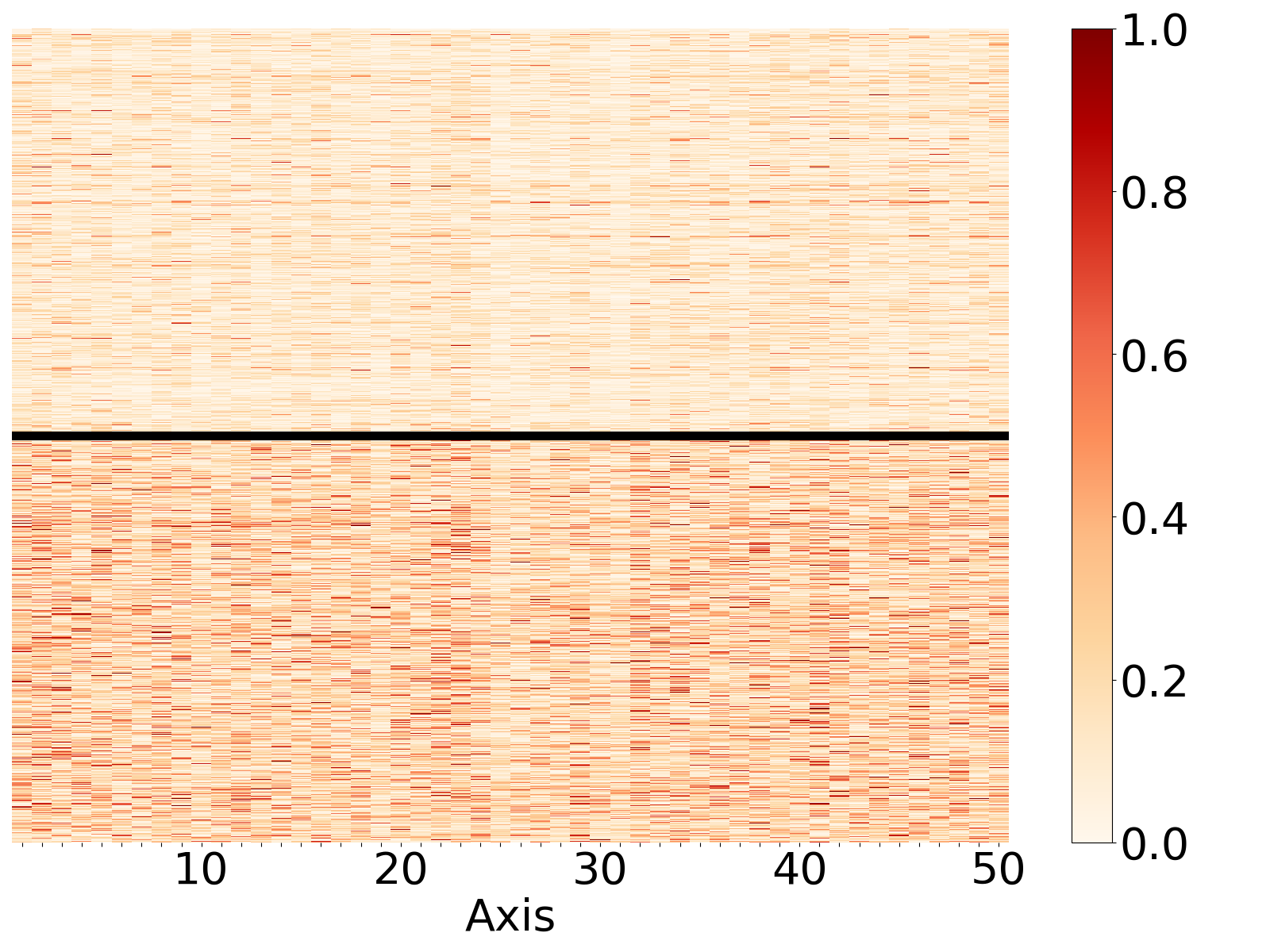}
        \subcaption{Fine-tuned \ac{SCWE}, Raw}
        \label{fig:wic_am2ico_ru_instances_raw_finetuned}
    \end{minipage}
    \begin{minipage}[b]{0.65\columnwidth}
        \centering
        \includegraphics[width=0.85\columnwidth]{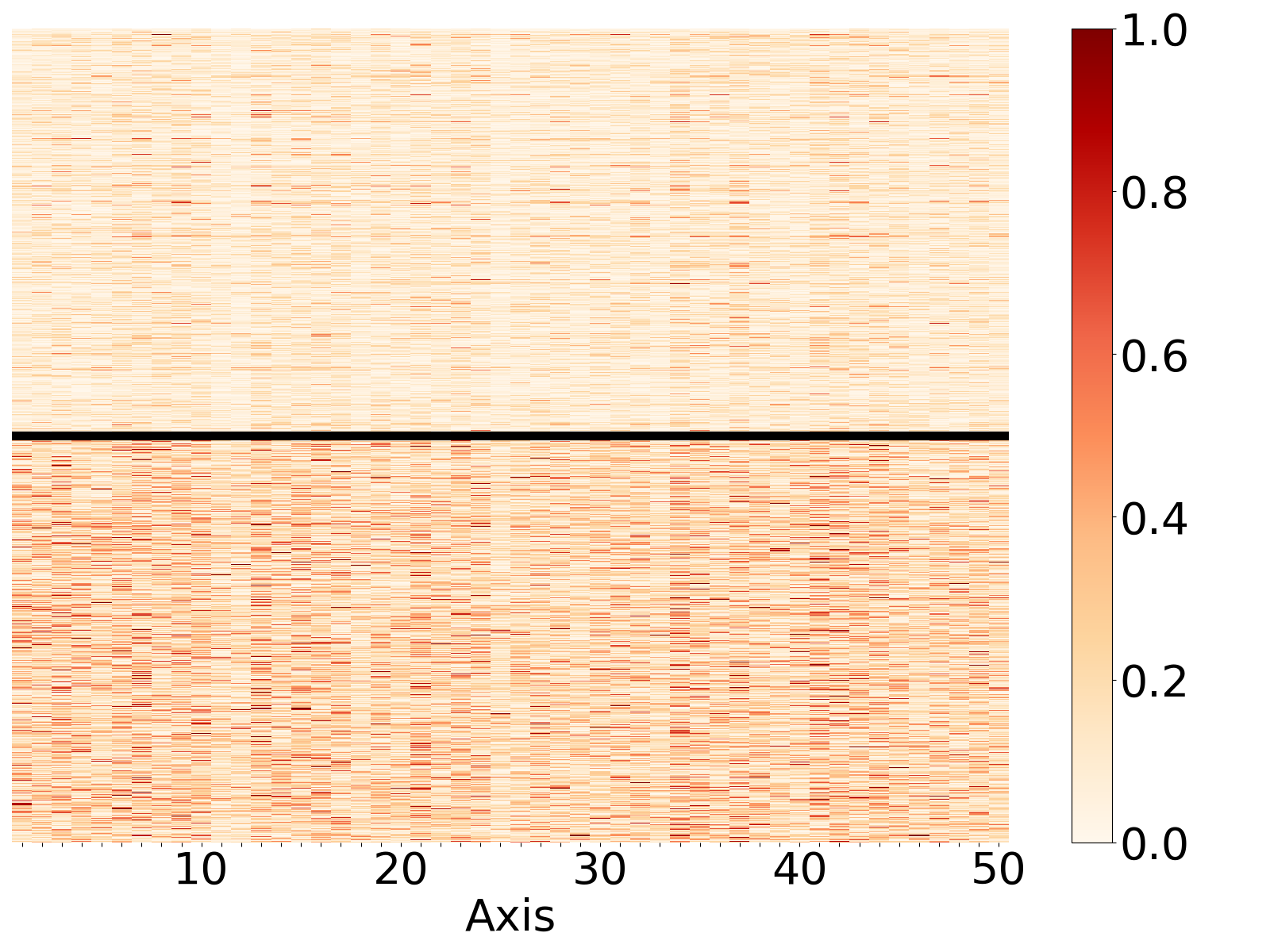}
        \subcaption{Fine-tuned \ac{SCWE}, PCA}
        \label{fig:wic_am2ico_ru_instances_pca_finetuned}
    \end{minipage}
    \begin{minipage}[b]{0.65\columnwidth}
        \centering
        \includegraphics[width=0.85\columnwidth]{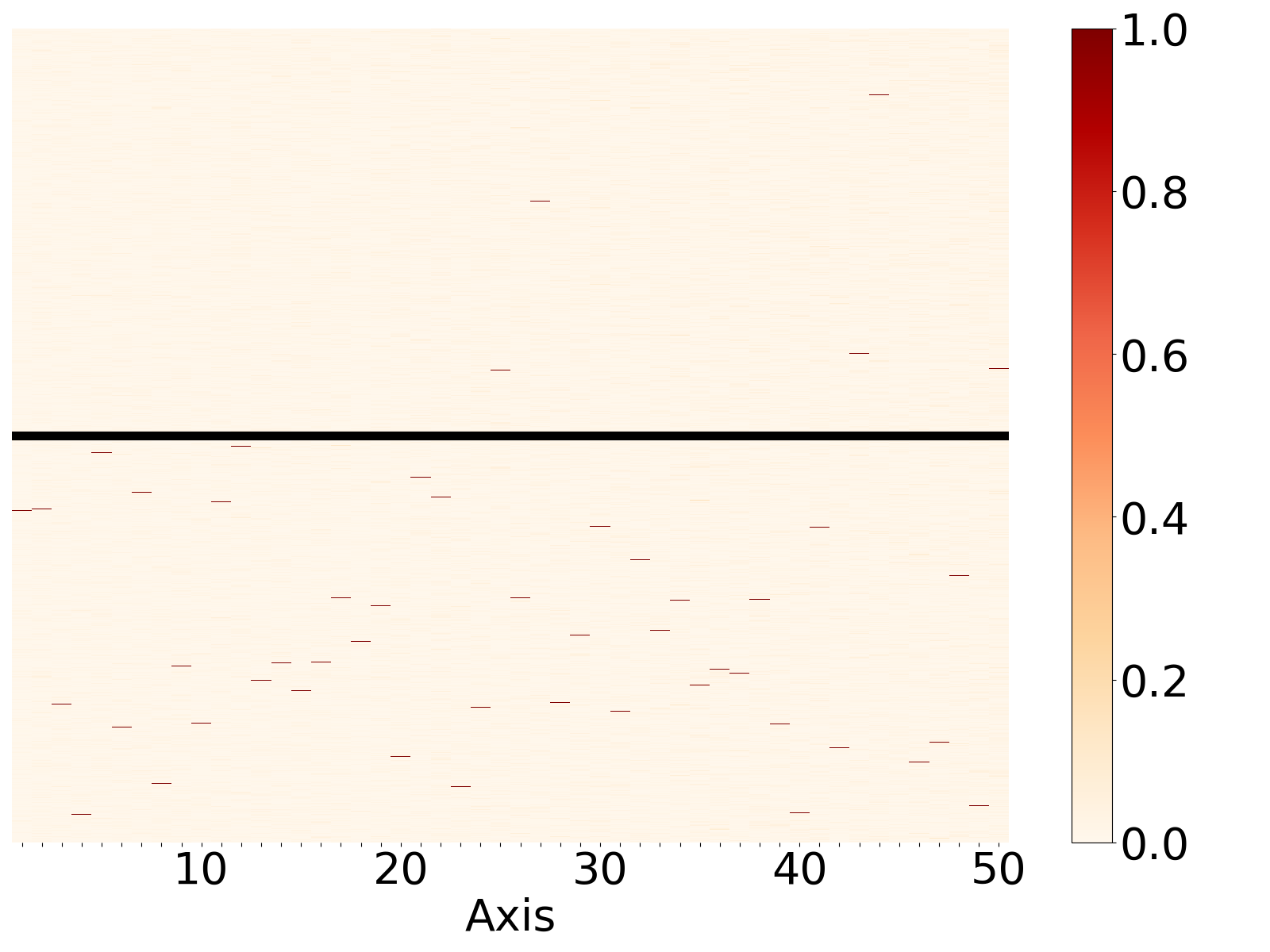}
        \subcaption{Fine-tuned \ac{SCWE}, ICA}
        \label{fig:wic_am2ico_ru_instances_ica_finetuned}
    \end{minipage}
    \caption{Visualisation of the top-50 dimensions of pre-trained \acp{CWE} (XLM-RoBERTa) and \acp{SCWE} (XL-LEXEME) for each instance in AM$^2$iCo (Russian) dataset, where the difference of vectors is calculated for (a/d) \textbf{Raw} vectors, (b/e) \ac{PCA}-transformed axes, and (c/f) \ac{ICA}-transformed axes. In each figure, the upper/lower half uses instances for the True/False labels.}
    \label{fig:wic_instance_am2ico_ru}
\end{figure*}

\begin{figure*}[t]
    \centering
    \begin{minipage}[b]{0.65\columnwidth}
        \centering
        \includegraphics[width=0.85\columnwidth]{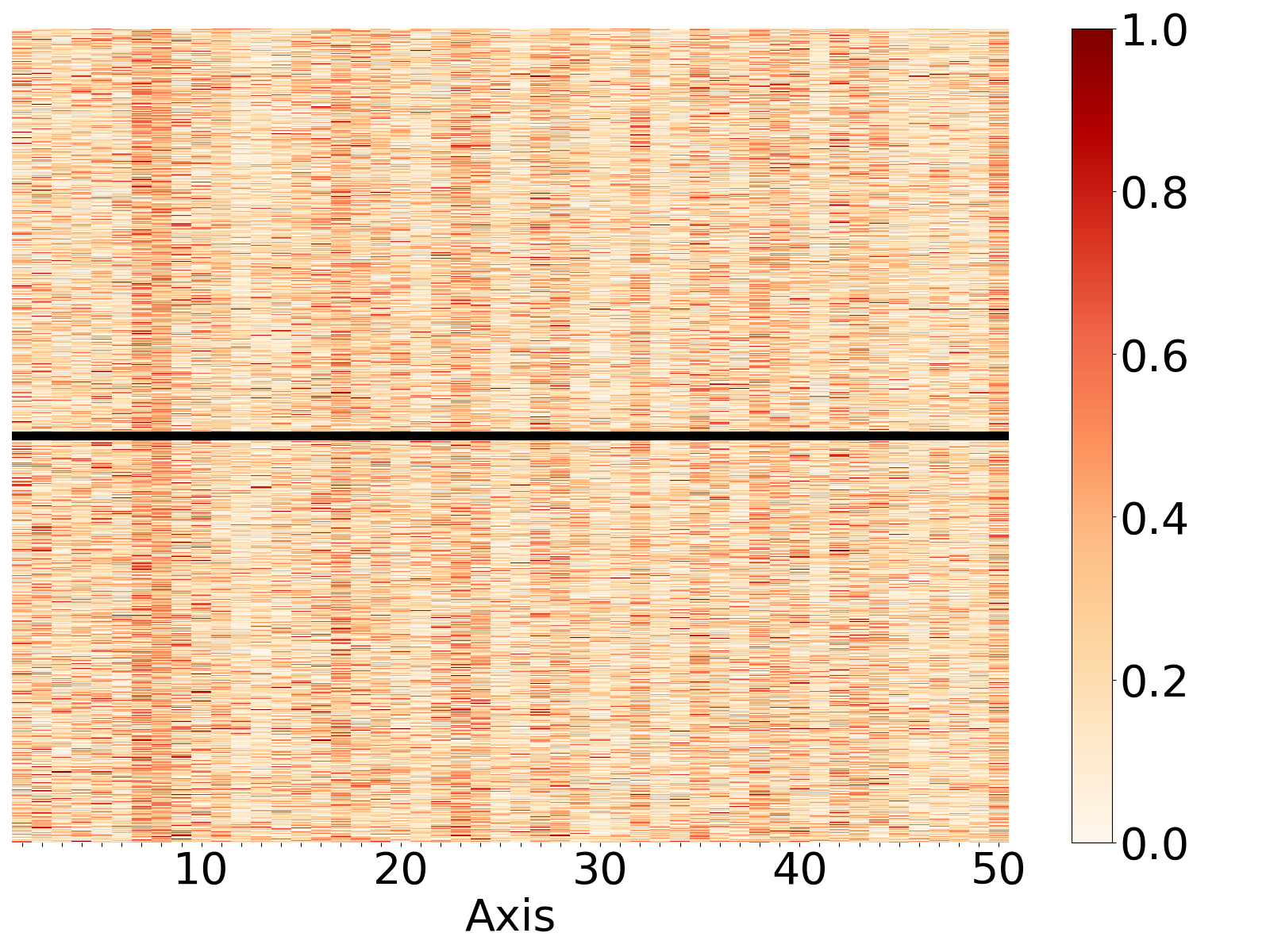}
        \subcaption{Pre-trained \ac{CWE}, Raw}
        \label{fig:wic_am2ico_ja_instances_raw_pretrained}
    \end{minipage}
    \begin{minipage}[b]{0.65\columnwidth}
        \centering
        \includegraphics[width=0.85\columnwidth]{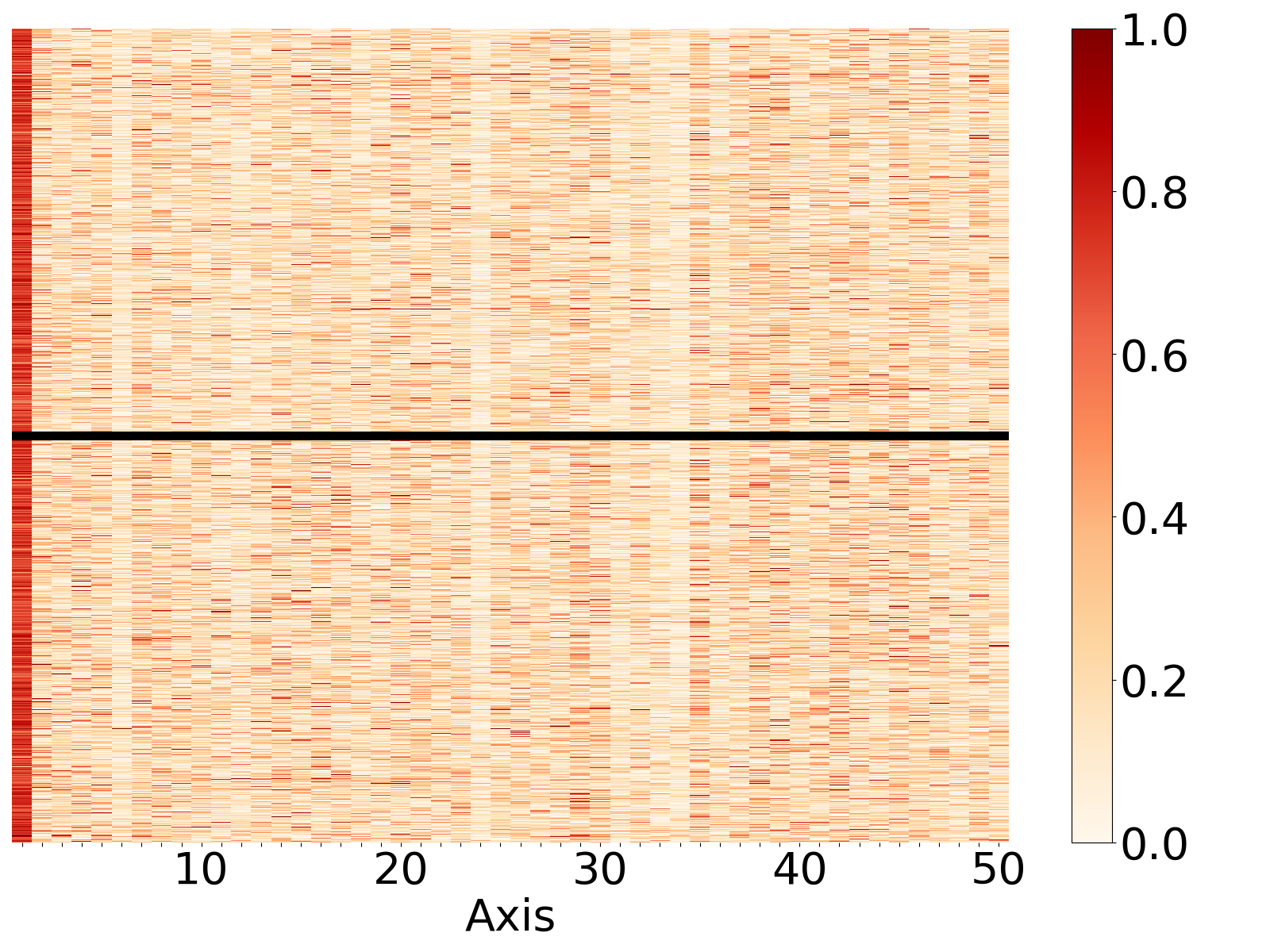}
        \subcaption{Pre-trained \ac{CWE}, PCA}
        \label{fig:wic_am2ico_ja_instances_pca_pretrained}
    \end{minipage}
    \begin{minipage}[b]{0.65\columnwidth}
        \centering
        \includegraphics[width=0.85\columnwidth]{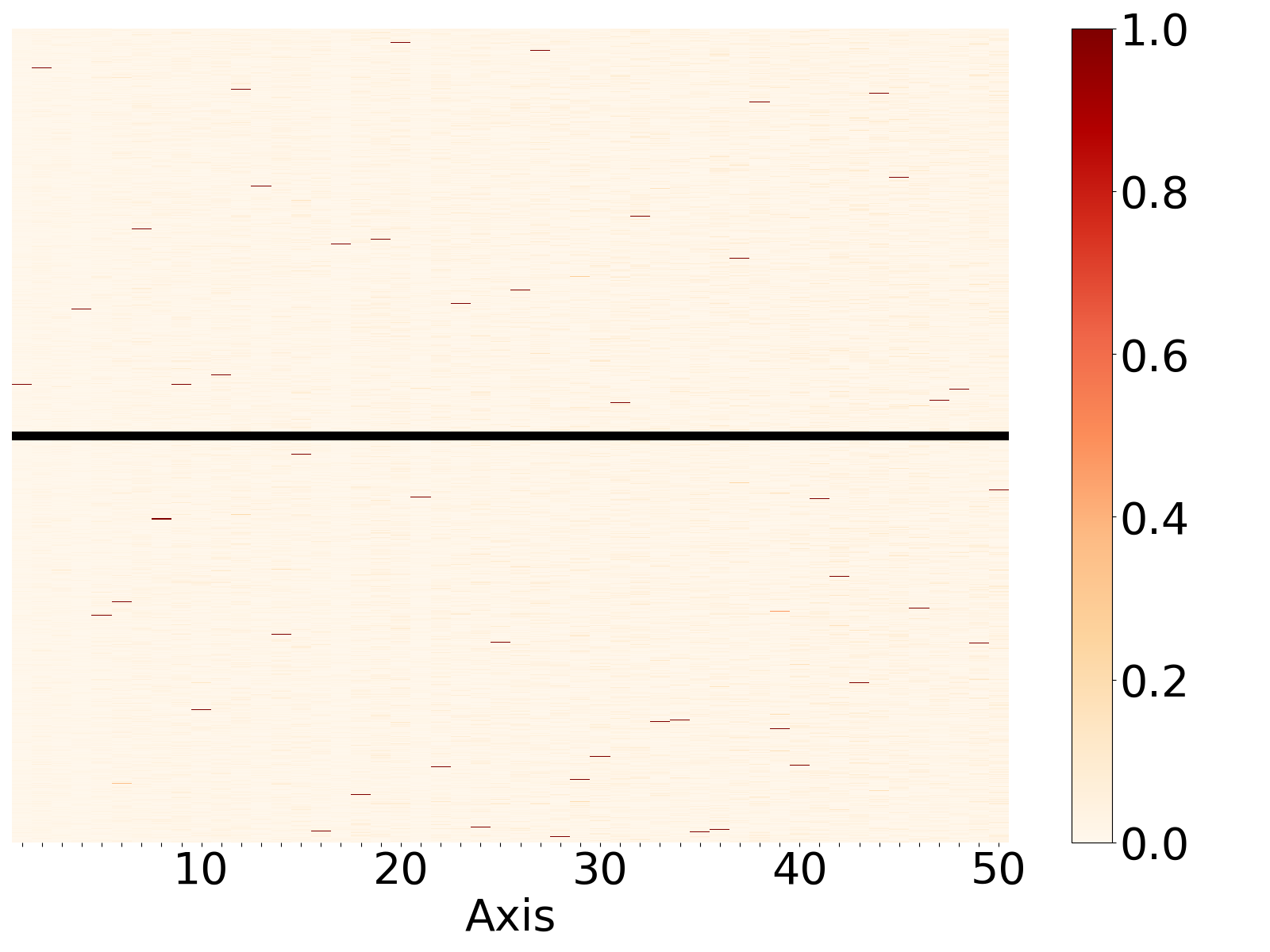}
        \subcaption{Pre-trained \ac{CWE}, ICA}
        \label{fig:wic_am2ico_ja_instances_ica_pretrained}
    \end{minipage} \\
    \begin{minipage}[b]{0.65\columnwidth}
        \centering
        \includegraphics[width=0.85\columnwidth]{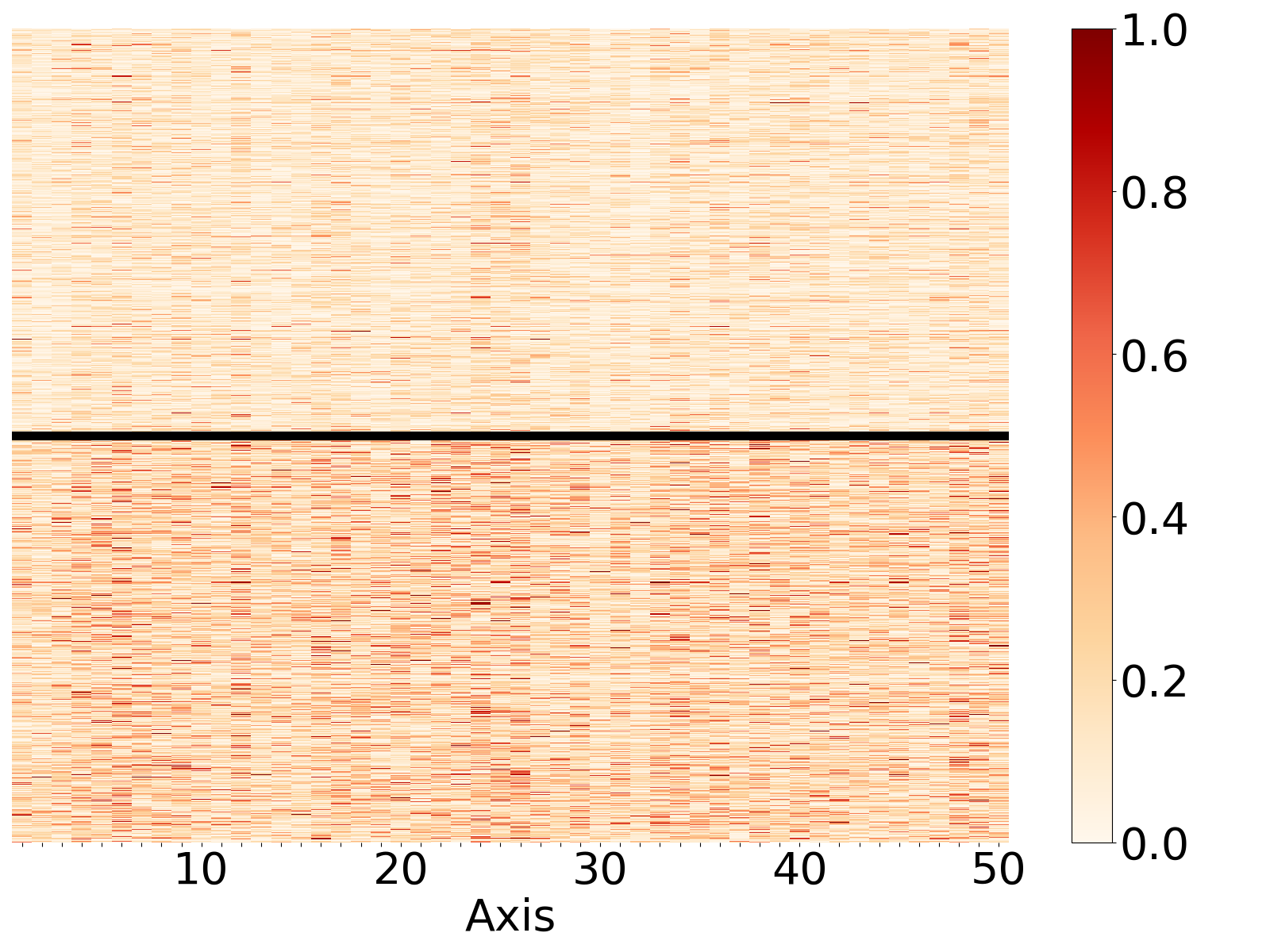}
        \subcaption{Fine-tuned \ac{SCWE}, Raw}
        \label{fig:wic_am2ico_ja_instances_raw_finetuned}
    \end{minipage}
    \begin{minipage}[b]{0.65\columnwidth}
        \centering
        \includegraphics[width=0.85\columnwidth]{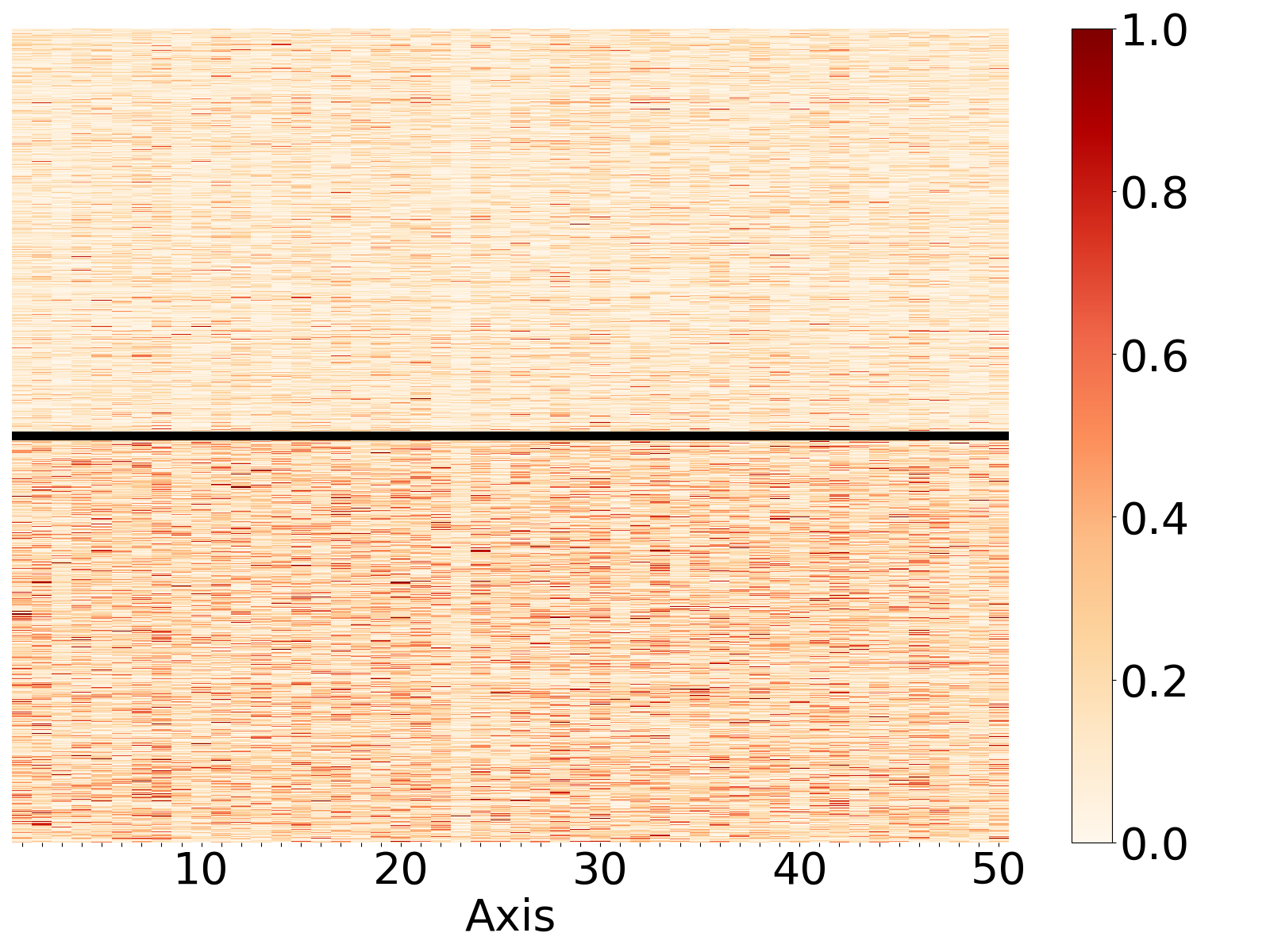}
        \subcaption{Fine-tuned \ac{SCWE}, PCA}
        \label{fig:wic_am2ico_ja_instances_pca_finetuned}
    \end{minipage}
    \begin{minipage}[b]{0.65\columnwidth}
        \centering
        \includegraphics[width=0.85\columnwidth]{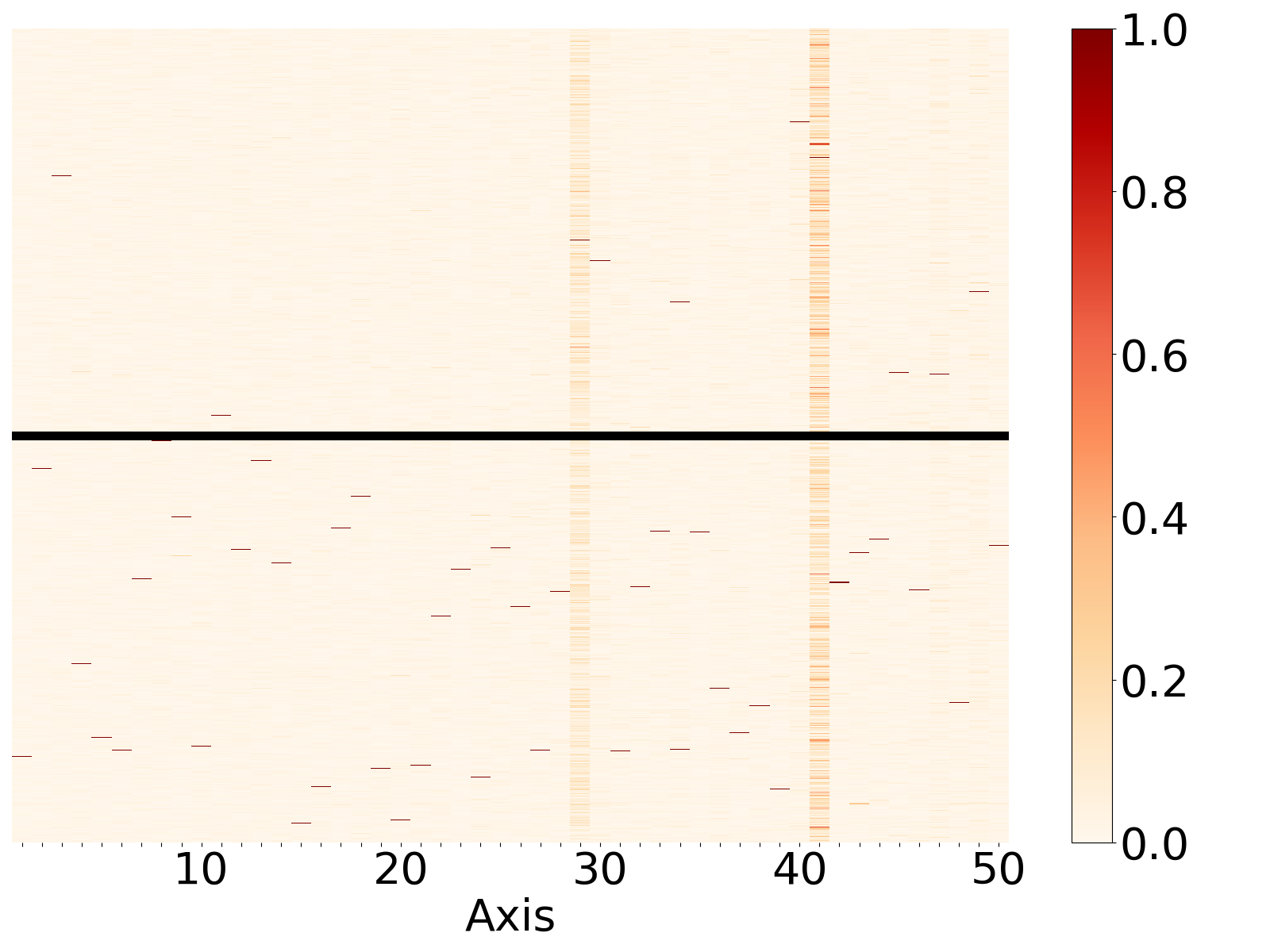}
        \subcaption{Fine-tuned \ac{SCWE}, ICA}
        \label{fig:wic_am2ico_ja_instances_ica_finetuned}
    \end{minipage}
    \caption{Visualisation of the top-50 dimensions of pre-trained \acp{CWE} (XLM-RoBERTa) and \acp{SCWE} (XL-LEXEME) for each instance in AM$^2$iCo (Japanese) dataset, where the difference of vectors is calculated for (a/d) \textbf{Raw} vectors, (b/e) \ac{PCA}-transformed axes, and (c/f) \ac{ICA}-transformed axes. In each figure, the upper/lower half uses instances for the True/False labels.}
    \label{fig:wic_instance_am2ico_ja}
\end{figure*}

\begin{figure*}[t]
    \centering
    \begin{minipage}[b]{0.65\columnwidth}
        \centering
        \includegraphics[width=0.85\columnwidth]{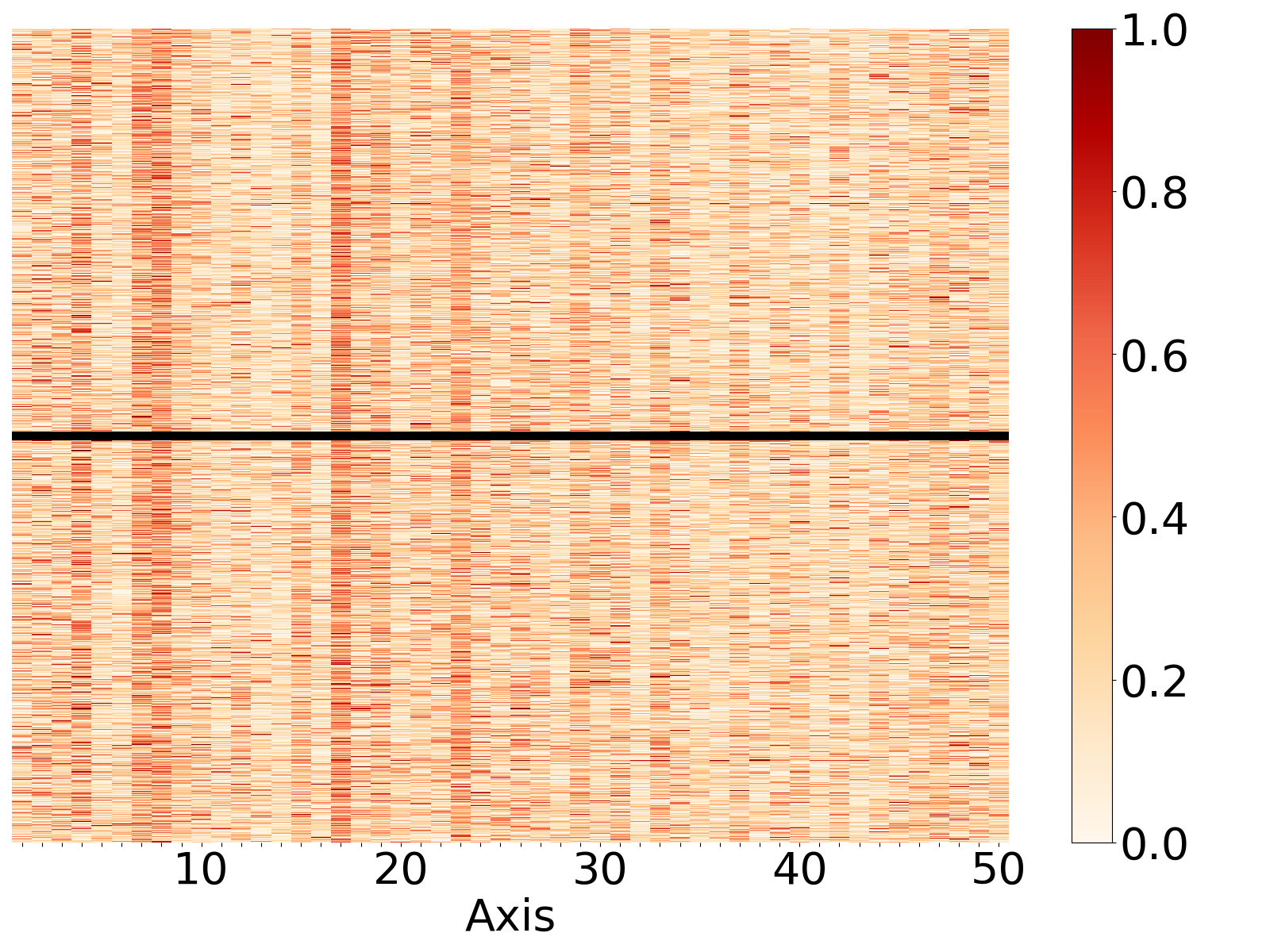}
        \subcaption{Pre-trained \ac{CWE}, Raw}
        \label{fig:wic_am2ico_zh_instances_raw_pretrained}
    \end{minipage}
    \begin{minipage}[b]{0.65\columnwidth}
        \centering
        \includegraphics[width=0.85\columnwidth]{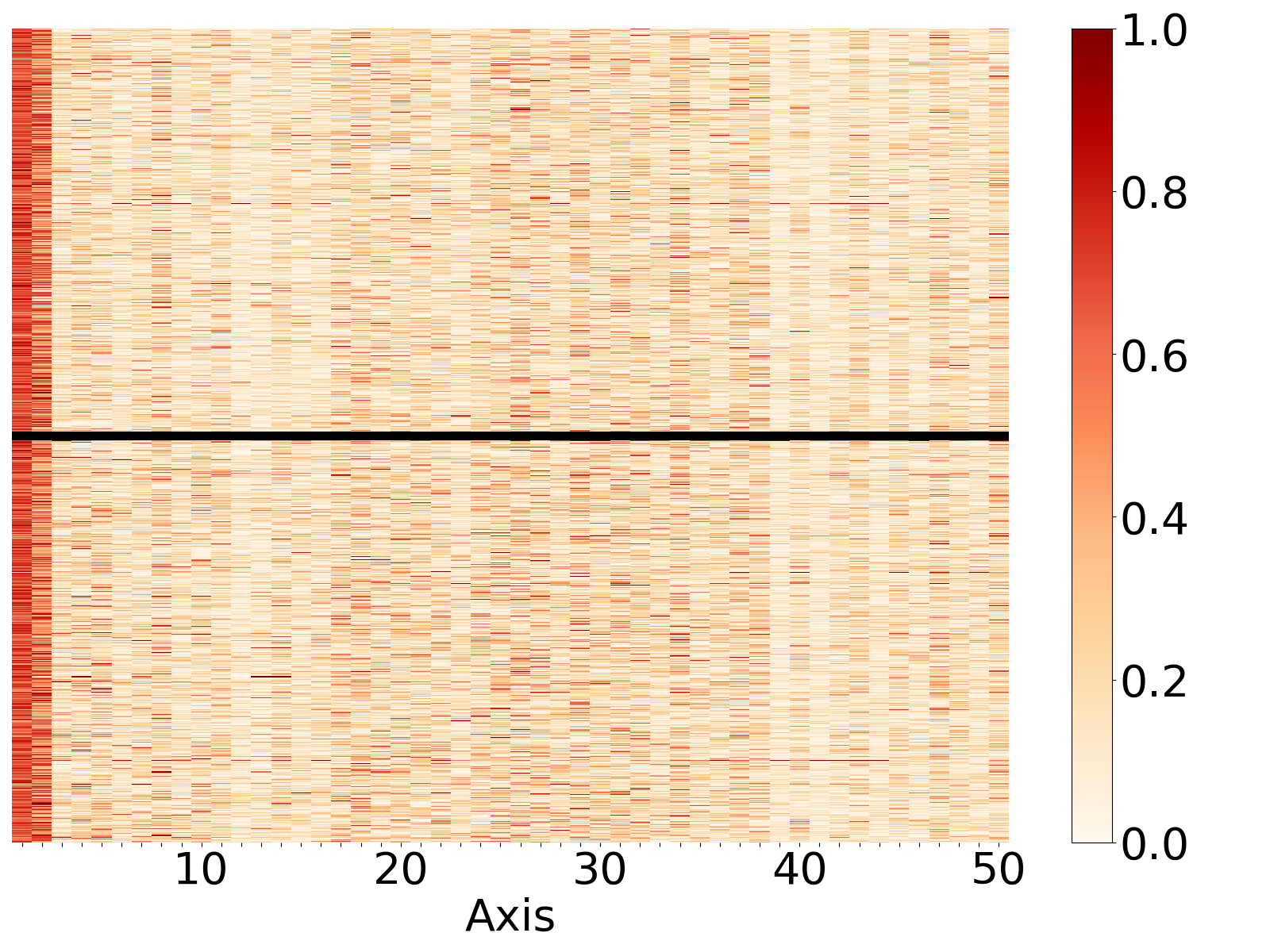}
        \subcaption{Pre-trained \ac{CWE}, PCA}
        \label{fig:wic_am2ico_zh_instances_pca_pretrained}
    \end{minipage}
    \begin{minipage}[b]{0.65\columnwidth}
        \centering
        \includegraphics[width=0.85\columnwidth]{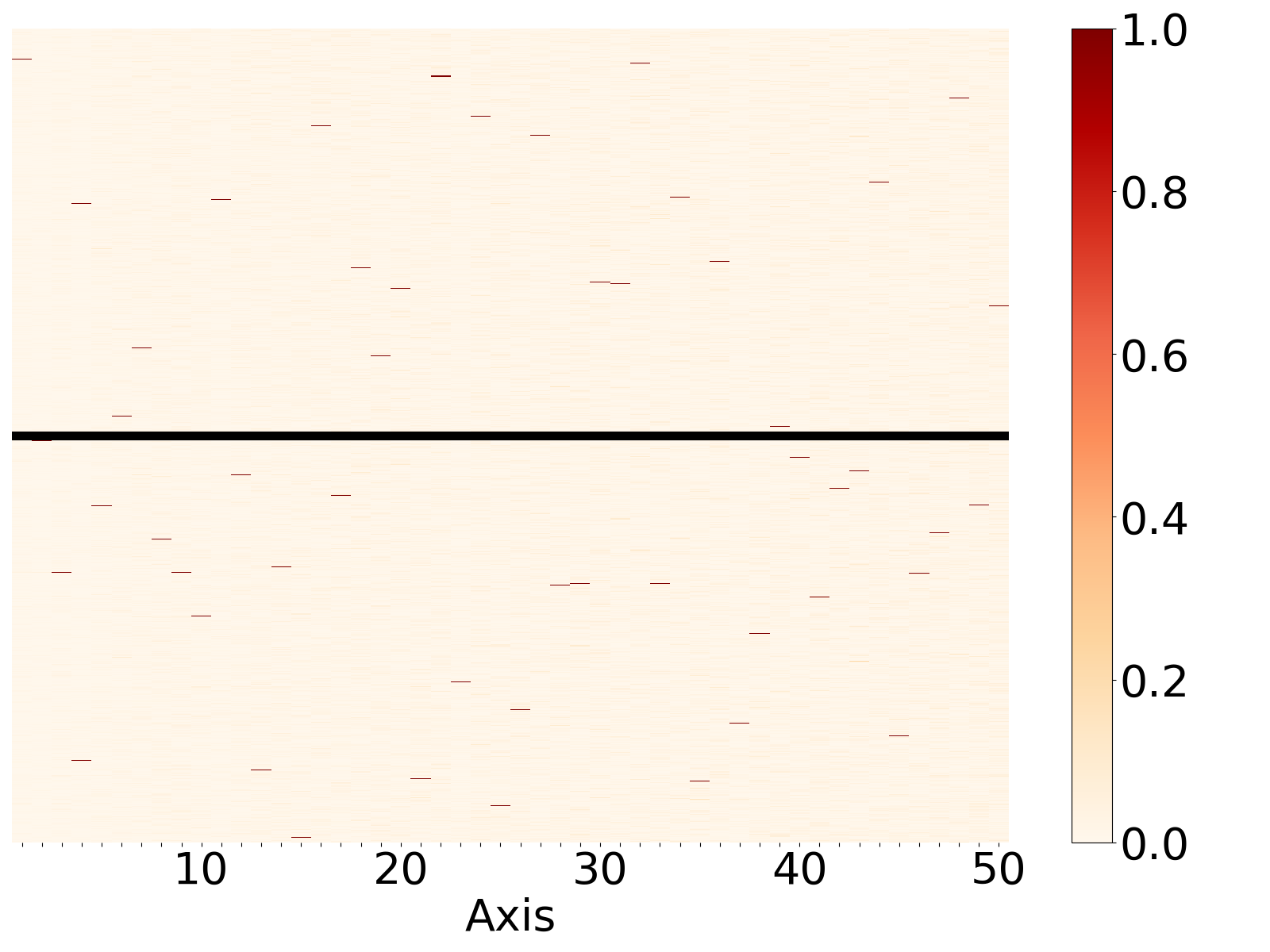}
        \subcaption{Pre-trained \ac{CWE}, ICA}
        \label{fig:wic_am2ico_zh_instances_ica_pretrained}
    \end{minipage} \\
    \begin{minipage}[b]{0.65\columnwidth}
        \centering
        \includegraphics[width=0.85\columnwidth]{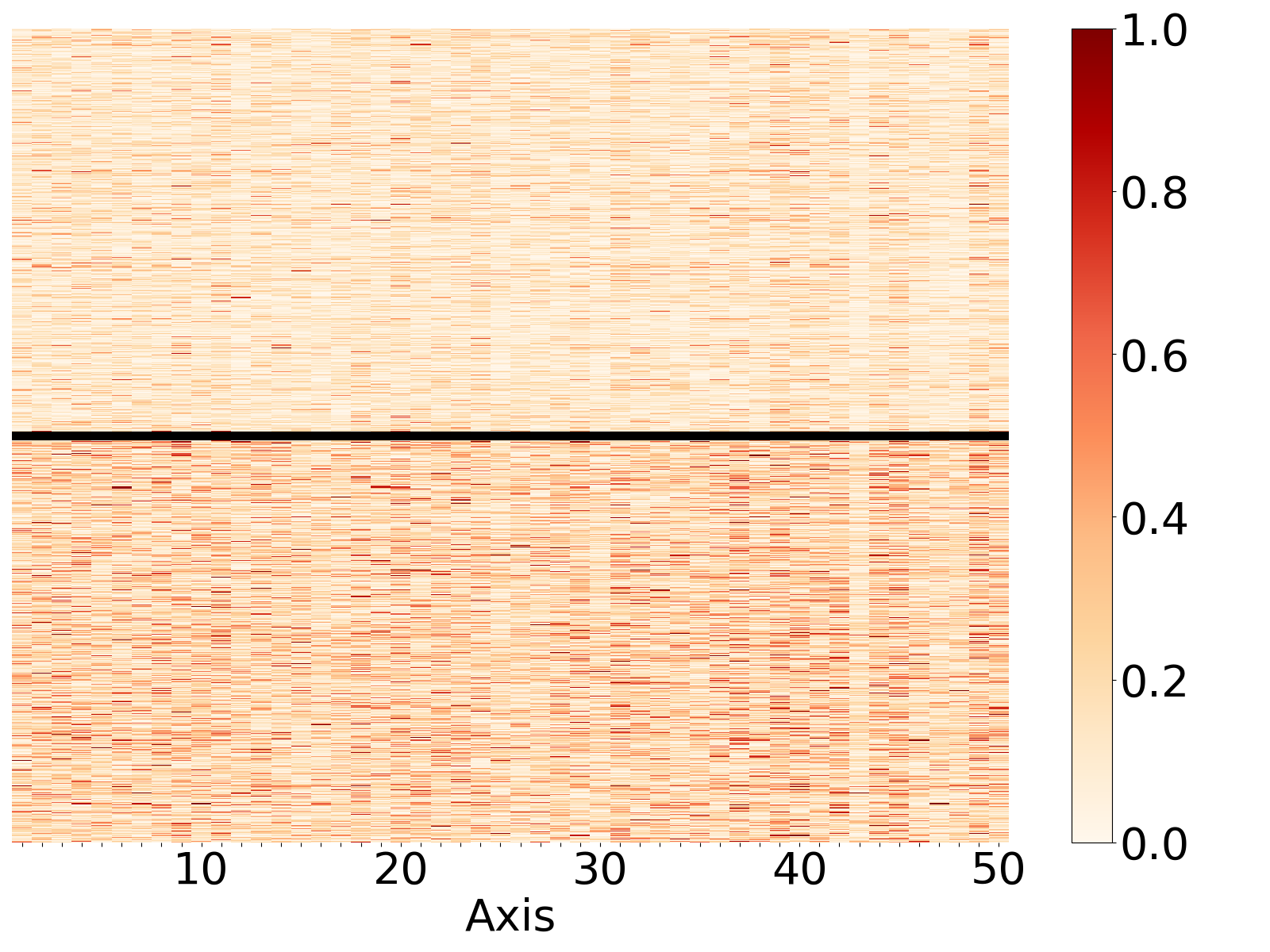}
        \subcaption{Fine-tuned \ac{SCWE}, Raw}
        \label{fig:wic_am2ico_zh_instances_raw_finetuned}
    \end{minipage}
    \begin{minipage}[b]{0.65\columnwidth}
        \centering
        \includegraphics[width=0.85\columnwidth]{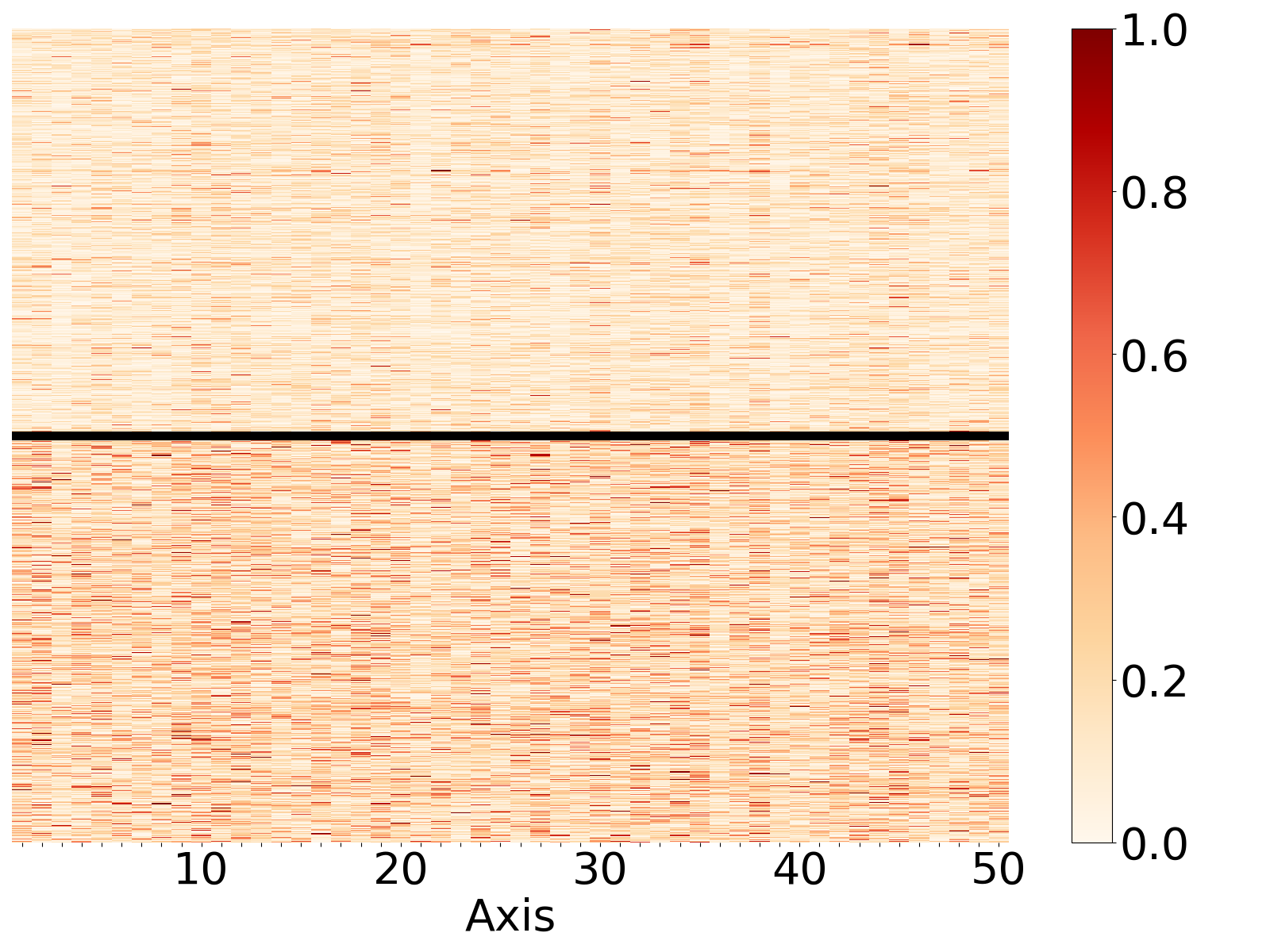}
        \subcaption{Fine-tuned \ac{SCWE}, PCA}
        \label{fig:wic_am2ico_zh_instances_pca_finetuned}
    \end{minipage}
    \begin{minipage}[b]{0.65\columnwidth}
        \centering
        \includegraphics[width=0.85\columnwidth]{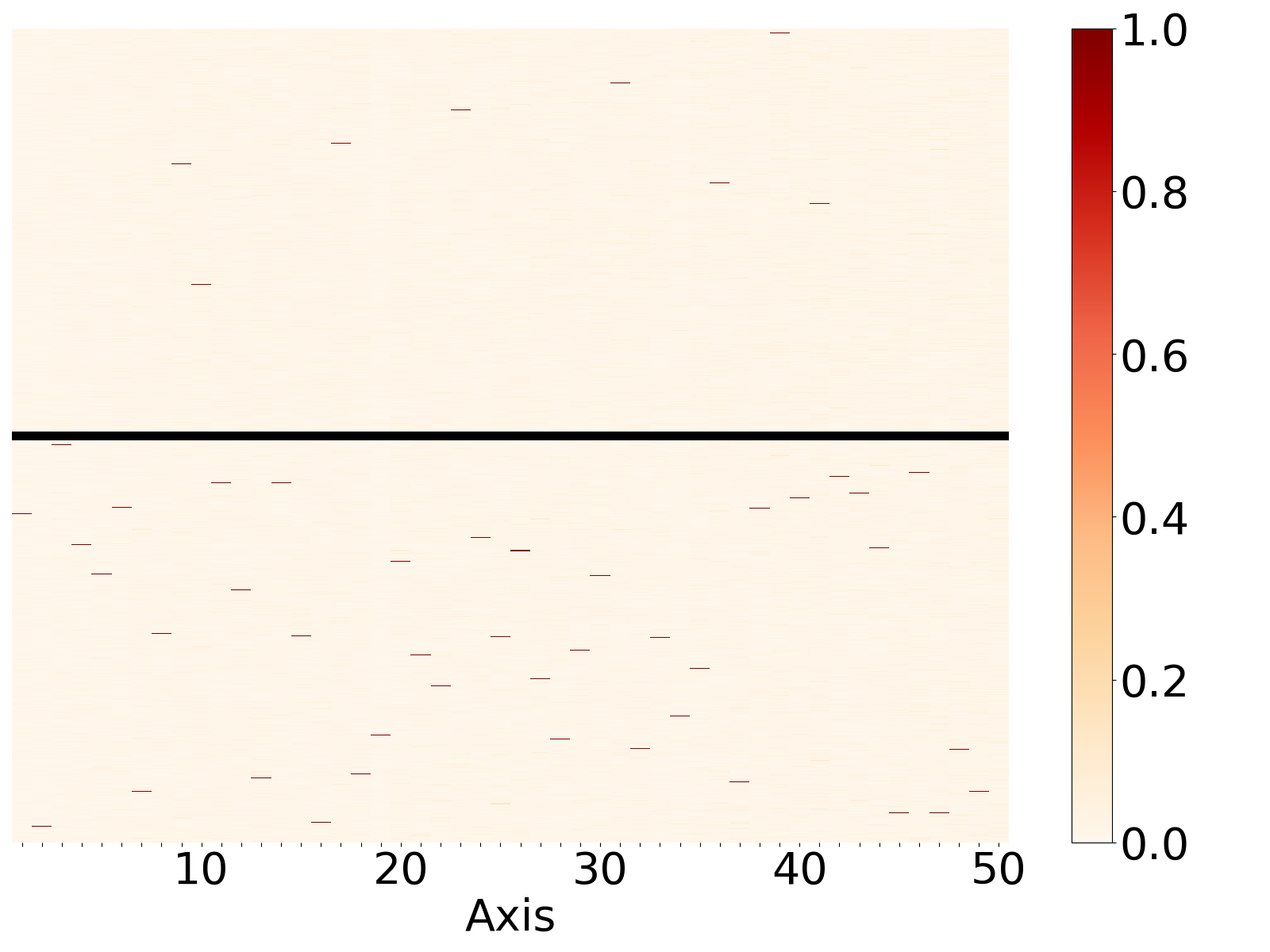}
        \subcaption{Fine-tuned \ac{SCWE}, ICA}
        \label{fig:wic_am2ico_zh_instances_ica_finetuned}
    \end{minipage}
    \caption{Visualisation of the top-50 dimensions of pre-trained \acp{CWE} (XLM-RoBERTa) and \acp{SCWE} (XL-LEXEME) for each instance in AM$^2$iCo (Chinese) dataset, where the difference of vectors is calculated for (a/d) \textbf{Raw} vectors, (b/e) \ac{PCA}-transformed axes, and (c/f) \ac{ICA}-transformed axes. In each figure, the upper/lower half uses instances for the True/False labels.}
    \label{fig:wic_instance_am2ico_zh}
\end{figure*}

\begin{figure*}[t]
    \centering
    \begin{minipage}[b]{0.65\columnwidth}
        \centering
        \includegraphics[width=0.85\columnwidth]{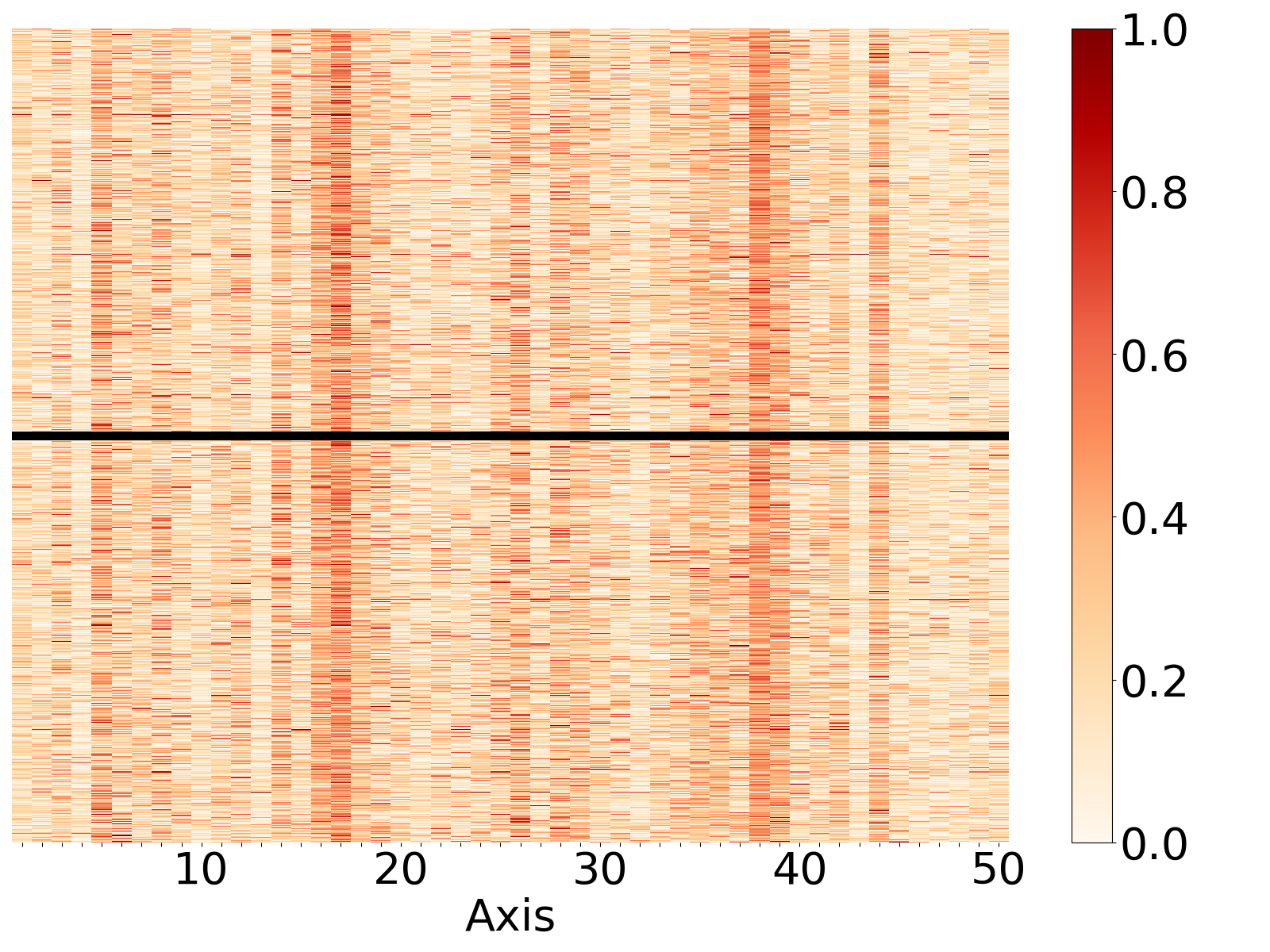}
        \subcaption{Pre-trained \ac{CWE}, Raw}
        \label{fig:wic_am2ico_ar_instances_raw_pretrained}
    \end{minipage}
    \begin{minipage}[b]{0.65\columnwidth}
        \centering
        \includegraphics[width=0.85\columnwidth]{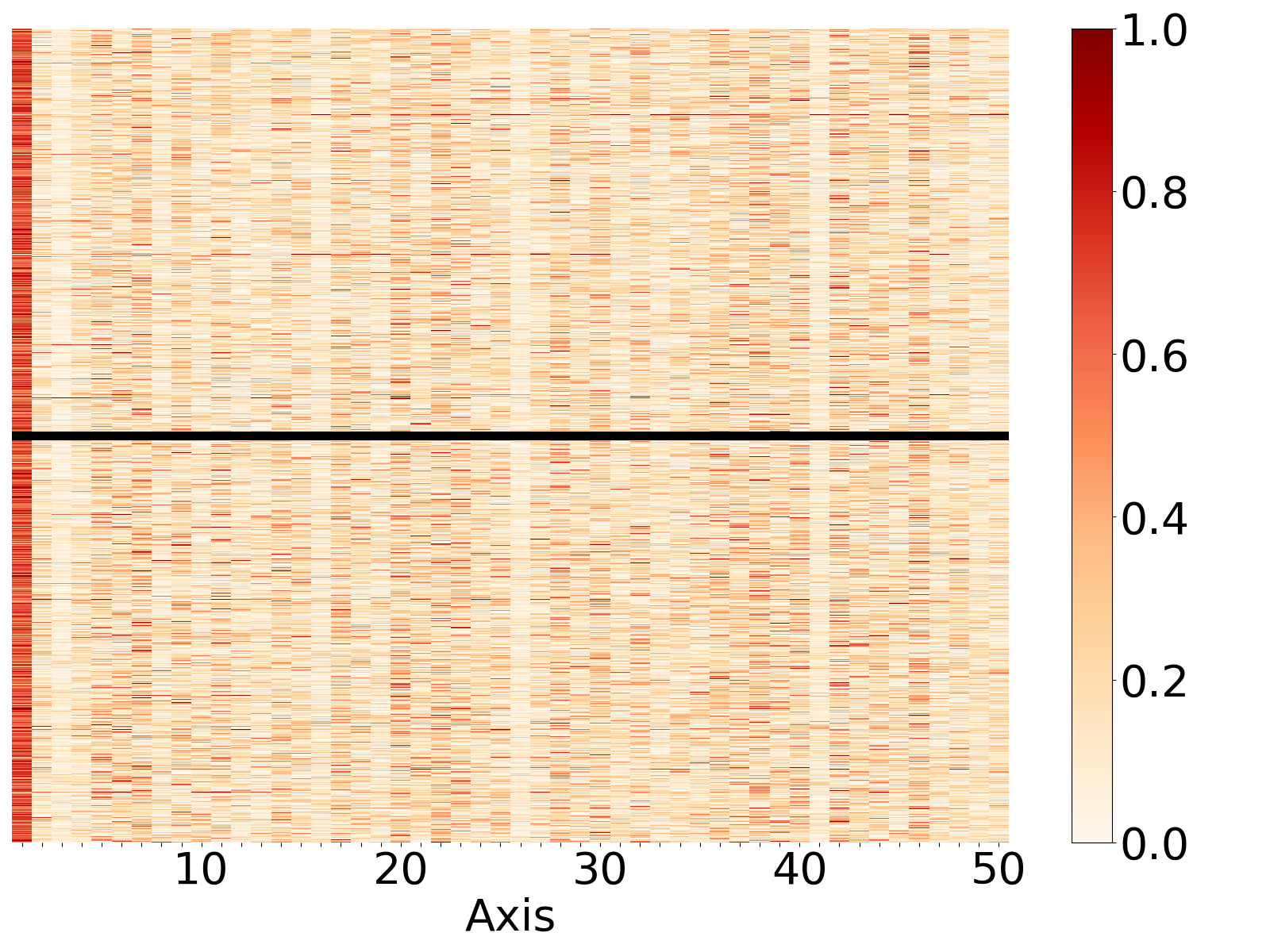}
        \subcaption{Pre-trained \ac{CWE}, PCA}
        \label{fig:wic_am2ico_ar_instances_pca_pretrained}
    \end{minipage}
    \begin{minipage}[b]{0.65\columnwidth}
        \centering
        \includegraphics[width=0.85\columnwidth]{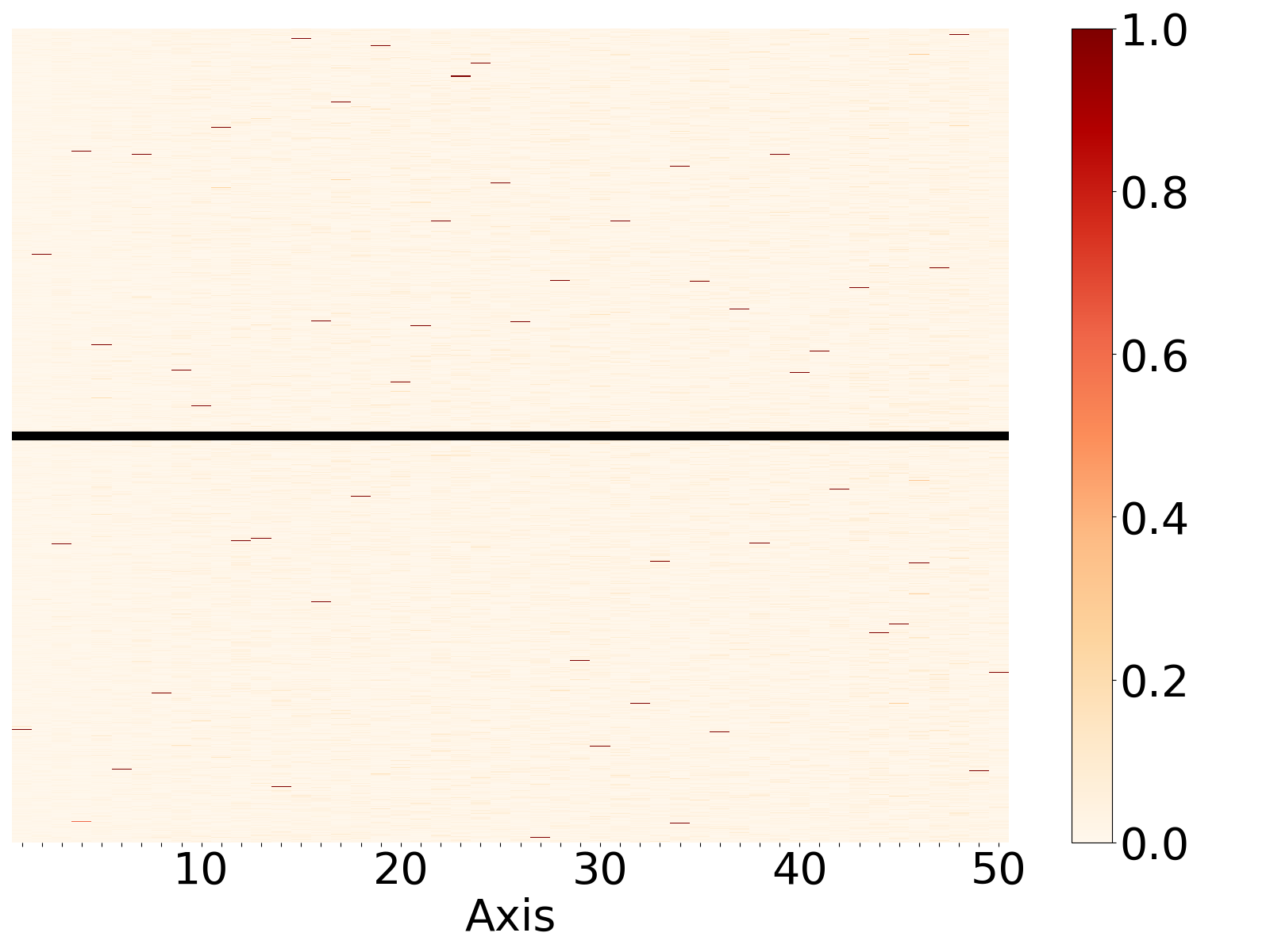}
        \subcaption{Pre-trained \ac{CWE}, ICA}
        \label{fig:wic_am2ico_ar_instances_ica_pretrained}
    \end{minipage} \\
    \begin{minipage}[b]{0.65\columnwidth}
        \centering
        \includegraphics[width=0.85\columnwidth]{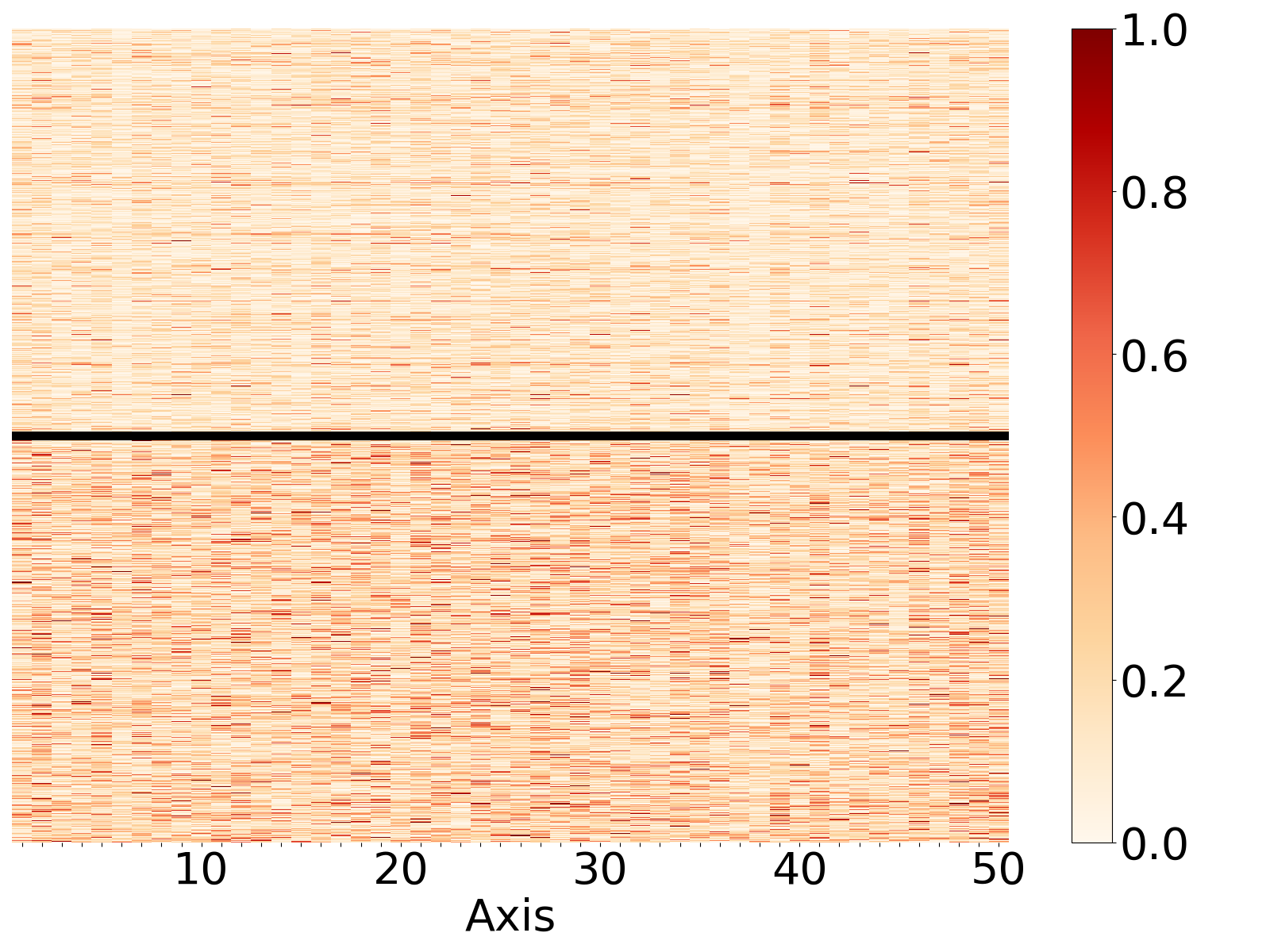}
        \subcaption{Fine-tuned \ac{SCWE}, Raw}
        \label{fig:wic_am2ico_ar_instances_raw_finetuned}
    \end{minipage}
    \begin{minipage}[b]{0.65\columnwidth}
        \centering
        \includegraphics[width=0.85\columnwidth]{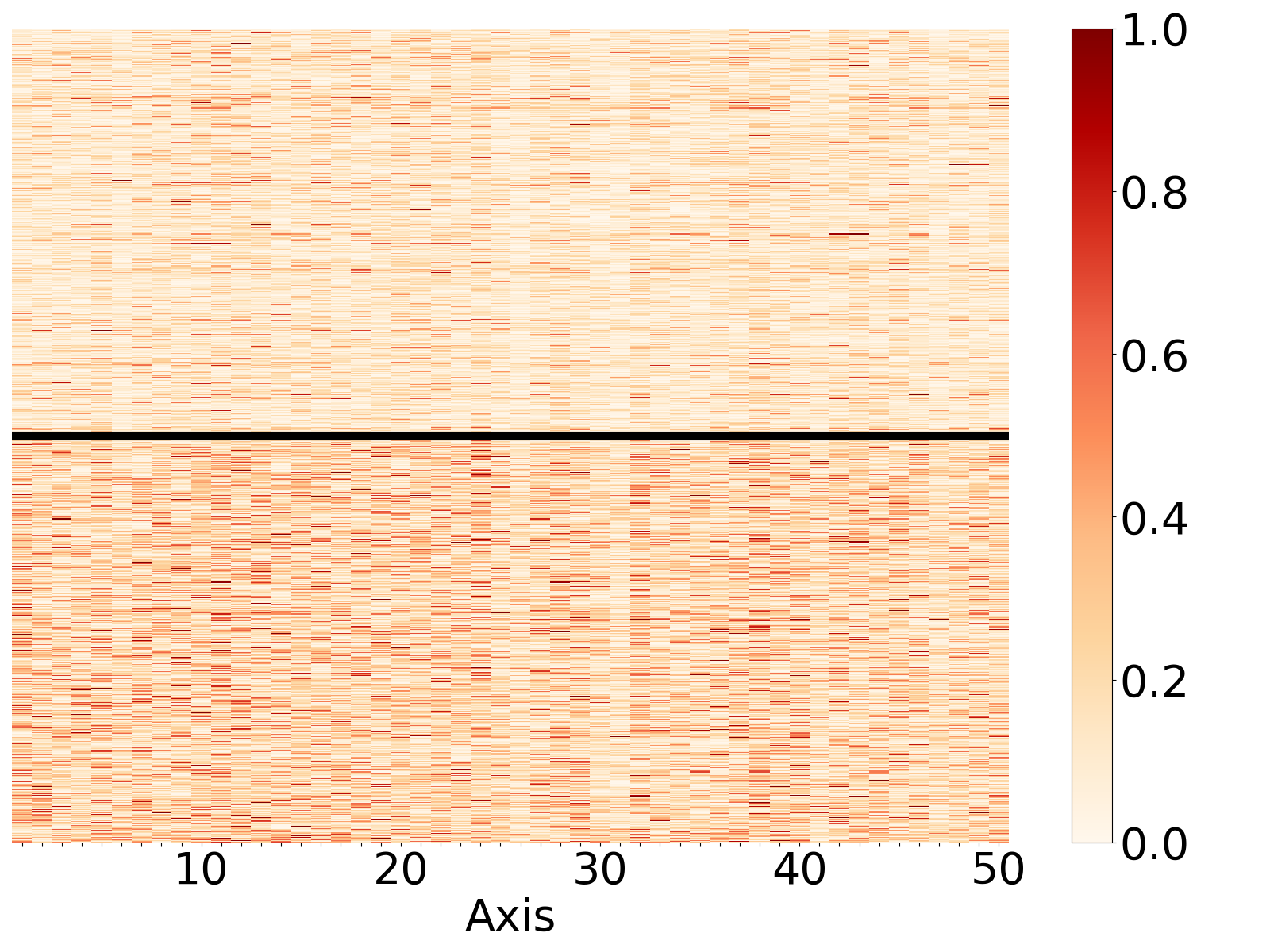}
        \subcaption{Fine-tuned \ac{SCWE}, PCA}
        \label{fig:wic_am2ico_ar_instances_pca_finetuned}
    \end{minipage}
    \begin{minipage}[b]{0.65\columnwidth}
        \centering
        \includegraphics[width=0.85\columnwidth]{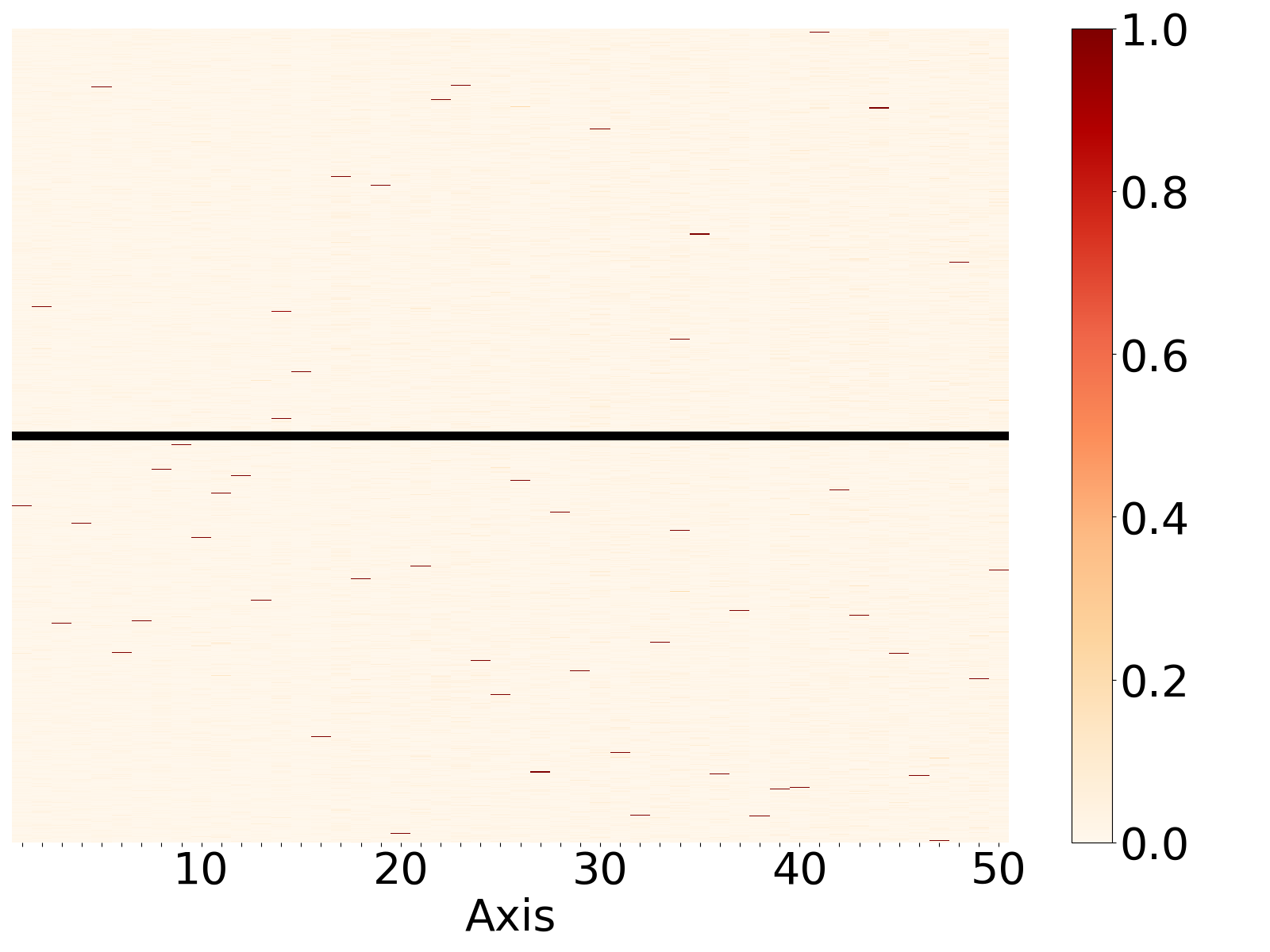}
        \subcaption{Fine-tuned \ac{SCWE}, ICA}
        \label{fig:wic_am2ico_ar_instances_ica_finetuned}
    \end{minipage}
    \caption{Visualisation of the top-50 dimensions of pre-trained \acp{CWE} (XLM-RoBERTa) and \acp{SCWE} (XL-LEXEME) for each instance in AM$^2$iCo (Arabic) dataset, where the difference of vectors is calculated for (a/d) \textbf{Raw} vectors, (b/e) \ac{PCA}-transformed axes, and (c/f) \ac{ICA}-transformed axes. In each figure, the upper/lower half uses instances for the True/False labels.}
    \label{fig:wic_instance_am2ico_ar}
\end{figure*}

\begin{figure*}[t]
    \centering
    \begin{minipage}[b]{0.65\columnwidth}
        \centering
        \includegraphics[width=0.85\columnwidth]{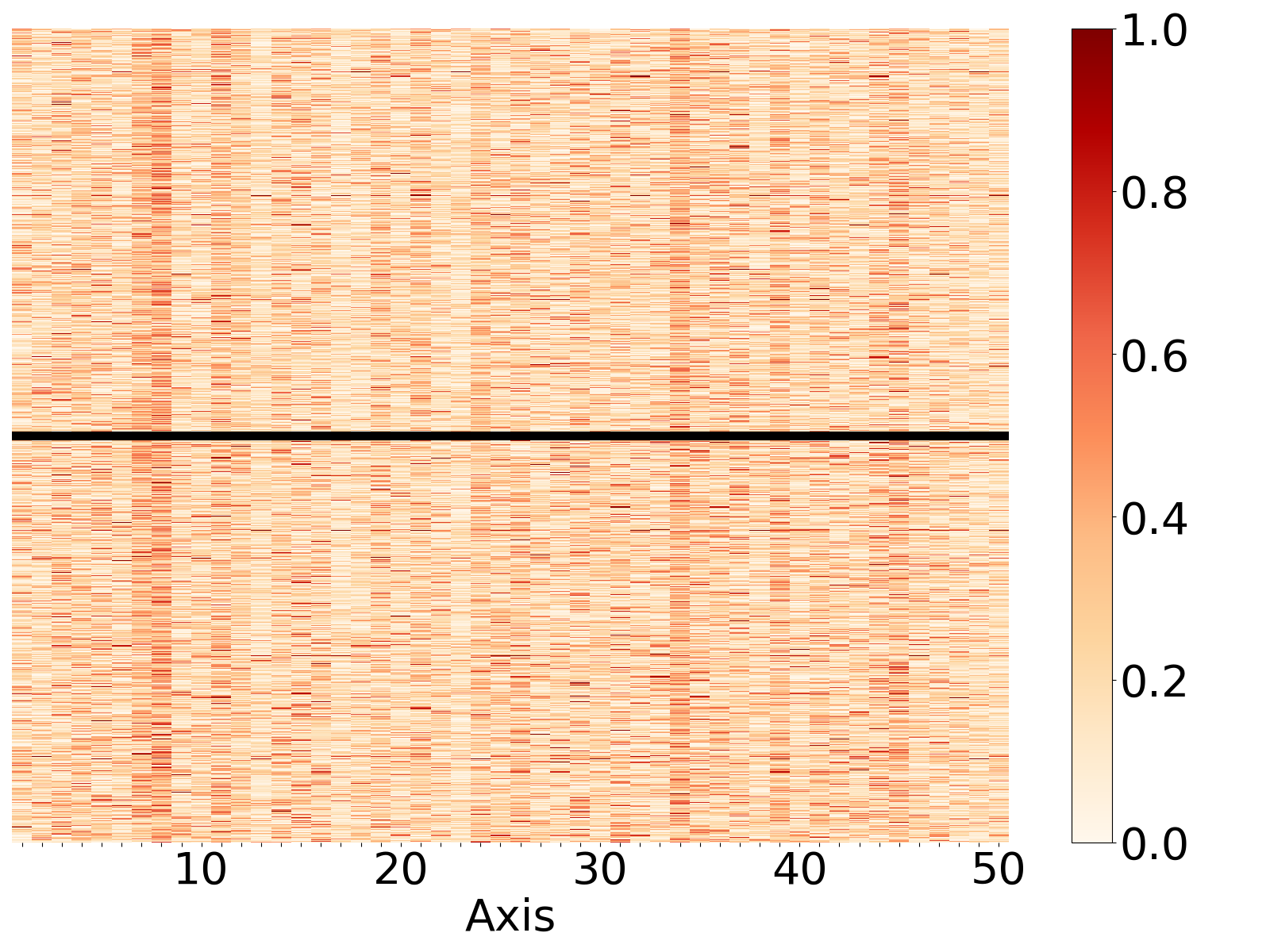}
        \subcaption{Pre-trained \ac{CWE}, Raw}
        \label{fig:wic_am2ico_ko_instances_raw_pretrained}
    \end{minipage}
    \begin{minipage}[b]{0.65\columnwidth}
        \centering
        \includegraphics[width=0.85\columnwidth]{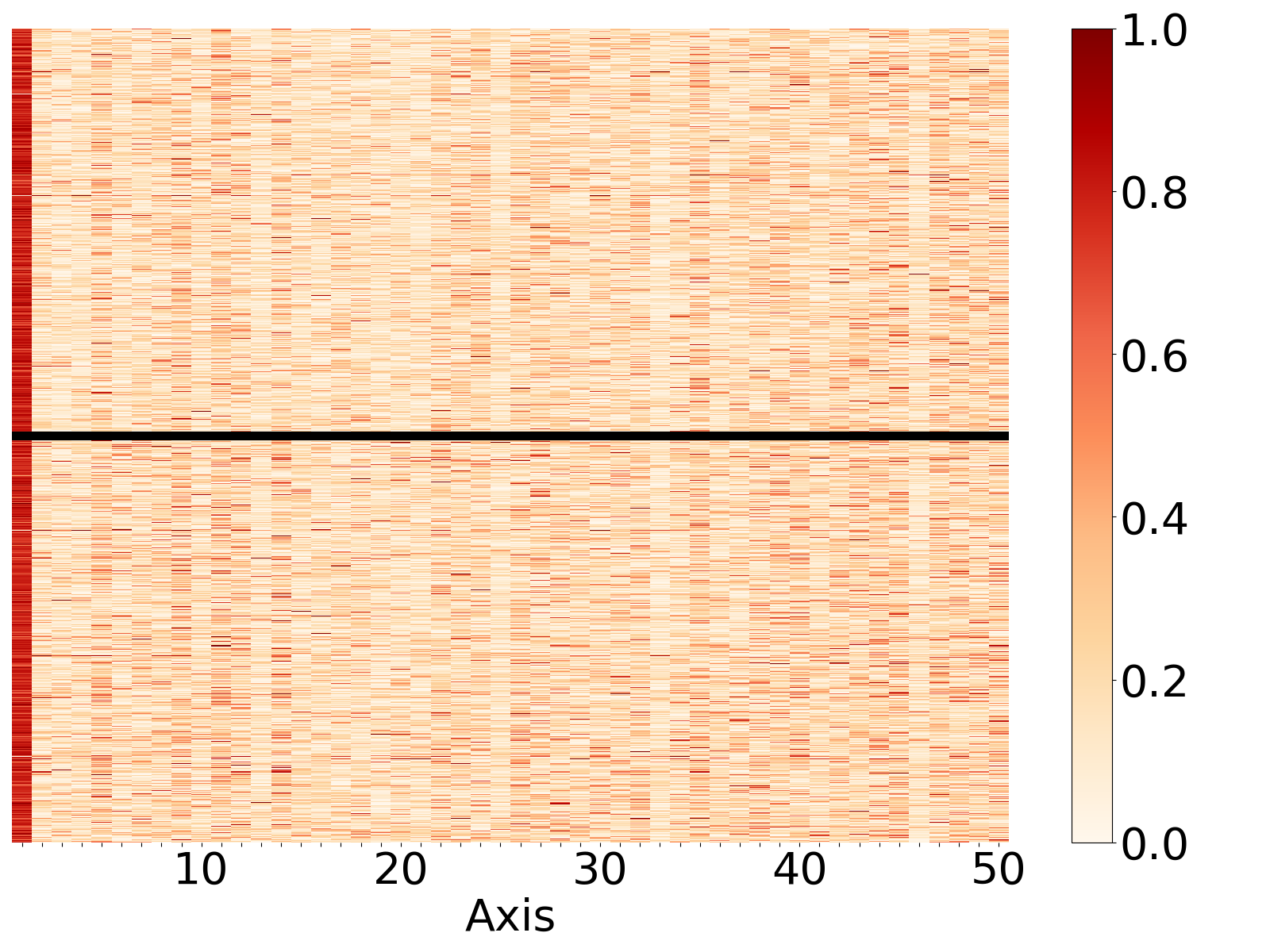}
        \subcaption{Pre-trained \ac{CWE}, PCA}
        \label{fig:wic_am2ico_ko_instances_pca_pretrained}
    \end{minipage}
    \begin{minipage}[b]{0.65\columnwidth}
        \centering
        \includegraphics[width=0.85\columnwidth]{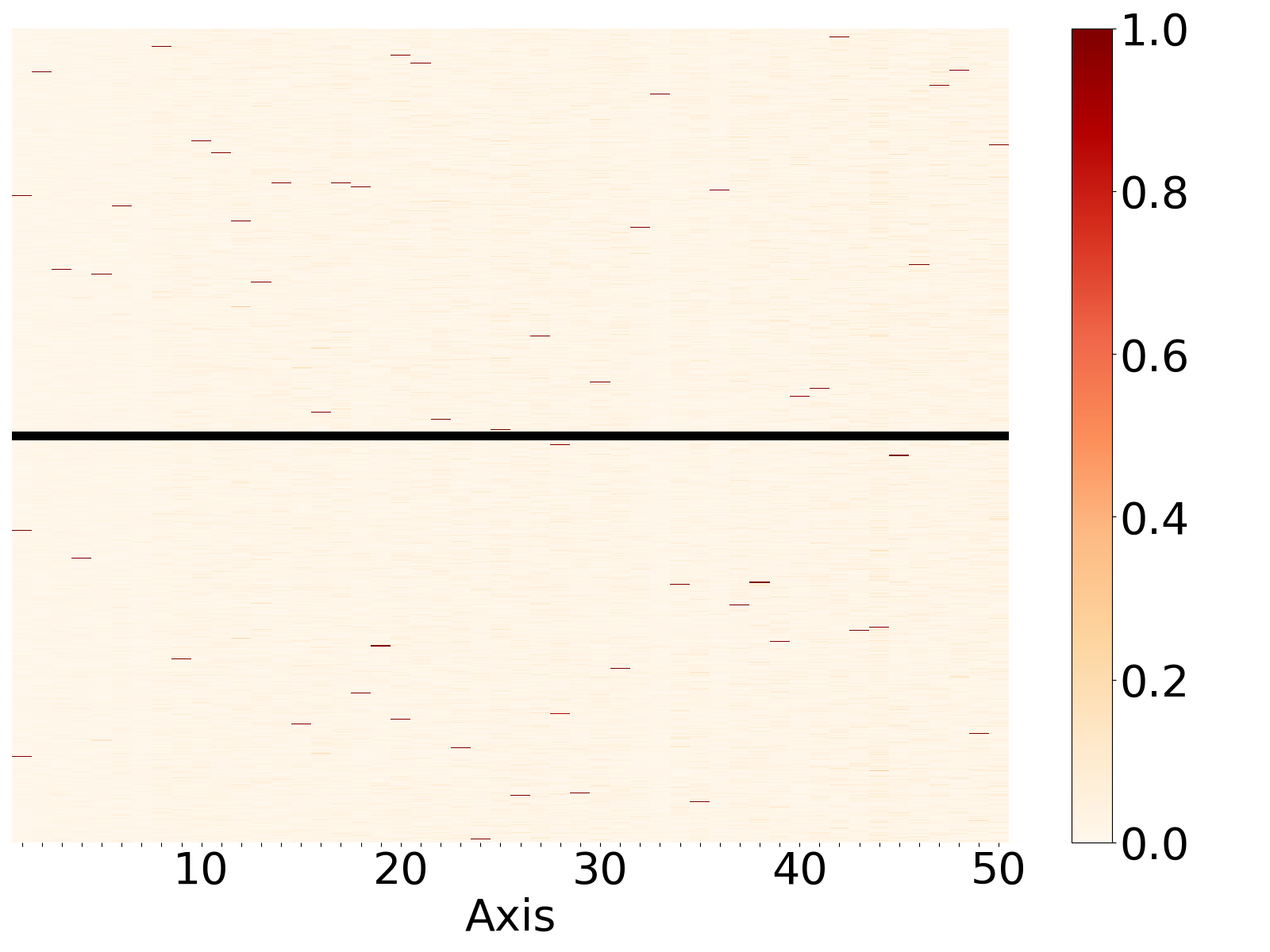}
        \subcaption{Pre-trained \ac{CWE}, ICA}
        \label{fig:wic_am2ico_ko_instances_ica_pretrained}
    \end{minipage} \\
    \begin{minipage}[b]{0.65\columnwidth}
        \centering
        \includegraphics[width=0.85\columnwidth]{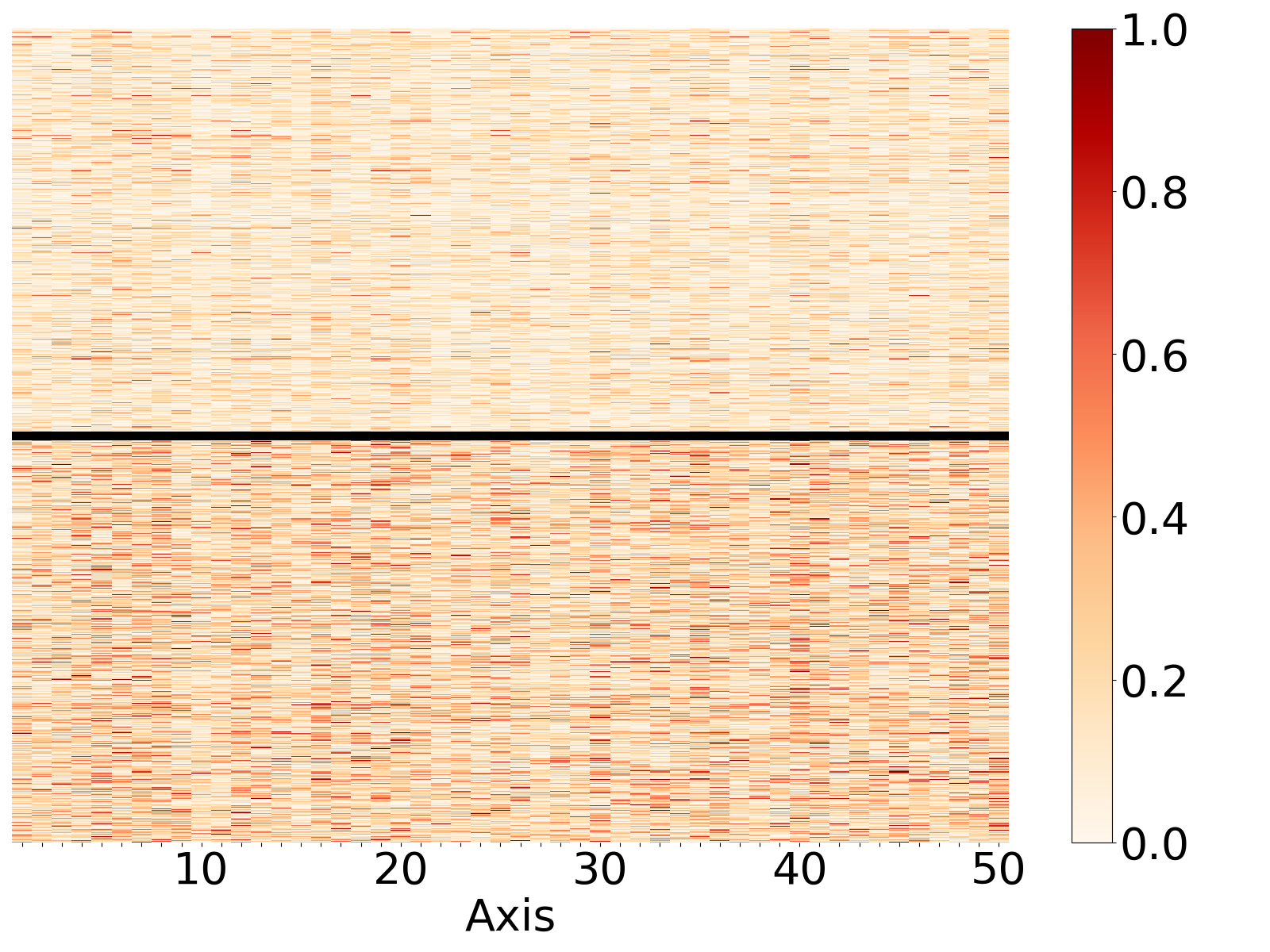}
        \subcaption{Fine-tuned \ac{SCWE}, Raw}
        \label{fig:wic_am2ico_ko_instances_raw_finetuned}
    \end{minipage}
    \begin{minipage}[b]{0.65\columnwidth}
        \centering
        \includegraphics[width=0.85\columnwidth]{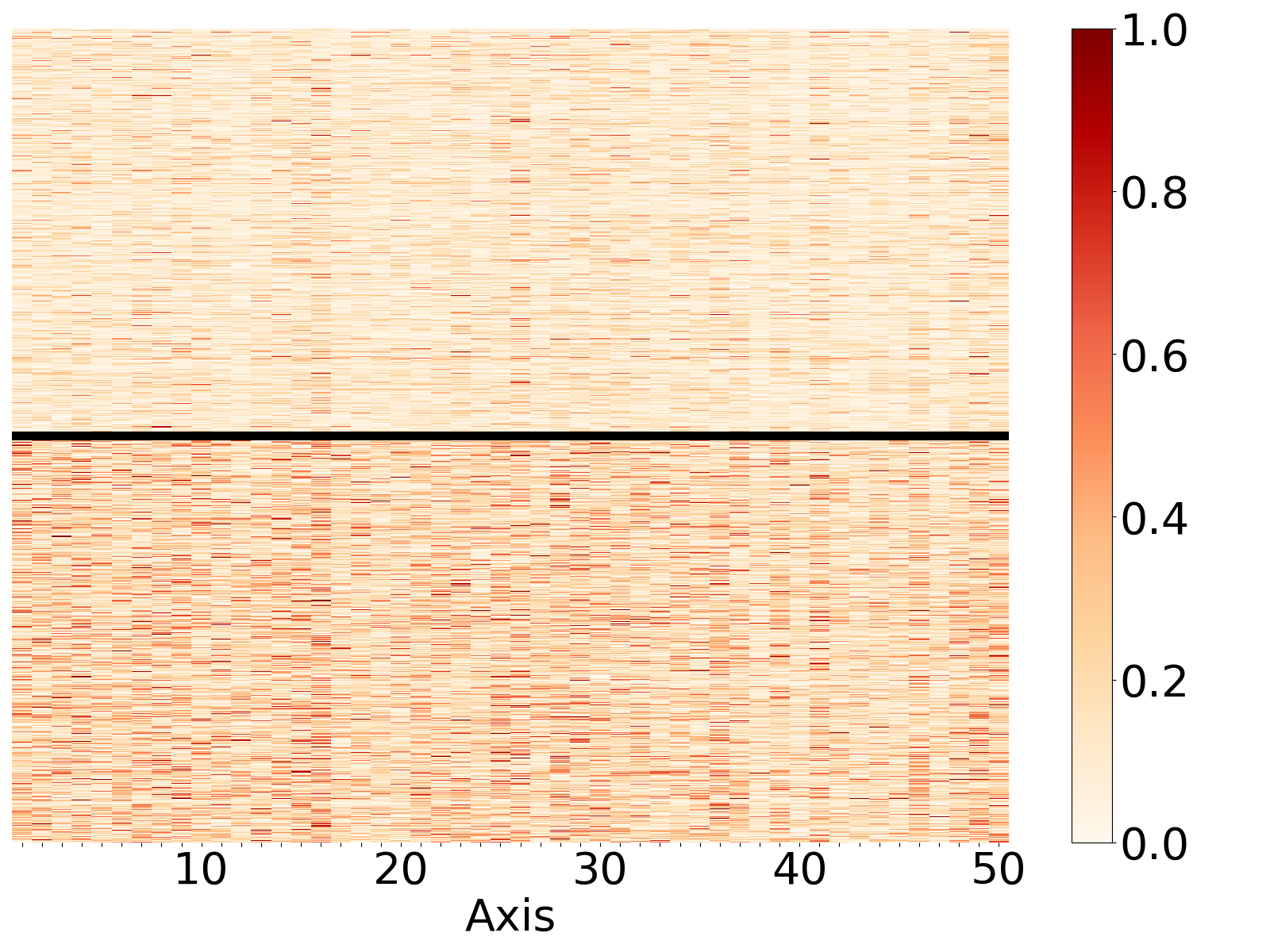}
        \subcaption{Fine-tuned \ac{SCWE}, PCA}
        \label{fig:wic_am2ico_ko_instances_pca_finetuned}
    \end{minipage}
    \begin{minipage}[b]{0.65\columnwidth}
        \centering
        \includegraphics[width=0.85\columnwidth]{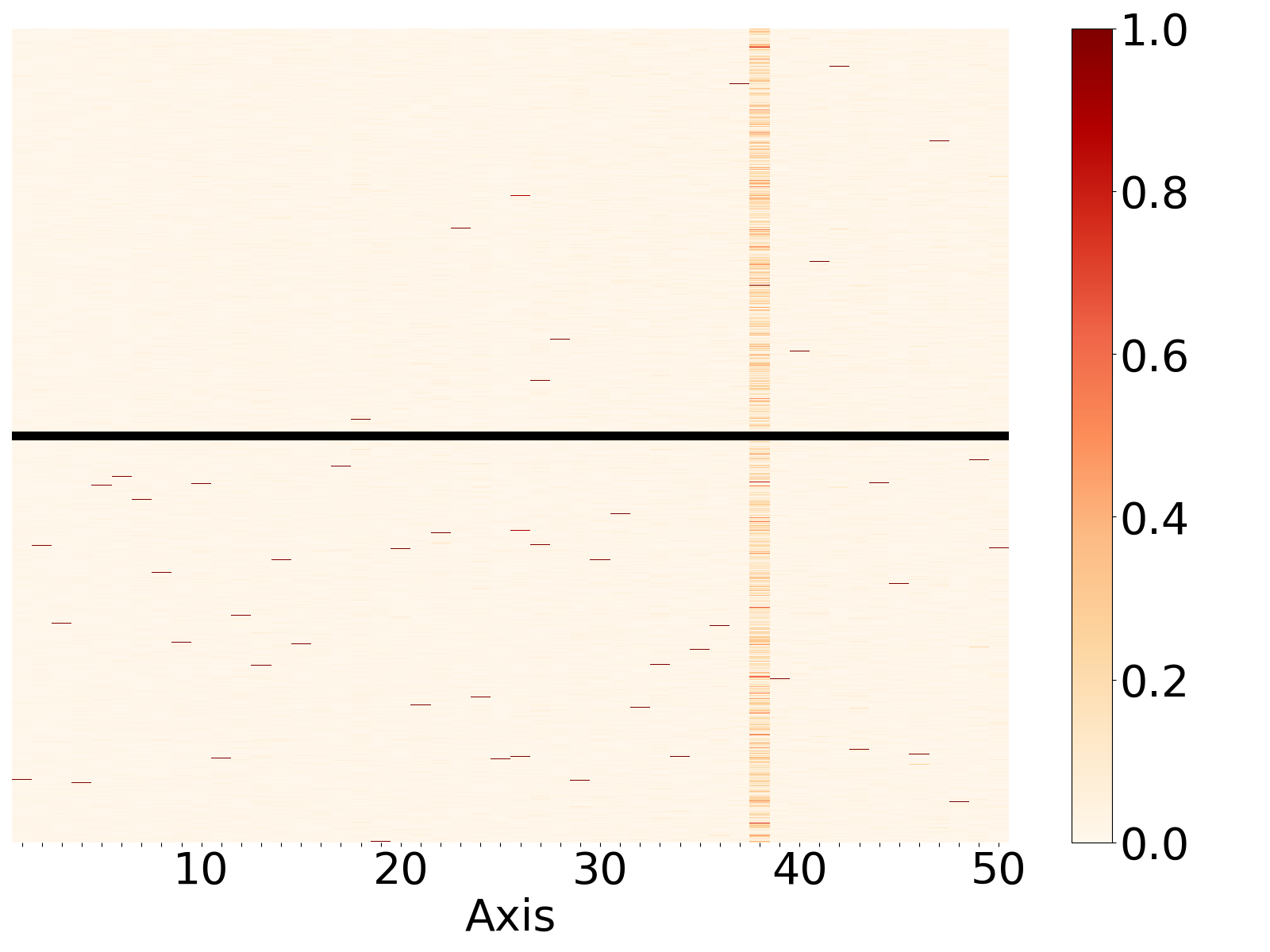}
        \subcaption{Fine-tuned \ac{SCWE}, ICA}
        \label{fig:wic_am2ico_ko_instances_ica_finetuned}
    \end{minipage}
    \caption{Visualisation of the top-50 dimensions of pre-trained \acp{CWE} (XLM-RoBERTa) and \acp{SCWE} (XL-LEXEME) for each instance in AM$^2$iCo (Korean) dataset, where the difference of vectors is calculated for (a/d) \textbf{Raw} vectors, (b/e) \ac{PCA}-transformed axes, and (c/f) \ac{ICA}-transformed axes. In each figure, the upper/lower half uses instances for the True/False labels.}
    \label{fig:wic_instance_am2ico_ko}
\end{figure*}

\begin{figure*}[t]
    \centering
    \begin{minipage}[b]{0.65\columnwidth}
        \centering
        \includegraphics[width=0.85\columnwidth]{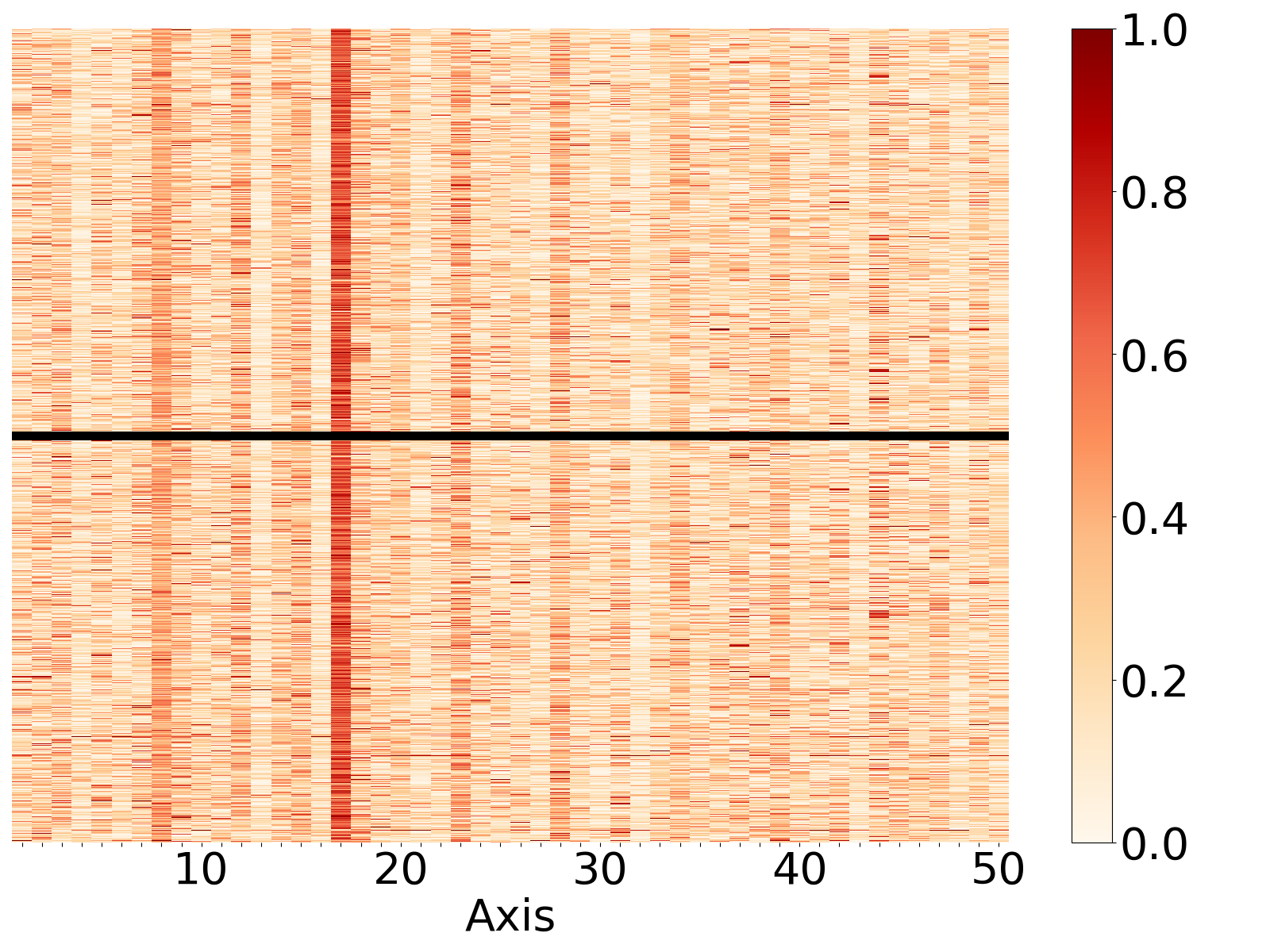}
        \subcaption{Pre-trained \ac{CWE}, Raw}
        \label{fig:wic_am2ico_fi_instances_raw_pretrained}
    \end{minipage}
    \begin{minipage}[b]{0.65\columnwidth}
        \centering
        \includegraphics[width=0.85\columnwidth]{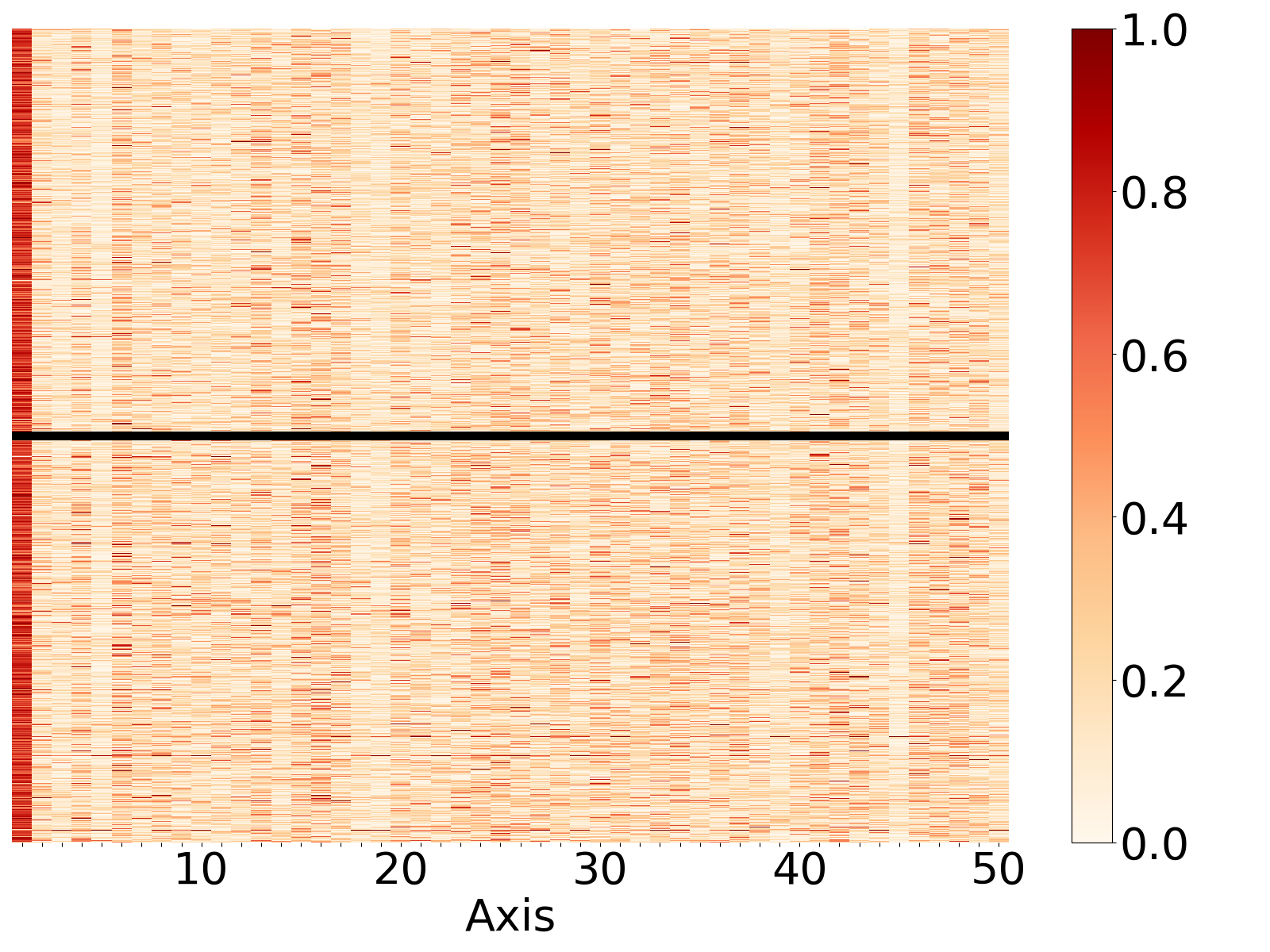}
        \subcaption{Pre-trained \ac{CWE}, PCA}
        \label{fig:wic_am2ico_fi_instances_pca_pretrained}
    \end{minipage}
    \begin{minipage}[b]{0.65\columnwidth}
        \centering
        \includegraphics[width=0.85\columnwidth]{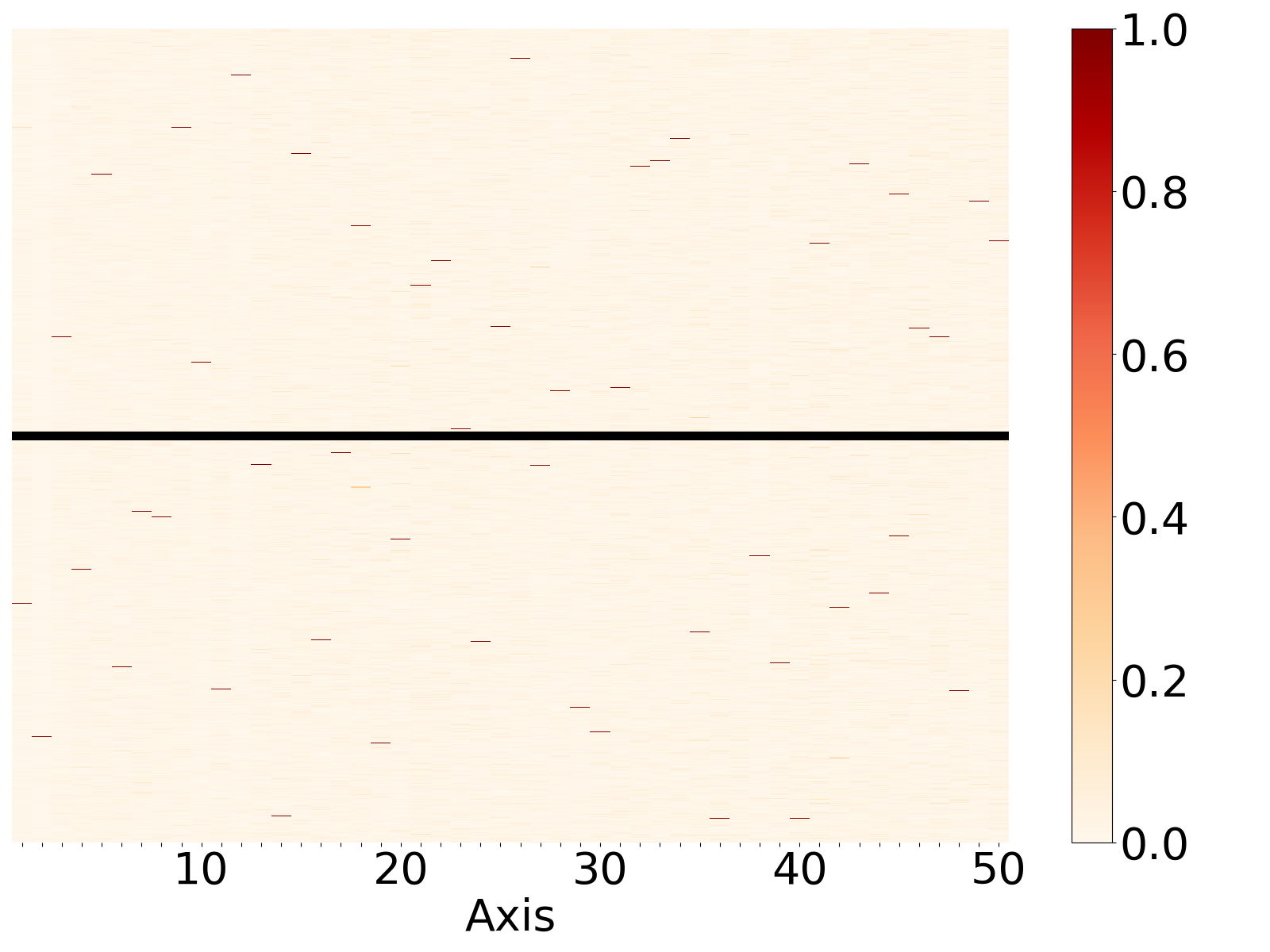}
        \subcaption{Pre-trained \ac{CWE}, ICA}
        \label{fig:wic_am2ico_fi_instances_ica_pretrained}
    \end{minipage} \\
    \begin{minipage}[b]{0.65\columnwidth}
        \centering
        \includegraphics[width=0.85\columnwidth]{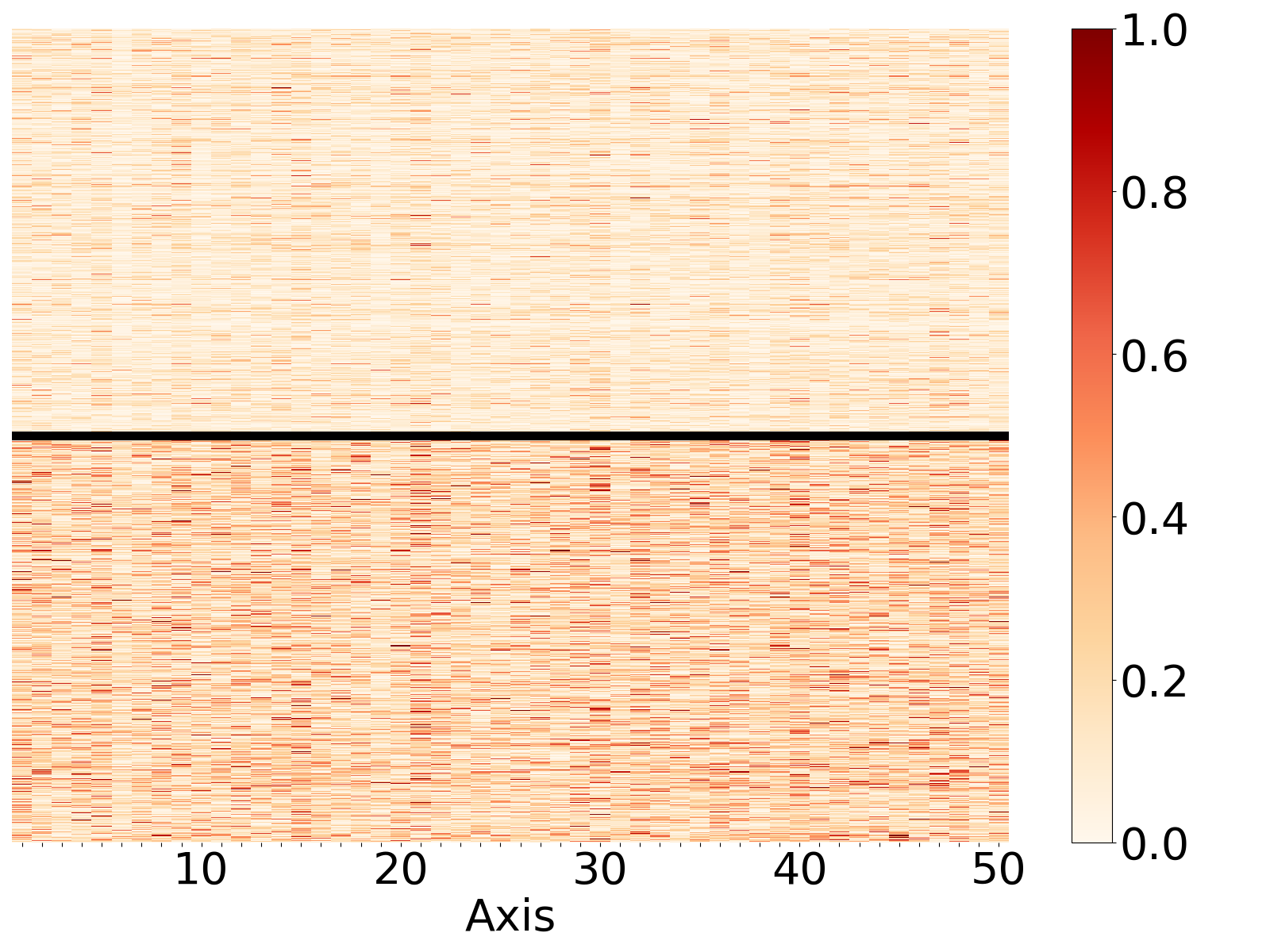}
        \subcaption{Fine-tuned \ac{SCWE}, Raw}
        \label{fig:wic_am2ico_fi_instances_raw_finetuned}
    \end{minipage}
    \begin{minipage}[b]{0.65\columnwidth}
        \centering
        \includegraphics[width=0.85\columnwidth]{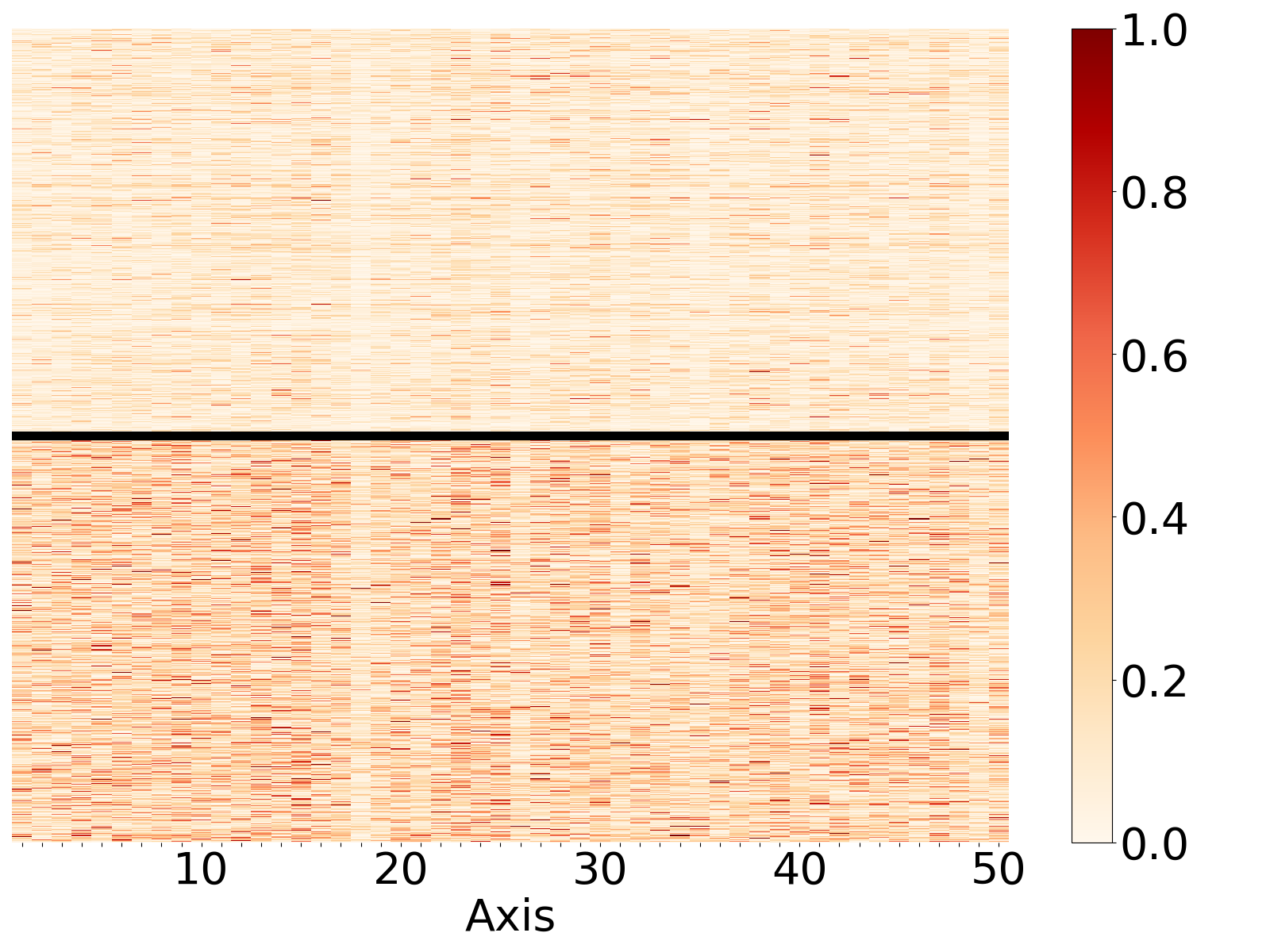}
        \subcaption{Fine-tuned \ac{SCWE}, PCA}
        \label{fig:wic_am2ico_fi_instances_pca_finetuned}
    \end{minipage}
    \begin{minipage}[b]{0.65\columnwidth}
        \centering
        \includegraphics[width=0.85\columnwidth]{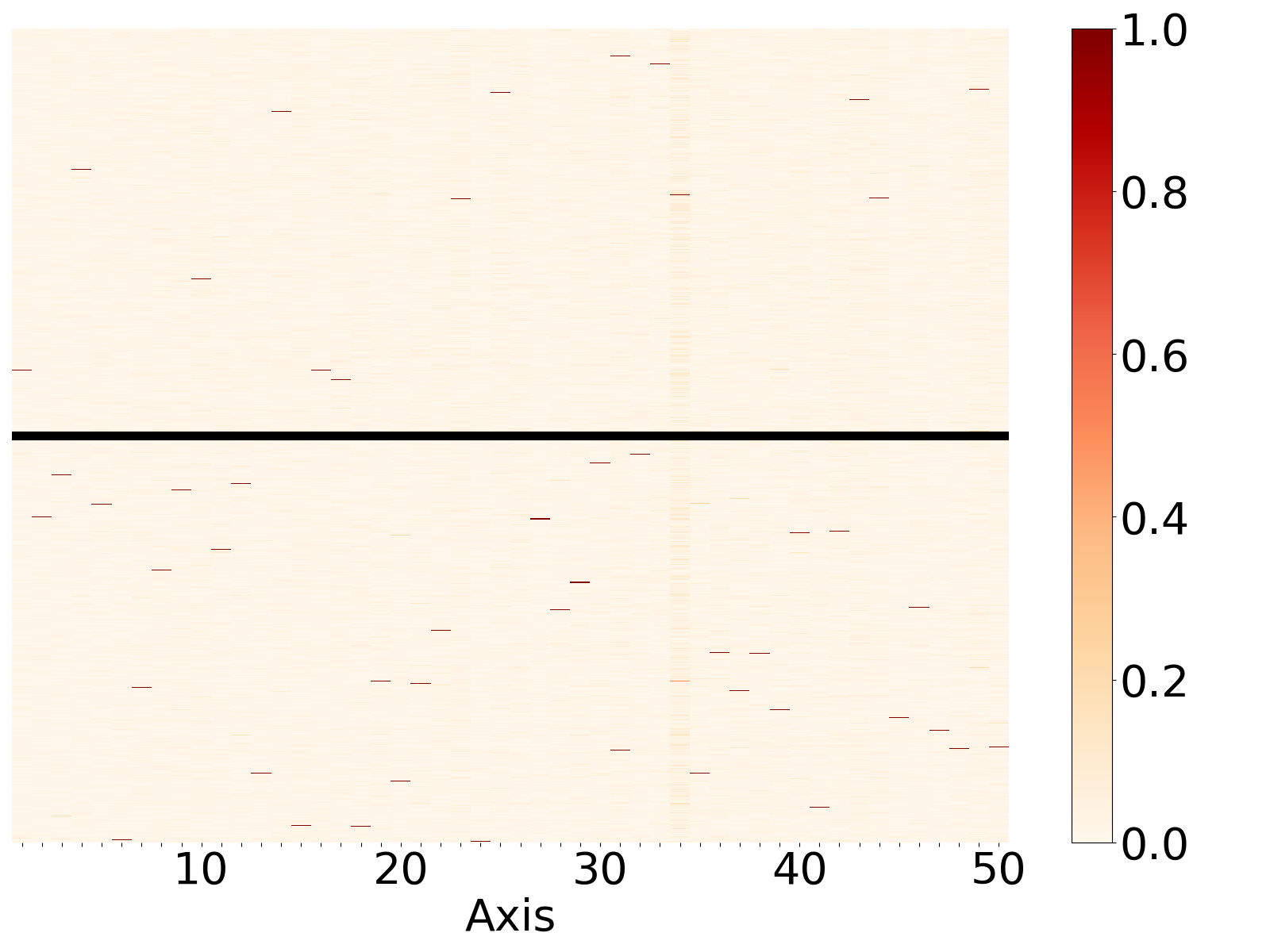}
        \subcaption{Fine-tuned \ac{SCWE}, ICA}
        \label{fig:wic_am2ico_fi_instances_ica_finetuned}
    \end{minipage}
    \caption{Visualisation of the top-50 dimensions of pre-trained \acp{CWE} (XLM-RoBERTa) and \acp{SCWE} (XL-LEXEME) for each instance in AM$^2$iCo (Finnish) dataset, where the difference of vectors is calculated for (a/d) \textbf{Raw} vectors, (b/e) \ac{PCA}-transformed axes, and (c/f) \ac{ICA}-transformed axes. In each figure, the upper/lower half uses instances for the True/False labels.}
    \label{fig:wic_instance_am2ico_fi}
\end{figure*}

\begin{figure*}[t]
    \centering
    \begin{minipage}[b]{0.65\columnwidth}
        \centering
        \includegraphics[width=0.85\columnwidth]{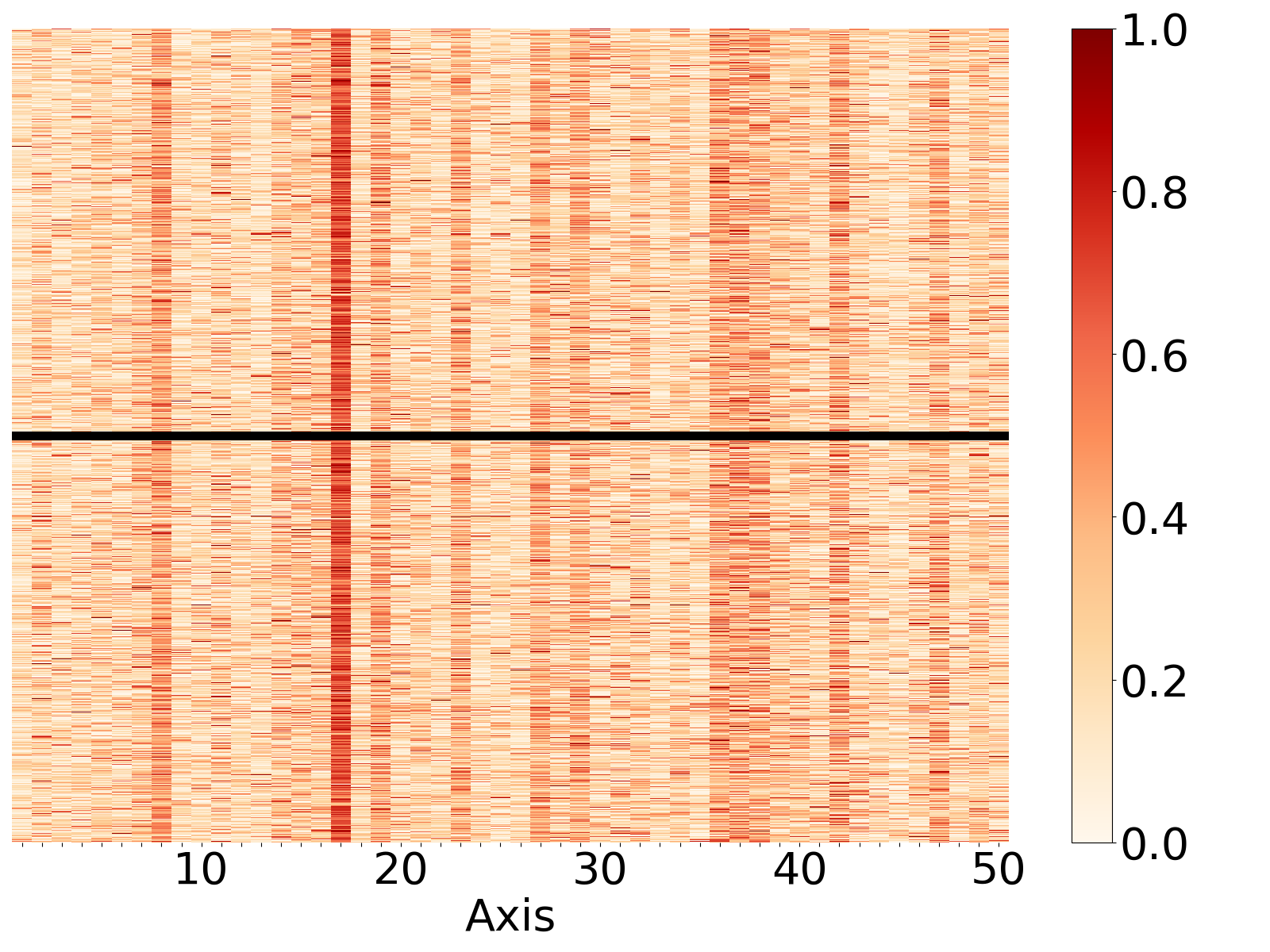}
        \subcaption{Pre-trained \ac{CWE}, Raw}
        \label{fig:wic_am2ico_tr_instances_raw_pretrained}
    \end{minipage}
    \begin{minipage}[b]{0.65\columnwidth}
        \centering
        \includegraphics[width=0.85\columnwidth]{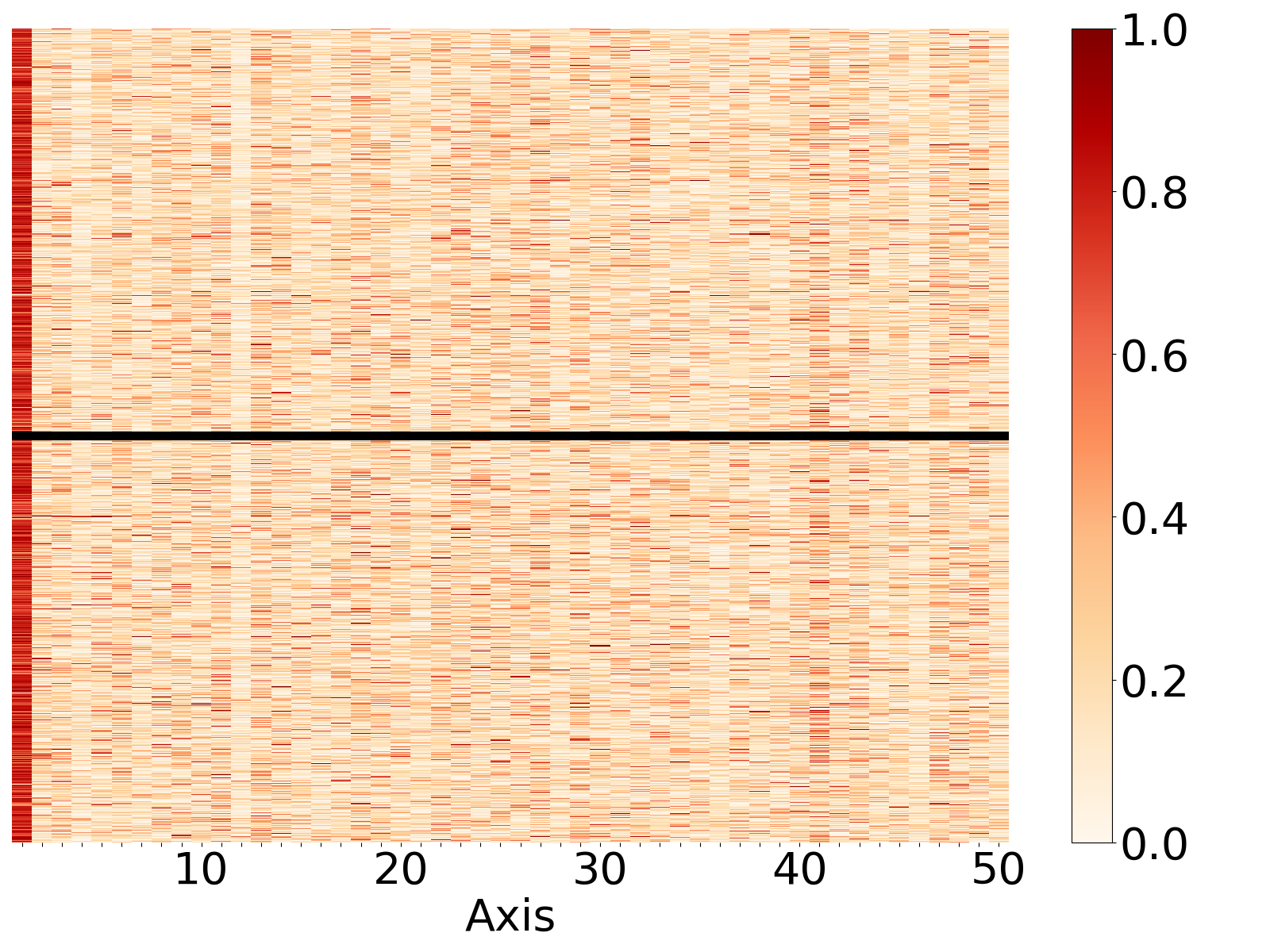}
        \subcaption{Pre-trained \ac{CWE}, PCA}
        \label{fig:wic_am2ico_tr_instances_pca_pretrained}
    \end{minipage}
    \begin{minipage}[b]{0.65\columnwidth}
        \centering
        \includegraphics[width=0.85\columnwidth]{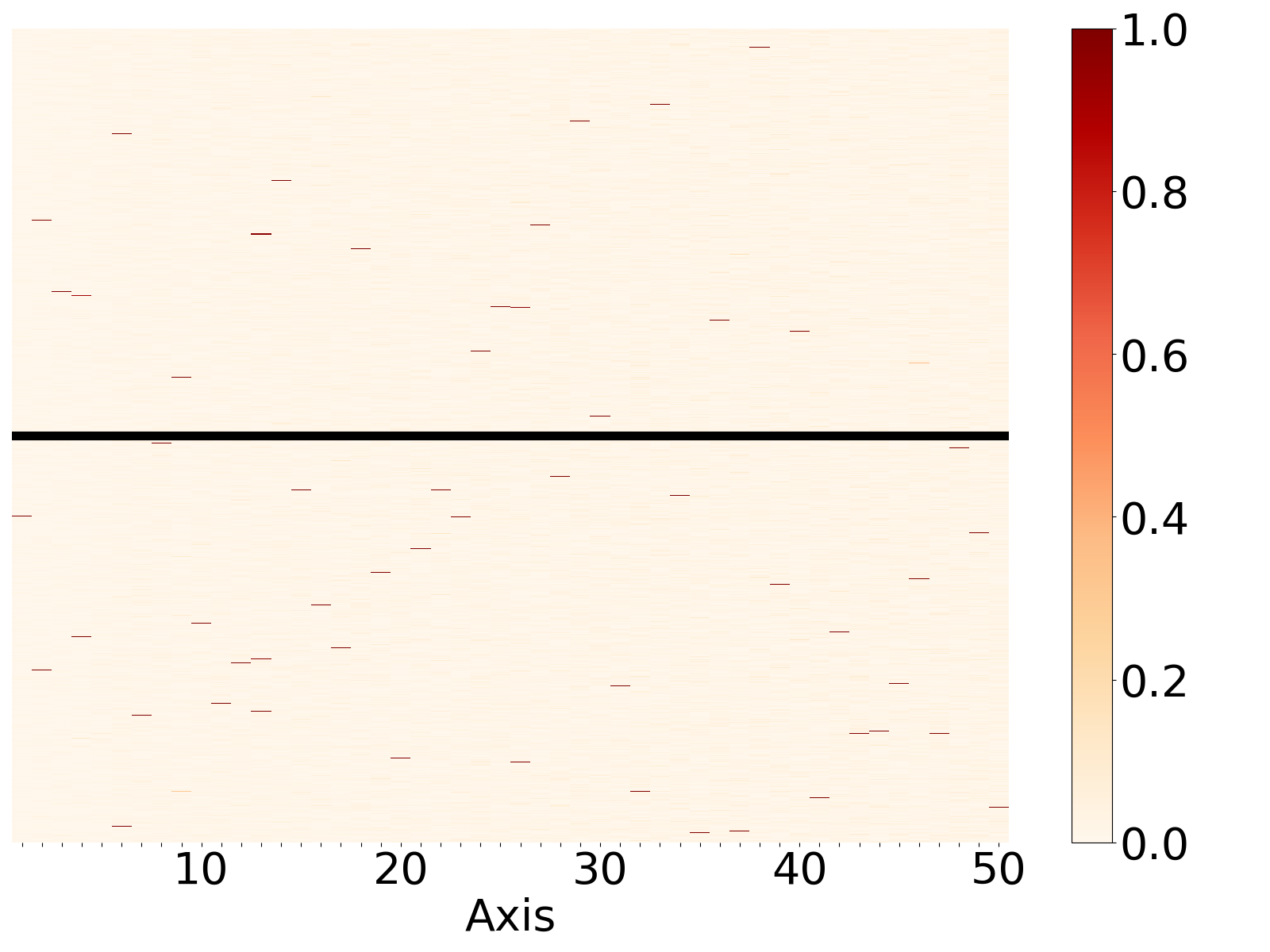}
        \subcaption{Pre-trained \ac{CWE}, ICA}
        \label{fig:wic_am2ico_tr_instances_ica_pretrained}
    \end{minipage} \\
    \begin{minipage}[b]{0.65\columnwidth}
        \centering
        \includegraphics[width=0.85\columnwidth]{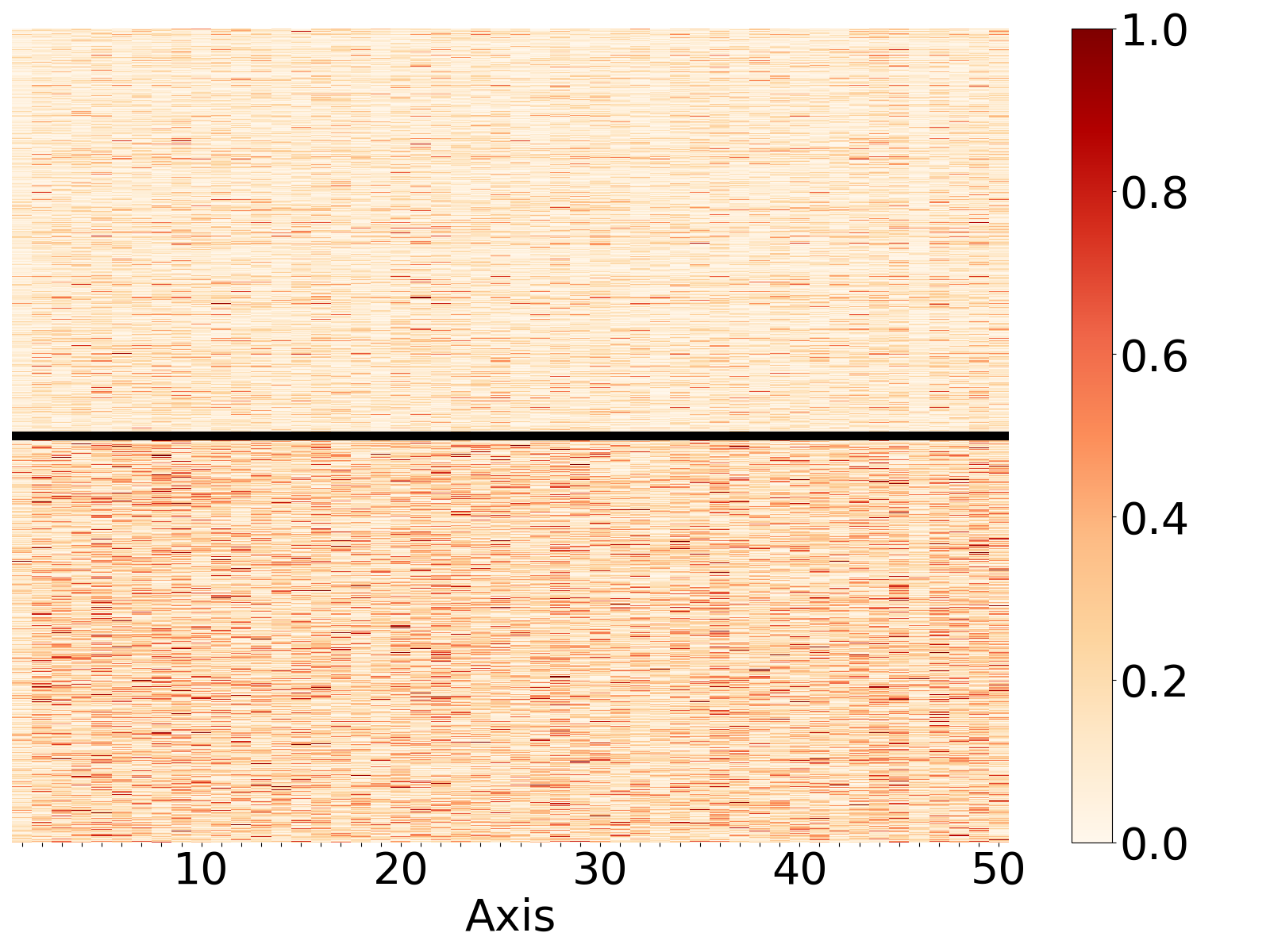}
        \subcaption{Fine-tuned \ac{SCWE}, Raw}
        \label{fig:wic_am2ico_tr_instances_raw_finetuned}
    \end{minipage}
    \begin{minipage}[b]{0.65\columnwidth}
        \centering
        \includegraphics[width=0.85\columnwidth]{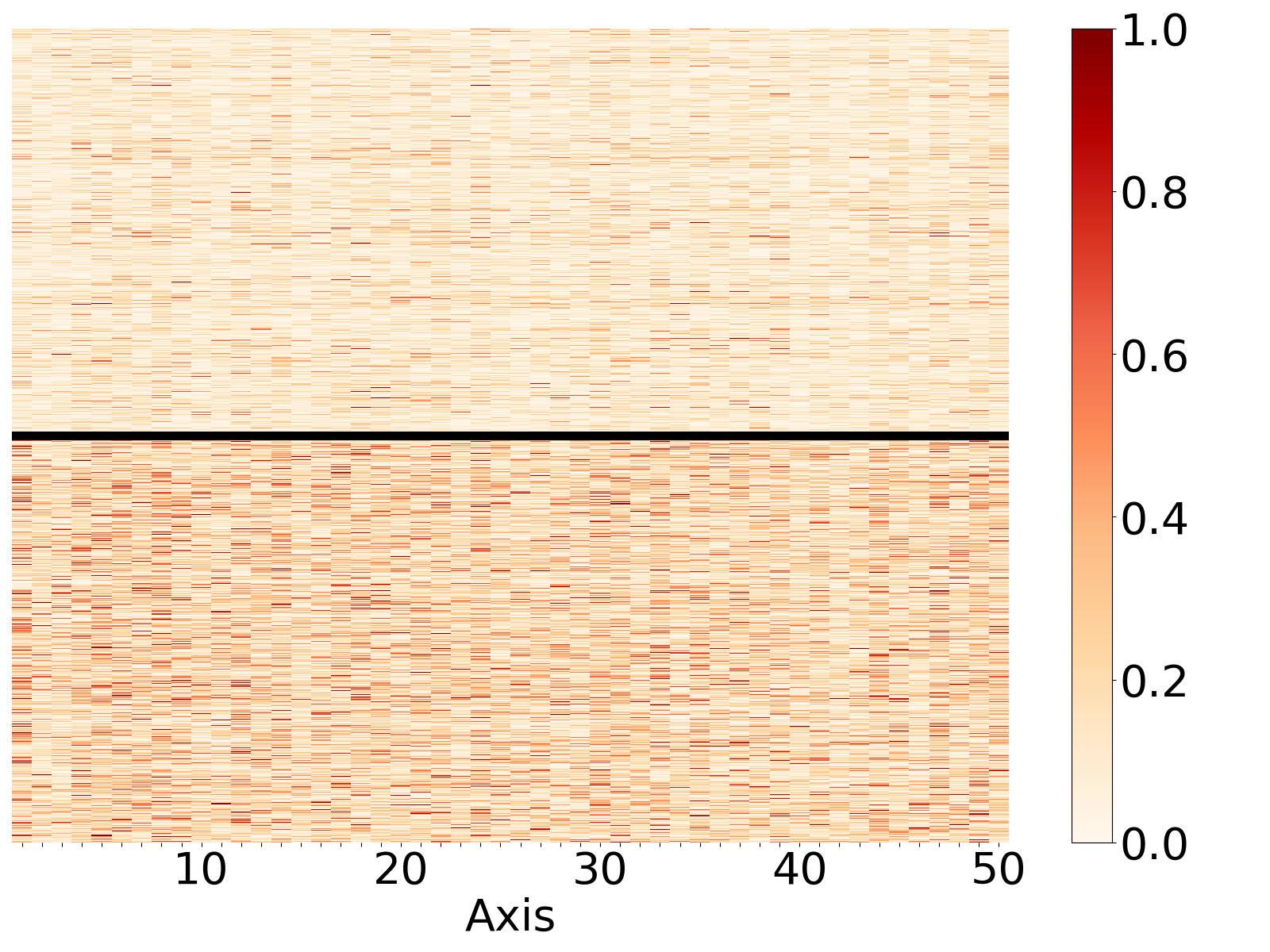}
        \subcaption{Fine-tuned \ac{SCWE}, PCA}
        \label{fig:wic_am2ico_tr_instances_pca_finetuned}
    \end{minipage}
    \begin{minipage}[b]{0.65\columnwidth}
        \centering
        \includegraphics[width=0.85\columnwidth]{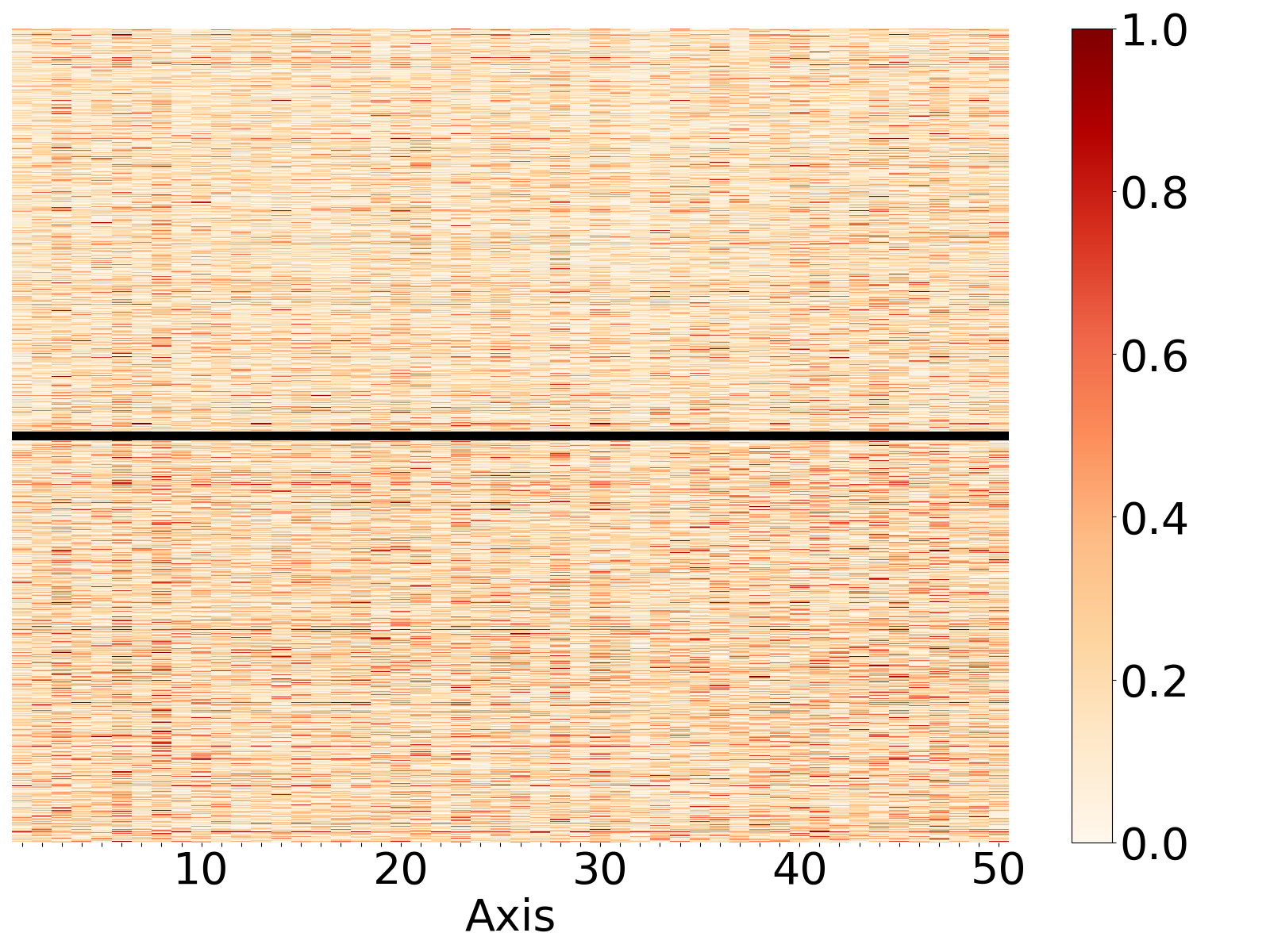}
        \subcaption{Fine-tuned \ac{SCWE}, ICA}
        \label{fig:wic_am2ico_tr_instances_ica_finetuned}
    \end{minipage}
    \caption{Visualisation of the top-50 dimensions of pre-trained \acp{CWE} (XLM-RoBERTa) and \acp{SCWE} (XL-LEXEME) for each instance in AM$^2$iCo (Turkish) dataset, where the difference of vectors is calculated for (a/d) \textbf{Raw} vectors, (b/e) \ac{PCA}-transformed axes, and (c/f) \ac{ICA}-transformed axes. In each figure, the upper/lower half uses instances for the True/False labels.}
    \label{fig:wic_instance_am2ico_tr}
\end{figure*}

\begin{figure*}[t]
    \centering
    \begin{minipage}[b]{0.65\columnwidth}
        \centering
        \includegraphics[width=0.85\columnwidth]{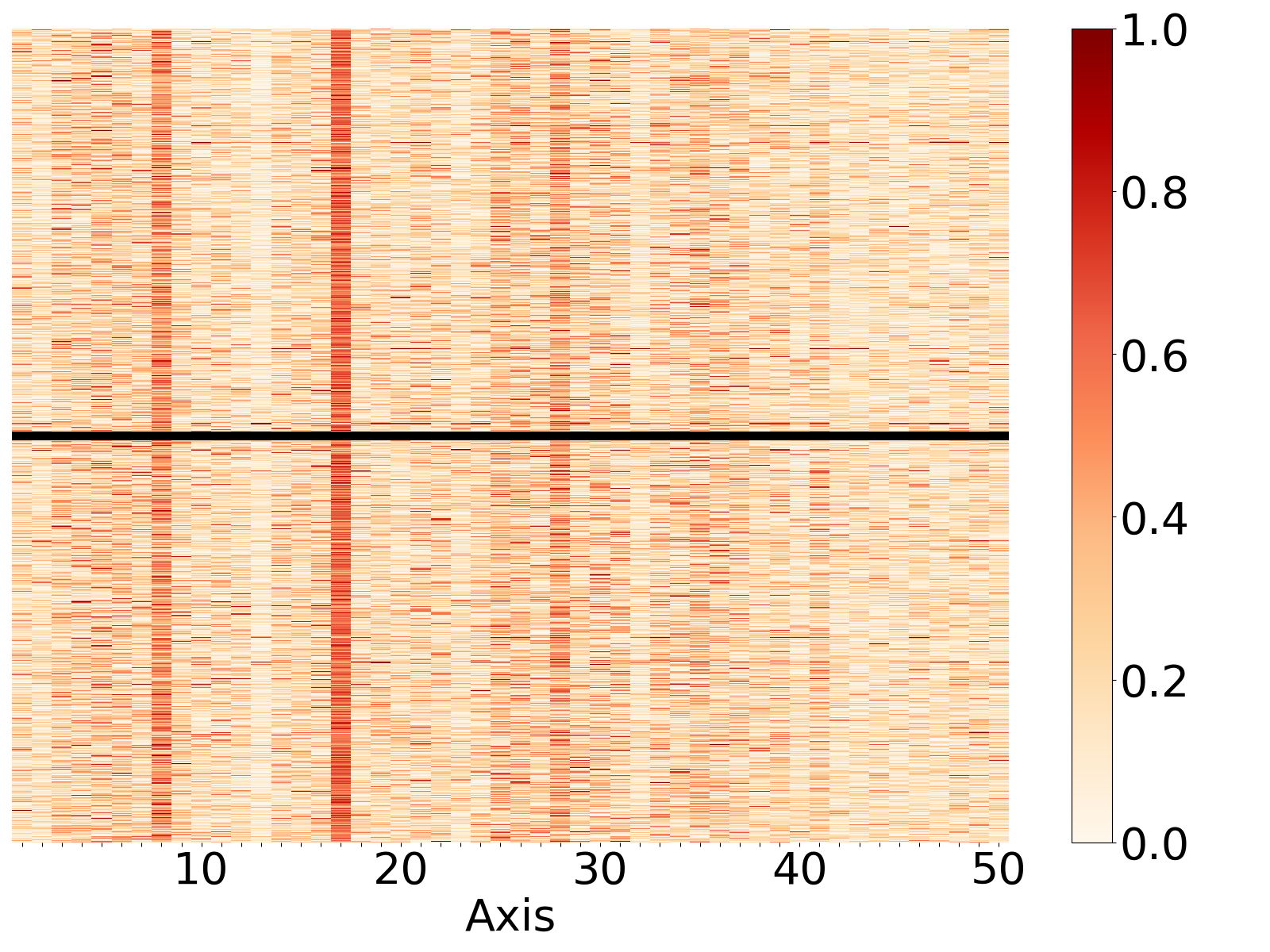}
        \subcaption{Pre-trained \ac{CWE}, Raw}
        \label{fig:wic_am2ico_id_instances_raw_pretrained}
    \end{minipage}
    \begin{minipage}[b]{0.65\columnwidth}
        \centering
        \includegraphics[width=0.85\columnwidth]{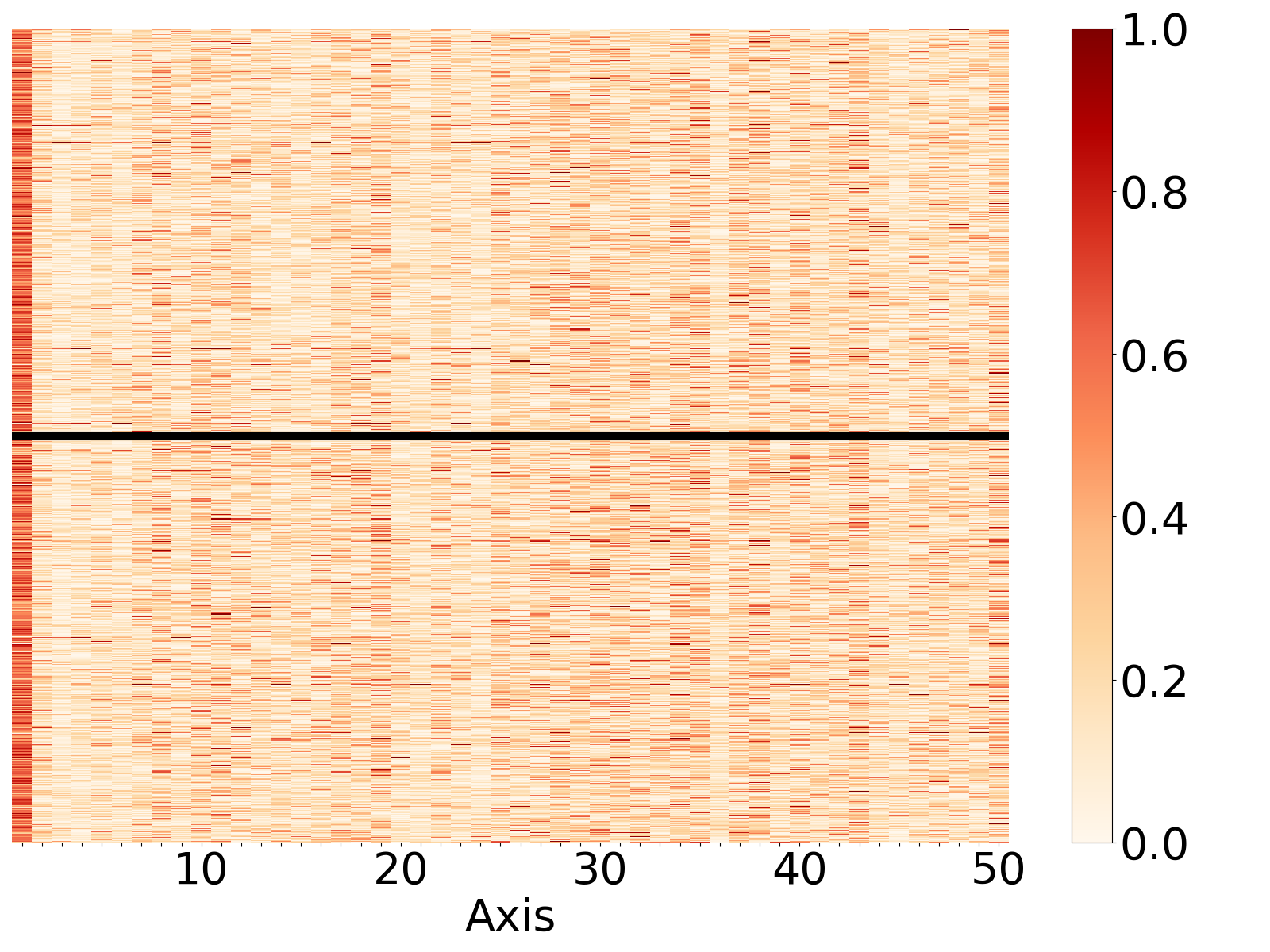}
        \subcaption{Pre-trained \ac{CWE}, PCA}
        \label{fig:wic_am2ico_id_instances_pca_pretrained}
    \end{minipage}
    \begin{minipage}[b]{0.65\columnwidth}
        \centering
        \includegraphics[width=0.85\columnwidth]{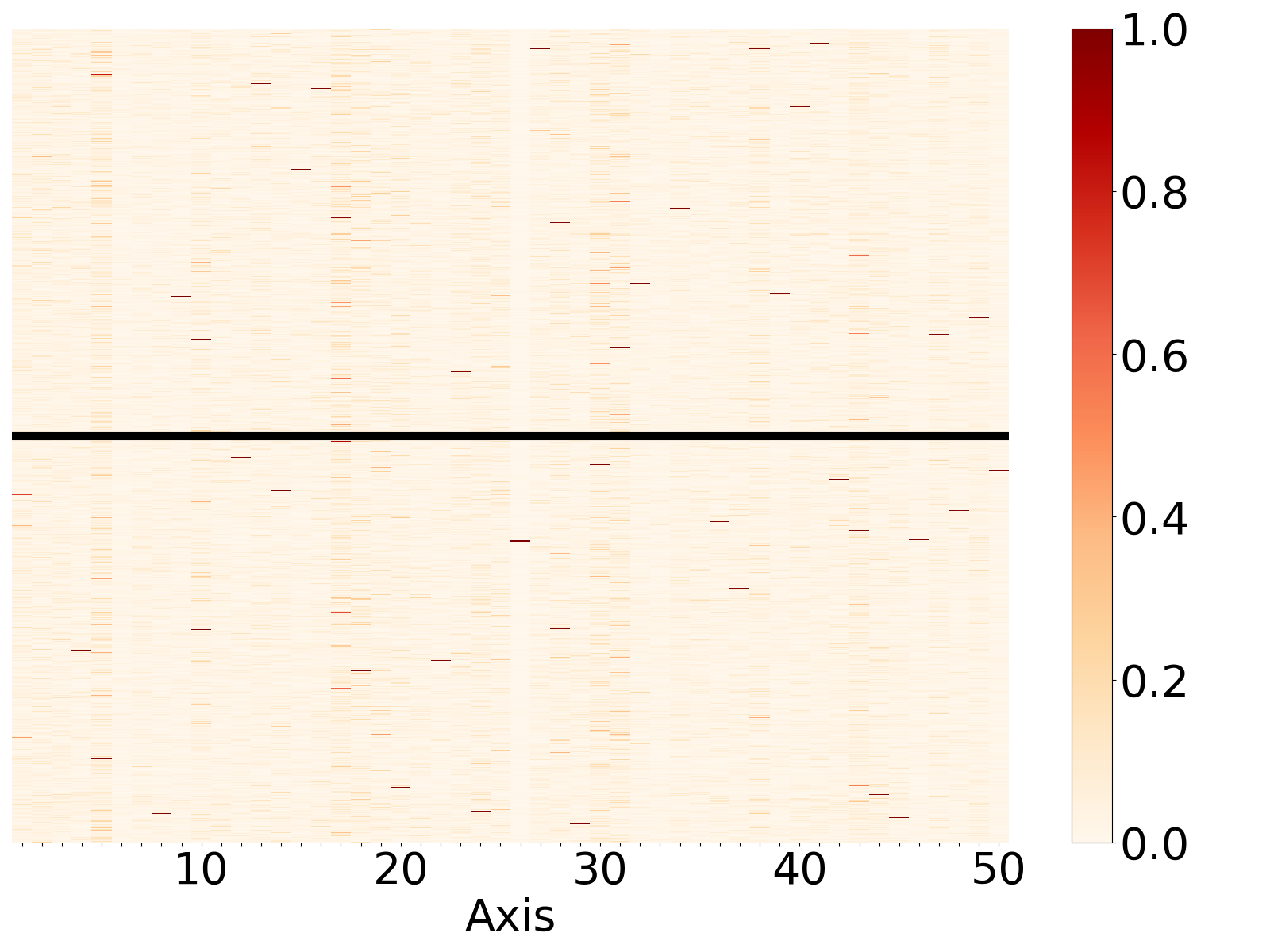}
        \subcaption{Pre-trained \ac{CWE}, ICA}
        \label{fig:wic_am2ico_id_instances_ica_pretrained}
    \end{minipage} \\
    \begin{minipage}[b]{0.65\columnwidth}
        \centering
        \includegraphics[width=0.85\columnwidth]{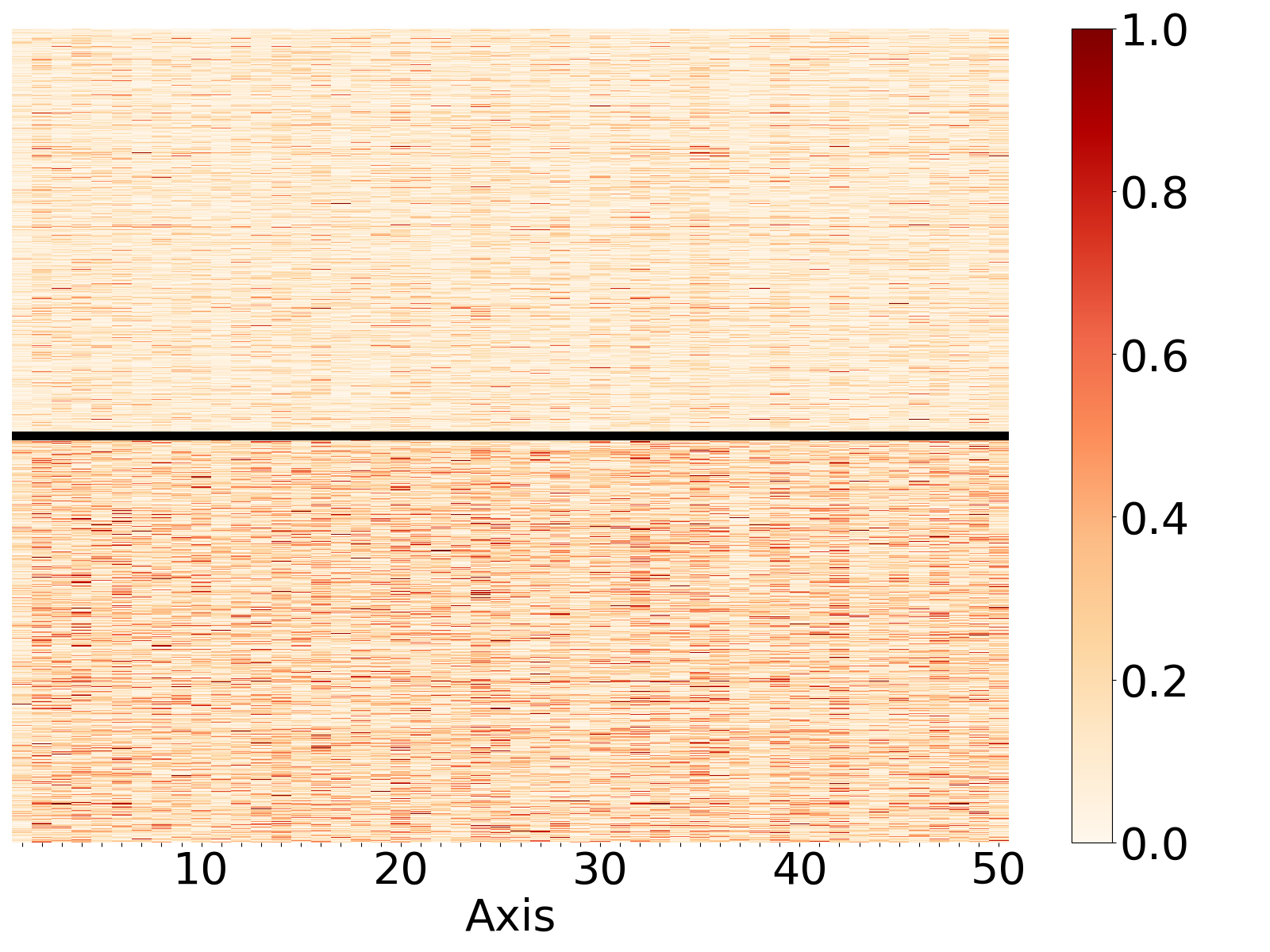}
        \subcaption{Fine-tuned \ac{SCWE}, Raw}
        \label{fig:wic_am2ico_id_instances_raw_finetuned}
    \end{minipage}
    \begin{minipage}[b]{0.65\columnwidth}
        \centering
        \includegraphics[width=0.85\columnwidth]{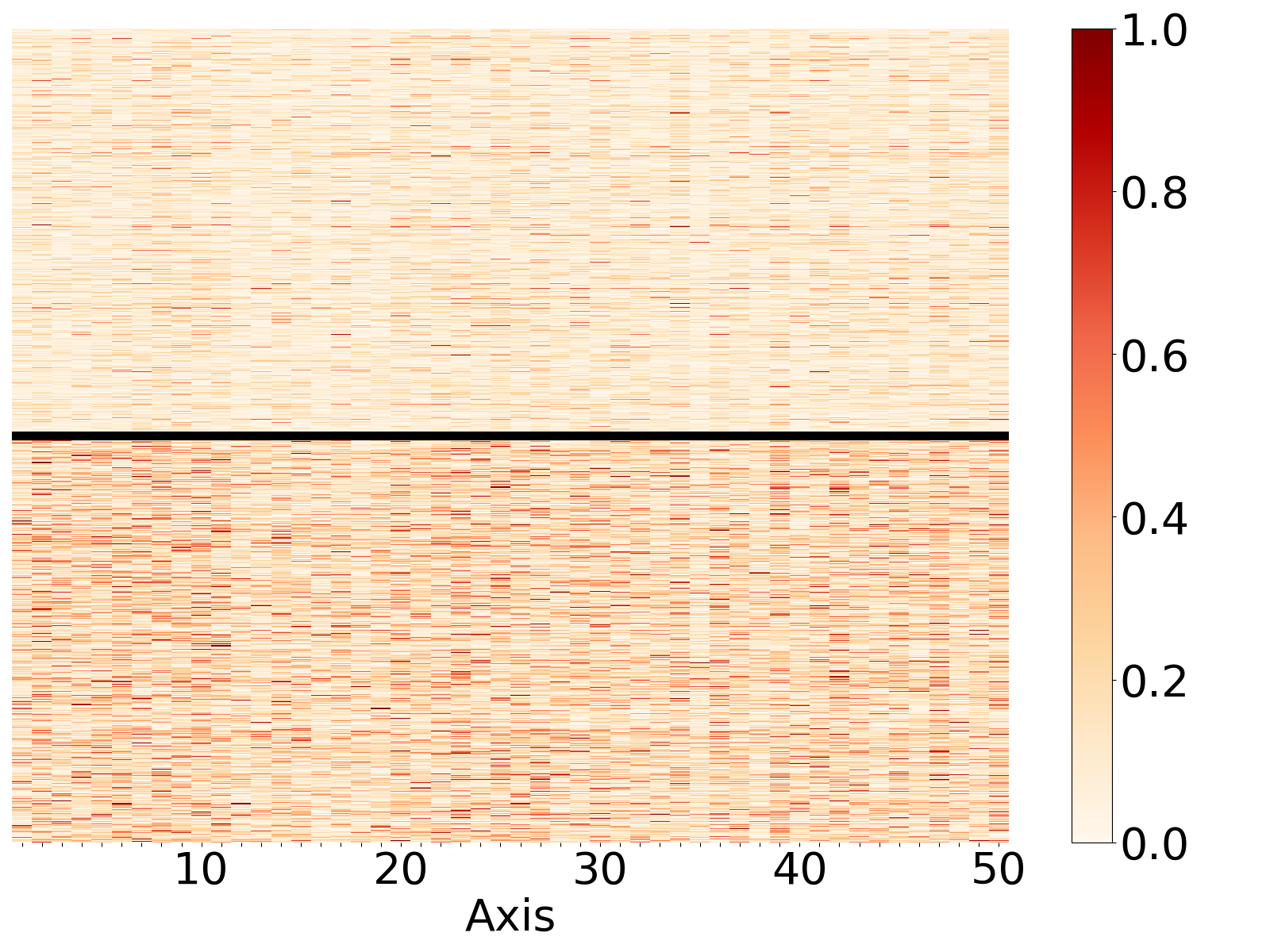}
        \subcaption{Fine-tuned \ac{SCWE}, PCA}
        \label{fig:wic_am2ico_id_instances_pca_finetuned}
    \end{minipage}
    \begin{minipage}[b]{0.65\columnwidth}
        \centering
        \includegraphics[width=0.85\columnwidth]{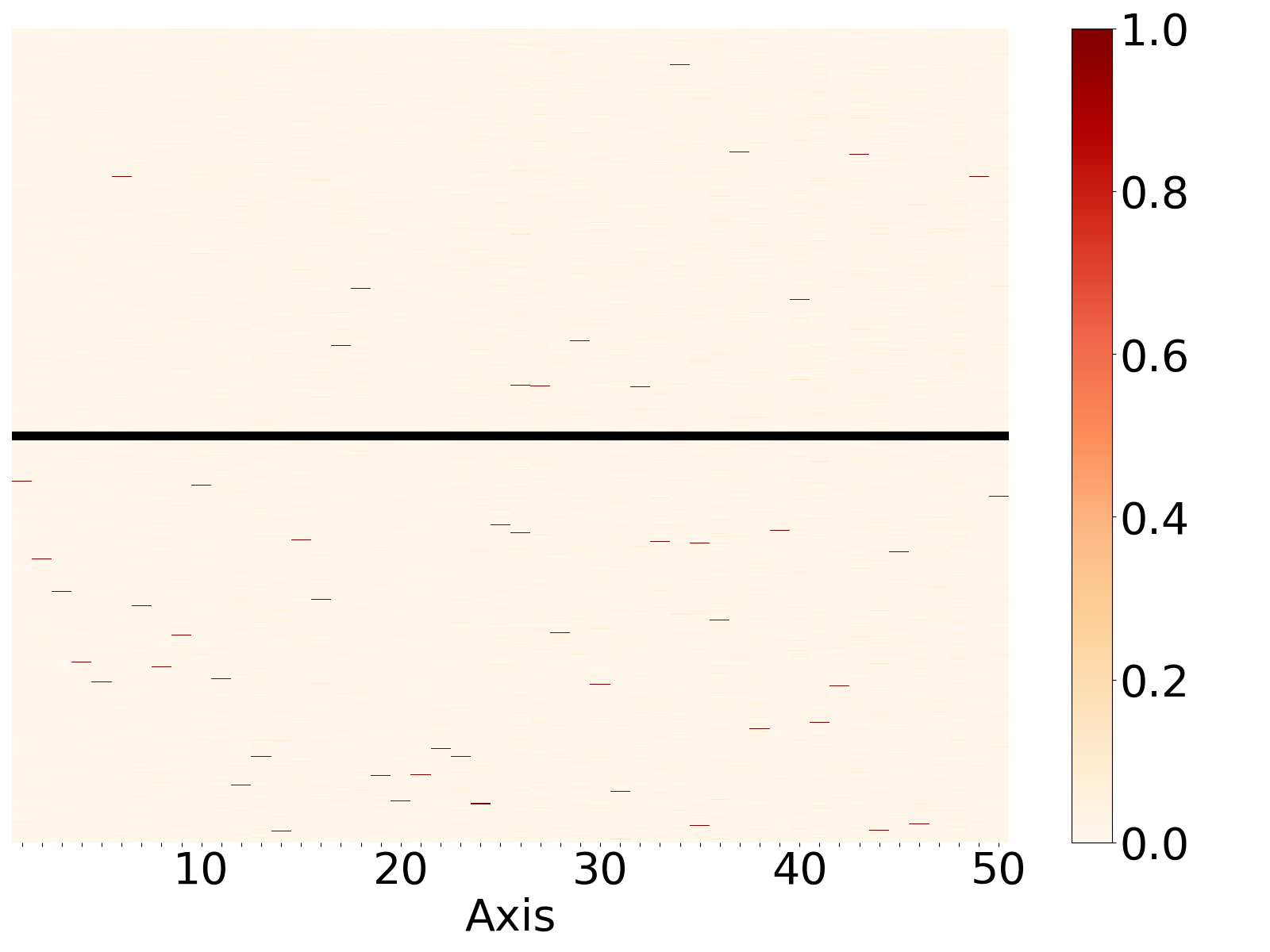}
        \subcaption{Fine-tuned \ac{SCWE}, ICA}
        \label{fig:wic_am2ico_id_instances_ica_finetuned}
    \end{minipage}
    \caption{Visualisation of the top-50 dimensions of pre-trained \acp{CWE} (XLM-RoBERTa) and \acp{SCWE} (XL-LEXEME) for each instance in AM$^2$iCo (Indonesian) dataset, where the difference of vectors is calculated for (a/d) \textbf{Raw} vectors, (b/e) \ac{PCA}-transformed axes, and (c/f) \ac{ICA}-transformed axes. In each figure, the upper/lower half uses instances for the True/False labels.}
    \label{fig:wic_instance_am2ico_id}
\end{figure*}

\begin{figure*}[t]
    \centering
    \begin{minipage}[b]{0.65\columnwidth}
        \centering
        \includegraphics[width=0.85\columnwidth]{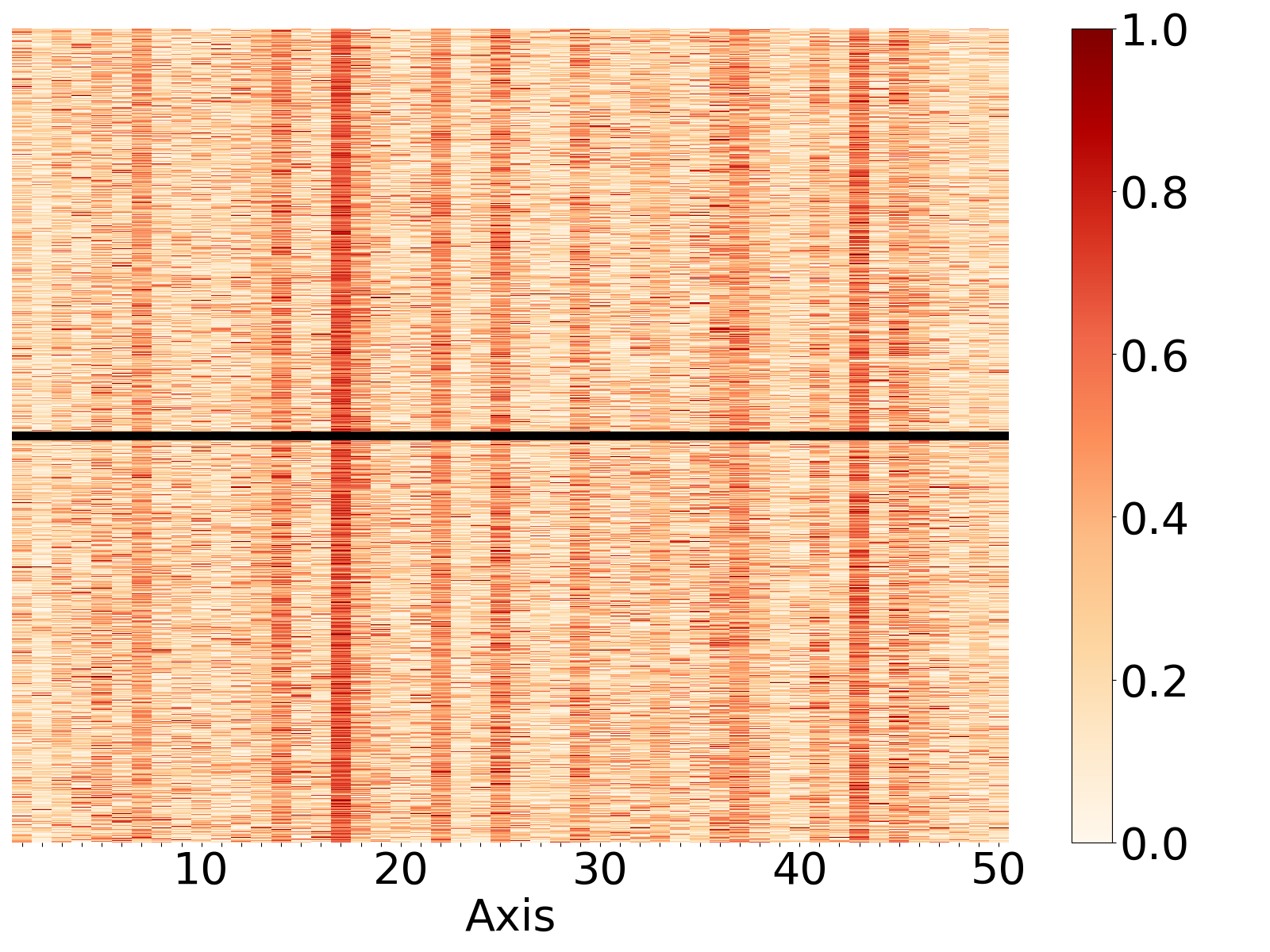}
        \subcaption{Pre-trained \ac{CWE}, Raw}
        \label{fig:wic_am2ico_eu_instances_raw_pretrained}
    \end{minipage}
    \begin{minipage}[b]{0.65\columnwidth}
        \centering
        \includegraphics[width=0.85\columnwidth]{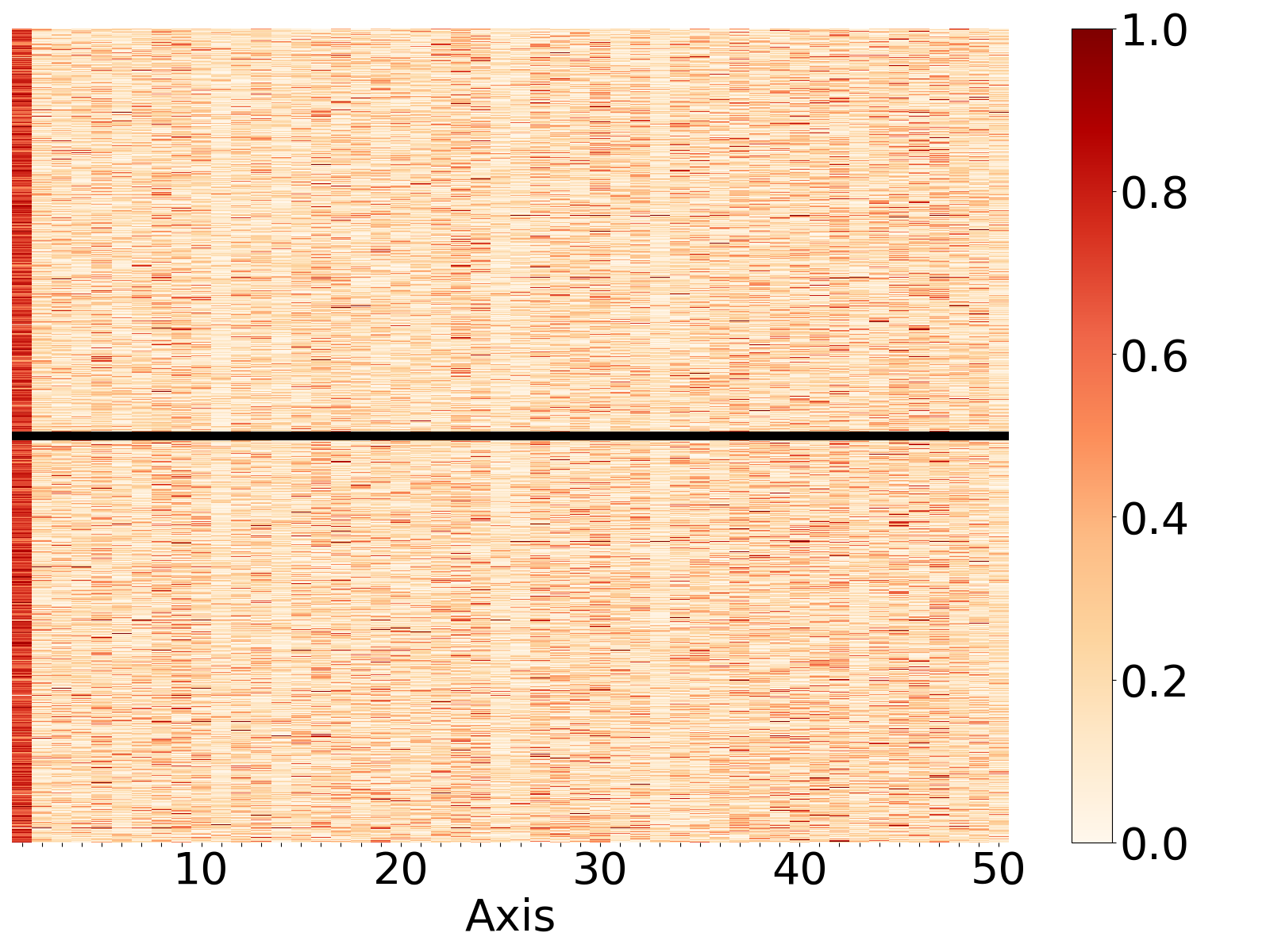}
        \subcaption{Pre-trained \ac{CWE}, PCA}
        \label{fig:wic_am2ico_eu_instances_pca_pretrained}
    \end{minipage}
    \begin{minipage}[b]{0.65\columnwidth}
        \centering
        \includegraphics[width=0.85\columnwidth]{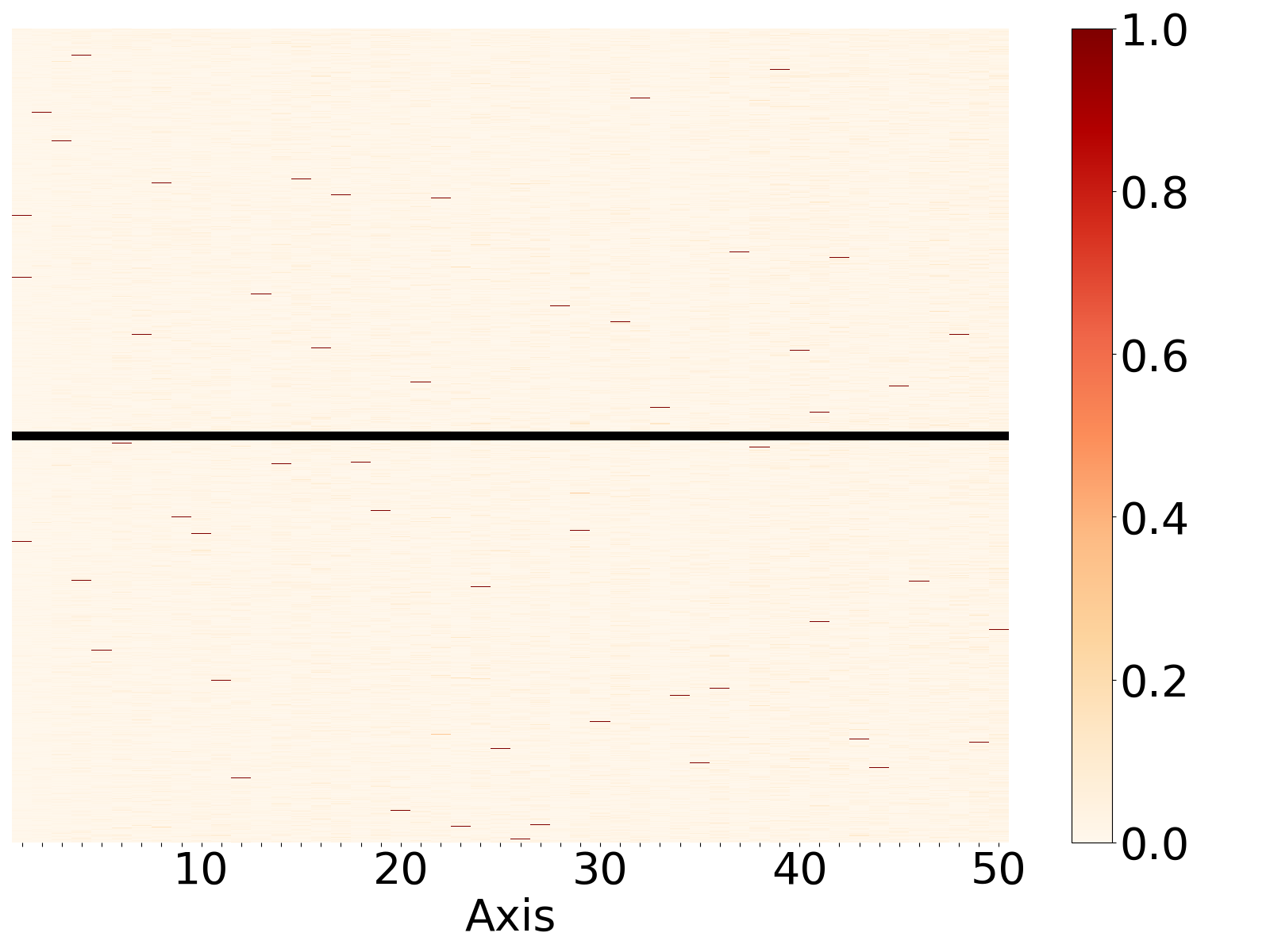}
        \subcaption{Pre-trained \ac{CWE}, ICA}
        \label{fig:wic_am2ico_eu_instances_ica_pretrained}
    \end{minipage} \\
    \begin{minipage}[b]{0.65\columnwidth}
        \centering
        \includegraphics[width=0.85\columnwidth]{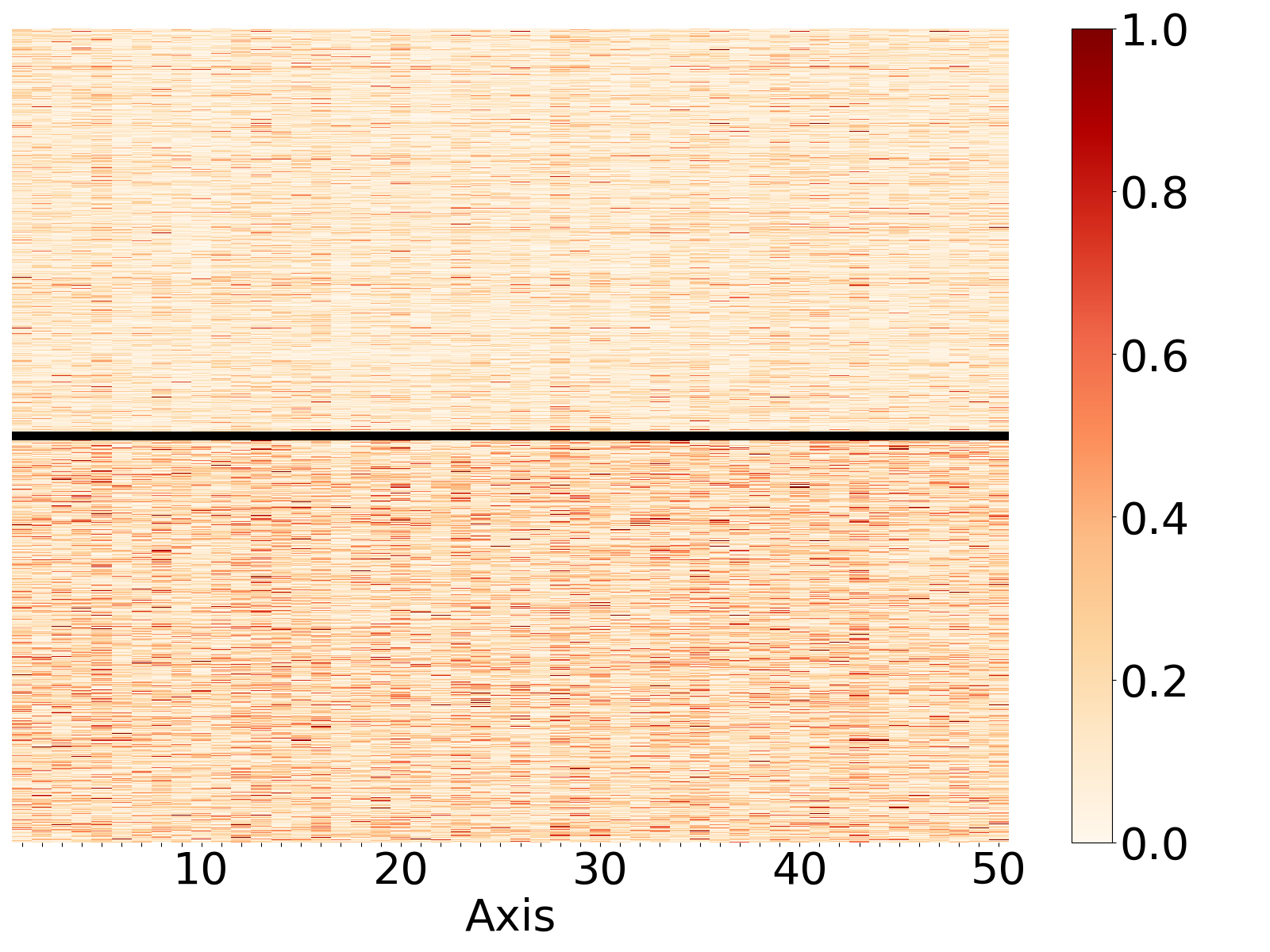}
        \subcaption{Fine-tuned \ac{SCWE}, Raw}
        \label{fig:wic_am2ico_eu_instances_raw_finetuned}
    \end{minipage}
    \begin{minipage}[b]{0.65\columnwidth}
        \centering
        \includegraphics[width=0.85\columnwidth]{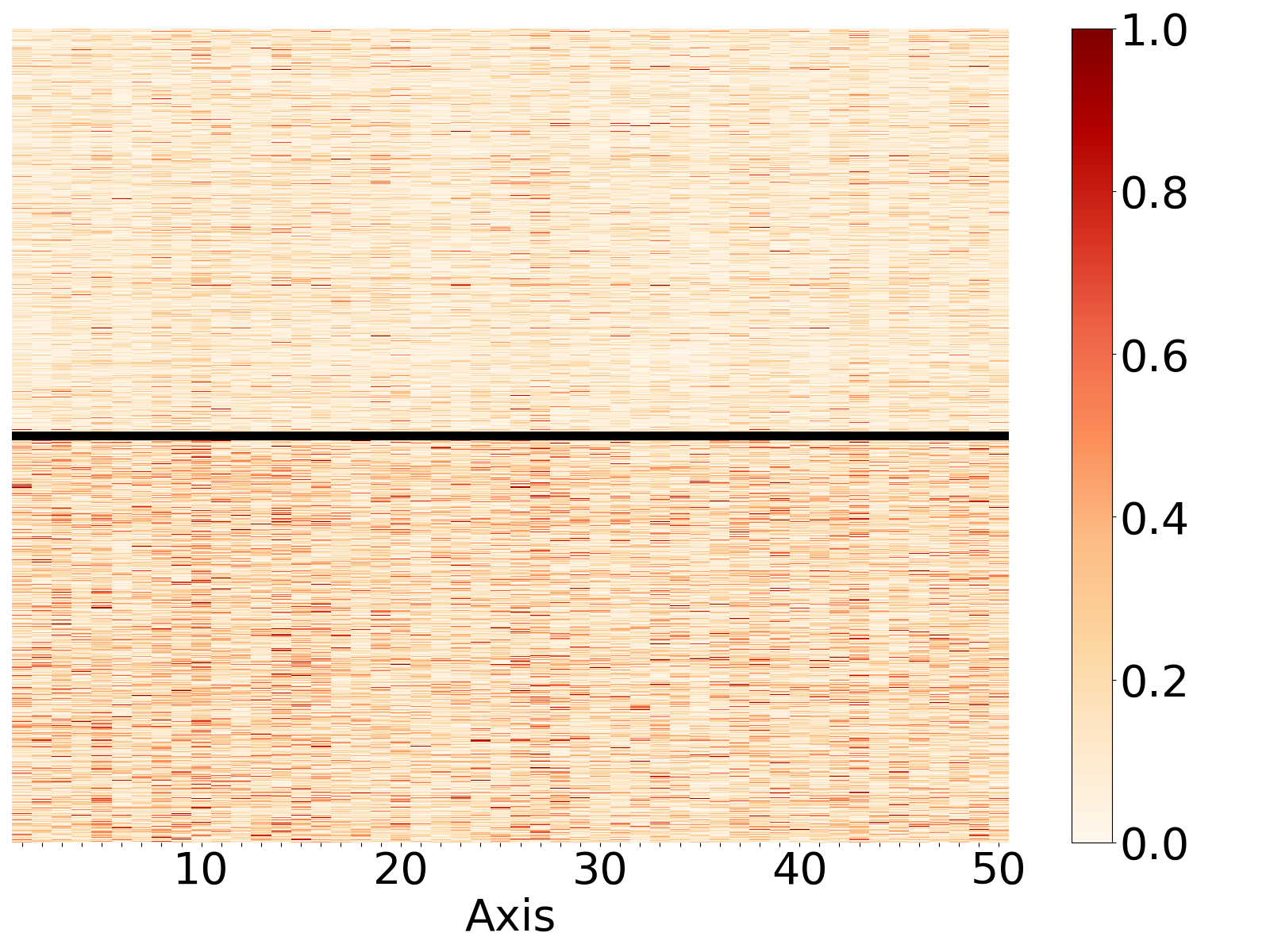}
        \subcaption{Fine-tuned \ac{SCWE}, PCA}
        \label{fig:wic_am2ico_eu_instances_pca_finetuned}
    \end{minipage}
    \begin{minipage}[b]{0.65\columnwidth}
        \centering
        \includegraphics[width=0.85\columnwidth]{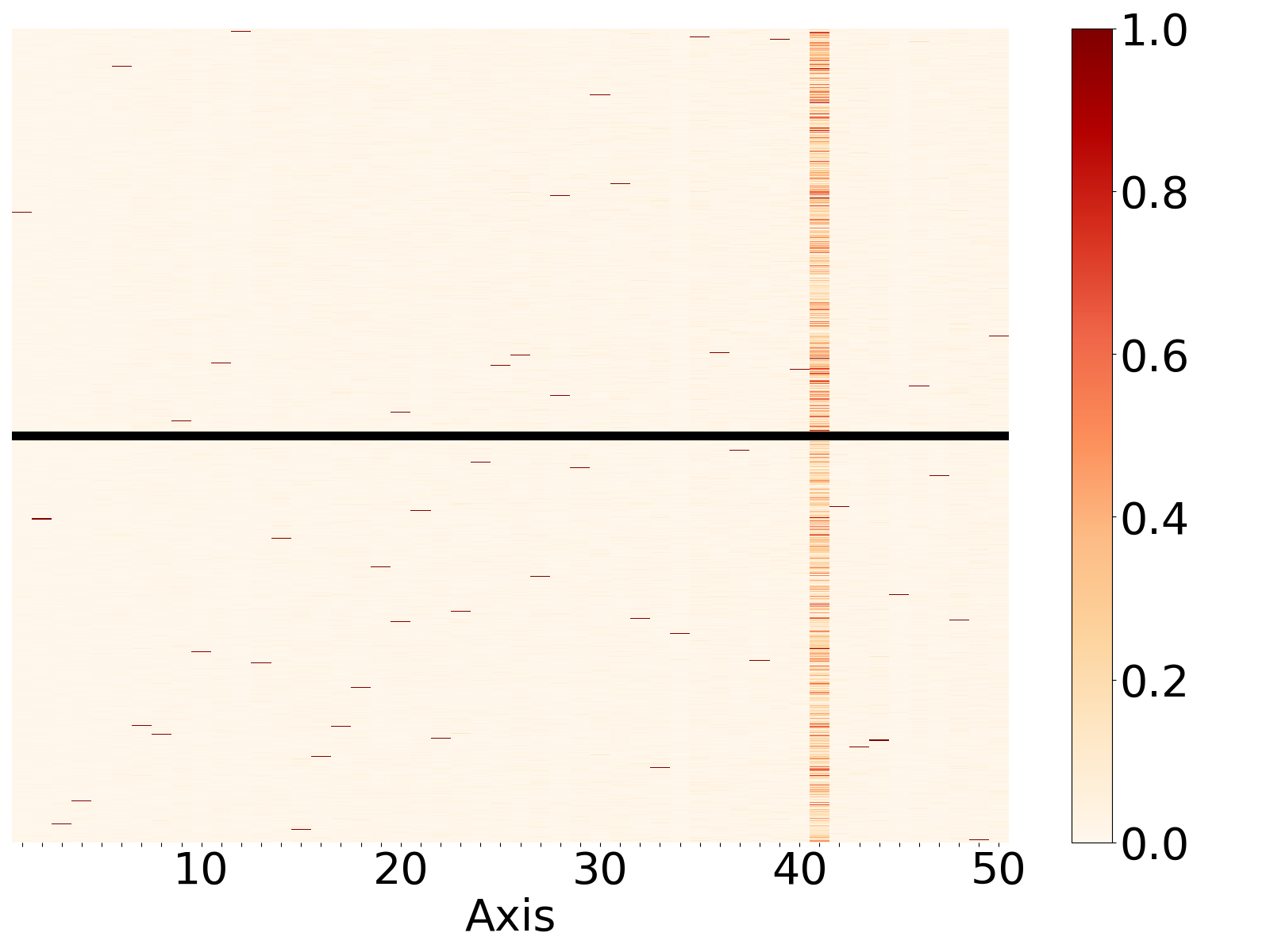}
        \subcaption{Fine-tuned \ac{SCWE}, ICA}
        \label{fig:wic_am2ico_eu_instances_ica_finetuned}
    \end{minipage}
    \caption{Visualisation of the top-50 dimensions of pre-trained \acp{CWE} (XLM-RoBERTa) and \acp{SCWE} (XL-LEXEME) for each instance in AM$^2$iCo (Basque) dataset, where the difference of vectors is calculated for (a/d) \textbf{Raw} vectors, (b/e) \ac{PCA}-transformed axes, and (c/f) \ac{ICA}-transformed axes. In each figure, the upper/lower half uses instances for the True/False labels.}
    \label{fig:wic_instance_am2ico_eu}
\end{figure*}

\begin{figure*}[t]
    \centering
    \begin{minipage}[b]{\columnwidth}
        \centering
        \includegraphics[width=0.85\columnwidth]{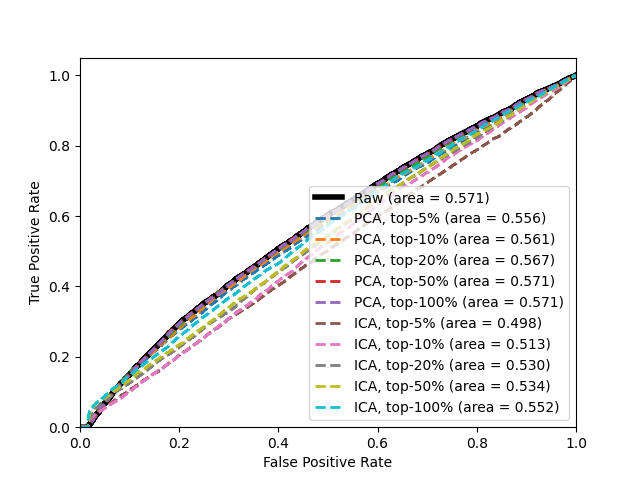}
        \subcaption{Pre-trained \ac{CWE}, De}
        \label{fig:xlwic_de_roc_pretrained}
    \end{minipage}
    \begin{minipage}[b]{\columnwidth}
        \centering
        \includegraphics[width=0.85\columnwidth]{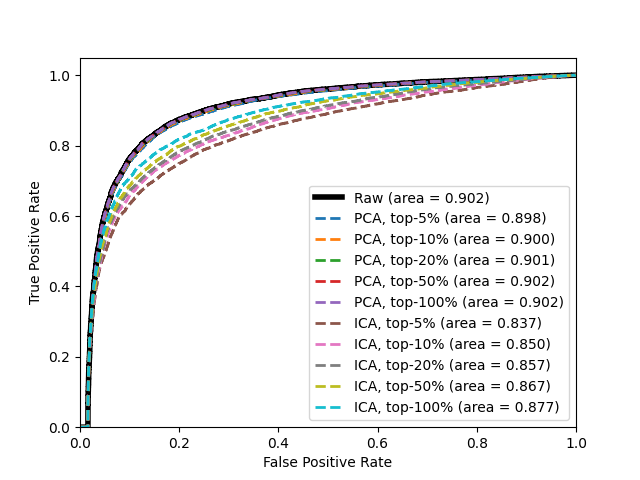}
        \subcaption{Fine-tuned \ac{SCWE}, De}
        \label{fig:xlwic_de_roc_finetuned}
    \end{minipage} \\
    \begin{minipage}[b]{\columnwidth}
        \centering
        \includegraphics[width=0.85\columnwidth]{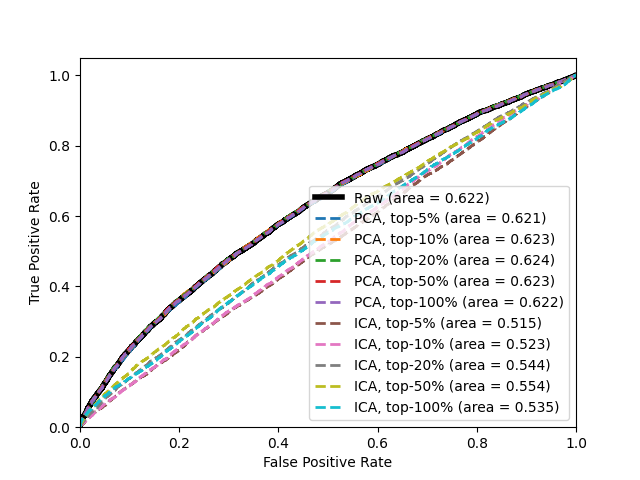}
        \subcaption{Pre-trained \ac{CWE}, Fr}
        \label{fig:xlwic_fr_roc_pretrained}
    \end{minipage}
    \begin{minipage}[b]{\columnwidth}
        \centering
        \includegraphics[width=0.85\columnwidth]{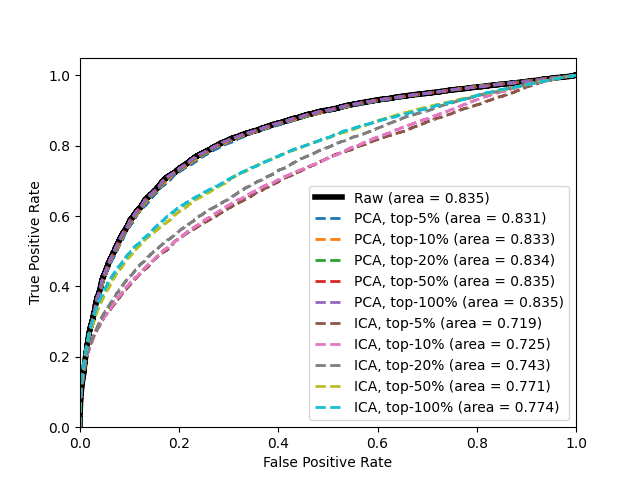}
        \subcaption{Fine-tuned \ac{SCWE}, Fr}
        \label{fig:xlwic_fr_roc_finetuned}
    \end{minipage} \\
    \begin{minipage}[b]{\columnwidth}
        \centering
        \includegraphics[width=0.85\columnwidth]{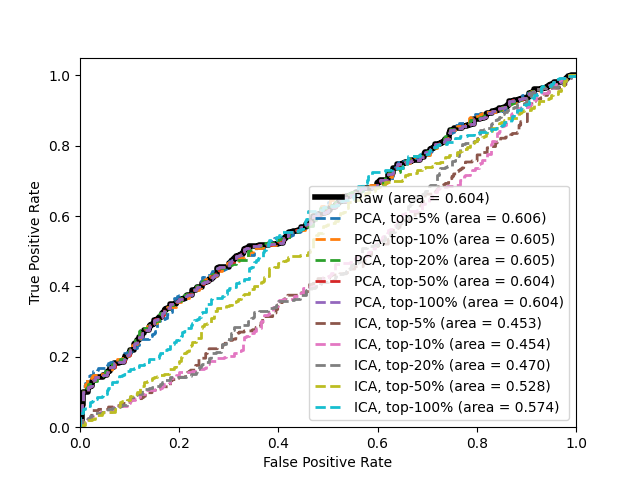}
        \subcaption{Pre-trained \ac{CWE}, It}
        \label{fig:xlwic_it_roc_pretrained}
    \end{minipage}
    \begin{minipage}[b]{\columnwidth}
        \centering
        \includegraphics[width=0.85\columnwidth]{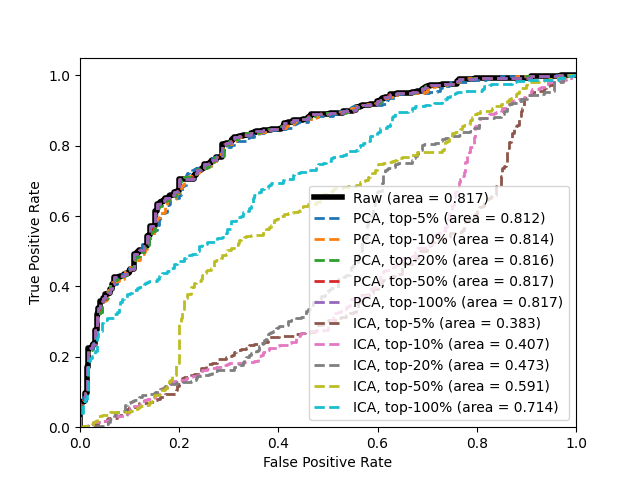}
        \subcaption{Fine-tuned \ac{SCWE}, It}
        \label{fig:xlwic_it_roc_finetuned}
    \end{minipage}
    \caption{The ROC curve on the contextual \ac{SCD} benchmark, XLWiC dataset (De: German, Fr: French, It: Italian).
    \textbf{Raw} indicates the performance of using full dimensions.
    \ac{PCA}/\ac{ICA} uses top-5/10/20/50/100\% of axes.}
    \label{fig:xlwic}
\end{figure*}

\begin{figure*}[t]
    \centering
    \begin{minipage}[b]{\columnwidth}
        \centering
        \includegraphics[width=0.85\columnwidth]{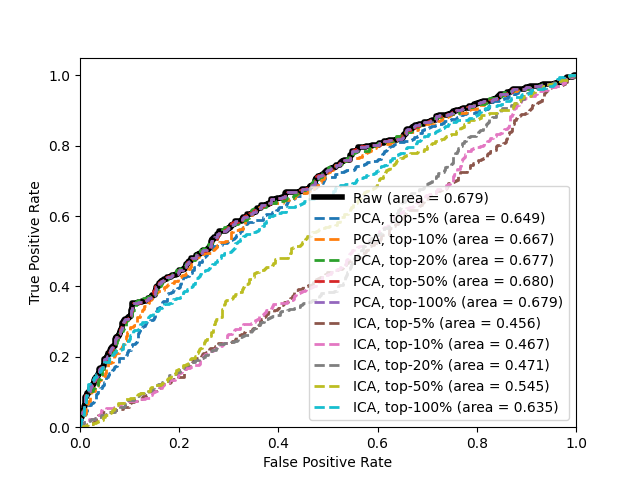}
        \subcaption{Pre-trained \ac{CWE}, Ar}
        \label{fig:mclwic_ar_roc_pretrained}
    \end{minipage}
    \begin{minipage}[b]{\columnwidth}
        \centering
        \includegraphics[width=0.85\columnwidth]{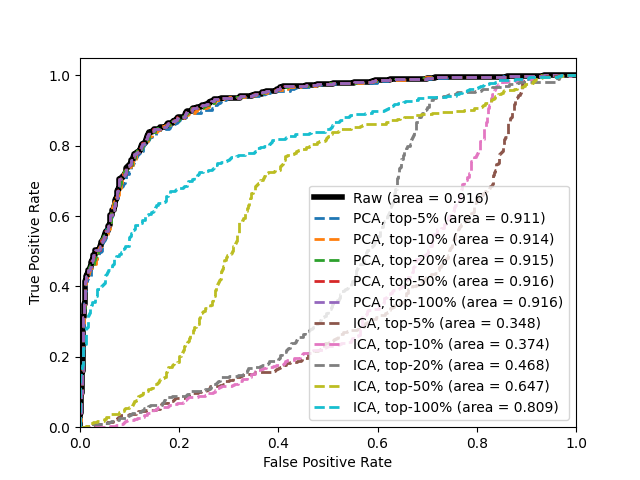}
        \subcaption{Fine-tuned \ac{SCWE}, Ar}
        \label{fig:mclwic_ar_roc_finetuned}
    \end{minipage} \\
    \begin{minipage}[b]{\columnwidth}
        \centering
        \includegraphics[width=0.85\columnwidth]{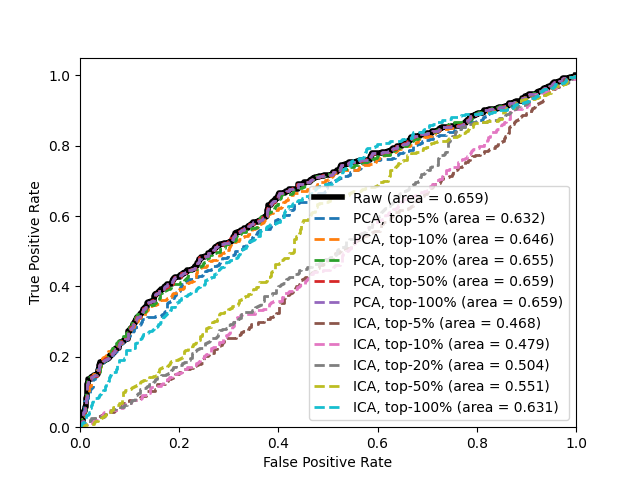}
        \subcaption{Pre-trained \ac{CWE}, En}
        \label{fig:mclwic_en_roc_pretrained}
    \end{minipage}
    \begin{minipage}[b]{\columnwidth}
        \centering
        \includegraphics[width=0.85\columnwidth]{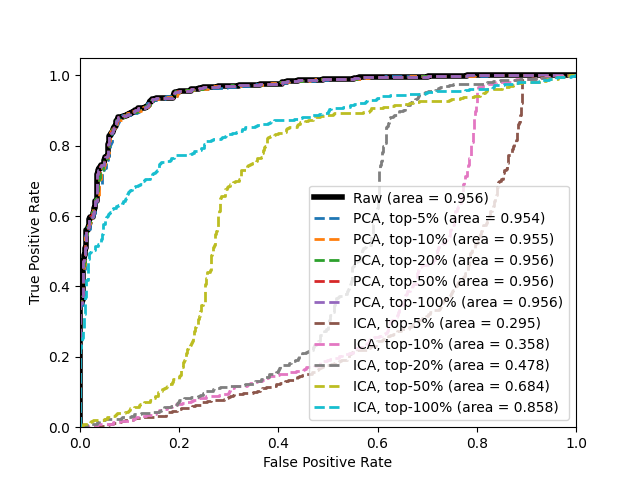}
        \subcaption{Fine-tuned \ac{SCWE}, En}
        \label{fig:mclwic_en_roc_finetuned}
    \end{minipage} \\
    \begin{minipage}[b]{\columnwidth}
        \centering
        \includegraphics[width=0.85\columnwidth]{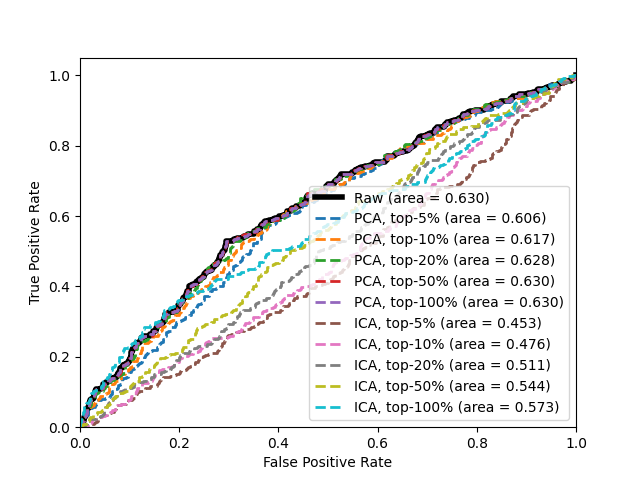}
        \subcaption{Pre-trained \ac{CWE}, Fr}
        \label{fig:mclwic_fr_roc_pretrained}
    \end{minipage}
    \begin{minipage}[b]{\columnwidth}
        \centering
        \includegraphics[width=0.85\columnwidth]{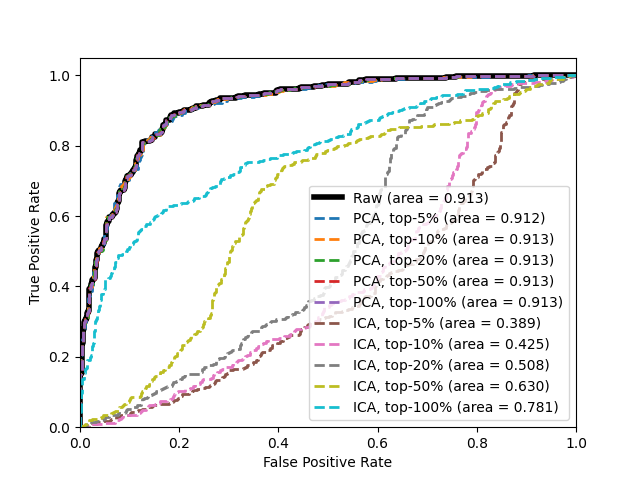}
        \subcaption{Fine-tuned \ac{SCWE}, Fr}
        \label{fig:mclwic_fr_roc_finetuned}
    \end{minipage}
    \caption{The ROC curve on the contextual \ac{SCD} benchmark, MCLWiC dataset (Ar: Arabic, En: English, Fr: French).
    \textbf{Raw} indicates the performance of using full dimensions.
    \ac{PCA}/\ac{ICA} uses top-5/10/20/50/100\% of axes.}
    \label{fig:mclwic_ar_en_fr}
\end{figure*}

\begin{figure*}[t]
    \centering
    \begin{minipage}[b]{\columnwidth}
        \centering
        \includegraphics[width=0.85\columnwidth]{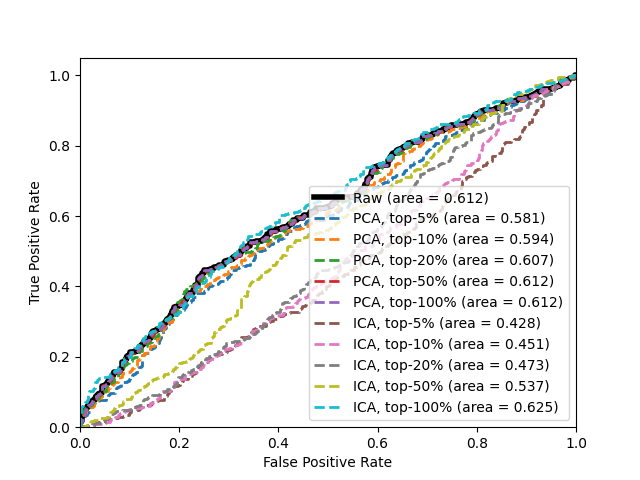}
        \subcaption{Pre-trained \ac{CWE}, Ru}
        \label{fig:mclwic_ru_roc_pretrained}
    \end{minipage}
    \begin{minipage}[b]{\columnwidth}
        \centering
        \includegraphics[width=0.85\columnwidth]{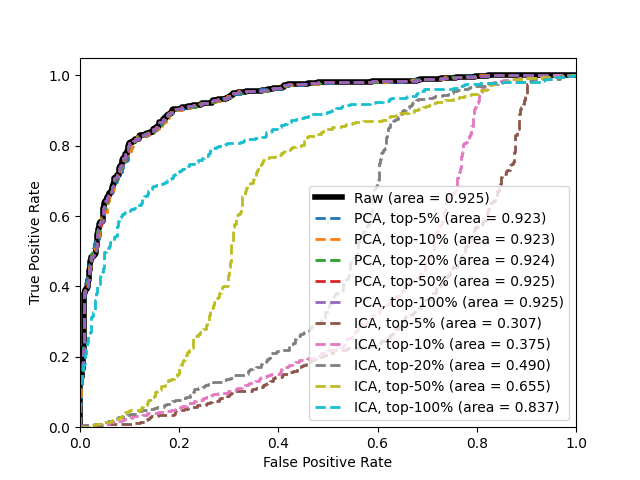}
        \subcaption{Fine-tuned \ac{SCWE}, Ru}
        \label{fig:mclwic_ru_roc_finetuned}
    \end{minipage} \\ 
    \begin{minipage}[b]{\columnwidth}
        \centering
        \includegraphics[width=0.85\columnwidth]{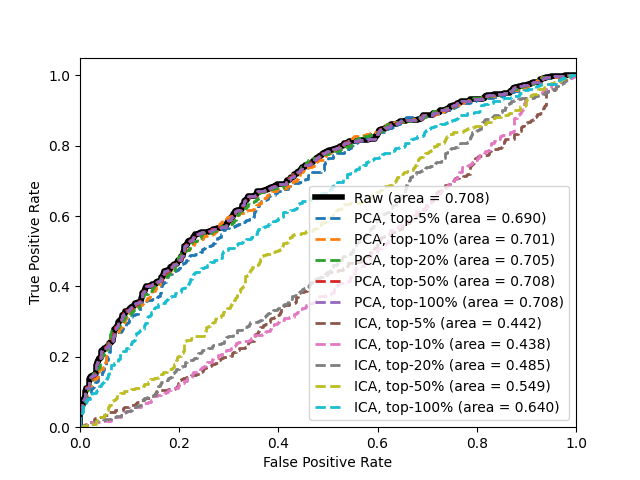}
        \subcaption{Pre-trained \ac{CWE}, Zh}
        \label{fig:mclwic_zh_roc_pretrained}
    \end{minipage}
    \begin{minipage}[b]{\columnwidth}
        \centering
        \includegraphics[width=0.85\columnwidth]{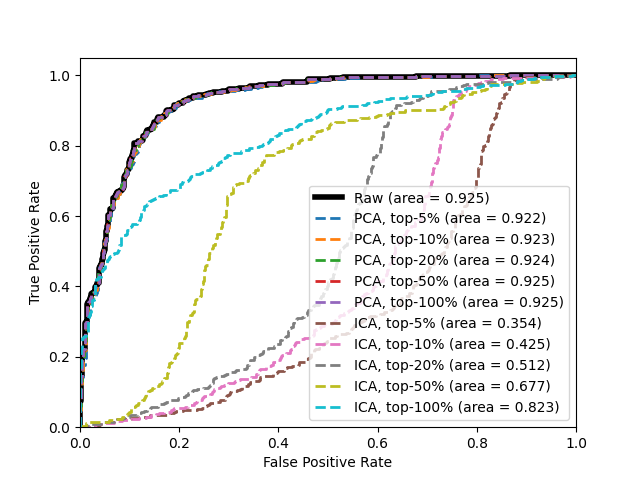}
        \subcaption{Fine-tuned \ac{SCWE}, Zh}
        \label{fig:mclwic_zh_roc_finetuned}
    \end{minipage}
    \caption{The ROC curve on the contextual \ac{SCD} benchmark, MCLWiC dataset (Ru: Russian, Zh: Chinese).
    \textbf{Raw} indicates the performance of using full dimensions.
    \ac{PCA}/\ac{ICA} uses top-5/10/20/50/100\% of axes.}
    \label{fig:mclwic_ru_zh}
\end{figure*}

\begin{figure*}[t]
    \centering
    \begin{minipage}[b]{\columnwidth}
        \centering
        \includegraphics[width=0.85\columnwidth]{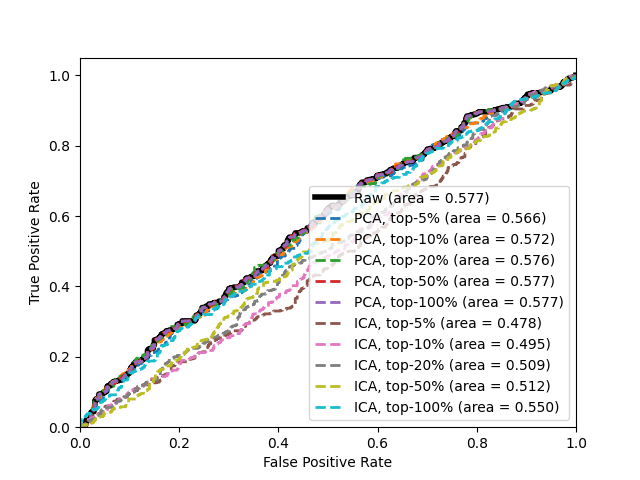}
        \subcaption{Pre-trained \ac{CWE}, De}
        \label{fig:am2ico_de_roc_pretrained}
    \end{minipage}
    \begin{minipage}[b]{\columnwidth}
        \centering
        \includegraphics[width=0.85\columnwidth]{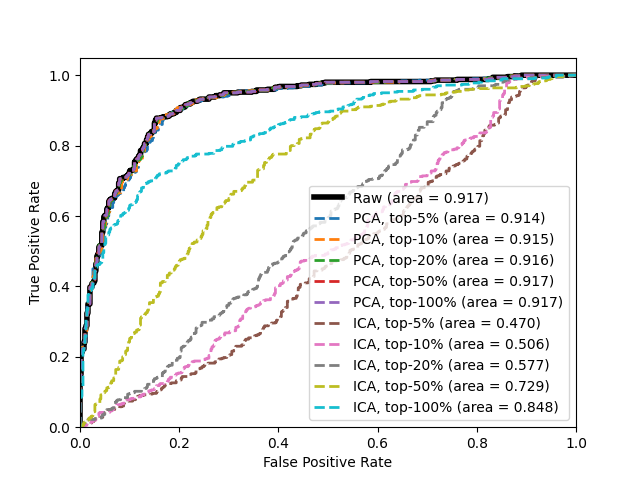}
        \subcaption{Fine-tuned \ac{SCWE}, De}
        \label{fig:am2ico_de_roc_finetuned}
    \end{minipage} \\
    \begin{minipage}[b]{\columnwidth}
        \centering
        \includegraphics[width=0.85\columnwidth]{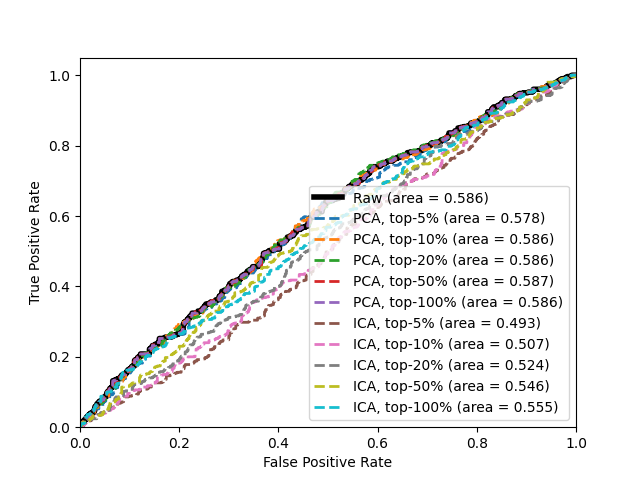}
        \subcaption{Pre-trained \ac{CWE}, Ru}
        \label{fig:am2ico_ru_roc_pretrained}
    \end{minipage}
    \begin{minipage}[b]{\columnwidth}
        \centering
        \includegraphics[width=0.85\columnwidth]{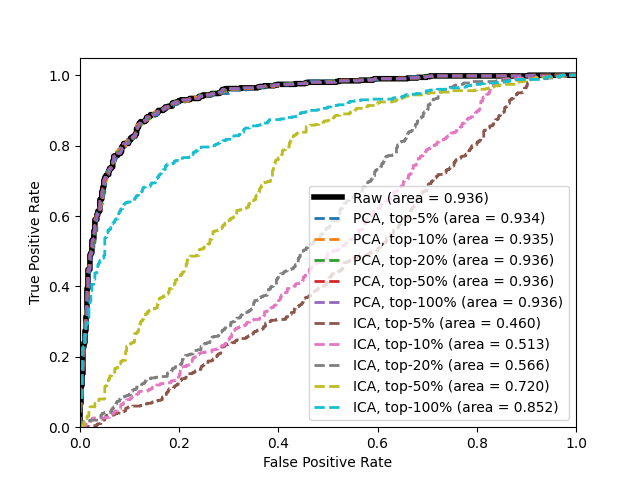}
        \subcaption{Fine-tuned \ac{SCWE}, Ru}
        \label{fig:am2ico_ru_roc_finetuned}
    \end{minipage} \\
    \begin{minipage}[b]{\columnwidth}
        \centering
        \includegraphics[width=0.85\columnwidth]{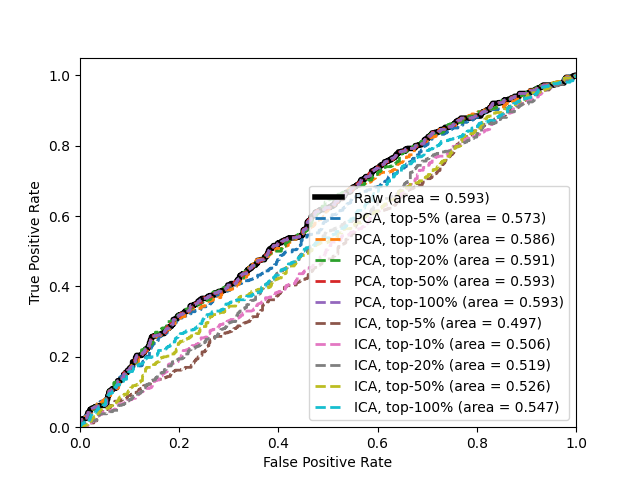}
        \subcaption{Pre-trained \ac{CWE}, Ja}
        \label{fig:am2ico_ja_roc_pretrained}
    \end{minipage}
    \begin{minipage}[b]{\columnwidth}
        \centering
        \includegraphics[width=0.85\columnwidth]{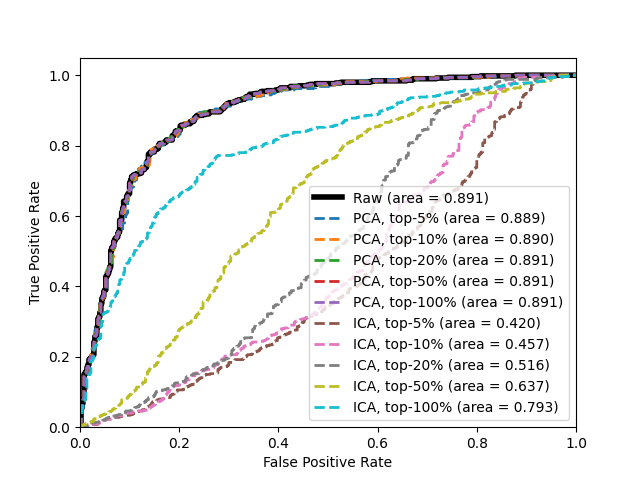}
        \subcaption{Fine-tuned \ac{SCWE}, Ja}
        \label{fig:am2ico_ja_roc_finetuned}
    \end{minipage} \\
    \begin{minipage}[b]{\columnwidth}
        \centering
        \includegraphics[width=0.85\columnwidth]{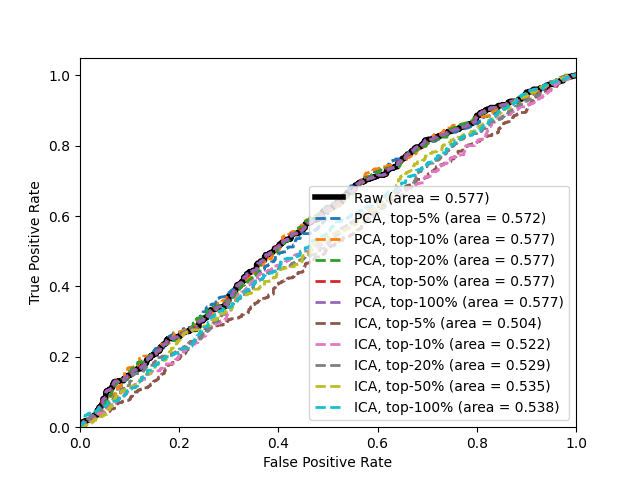}
        \subcaption{Pre-trained \ac{CWE}, Zh}
        \label{fig:am2ico_zh_roc_pretrained}
    \end{minipage}
    \begin{minipage}[b]{\columnwidth}
        \centering
        \includegraphics[width=0.85\columnwidth]{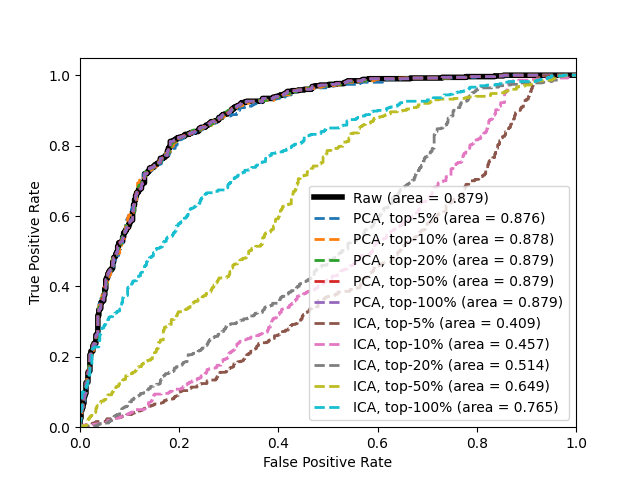}
        \subcaption{Fine-tuned \ac{SCWE}, Zh}
        \label{fig:am2ico_zh_roc_finetuned}
    \end{minipage}
    \caption{The ROC curve on the contextual \ac{SCD} benchmark, AM$^2$iCo dataset (De: German, Ru: Russian, Ja: Japanese, Zh: Chinese).
    \textbf{Raw} indicates the performance of using full dimensions.
    \ac{PCA}/\ac{ICA} uses top-5/10/20/50/100\% of axes.}
    \label{fig:am2ico_de_ru_ja_zh}
\end{figure*}

\begin{figure*}[t]
    \centering
    \begin{minipage}[b]{\columnwidth}
        \centering
        \includegraphics[width=0.85\columnwidth]{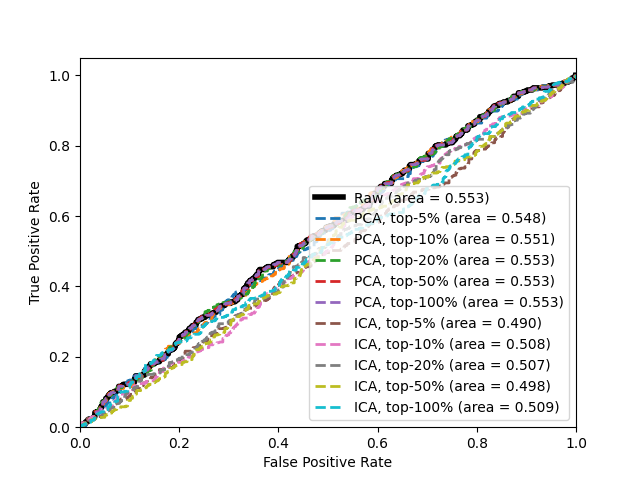}
        \subcaption{Pre-trained \ac{CWE}, Ar}
        \label{fig:am2ico_ar_roc_pretrained}
    \end{minipage}
    \begin{minipage}[b]{\columnwidth}
        \centering
        \includegraphics[width=0.85\columnwidth]{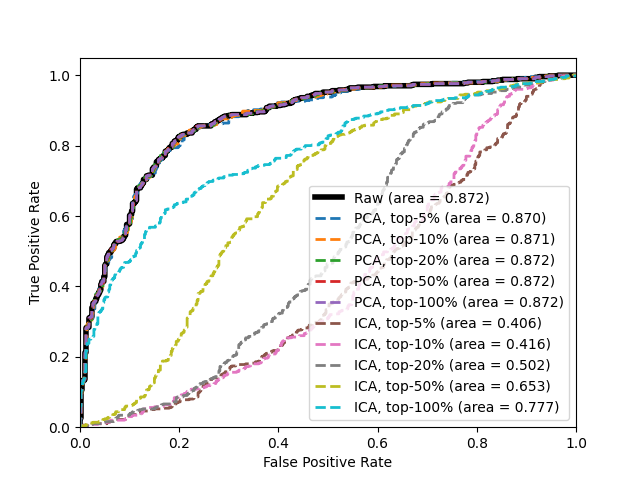}
        \subcaption{Fine-tuned \ac{SCWE}, Ar}
        \label{fig:am2ico_ar_roc_finetuned}
    \end{minipage} \\
    \begin{minipage}[b]{\columnwidth}
        \centering
        \includegraphics[width=0.85\columnwidth]{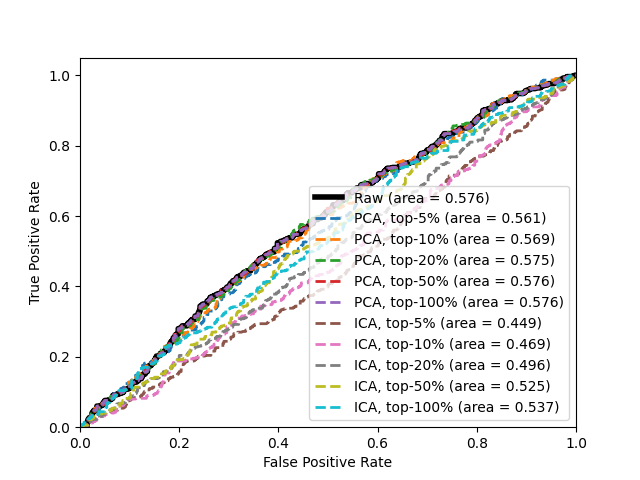}
        \subcaption{Pre-trained \ac{CWE}, Ko}
        \label{fig:am2ico_ko_roc_pretrained}
    \end{minipage}
    \begin{minipage}[b]{\columnwidth}
        \centering
        \includegraphics[width=0.85\columnwidth]{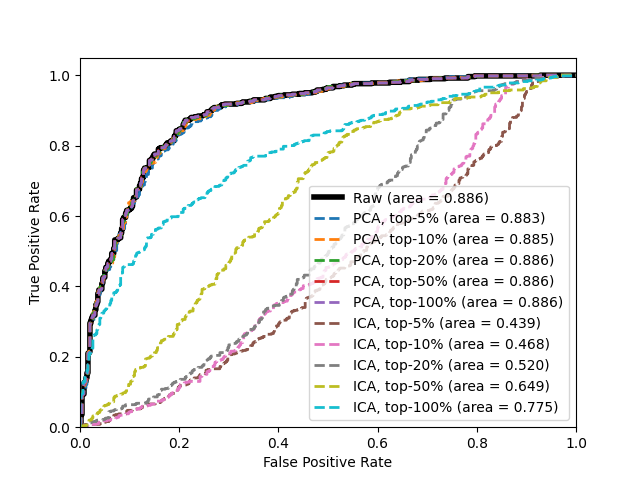}
        \subcaption{Fine-tuned \ac{SCWE}, Ko}
        \label{fig:am2ico_ko_roc_finetuned}
    \end{minipage} \\
    \begin{minipage}[b]{\columnwidth}
        \centering
        \includegraphics[width=0.85\columnwidth]{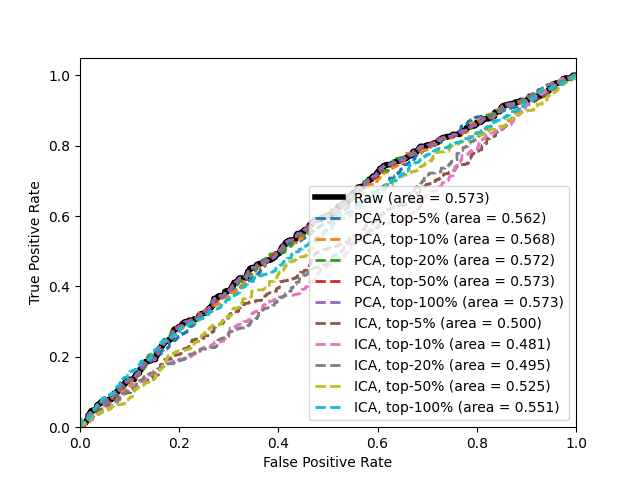}
        \subcaption{Pre-trained \ac{CWE}, Fi}
        \label{fig:am2ico_fi_roc_pretrained}
    \end{minipage}
    \begin{minipage}[b]{\columnwidth}
        \centering
        \includegraphics[width=0.85\columnwidth]{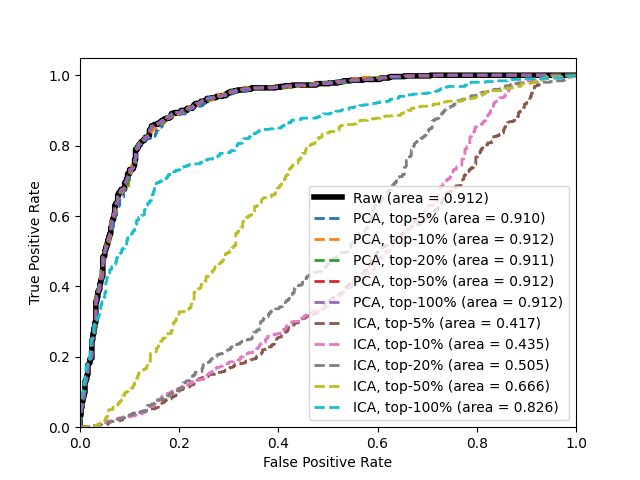}
        \subcaption{Fine-tuned \ac{SCWE}, Fi}
        \label{fig:am2ico_fi_roc_finetuned}
    \end{minipage} 
    \caption{The ROC curve on the contextual \ac{SCD} benchmark, AM$^2$iCo dataset (Ar: Arabic, Ko: Korean, Fi: Finnish).
    \textbf{Raw} indicates the performance of using full dimensions.
    \ac{PCA}/\ac{ICA} uses top-5/10/20/50/100\% of axes.}
    \label{fig:am2ico_ar_ko_fi}
\end{figure*}

\begin{figure*}[t]
    \centering
    \begin{minipage}[b]{\columnwidth}
        \centering
        \includegraphics[width=0.85\columnwidth]{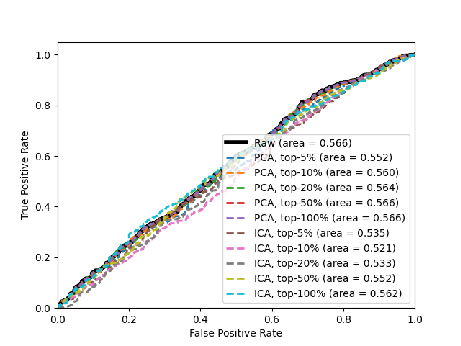}
        \subcaption{Pre-trained \ac{CWE}, Tr}
        \label{fig:am2ico_tr_roc_pretrained}
    \end{minipage}
    \begin{minipage}[b]{\columnwidth}
        \centering
        \includegraphics[width=0.85\columnwidth]{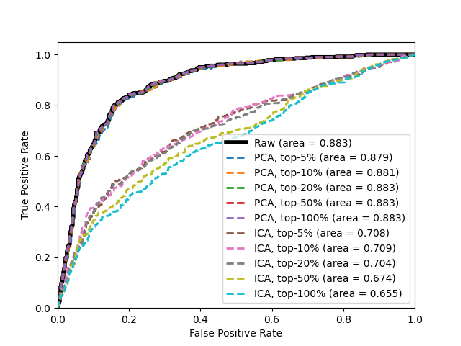}
        \subcaption{Fine-tuned \ac{SCWE}, Tr}
        \label{fig:am2ico_tr_roc_finetuned}
    \end{minipage} \\
    \begin{minipage}[b]{\columnwidth}
        \centering
        \includegraphics[width=0.85\columnwidth]{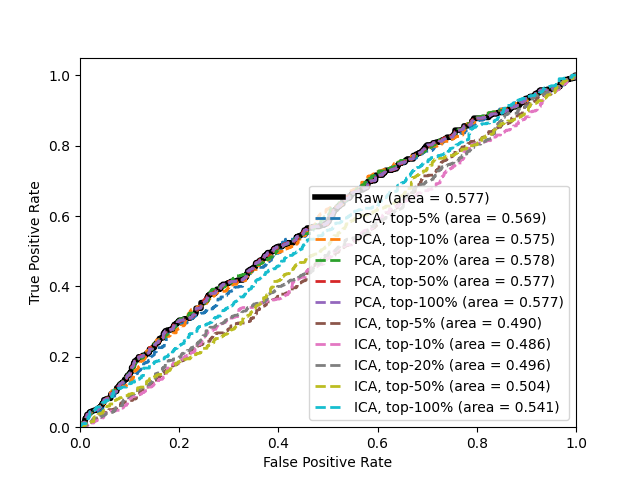}
        \subcaption{Pre-trained \ac{CWE}, Id}
        \label{fig:am2ico_id_roc_pretrained}
    \end{minipage}
    \begin{minipage}[b]{\columnwidth}
        \centering
        \includegraphics[width=0.85\columnwidth]{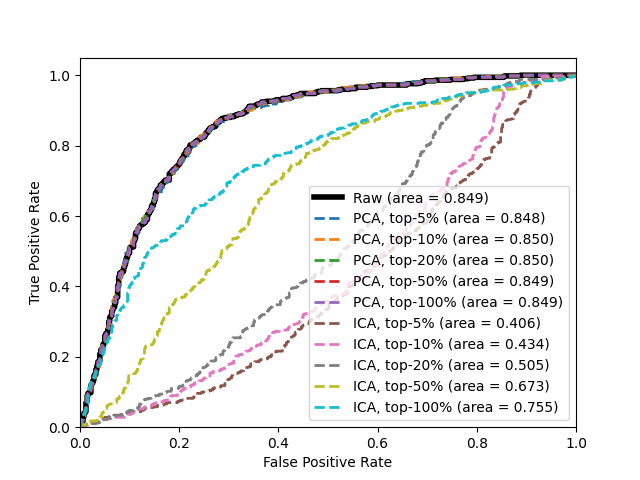}
        \subcaption{Fine-tuned \ac{SCWE}, Id}
        \label{fig:am2ico_id_roc_finetuned}
    \end{minipage} \\
    \begin{minipage}[b]{\columnwidth}
        \centering
        \includegraphics[width=0.85\columnwidth]{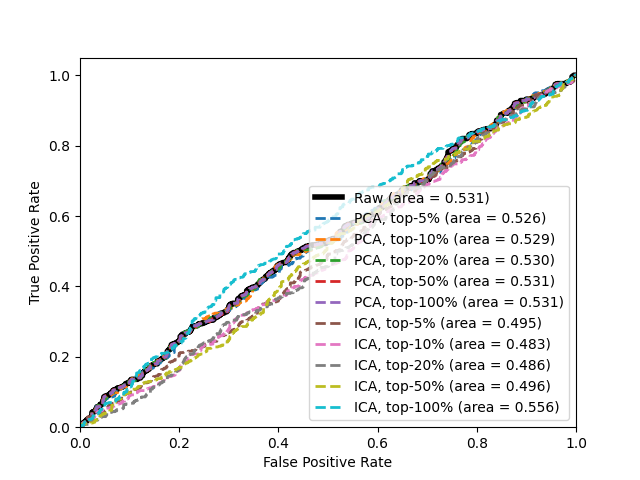}
        \subcaption{Pre-trained \ac{CWE}, Eu}
        \label{fig:am2ico_eu_roc_pretrained}
    \end{minipage}
    \begin{minipage}[b]{\columnwidth}
        \centering
        \includegraphics[width=0.85\columnwidth]{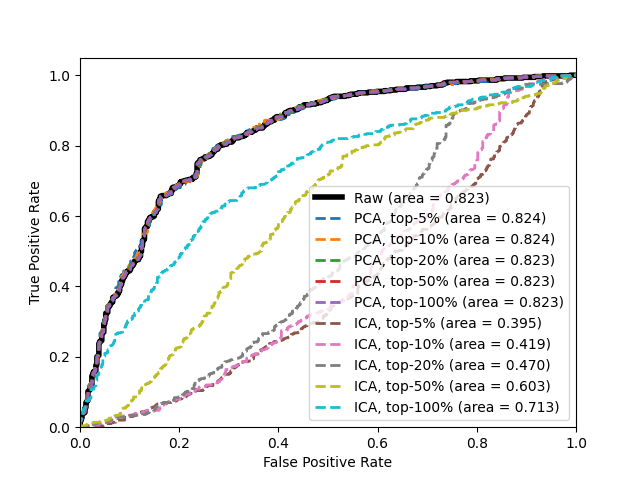}
        \subcaption{Fine-tuned \ac{SCWE}, Eu}
        \label{fig:am2ico_eu_roc_finetuned}
    \end{minipage} 
    \caption{The ROC curve on the contextual \ac{SCD} benchmark, AM$^2$iCo dataset (Tr: Turkish, Id: Indonesian, Eu: Basque).
    \textbf{Raw} indicates the performance of using full dimensions.
    \ac{PCA}/\ac{ICA} uses top-5/10/20/50/100\% of axes.}
    \label{fig:am2ico_tr_id_eu}
\end{figure*}

In this section, we present the full results of contextual and temporal \ac{SCD} tasks.
For the contextual \ac{SCD}, visualisations of instances in all datasets are as follows: XLWiC~(\autoref{fig:wic_instance_xlwic_de}, \autoref{fig:wic_instance_xlwic_fr}, and \autoref{fig:wic_instance_xlwic_it}), MCLWiC~(Figures \ref{fig:wic_instance_mclwic_ar}, \ref{fig:wic_instance_mclwic_en}, \ref{fig:wic_instance_mclwic_fr}, \ref{fig:wic_instance_mclwic_ru}, and \ref{fig:wic_instance_mclwic_zh}), and AM$^2$iCo~(Figures \ref{fig:wic_instance_am2ico_de}, \ref{fig:wic_instance_am2ico_ru}, \ref{fig:wic_instance_am2ico_ja}, \ref{fig:wic_instance_am2ico_zh}, \ref{fig:wic_instance_am2ico_ar}, \ref{fig:wic_instance_am2ico_ko}, \ref{fig:wic_instance_am2ico_fi}, \ref{fig:wic_instance_am2ico_tr}, \ref{fig:wic_instance_am2ico_id}, and \ref{fig:wic_instance_am2ico_eu}).
Similar to \autoref{sec:contextual}, the contextual semantic change-aware axes emerged after the fine-tuning process.
Moreover, full results related to the prediction task are as follows: XLWiC~(\autoref{fig:xlwic}), MCLWiC~(\autoref{fig:mclwic_ar_en_fr} and \autoref{fig:mclwic_ru_zh}), AM$^2$iCo~(\autoref{fig:am2ico_de_ru_ja_zh}, \autoref{fig:am2ico_ar_ko_fi}, and \autoref{fig:am2ico_tr_id_eu}).
As shown in \autoref{sec:contextual}, 10\% \ac{PCA}-transformed axes are able to obtain contextual semantic change-aware dimensions.

\begin{figure*}[t]
    \centering
    \begin{minipage}[b]{\columnwidth}
        \centering
        \includegraphics[width=0.85\columnwidth]{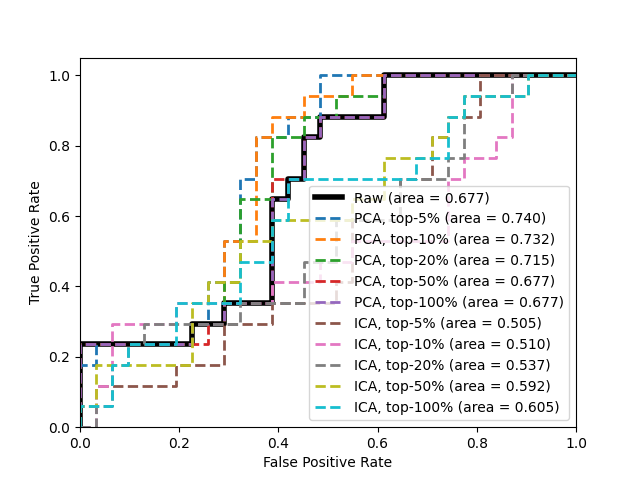}
        \subcaption{Pre-trained \ac{CWE}, De}
        \label{fig:scd_de_roc_pretrained}
    \end{minipage}
    \begin{minipage}[b]{\columnwidth}
        \centering
        \includegraphics[width=0.85\columnwidth]{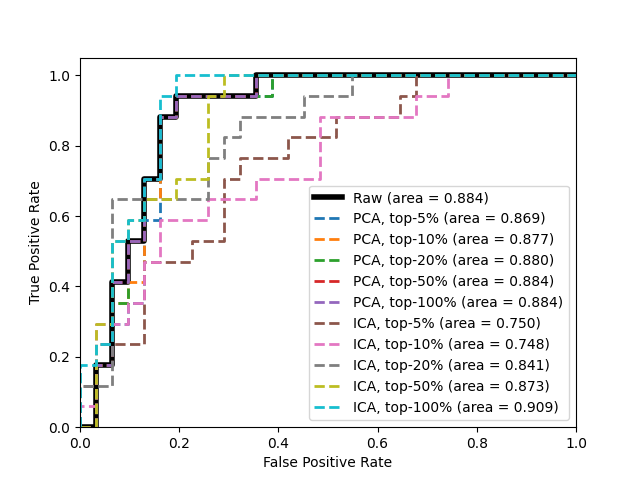}
        \subcaption{Fine-tuned \ac{SCWE}, De}
        \label{fig:scd_de_roc_finetuned}
    \end{minipage} \\
    \begin{minipage}[b]{\columnwidth}
        \centering
        \includegraphics[width=0.85\columnwidth]{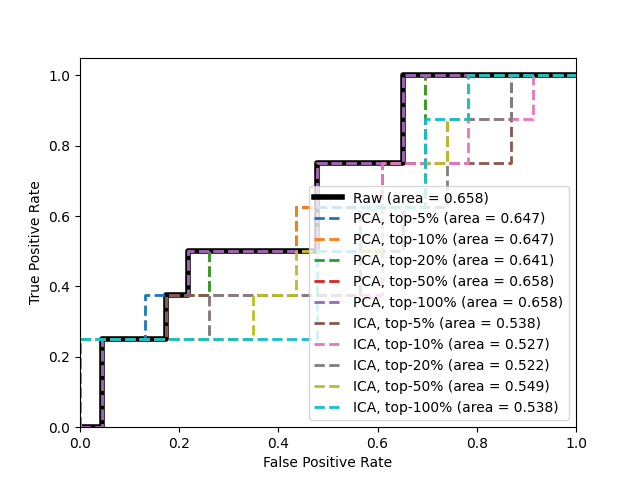}
        \subcaption{Pre-trained \ac{CWE}, Sv}
        \label{fig:scd_sv_roc_pretrained}
    \end{minipage}
    \begin{minipage}[b]{\columnwidth}
        \centering
        \includegraphics[width=0.85\columnwidth]{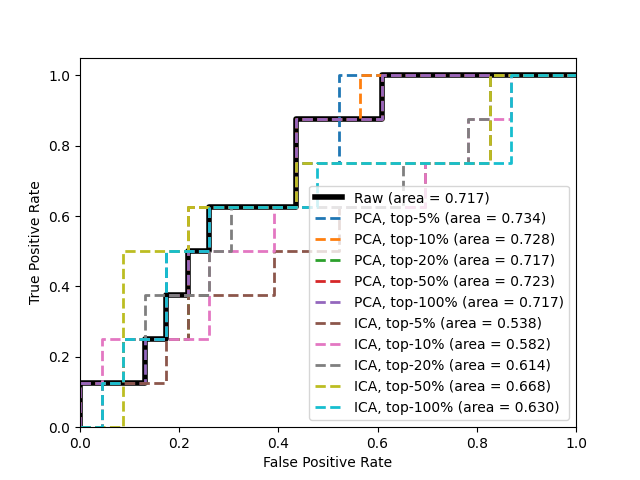}
        \subcaption{Fine-tuned \ac{SCWE}, Sv}
        \label{fig:scd_sv_roc_finetuned}
    \end{minipage} \\
    \begin{minipage}[b]{\columnwidth}
        \centering
        \includegraphics[width=0.85\columnwidth]{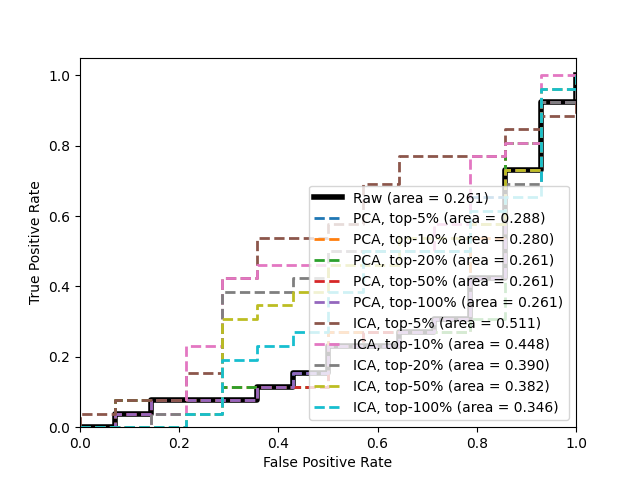}
        \subcaption{Pre-trained \ac{CWE}, La}
        \label{fig:scd_la_roc_pretrained}
    \end{minipage}
    \begin{minipage}[b]{\columnwidth}
        \centering
        \includegraphics[width=0.85\columnwidth]{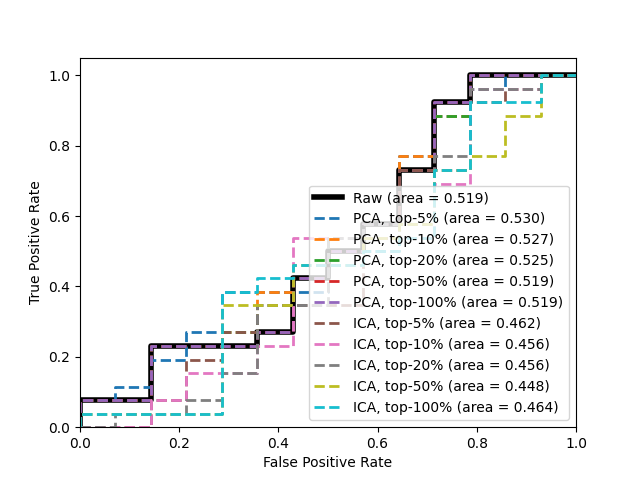}
        \subcaption{Fine-tuned \ac{SCWE}, La}
        \label{fig:scd_la_roc_finetuned}
    \end{minipage}
    \caption{The ROC curve on the temporal \ac{SCD} benchmark, SemEval-2020 Task 1 (De: German, Sv: Swedish, La: Latin).
    \textbf{Raw} indicates the performance of using full dimensions.
    \ac{PCA}/\ac{ICA} uses top-5/10/20/50/100\% of axes.}
    \label{fig:scd_roc_full}
\end{figure*}

\begin{figure*}[t]
    \centering
    \begin{minipage}[b]{\columnwidth}
        \centering
        \includegraphics[width=0.85\columnwidth]{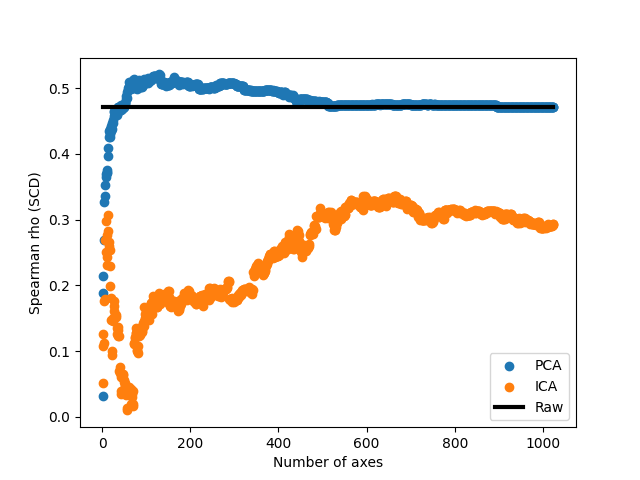}
        \subcaption{Pre-trained \ac{CWE}, De}
        \label{fig:scd_de_spearman_pretrained}
    \end{minipage}
    \begin{minipage}[b]{\columnwidth}
        \centering
        \includegraphics[width=0.85\columnwidth]{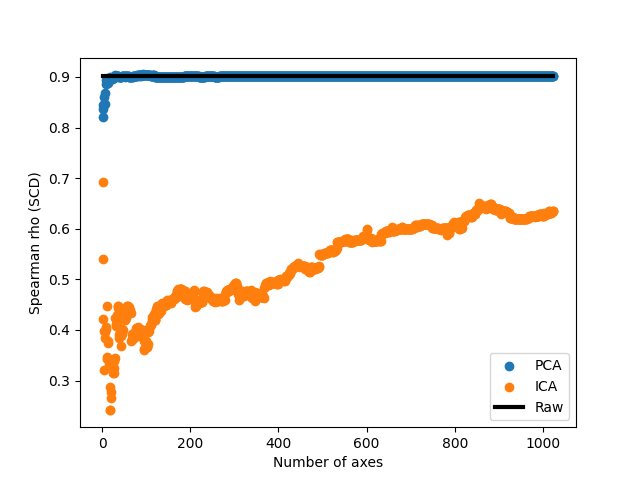}
        \subcaption{Fine-tuned \ac{SCWE}, De}
        \label{fig:scd_de_spearman_finetuned}
    \end{minipage} \\
    \begin{minipage}[b]{\columnwidth}
        \centering
        \includegraphics[width=0.85\columnwidth]{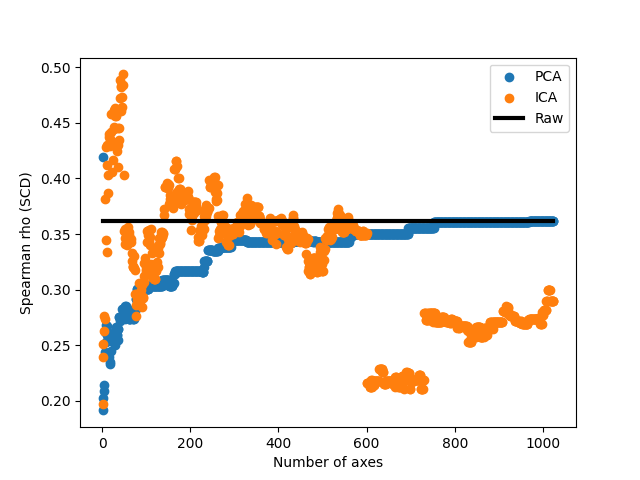}
        \subcaption{Pre-trained \ac{CWE}, Sv}
        \label{fig:scd_sv_spearman_pretrained}
    \end{minipage}
    \begin{minipage}[b]{\columnwidth}
        \centering
        \includegraphics[width=0.85\columnwidth]{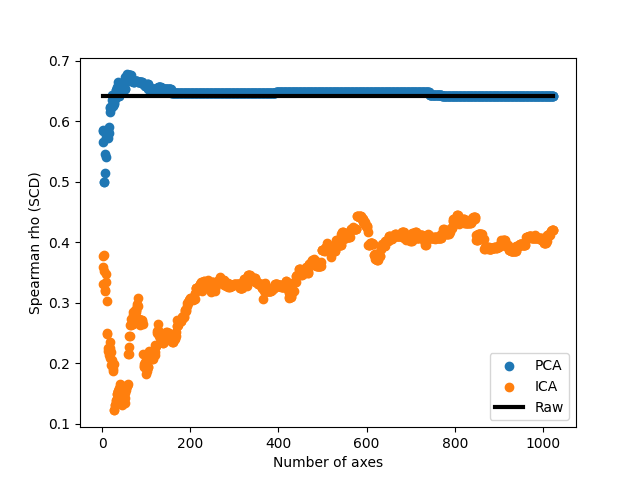}
        \subcaption{Fine-tuned \ac{SCWE}, Sv}
        \label{fig:scd_sv_spearman_finetuned}
    \end{minipage} \\
    \begin{minipage}[b]{\columnwidth}
        \centering
        \includegraphics[width=0.85\columnwidth]{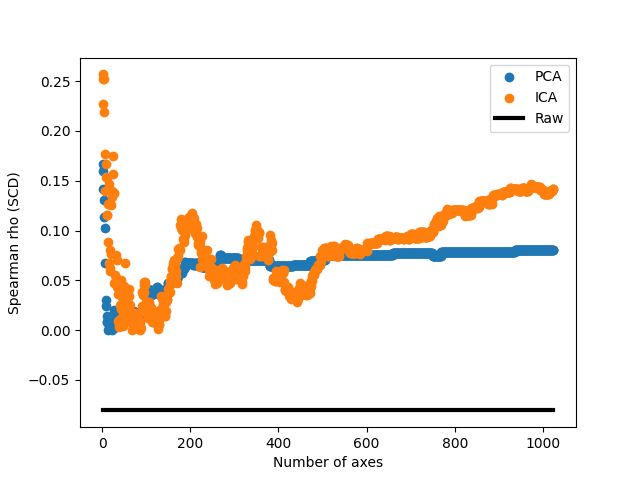}
        \subcaption{Pre-trained \ac{CWE}, La}
        \label{fig:scd_la_spearman_pretrained}
    \end{minipage}
    \begin{minipage}[b]{\columnwidth}
        \centering
        \includegraphics[width=0.85\columnwidth]{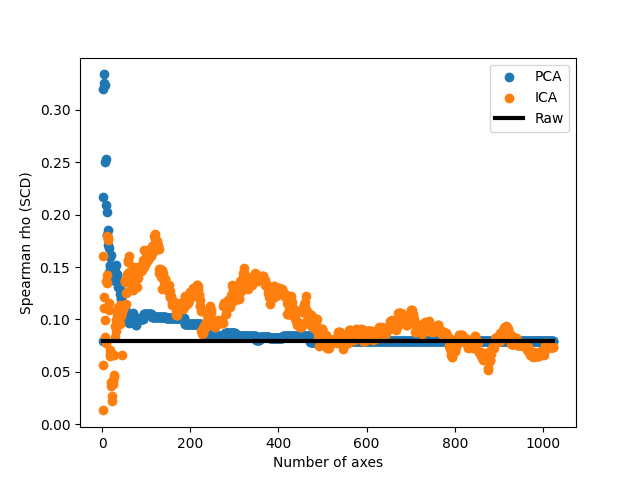}
        \subcaption{Fine-tuned \ac{SCWE}, La}
        \label{fig:scd_la_spearman_finetuned}
    \end{minipage}
    \caption{Spearman's rank correlation on the temporal \ac{SCD} benchmark, SemEval-2020 Task 1 (De: German, Sv: Swedish, La: Latin).
    \textbf{Raw} indicates the performance of using full dimensions. 
    \ac{PCA}/\ac{ICA} cumulatively uses sorted axes.}
    \label{fig:scd_spearman_full}
\end{figure*}

On the other hand, for the temporal \ac{SCD}, results for other languages (German, Swedish, and Latin) are shown in \autoref{fig:scd_roc_full} and \autoref{fig:scd_spearman_full}.
Similar to \autoref{sec:temporal}, temporal semantic change-aware dimensions are observed within 10\% \ac{PCA}-transformed axes.
However, there are some difficulties in obtaining these dimensions by PCA-transformed axes with insufficient pretraining data (Swedish)~\cite{conneau-etal-2020-xlmroberta} or lack of supervision for fine-tuning (Latin) shown in \autoref{tab:data_wic}.
In those cases, the use of \ac{ICA}-transformed axes proved to be effective.
More detailed analysis and understanding of those axes for interpretability will be addressed in future work.

\end{document}